\documentclass[10pt,twocolumn,letterpaper]{article}

\usepackage{cvpr}
\usepackage{times}
\usepackage{epsfig}
\usepackage{graphicx}
\usepackage{amsmath}
\usepackage{amssymb}

\usepackage{caption}
\usepackage{subcaption}
\usepackage{booktabs}
\usepackage{multirow}
\usepackage{gensymb}
\usepackage{array}
\usepackage[export]{adjustbox}
\usepackage{xcolor}
\usepackage[inline]{enumitem}
\usepackage{dblfloatfix}  
\usepackage{cuted}  
\usepackage{xr}  

\usepackage[pagebackref=true,breaklinks=true,letterpaper=true,colorlinks,bookmarks=false]{hyperref}

\newcommand\blfootnote[1]{%
  \begingroup
  \renewcommand\thefootnote{}\footnote{#1}%
  \addtocounter{footnote}{-1}%
  \endgroup
}
\usepackage[hang,flushmargin]{footmisc}  

\cvprfinalcopy 
\newif\ifproceedings  
\proceedingsfalse
\newif\ifsupponly
\supponlyfalse

\def\cvprPaperID{7621} 
\newcommand{\supp}{Appendix}

\ifproceedings
\else
\let\OLDthebibliography\thebibliography
\renewcommand\thebibliography[1]{
  \OLDthebibliography{#1}
  \linespread{0.935}\selectfont
}
\fi

\newcommand{\PAR}[1]{\vskip4pt \noindent{\bf #1~}}

\renewcommand{\b}[1]{\textbf{#1}}
\newcommand{\0}{\phantom{0}}

\ifproceedings\ifcvprfinal\fi\fi

\begin{document}

\title{SuperGlue: Learning Feature Matching with Graph Neural Networks}

\author{%
Paul-Edouard Sarlin$^{1}$\footnotemark
\hspace{.1in} Daniel DeTone$^2$ 
\hspace{.05in} Tomasz Malisiewicz$^2$ 
\hspace{.05in} Andrew Rabinovich$^2$ \\
$^1$ ETH Zurich
\hspace{.2in} $^2$ Magic Leap, Inc.
}

\ifsupponly\else
\maketitle
\ifproceedings\ifcvprfinal\thispagestyle{empty}\fi\fi

\begin{abstract}
This paper introduces SuperGlue, a neural network that matches two sets of local features by jointly finding correspondences and rejecting non-matchable points. Assignments are estimated by solving a differentiable optimal transport problem, whose costs are predicted by a graph neural network. We introduce a flexible context aggregation mechanism based on attention, enabling SuperGlue to reason about the underlying 3D scene and feature assignments jointly. Compared to traditional, hand-designed heuristics, our technique learns priors over geometric transformations and regularities of the 3D world through end-to-end training from image pairs.
SuperGlue outperforms other learned approaches and achieves state-of-the-art results on the task of pose estimation in challenging real-world indoor and outdoor environments. The proposed method performs matching in real-time on a modern GPU and can be readily integrated into modern SfM or SLAM systems. The code and trained weights are publicly available at \footnotesize{\href{https://github.com/magicleap/SuperGluePretrainedNetwork}{\texttt{github.com/magicleap/SuperGluePretrainedNetwork}}}.
\end{abstract}

\vspace{-3mm}
\section{Introduction}

\begin{figure}[t]
   \centering
   \includegraphics[width=1.0\linewidth]{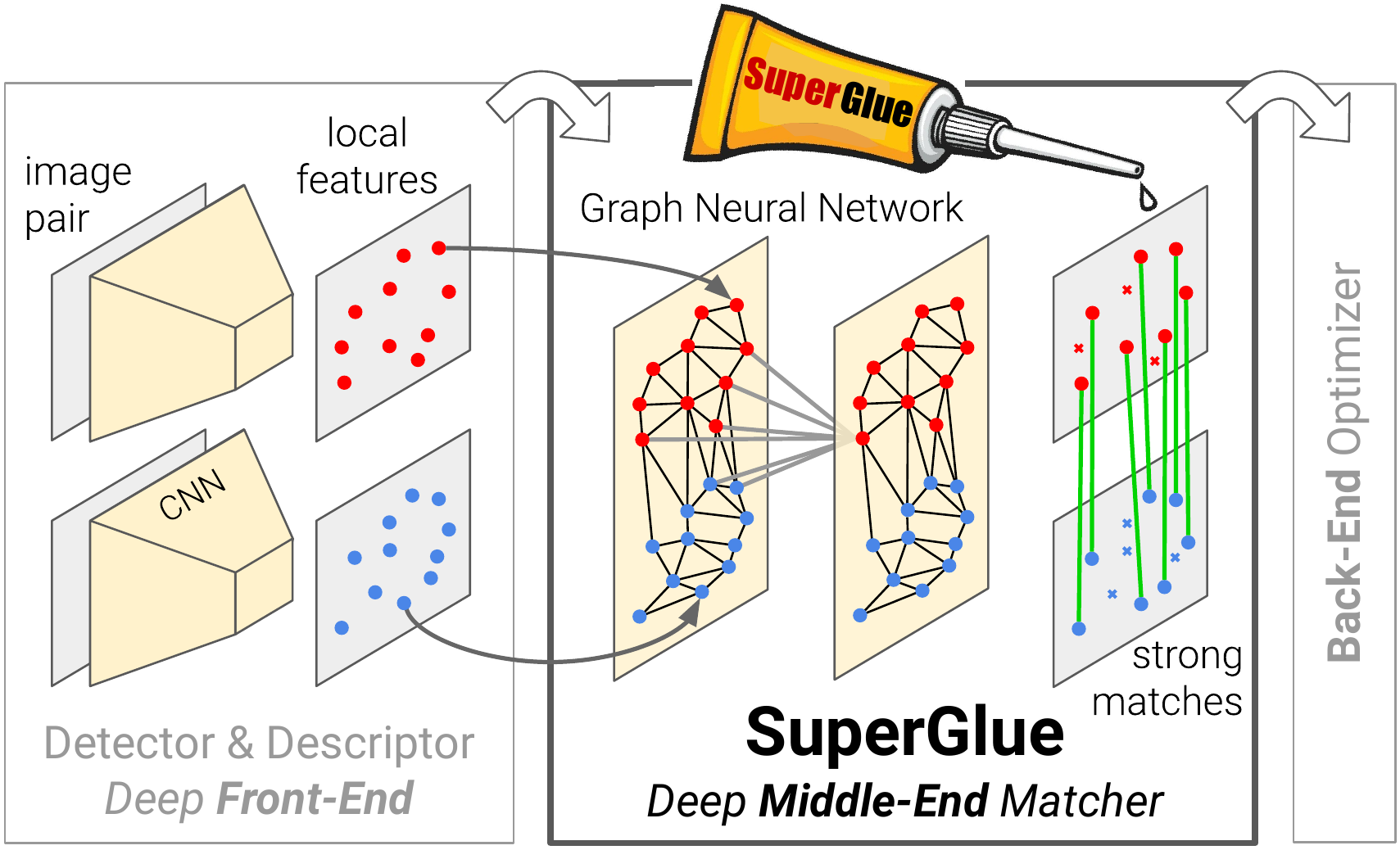}%
\caption{\textbf{Feature matching with SuperGlue.}
Our approach establishes pointwise correspondences from off-the-shelf local features: it acts as a middle-end between handcrafted or learned front-end and back-end. SuperGlue uses a graph neural network and attention to solve an assignment optimization problem, and handles partial point visibility and occlusion elegantly, producing a partial assignment.}
    \label{fig:teaser}%
\end{figure}

Correspondences between points in images are essential for estimating the 3D structure and camera poses in geometric computer vision tasks such as Simultaneous Localization and Mapping (SLAM) and Structure-from-Motion (SfM). Such correspondences are generally estimated by matching local features, a process known as data association. Large viewpoint and lighting changes, occlusion, blur, and lack of texture are factors that make 2D-to-2D data association particularly challenging.

In this paper, we present a new way of thinking about the feature matching problem. Instead of learning better task-agnostic local features followed by simple matching heuristics and tricks, we propose to learn the matching process from pre-existing local features using a novel neural architecture called SuperGlue. In the context of SLAM, which typically~\cite{cadena2016past} decomposes the problem into the visual feature extraction \emph{front-end} and the bundle adjustment or pose estimation \emph{back-end}, our network lies directly in the middle --~SuperGlue is a learnable \emph{middle-end} (see Figure~\ref{fig:teaser}).

In this work, \emph{learning feature matching} is viewed as finding the partial assignment between two sets of local features. We revisit the classical graph-based strategy of matching by solving a linear assignment problem, which, when relaxed to an optimal transport problem, can be solved differentiably. The cost function of this optimization is predicted by a Graph Neural Network (GNN). Inspired by the success of the Transformer~\cite{vaswani2017attention}, it uses self- (intra-image) and cross- (inter-image) attention to leverage both spatial relationships of the keypoints and their visual appearance. This formulation enforces the assignment structure of the predictions while enabling the cost to learn complex priors, elegantly handling occlusion and non-repeatable keypoints. Our method is trained end-to-end from image pairs -- we learn priors for pose estimation from a large annotated dataset, enabling SuperGlue to reason about the 3D scene and the assignment. Our work can be applied to a variety of multiple-view geometry problems that require high-quality feature correspondences~(see Figure~\ref{fig:examples}).\blfootnote{$^*$Work done at Magic Leap, Inc.\ for a Master's degree. The author thanks his academic supervisors: Cesar Cadena, Marcin Dymczyk, Juan Nieto.} 
\newpage
We show the superiority of SuperGlue compared to both handcrafted matchers and learned inlier classifiers. When combined with SuperPoint~\cite{superpoint}, a deep front-end, SuperGlue advances the state-of-the-art on the tasks of indoor and outdoor pose estimation and paves the way towards end-to-end deep SLAM.

\begin{figure}[t]
    \centering
   \includegraphics[width=1.0\linewidth]{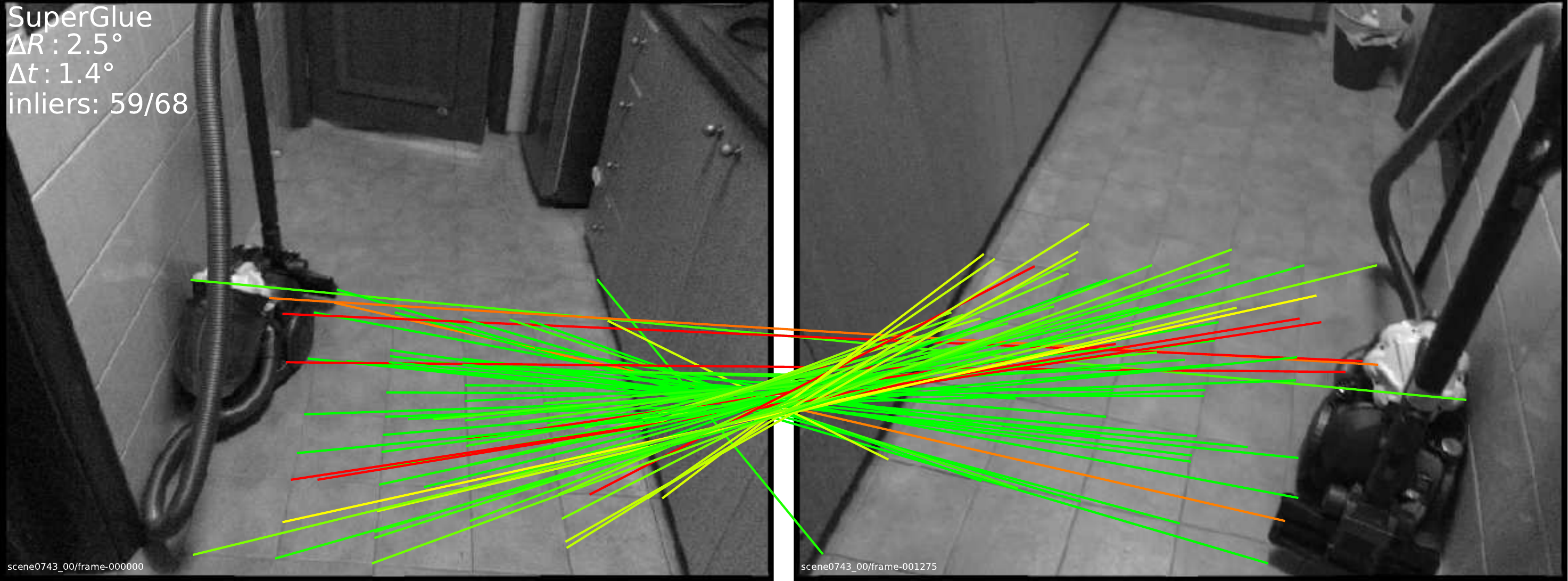}
   
   \vspace{1mm}
   \includegraphics[width=1.0\linewidth]{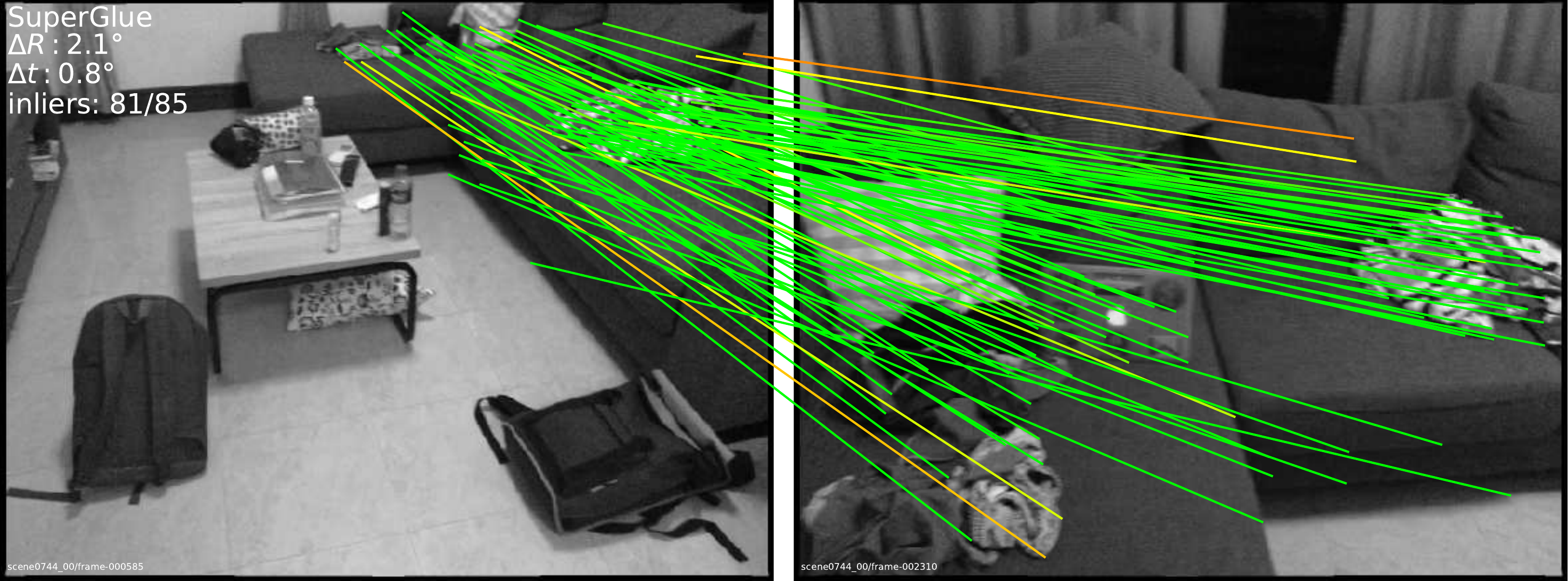}%
   \caption{{\bf SuperGlue correspondences.} For these two challenging indoor image pairs, matching with SuperGlue results in accurate poses while other learned or handcrafted methods fail (correspondences colored by epipolar error).}%
   \label{fig:examples}%
   \vspace{-3mm}%
\end{figure}

\section{Related work}
\PAR{Local feature matching} is generally performed by
\begin{enumerate*}[label=\roman*)]
\item detecting interest points,
\item computing visual descriptors,
\item matching these with a Nearest Neighbor (NN) search,
\item filtering incorrect matches, and finally
\item estimating a geometric transformation.
\end{enumerate*}
The classical pipeline developed in the 2000s is often based on SIFT~\cite{lowe2004distinctive}, filters matches with Lowe's ratio test~\cite{lowe2004distinctive}, the mutual check, and heuristics such as neighborhood consensus~\cite{tuytelaars2000wide, cech2010efficient, bian2017gms, sattler2009scramsac}, and finds a transformation with a robust solver like RANSAC~\cite{fischler1981random, raguram2008comparative}.

Recent works on deep learning for matching often focus on learning better sparse detectors and local descriptors~\cite{superpoint, dusmanu2019d2, ono2018lf, revaud2019r2d2, yi2016lift} from data using Convolutional Neural Networks (CNNs). To improve their discriminativeness, some works explicitly look at a wider context using regional features~\cite{luo2019contextdesc} or log-polar patches~\cite{ebel2019beyond}. Other approaches learn to filter matches by classifying them into inliers and outliers~\cite{moo2018learning, ranftl2018deep, brachmann2019neural, zhang2019learning}. These operate on sets of matches, still estimated by NN search, and thus ignore the assignment structure and discard visual information. Works that learn to perform matching have so far focused on dense matching~\cite{rocco2018neighbourhood} or 3D point clouds~\cite{wang2019deep}, and still exhibit the same limitations. In contrast, our learnable middle-end simultaneously performs context aggregation, matching, and filtering in a single end-to-end architecture.

\PAR{Graph matching} problems are usually formulated as quadratic assignment problems, which are NP-hard, requiring expensive, complex, and thus impractical solvers~\cite{loiola2007survey}. For local features, the computer vision literature of the 2000s~\cite{berg2005shape, leordeanu2005spectral, torresani2008feature} uses handcrafted costs with many heuristics, making it complex and brittle. Caetano \etal~\cite{caetano2009learning} learn the cost of the optimization for a simpler linear assignment, but only use a shallow model, while our SuperGlue learns a flexible cost using a deep neural network. Related to graph matching is the problem of \emph{optimal transport}~\cite{villani2008optimal} -- it is a generalized linear assignment with an efficient yet simple approximate solution, the Sinkhorn algorithm~\cite{sinkhorn1967concerning, cuturi2013sinkhorn, peyre2019computational}.

\PAR{Deep learning for sets} such as point clouds aims at designing permutation equi- or invariant functions by aggregating information across elements. Some works treat all elements equally, through global pooling~\cite{zaheer2017deep, qi2017pointnet, deng2018ppfnet} or instance normalization~\cite{ulyanov2016instance, moo2018learning, luo2019contextdesc}, while others focus on a local neighborhood in coordinate or feature space~\cite{qi2017pointnet++, dgcnn}. Attention~\cite{vaswani2017attention, wang2018non, velickovic2018graph, lee2019set} can perform both global and data-dependent local aggregation by focusing on specific elements and attributes, and is thus more flexible.
By observing that self-attention can be seen as an instance of a Message Passing Graph Neural Network~\cite{gilmer2017neural, battaglia2018relational} on a complete graph, we apply attention to graphs with multiple types of edges, similar to~\cite{li2019graph, zhang2019dual}, and enable SuperGlue to learn complex reasoning about the two sets of local features.
 
\begin{figure*}[!t]
\vspace{-.2in}
    \centering
    \includegraphics[width=0.98\linewidth]{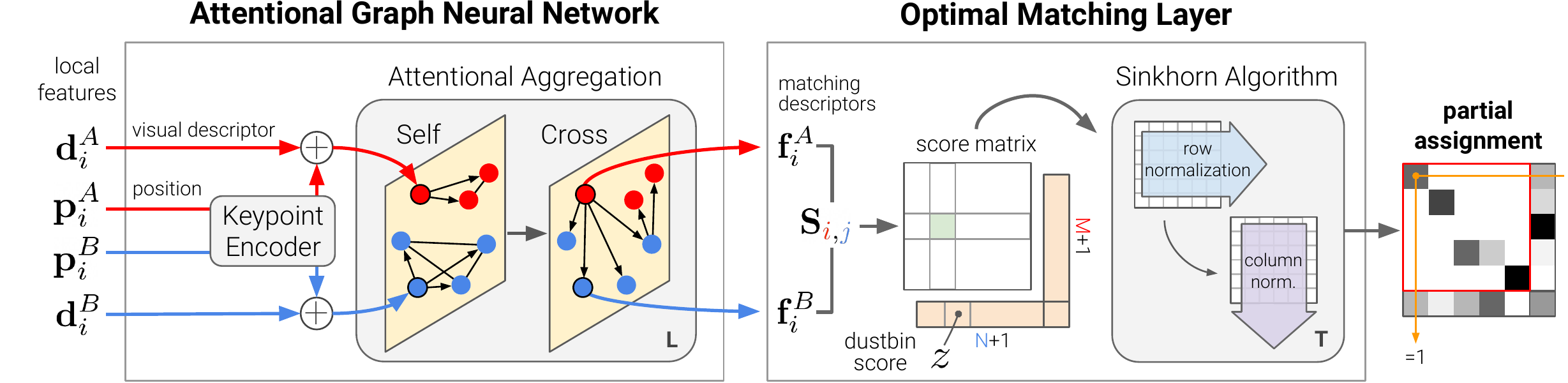}%
    \begin{subfigure}{0.5\linewidth}
        \caption{\label{fig:detail-a}}%
    \end{subfigure}%
    \begin{subfigure}{0.5\linewidth}
        \caption{\label{fig:detail-b}}%
    \end{subfigure}%
    \vspace{-3mm}%
    \caption{\textbf{The SuperGlue architecture.} SuperGlue is made up of two major components: the \emph{attentional graph neural network} (Section~\ref{sec:gnn}), and the \emph{optimal matching layer} (Section~\ref{sec:optimal}). The first component uses a \emph{keypoint encoder} to map keypoint positions $\mathbf{p}$ and their visual descriptors ${\bf d}$ into a single vector, and then uses alternating self- and cross-attention layers (repeated $L$ times) to create more powerful representations ${\bf f}$. The optimal matching layer creates an $M$ by $N$ score matrix, augments it with dustbins, then finds the optimal partial assignment using the Sinkhorn algorithm (for $T$ iterations).}
\label{fig:detail}%
\end{figure*}

\section{The SuperGlue Architecture}

\PAR{Motivation:}
In the image matching problem, some regularities of the world could be leveraged: the 3D world is largely smooth and sometimes planar, all correspondences for a given image pair derive from a single epipolar transform if the scene is static, and some poses are more likely than others. In addition, 2D keypoints are usually projections of salient 3D points, like corners or blobs, thus correspondences across images must adhere to certain physical constraints: \begin{enumerate*}[label=\textcolor{red}{\roman*)}]
\item\label{item:uniqueness} a keypoint can have at most a single correspondence in the other image; and
\item\label{item:occlusion} some keypoints will be unmatched due to occlusion and failure of the detector.
\end{enumerate*}
An effective model for feature matching should aim at finding all correspondences between reprojections of the same 3D points and identifying keypoints that have no matches. We formulate SuperGlue (see Figure~\ref{fig:detail}) as solving an optimization problem, whose cost is predicted by a deep neural network. This alleviates the need for domain expertise and heuristics -- we learn relevant priors directly from the data.

\PAR{Formulation:} Consider two images $A$ and $B$, each with a set of keypoint \emph{positions} ${\bf p}$ and associated \emph{visual descriptors} ${\bf d}$ -- we refer to them jointly $({\bf p},{\bf d})$ as the \emph{local features}.
Positions consist of $x$ and $y$ image coordinates as well as a detection confidence $c$, $\mathbf p_i := (x, y, c)_i$. Visual descriptors $\mathbf d_i \in \mathbb R^D$ can be those extracted by a CNN like SuperPoint or traditional descriptors like SIFT.
Images $A$ and $B$ have $M$ and $N$ local features, indexed by $\mathcal{A} := \{1, ..., M\}$ and $\mathcal{B} := \{1, ..., N\}$, respectively.

\PAR{Partial Assignment:}
Constraints \ref{item:uniqueness} and \ref{item:occlusion} mean that correspondences derive from a partial assignment between the two sets of keypoints. For the integration into downstream tasks and better interpretability, each possible correspondence should have a confidence value.
We consequently define a partial soft assignment matrix $\mathbf P \in [0,1]^{M \times N}$ as:
    \begin{equation}
        \mathbf P \mathbf 1_N \leq \mathbf 1_M
        \quad\text{and}\quad
        \mathbf P^\top \mathbf 1_M \leq \mathbf 1_N.
        \label{eq:constraints}
    \end{equation}
Our goal is to design a neural network that predicts the assignment $\mathbf P$ from two sets of local features.

\subsection{Attentional Graph Neural Network}
\label{sec:gnn}
Besides the position of a keypoint and its visual appearance, integrating other contextual cues can intuitively increase its distinctiveness. We can for example consider its spatial and visual relationship with other co-visible keypoints, such as ones that are salient~\cite{luo2019contextdesc}, self-similar~\cite{shechtman2007matching}, statistically co-occurring~\cite{zhang2011image}, or adjacent~\cite{trzcinski2018scone}. On the other hand, knowledge of keypoints in the second image can help to resolve ambiguities by comparing candidate matches or estimating the relative photometric or geometric transformation from global and unambiguous cues.

When asked to match a given ambiguous keypoint, humans look back-and-forth at both images: they sift through tentative matching keypoints, examine each, and look for contextual cues that help disambiguate the true match from other self-similarities~\cite{chun2000contextual}. This hints at an iterative process that can focus its attention on specific locations.

We consequently design the first major block of SuperGlue as an Attentional Graph Neural Network (see Figure~\ref{fig:detail}). Given initial local features, it computes \emph{matching descriptors} $\mathbf f_i \in \mathbb R^{D}$ by letting the features communicate with each other. 
As we will show, long-range feature aggregation within and across images is vital for robust matching.

\PAR{Keypoint Encoder:}
The initial representation ${}^{(0)}\mathbf x_i$ for each keypoint $i$ combines its visual appearance and location. We embed the keypoint position into a high-dimensional vector with a Multilayer Perceptron (MLP) as:
    \begin{equation}
    \label{eq:keypoint-encoder}
        {}^{(0)}\mathbf x_i = \mathbf d_i + \text{MLP}_{\text{enc}}\left(\mathbf{p}_i\right).
    \end{equation}
This encoder enables the graph network to later reason about both appearance and position jointly, especially when combined with attention, and is an instance of the ``positional encoder'' popular in language processing~\cite{gehring2017convolutional, vaswani2017attention}.

\PAR{Multiplex Graph Neural Network:}
We consider a single complete graph whose nodes are the keypoints of both images. The graph has two types of undirected edges -- it is a \emph{multiplex graph}~\cite{mucha2010community, nicosia2013growing}. Intra-image edges, or \emph{self} edges, $\mathcal E_{\text{self}}$, connect keypoints $i$ to all other keypoints within the same image. Inter-image edges, or \emph{cross} edges, $\mathcal E_{\text{cross}}$, connect keypoints $i$ to all keypoints in the other image. We use the message passing formulation~\cite{gilmer2017neural, battaglia2018relational} to propagate information along both types of edges. The resulting multiplex Graph Neural Network starts with a high-dimensional state for each node and computes at each layer an updated representation by simultaneously aggregating messages across all given edges for all nodes. 

Let ${}^{(\ell)}\mathbf{x}_i^A$ be the intermediate representation for element $i$ in image $A$ at layer $\ell$. The message $\mathbf{m}_{\mathcal E \rightarrow i}$ is the result of the aggregation from all keypoints $\{j: (i, j)\in\mathcal{E}\}$, where $\mathcal E \in \{\mathcal E_{\text{self}}, \mathcal E_{\text{cross}}\}$. The residual message passing update for all $i$ in $A$ is:
    \begin{equation}
    \label{eqn:message-passing}
        {}^{(\ell+1)}\mathbf{x}_i^A =
        {}^{(\ell)}\mathbf{x}_i^A
        + \text{MLP}\left( \left[{}^{(\ell)}\mathbf{x}_i^{A}\,||\,\mathbf{m}_{\mathcal{E}\rightarrow i}\right]\right),
    \end{equation}
where $[\cdot\,||\,\cdot]$ denotes concatenation.  A similar update can be simultaneously performed for all keypoints in image~$B$. A fixed number of layers $L$ with different parameters are chained and alternatively aggregate along the self and cross edges. As such, starting from $\ell = 1$, $\mathcal E = \mathcal E_{\text{self}}$ if $\ell$ is odd and $\mathcal E = \mathcal E_{\text{cross}}$ if $\ell$ is even.

\begin{figure}[!t]
\centering
\begin{minipage}{0.9\linewidth}
    \begin{minipage}{0.05\linewidth}
    \hfill
    \end{minipage}%
    \begin{minipage}{0.475\linewidth}
    \centering
    \footnotesize{image $A$}
    \end{minipage}%
    \begin{minipage}{0.475\linewidth}
    \centering
    \footnotesize{image $B$}
    \end{minipage}%

    \begin{minipage}{0.05\linewidth}
    \rotatebox[origin=c]{90}{\footnotesize{Self-Attention}}
    \end{minipage}%
    \begin{minipage}{0.95\linewidth}
    \includegraphics[width=1.0\linewidth]{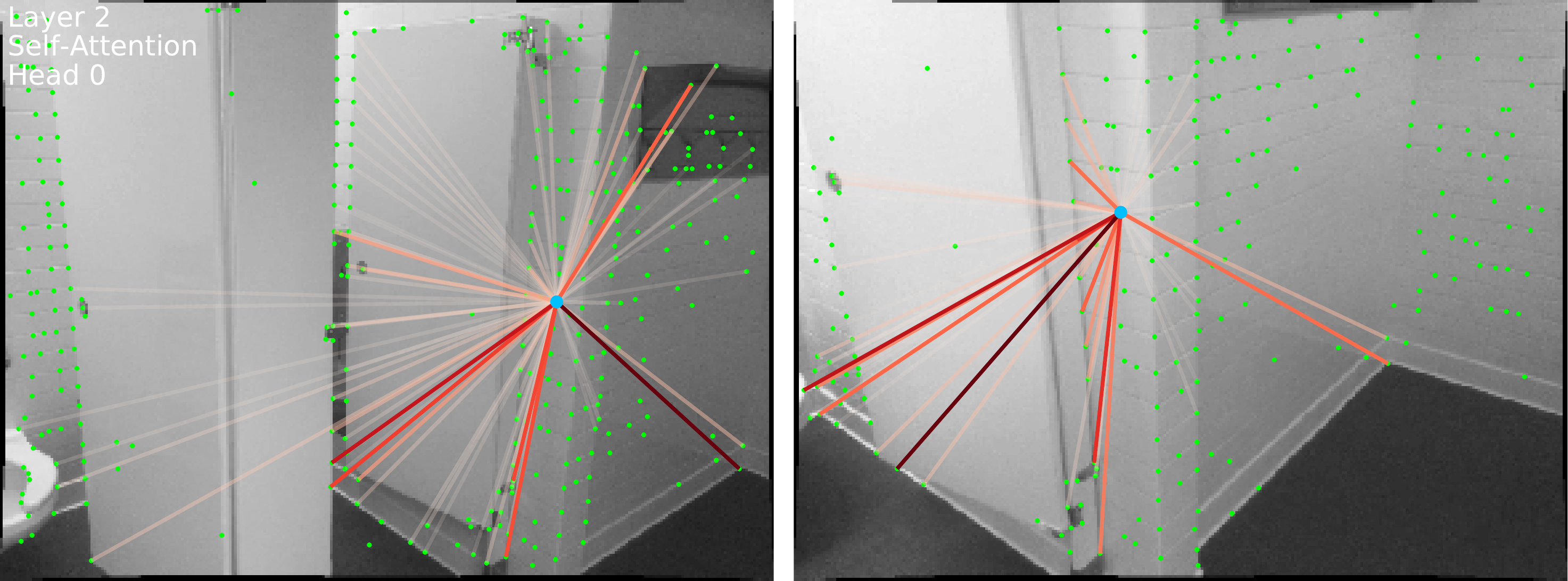}%
    \end{minipage}
    
    \vspace{1mm}
    \begin{minipage}{0.05\linewidth}
    \rotatebox[origin=c]{90}{\footnotesize{Cross-Attention}}
    \end{minipage}%
    \begin{minipage}{0.95\linewidth}
    \includegraphics[width=1.0\linewidth]{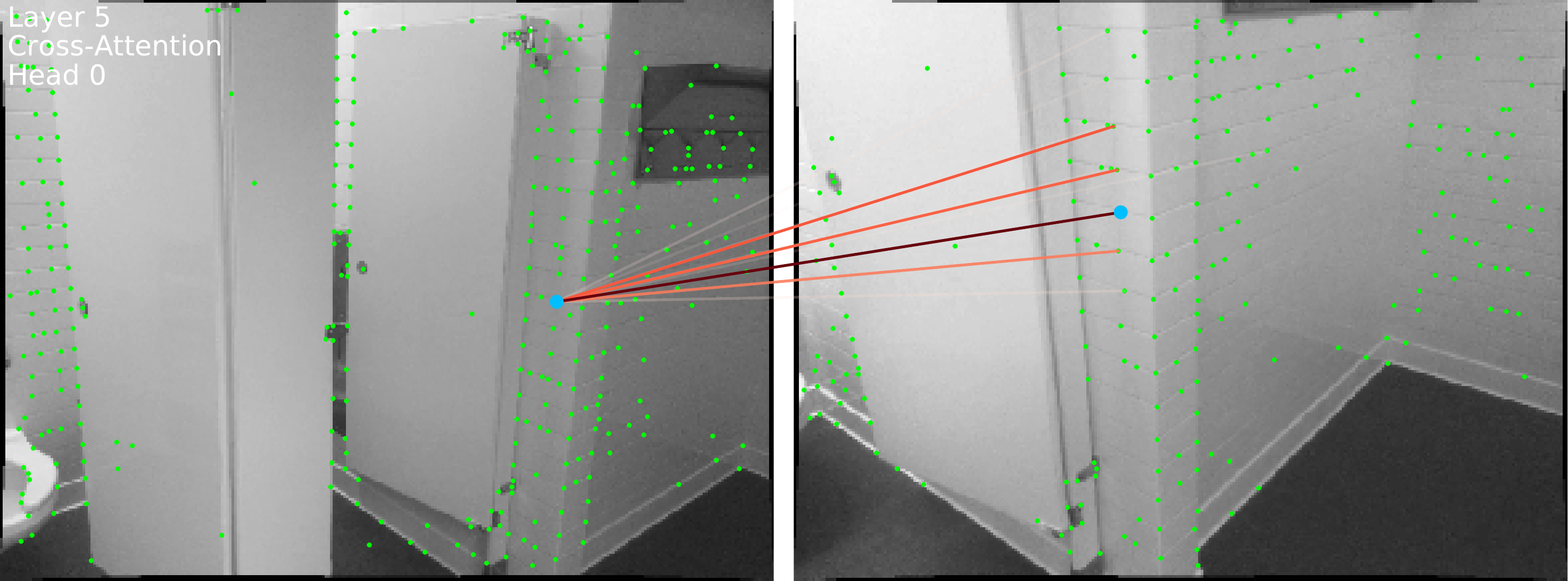}%
    \end{minipage}
\end{minipage}%
\hspace{1mm}
\begin{minipage}{0.05\linewidth}
\centering
    \centering
    \includegraphics[width=0.58\linewidth]{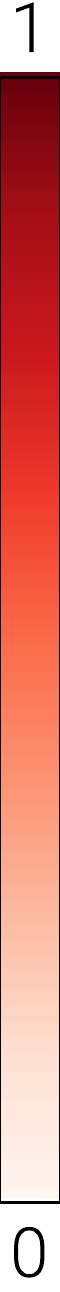}%
\end{minipage}
\vspace{-1mm}
\caption{{\bf Visualizing self- and cross-attention.} Attentional aggregation builds a dynamic graph between keypoints. Weights $\alpha_{ij}$ are shown as rays. Self-attention (top) can attend anywhere in the same image, e.g.\ distinctive locations, and is thus not restricted to nearby locations. Cross-attention (bottom) attends to locations in the other image, such as potential matches that have a similar appearance.
}%
\label{fig:attention}%
\vspace{-1mm}%
\end{figure}

\PAR{Attentional Aggregation:}
An attention mechanism performs the aggregation and computes the message $\mathbf{m}_{\mathcal E \rightarrow i}$. Self edges are based on self-attention~\cite{vaswani2017attention} and cross edges are based on \emph{cross-attention}. Akin to database retrieval, a representation of $i$, the query $\mathbf q_i$, retrieves the values $\mathbf v_j$ of some elements based on their attributes, the keys $\mathbf{k}_j$. The message is computed as a weighted average of the values:
    \begin{equation}
        \mathbf{m}_{\mathcal E\rightarrow i} =  \sum_{j: (i, j)\in\mathcal{E}} \alpha_{ij} \mathbf{v}_j,
    \end{equation}
    where the attention weight $\alpha_{ij}$ is the Softmax over the key-query similarities: $\alpha_{ij} = \text{Softmax}_j\left(\mathbf q_i^\top \mathbf k_j\right)$.

The key, query, and value are computed as linear projections of deep features of the graph neural network. Considering that query keypoint $i$ is in the image $Q$ and all source keypoints are in image $S$, $(Q, S) \in \{A, B\}^2$, we can write:
    \begin{equation}
    \label{eq:gnn-projections}
    \begin{split}
        \mathbf{q}_i &= \mathbf{W}_1\ {}^{(\ell)}\mathbf{x}_i^Q + \mathbf{b}_1\\
        \begin{bmatrix}\mathbf{k}_j\\\mathbf{v}_j\end{bmatrix} &= 
        \begin{bmatrix}\mathbf{W}_2\\\mathbf{W}_3\end{bmatrix} {}^{(\ell)}\mathbf{x}_j^S
        + \begin{bmatrix}\mathbf{b}_2\\\mathbf{b}_3\end{bmatrix}.
    \end{split}
    \end{equation}
Each layer $\ell$ has its own projection parameters, learned and shared for all keypoints of both images. In practice, we improve the expressivity with multi-head attention~\cite{vaswani2017attention}.

Our formulation provides maximum flexibility as the network can learn to focus on a subset of keypoints based on specific attributes (see Figure~\ref{fig:attention}). SuperGlue can retrieve or attend based on both appearance and keypoint location as they are encoded in the representation $\mathbf{x}_i$. This includes attending to a nearby keypoint and retrieving the relative positions of similar or salient keypoints. This enables representations of the geometric transformation and the assignment. The final matching descriptors are linear projections: 
\begin{equation}
\label{eq:final-projection}
    \mathbf{f}^A_i = \mathbf{W}\cdot{}^{(L)}\mathbf{x}^A_i+\mathbf b, \quad \forall i \in \mathcal A,
\end{equation}
and similarly for keypoints in $B$.
 
\subsection{Optimal matching layer}
\label{sec:optimal}
The second major block of SuperGlue (see Figure~\ref{fig:detail}) is the optimal matching layer, which produces a partial assignment matrix. As in the standard graph matching formulation, the assignment $\mathbf P$ can be obtained by computing a score matrix $\mathbf S \in \mathbb{R}^{M\times N}$ for all possible matches and maximizing the total score $\sum_{i,j}\mathbf S_{i,j} \mathbf P_{i, j}$ under the constraints in Equation~\ref{eq:constraints}. This is equivalent to solving a linear assignment problem.

\PAR{Score Prediction:}
Building a separate representation for all $M\times N$ potential matches would be prohibitive. We instead express the pairwise score as the similarity of matching descriptors:
\begin{equation}
    \mathbf{S}_{i,j} = <\mathbf f^A_i, \mathbf f^B_j>,\ \forall (i, j) \in \mathcal{A} \times \mathcal{B},
\end{equation}
where $<\cdot,\cdot>$ is the inner product.
As opposed to learned visual descriptors, the matching descriptors are not normalized, and their magnitude can change per feature and during training to reflect the prediction confidence.

\PAR{Occlusion and Visibility:}
To let the network suppress some keypoints, we augment each set with a dustbin so that unmatched keypoints are explicitly assigned to it. This technique is common in graph matching, and dustbins have also been used by SuperPoint~\cite{superpoint} to account for image cells that might not have a detection. We augment the scores $\mathbf S$ to $\bar{\mathbf S}$ by appending a new row and column, the point-to-bin and bin-to-bin scores, filled with a single learnable parameter:
\begin{equation}
    \bar{\mathbf S}_{i,N+1} = \bar{\mathbf S}_{M+1,j}
    = \bar{\mathbf S}_{M+1, N+1} = z \in \mathbb{R}.
\end{equation}
While keypoints in $A$ will be assigned to a single keypoint in $B$ or the dustbin, each dustbin has as many matches as there are keypoints in the other set: $N$, $M$ for dustbins in $A$, $B$ respectively. We denote as $\mathbf{a} = \begin{bmatrix}\mathbf{1}_M^\top & N\end{bmatrix}^\top$ and $\mathbf{b} = \begin{bmatrix}\mathbf{1}_N^\top & M\end{bmatrix}^\top$ the number of expected matches for each keypoint and dustbin in $A$ and $B$. The augmented assignment $\bar{\mathbf P}$ now has the constraints:
    \begin{equation}
        \bar{\mathbf P}\mathbf{1}_{N+1} = \mathbf a
        \quad\text{and}\quad
        \bar{\mathbf P}^\top\mathbf{1}_{M+1} = \mathbf b.
        \label{eq:constraints_aug}
    \end{equation}

\PAR{Sinkhorn Algorithm:} The solution of the above optimization problem corresponds to the optimal transport~\cite{peyre2019computational} between discrete distributions $\mathbf{a}$ and $\mathbf{b}$ with scores $\bar{\mathbf S}$.
Its entropy-regularized formulation naturally results in the desired soft assignment, and can be efficiently solved on GPU with the Sinkhorn algorithm~\cite{sinkhorn1967concerning, cuturi2013sinkhorn}. It is a differentiable version of the Hungarian algorithm~\cite{munkres1957algorithms}, classically used for bipartite matching, that consists in iteratively normalizing $\exp(\bar{\mathbf{S}})$ along rows and columns, similar to row and column Softmax.
After $T$ iterations, we drop the dustbins and recover $\mathbf{P} = \bar{\mathbf{P}}_{1:M,1:N}$.

\subsection{Loss}
By design, both the graph neural network and the optimal matching layer are differentiable -- this enables backpropagation from matches to visual descriptors. SuperGlue is trained in a supervised manner from ground truth matches $\mathcal M = \{(i, j)\} \subset \mathcal{A} \times \mathcal{B}$. These are estimated from ground truth relative transformations -- using poses and depth maps or homographies. This also lets us label some keypoints $\mathcal I \subseteq \mathcal{A}$ and $\mathcal J \subseteq \mathcal{B}$ as unmatched if they do not have any reprojection in their vicinity. Given these labels, we minimize the negative log-likelihood of the assignment $\bar{\mathbf P}$:
\begin{equation}
\begin{split}
    \text{Loss} =
    &-\sum_{(i, j) \in \mathcal M} \log \bar{\mathbf{P}}_{i,j}\\
    &- \sum_{i \in \mathcal I} \log \bar{\mathbf{P}}_{i,N+1}
    - \sum_{j \in \mathcal J} \log \bar{\mathbf{P}}_{M+1,j}.
\end{split}
\end{equation}
This supervision aims at simultaneously maximizing the precision and the recall of the matching.

\subsection{Comparisons to related work}
The SuperGlue architecture is equivariant to permutation of the keypoints within an image. Unlike other handcrafted or learned approaches, it is also equivariant to permutation \emph{of the images}, which better reflects the symmetry of the problem and provides a beneficial inductive bias.
Additionally, the optimal transport formulation enforces reciprocity of the matches, like the mutual check, but in a soft manner, similar to~\cite{rocco2018neighbourhood}, thus embedding it into the training process.

\PAR{SuperGlue vs.\ Instance Normalization~\cite{ulyanov2016instance}:}
Attention, as used by SuperGlue, is a more flexible and powerful context aggregation mechanism than instance normalization, which treats all keypoints equally, as used by previous work on feature matching~\cite{moo2018learning, zhang2019learning, luo2019contextdesc, ranftl2018deep, brachmann2019neural}. 

\PAR{SuperGlue vs.\ ContextDesc~\cite{luo2019contextdesc}:}
SuperGlue can jointly reason about appearance and position while ContextDesc processes them separately. Moreover, ContextDesc is a front-end that additionally requires a larger regional extractor, and a loss for keypoints scoring. SuperGlue only needs local features, learned or handcrafted, and can thus be a simple drop-in replacement for existing matchers.

\PAR{SuperGlue vs.\ Transformer~\cite{vaswani2017attention}:}
SuperGlue borrows the self-attention from the Transformer, but embeds it into a graph neural network, and additionally introduces the cross-attention, which is symmetric. This simplifies the architecture and results in better feature reuse across layers.

\section{Implementation details}
\label{sec:implementation}
SuperGlue can be combined with any local feature detector and descriptor but works particularly well with SuperPoint~\cite{superpoint}, which produces repeatable and sparse keypoints -- enabling very efficient matching. Visual descriptors are bilinearly sampled from the semi-dense feature map. For a fair comparison to other matchers, unless explicitly mentioned, we do not train the visual descriptor network when training SuperGlue. At test time, one can use a confidence threshold (we choose 0.2) to retain some matches from the soft assignment, or use all of them and their confidence in a subsequent step, such as weighted pose estimation.

\PAR{Architecture details:}
All intermediate representations (key, query value, descriptors) have the same dimension $D=256$ as the SuperPoint descriptors.
We use $L = 9$ layers of alternating multi-head self- and cross-attention with 4 heads each, and perform $T=100$ Sinkhorn iterations.
The model is implemented in PyTorch~\cite{paszke2017automatic}, contains 12M parameters, and runs in real-time on an NVIDIA GTX 1080 GPU: a~forward pass takes on average \b{69~ms~(15~FPS)} for an indoor image pair (see \supp~\ref{sec:supp-timings}).

\PAR{Training details:}
To allow for data augmentation, SuperPoint detect and describe steps are performed on-the-fly as batches during training. A number of random keypoints are further added for efficient batching and increased robustness. More details are provided in \supp~\ref{sec:details-supp}.

\section{Experiments}

\begin{table}[b]
\centering
\vspace{-3mm}
\footnotesize{
\setlength\tabcolsep{3.0pt}
\begin{tabular}{ll>{\centering}p{15mm}>{\centering}p{10mm}cc}
    \toprule
    \multirow{2}{1cm}[-.4em]{Local features} & \multirow{2}{*}[-.4em]{Matcher} &
    \multicolumn{2}{c}{Homography estimation AUC} & 
    \multirow{2}{*}[-.4em]{P} & \multirow{2}{*}[-.4em]{R} \\
    \cmidrule(lr){3-4}
    && RANSAC & DLT && \\
    \midrule
    \multirow{5}{*}{SuperPoint}
    & NN & 39.47 & 0.00 & 21.7 & 65.4\\
    & NN + mutual & 42.45 & 0.24 & 43.8 & 56.5\\
    & NN + PointCN & 43.02 & 45.40 & 76.2 & 64.2\\
    & NN + OANet & 44.55 & 52.29 & 82.8 & 64.7\\
    & \b{SuperGlue} & \b{53.67} & \b{65.85} & \b{90.7} & \b{98.3}\\
    \bottomrule
\end{tabular}
}
\vspace{-.1in}
\caption{\textbf{Homography estimation.}
SuperGlue recovers almost all possible matches while suppressing most outliers. Because SuperGlue correspondences are high-quality, the Direct Linear Transform (DLT), a least-squares based solution with no robustness mechanism, outperforms RANSAC.}
\label{tab:planar}
\vspace{-1mm}
\end{table}

\subsection{Homography estimation}
\label{sec:homography}
We perform a large-scale homography estimation experiment using real images and synthetic homographies with both robust (RANSAC) and non-robust (DLT) estimators.

\PAR{Dataset:} We generate image pairs by sampling random homographies and applying random photometric distortions to real images, following a recipe similar to~\cite{detone2016deep, superpoint, revaud2019r2d2, ranftl2018deep}. The underlying images come from the set of 1M distractor images in the Oxford and Paris dataset~\cite{radenovic2018revisiting}, split into training, validation, and test sets.

\PAR{Baselines:}  We compare SuperGlue against several matchers applied to SuperPoint local features -- the Nearest Neighbor (NN) matcher and various outlier rejectors: the mutual NN constraint, PointCN~\cite{moo2018learning}, and Order-Aware Network (OANet)~\cite{zhang2019learning}. All learned methods, including SuperGlue, are trained on ground truth correspondences, found by projecting keypoints from one image to the other. We generate homographies and photometric distortions on-the-fly -- an image pair is never seen twice during training.

\PAR{Metrics:} Match precision (P) and recall (R) are computed from the ground truth correspondences. Homography estimation is performed with both RANSAC and the Direct Linear Transformation~\cite{hartley2003multiple} (DLT), which has a direct least-squares solution. We compute the mean reprojection error of the four corners of the image and report the area under the cumulative error curve (AUC) up to a value of 10 pixels.

\PAR{Results:} SuperGlue is sufficiently expressive to master homographies, achieving 98\% recall and high precision (see Table~\ref{tab:planar}). The estimated correspondences are so good that a robust estimator is not required -- SuperGlue works even better with DLT than RANSAC. Outlier rejection methods like PointCN and OANet cannot predict more correct matches than the NN matcher itself, overly relying on the initial descriptors (see Figure~\ref{fig:qual-all} and \supp~\ref{sec:results-supp}).

\subsection{Indoor pose estimation}
\label{sec:indoor}
Indoor image matching is very challenging due to the lack of texture, the abundance of self-similarities, the complex 3D geometry of scenes, and large viewpoint changes. As we show in the following, SuperGlue can effectively learn priors to overcome these challenges.

\PAR{Dataset:} We use ScanNet~\cite{dai2017scannet}, a large-scale indoor dataset composed of monocular sequences with ground truth poses and depth images, and well-defined training, validation, and test splits corresponding to different scenes.
Previous works select training and evaluation pairs based on time difference~\cite{ono2018lf, detone2018self} or SfM covisibility~\cite{moo2018learning, zhang2019learning, brachmann2019neural}, usually computed using SIFT. We argue that this limits the difficulty of the pairs, and instead select these based on an overlap score computed for all possible image pairs in a given sequence using only ground truth poses and depth. This results in significantly wider-baseline pairs, which corresponds to the current frontier for real-world indoor image matching.
Discarding pairs with too small or too large overlap, we select 230M training and 1500 test pairs.

\PAR{Metrics:} As in previous work~\cite{moo2018learning, zhang2019learning, brachmann2019neural}, we report the AUC of the pose error at the thresholds (5\degree$,\, $10\degree$,\, $20\degree), where the pose error is the maximum of the angular errors in rotation and translation. Relative poses are obtained from essential matrix estimation with RANSAC. We also report the match precision and the matching score~\cite{superpoint, yi2016lift}, where a match is deemed correct based on its epipolar distance.

\begin{table}[tb]
\centering
\scriptsize{
\setlength\tabcolsep{4.0pt}
\begin{tabular}{llccccc}
    \toprule
    \multirow{2}{1cm}[-.4em]{Local features} & \multirow{2}{*}[-.4em]{Matcher}
    & \multicolumn{3}{c}{Pose estimation AUC} & \multirow{2}{*}[-.4em]{P} & \multirow{2}{*}[-.4em]{MS} \\
    \cmidrule(lr){3-5}
    && @5\degree & @10\degree & @20\degree & & \\
    \midrule
    ORB & NN + GMS & \05.21 & 13.65 & 25.36 & 72.0 & \05.7 \\
    D2-Net & NN + mutual & \05.25 & 14.53 & 27.96 & 46.7 & 12.0 \\
    ContextDesc & NN + ratio test & \06.64 & 15.01 & 25.75 & 51.2 & \09.2\\
    \midrule
    \multirow{4}{*}{SIFT}
    & NN + ratio test & \05.83 & 13.06 & 22.47 & 40.3 & \01.0 \\
    & NN + NG-RANSAC & \06.19 & 13.80 & 23.73 & 61.9 & \00.7\\
    & NN + OANet & \06.00 & 14.33 & 25.90 & 38.6 & \04.2 \\
    & \b{SuperGlue} & \b{\06.71} & \b{15.70} & \b{28.67} & \b{74.2} & \b{\09.8}\\
    \midrule
    \multirow{6}{*}{SuperPoint}
    & NN + mutual & \09.43 & 21.53 & 36.40 & 50.4 & 18.8\\
    & NN + distance + mutual & \09.82 & 22.42 & 36.83 & 63.9 & 14.6\\
    & NN + GMS & \08.39 & 18.96 & 31.56 & 50.3 & 19.0\\
    & NN + PointCN & 11.40 & 25.47 & 41.41 & 71.8 & 25.5 \\
    & NN + OANet & 11.76 & 26.90 & 43.85 & 74.0 & 25.7 \\
    & \b{SuperGlue} & \b{16.16} & \b{33.81} & \b{51.84} & \b{84.4} & \b{31.5} \\
    \bottomrule
\end{tabular}
}
\caption{\textbf{Wide-baseline indoor pose estimation.} We report the AUC of the pose error, the matching score (MS) and precision (P), all in percents \%. SuperGlue outperforms all handcrafted and learned matchers when applied to both SIFT and SuperPoint.}
\label{tab:scannet}
\end{table}

\begin{figure}[t]
    \centering
    \includegraphics[width=1.0\linewidth, trim=0cm 0cm 0cm 0, clip]{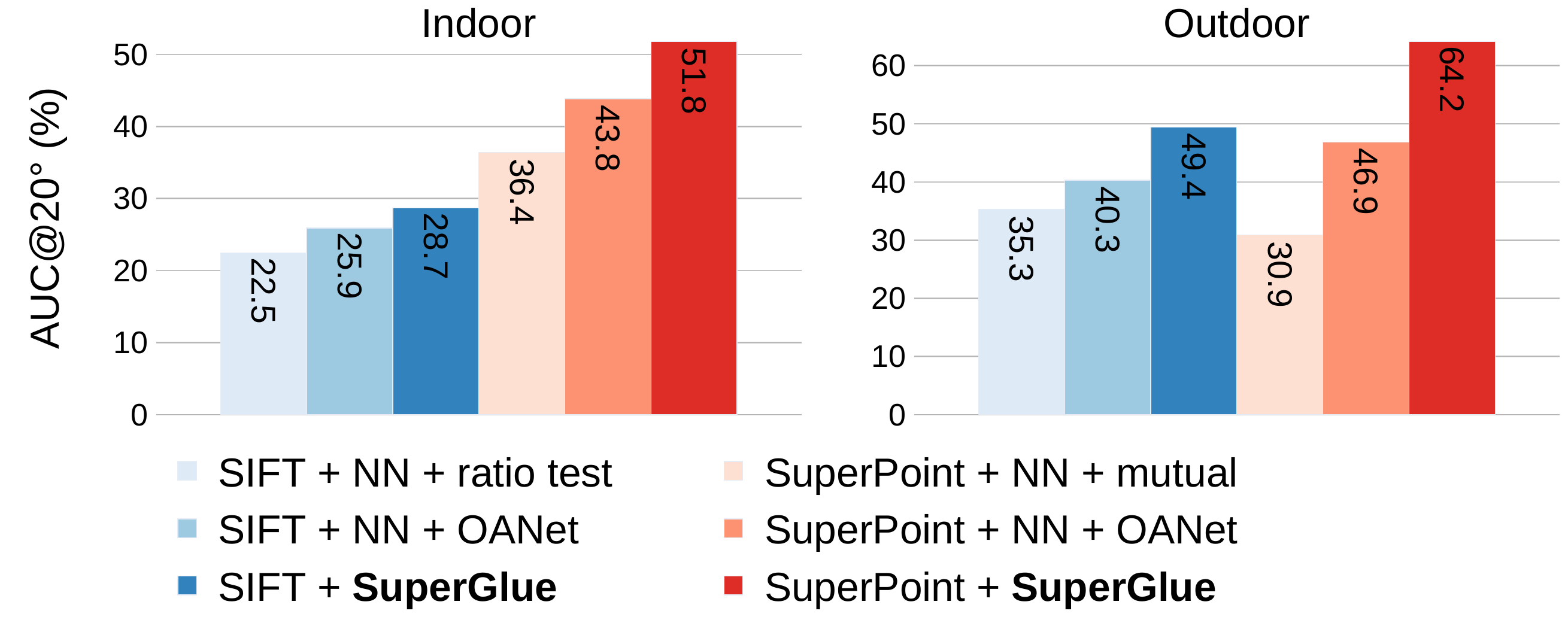}%
    \vspace{-.1cm}%
    \caption{\textbf{Indoor and outdoor pose estimation.}
    SuperGlue works with SIFT or SuperPoint local features and consistently improves by a large margin the pose accuracy over OANet, a state-of-the-art outlier rejection neural network.
    }
    \label{fig:scannet-bars}
\end{figure}

\PAR{Baselines:} We evaluate SuperGlue and various baseline matchers using both root-normalized SIFT~\cite{lowe2004distinctive, arandjelovic2012three} and SuperPoint~\cite{superpoint} features.
SuperGlue is trained with correspondences and unmatched keypoints derived from ground truth poses and depth.
All baselines are based on the Nearest Neighbor (NN) matcher and potentially an outlier rejection method. In the ``Handcrafted'' category, we consider the mutual check, the ratio test~\cite{lowe2004distinctive}, thresholding by descriptor distance, and the more complex GMS~\cite{bian2017gms}. Methods in the ``Learned'' category are PointCN~\cite{moo2018learning}, and its follow-ups OANet~\cite{zhang2019learning} and NG-RANSAC~\cite{brachmann2019neural}. We retrain PointCN and OANet on ScanNet for both SuperPoint and SIFT with the classification loss using the above-defined correctness criterion and their respective regression losses. For NG-RANSAC, we use the original trained model.
We do not include any graph matching methods as they are orders of magnitude too slow for the number of keypoints that we consider (\textgreater 500). Other local features are evaluated as reference: ORB~\cite{rublee2011orb} with GMS, D2-Net~\cite{dusmanu2019d2}, and ContextDesc~\cite{luo2019contextdesc} using the publicly available trained models.

\PAR{Results:} SuperGlue enables significantly higher pose accuracy compared to both handcrafted and learned matchers (see Table~\ref{tab:scannet} and Figure~\ref{fig:scannet-bars}), and works well with both SIFT and SuperPoint. It has a significantly higher precision than other learned matchers, demonstrating its higher representation power. It also produces a larger number of correct matches -- up to 10 times more than the ratio test when applied to SIFT, because it operates on the full set of possible matches, rather than the limited set of nearest neighbors.
SuperGlue with SuperPoint achieves state-of-the-art results on indoor pose estimation. They complement each other well since repeatable keypoints make it possible to estimate a larger number of correct matches even in very challenging situations (see Figure~\ref{fig:examples}, Figure~\ref{fig:qual-all}, and \supp~\ref{sec:results-supp}).

\subsection{Outdoor pose estimation}
\label{sec:outdoor}
As outdoor image sequences present their own set of challenges (e.g., lighting changes and occlusion), we train and evaluate SuperGlue for pose estimation in an outdoor setting. We use the same evaluation metrics and baseline methods as in the indoor pose estimation task.

\PAR{Dataset:} We evaluate on the PhotoTourism dataset, which is part of the CVPR'19 Image Matching Challenge~\cite{imwchallenge2019}. It is a subset of the YFCC100M dataset~\cite{thomee2016yfcc100m} and has ground truth poses and sparse 3D models obtained from an off-the-shelf SfM tool~\cite{ono2018lf, schoenberger2016sfm, schoenberger2016mvs}. All learned methods are trained on the larger MegaDepth dataset~\cite{li2018megadepth}, which also has depth maps computed with multi-view stereo. Scenes that are in the PhotoTourism test set are removed from the training set.
Similarly as in the indoor case, we select challenging image pairs for training and evaluation using an overlap score computed from the SfM covisibility as in~\cite{dusmanu2019d2, ono2018lf}.

\PAR{Results:} As shown in Table~\ref{tab:phototourism}, SuperGlue outperforms all baselines, at all relative pose thresholds, when applied to both SuperPoint and SIFT. Most notably, the precision of the resulting matching is very high (84.9\%), reinforcing the analogy that SuperGlue ``glues'' together local features.

\begin{table}[htb]
\centering
\scriptsize{
\setlength\tabcolsep{4.0pt}
\begin{tabular}{llccccc}
    \toprule
    \multirow{2}{1cm}[-.4em]{Local features} & \multirow{2}{*}[-.4em]{Matcher}
    & \multicolumn{3}{c}{Pose estimation AUC} & \multirow{2}{*}[-.4em]{P} & \multirow{2}{*}[-.4em]{MS} \\
    \cmidrule(lr){3-5}
    && @5\degree & @10\degree & @20\degree & & \\
    \midrule
    \multirow{1}{*}{ContextDesc}
    & NN + ratio test & 20.16 & 31.65 & 44.05 & 56.2 & \03.3 \\
    \midrule
    \multirow{4}{*}{SIFT}
    & NN + ratio test & 15.19 & 24.72 & 35.30 & 43.4 & \01.7\\
    & NN + NG-RANSAC & 15.61 & 25.28 & 35.87 & 64.4 & \01.9\\
    & NN + OANet & 18.02 & 28.76 & 40.31 & 55.0 & \03.7\\
    & \b{SuperGlue} & \b{23.68} & \b{36.44} & \b{49.44} & \b{74.1} & \0\b{7.2}\\
    \midrule
    \multirow{4}{*}{SuperPoint}
    & NN + mutual & 9.80 & 18.99 & 30.88 & 22.5 & \04.9\\
    & NN + GMS & 13.96 & 24.58 & 36.53 & 47.1 & \04.7\\
    & NN + OANet & 21.03 & 34.08 & 46.88 & 52.4 & \08.4\\
    & \b{SuperGlue} & \b{34.18} & \b{50.32} & \b{64.16} & \b{84.9} & \b{11.1}\\
    \bottomrule
\end{tabular}
}
\caption{\textbf{Outdoor pose estimation.} Matching SuperPoint and SIFT features with SuperGlue results in significantly higher pose accuracy (AUC), precision (P), and matching score (MS) than with handcrafted or other learned methods.}%
\label{tab:phototourism}%
\end{table}

\begin{table}[tb]
\centering
\scriptsize{
\setlength\tabcolsep{3.0pt}
\begin{tabular}{ll>{\centering}m{14mm}>{\centering}m{10mm}m{10mm}<{\centering}}
    \toprule
    \multicolumn{2}{l}{Matcher} & Pose AUC@20\degree & Match precision & Matching score\\
    \midrule
    \multicolumn{2}{l}{NN + mutual}     & 36.40 & 50.4 & 18.8 \\
    \midrule
    \multirow{5}{*}{\b{SuperGlue}} & No Graph Neural Net    & 38.56 & 66.0 & 17.2 \\
    & No cross-attention                                    & 42.57 & 74.0 & 25.3 \\
    & No positional encoding                                  & 47.12 & 75.8 & 26.6 \\
    & Smaller (3 layers)                                    & 46.93 & 79.9 & 30.0 \\
    & \b{Full} (9 layers)                                   & \b{51.84} & \b{84.4} & \b{31.5} \\
    \bottomrule
\end{tabular}
}
\caption{\textbf{Ablation of SuperGlue.} While the optimal matching layer alone improves over the baseline Nearest Neighbor matcher, the Graph Neural Network explains the majority of the gains brought by SuperGlue. Both cross-attention and positional encoding are critical for strong gluing, and a deeper network further improves the precision.}%
\label{tab:ablation}%
\vspace{-2mm}%
\end{table}

\subsection{Understanding SuperGlue}
\label{sec:understanding}
\PAR{Ablation study:}
To evaluate our design decisions, we repeat the indoor experiments with SuperPoint features, but this time focusing on different SuperGlue variants. This ablation study, presented in Table~\ref{tab:ablation}, shows that all SuperGlue blocks are useful and bring substantial performance gains.

When we additionally backpropagate through the SuperPoint descriptor network while training SuperGlue, we observe an improvement in AUC@20\degree from 51.84 to 53.38.
This confirms that SuperGlue is suitable for end-to-end learning beyond matching.

\PAR{Visualizing Attention:}
The extensive diversity of self- and cross-attention patterns is shown in Figure~\ref{fig:qualitative-attention} and reflects the complexity of the learned behavior. A detailed analysis of the trends and inner-workings is performed in \supp~\ref{sec:sup:attention}.

\section{Conclusion}
This paper demonstrates the power of attention-based graph neural networks for local feature matching. SuperGlue's architecture uses two kinds of attention:
\begin{enumerate*}[label=(\roman*)]
\item self-attention, which boosts the receptive field of local descriptors, and
\item cross-attention, which enables cross-image communication and is inspired by the way humans look back-and-forth when matching images.
\end{enumerate*}
Our method elegantly handles partial assignments and occluded points by solving an optimal transport problem. Our experiments show that SuperGlue achieves significant improvement over existing approaches, enabling highly accurate relative pose estimation on extreme wide-baseline indoor and outdoor image pairs. In addition, SuperGlue runs in real-time and works well with both classical and learned features.

In summary, our learnable middle-end replaces hand-crafted heuristics with a powerful neural model that simultaneously performs context aggregation, matching, and filtering in a single unified architecture. We believe that, when combined with a deep front-end, SuperGlue is a major milestone towards end-to-end deep SLAM.

\begin{figure*}[ht!]
\centering
\begin{minipage}{0.02\textwidth}
    \hfill
\end{minipage}%
\hspace{1mm}%
\begin{minipage}{0.32\textwidth}
    \centering
    \small{SuperPoint + NN + distance threshold}
\end{minipage}%
\hspace{1mm}%
\begin{minipage}{0.32\textwidth}
    \centering
    \small{SuperPoint + NN + OANet}
\end{minipage}%
\hspace{1mm}%
\begin{minipage}{0.32\textwidth}
    \centering
    \small{SuperPoint + \b{SuperGlue}}
\end{minipage}%

\begin{minipage}{0.02\textwidth}
\rotatebox[origin=c]{90}{Indoor}
\end{minipage}%
\hfill{\vline width 1pt}\hfill
\hspace{1mm}%
\begin{minipage}{0.32\textwidth}
    \includegraphics[width=\linewidth]{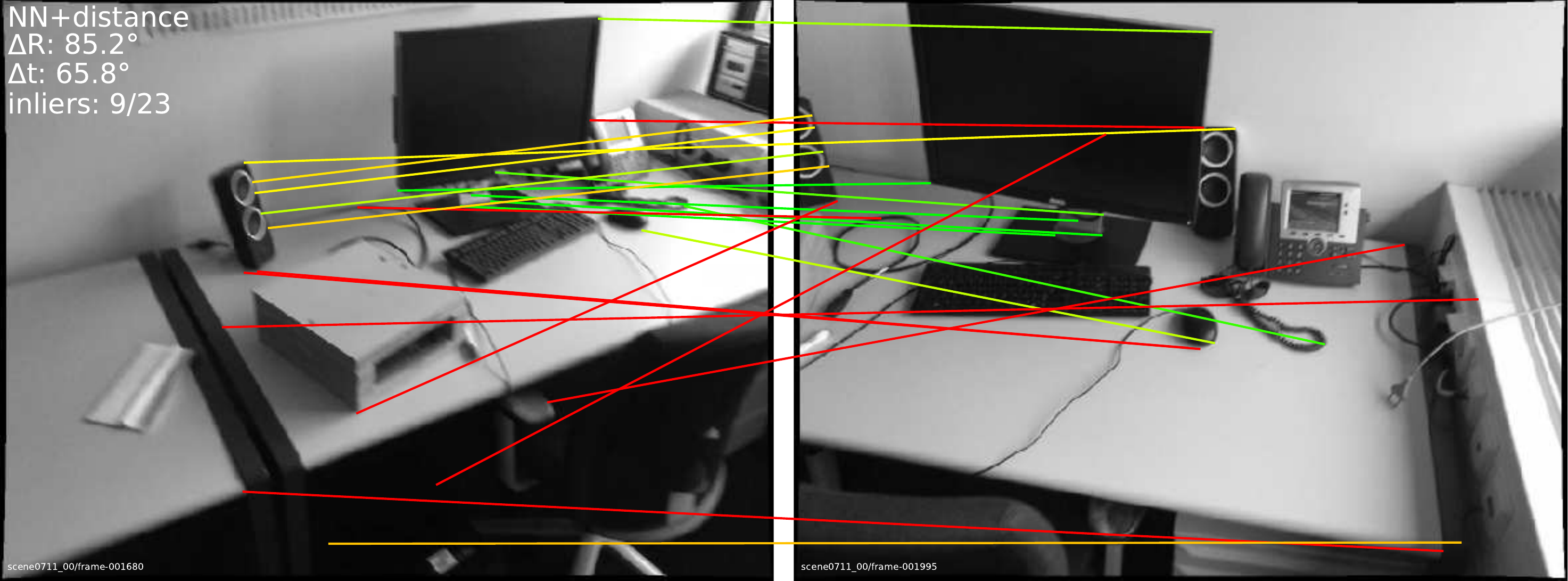}
    
    \vspace{.5mm}
    \includegraphics[width=\linewidth]{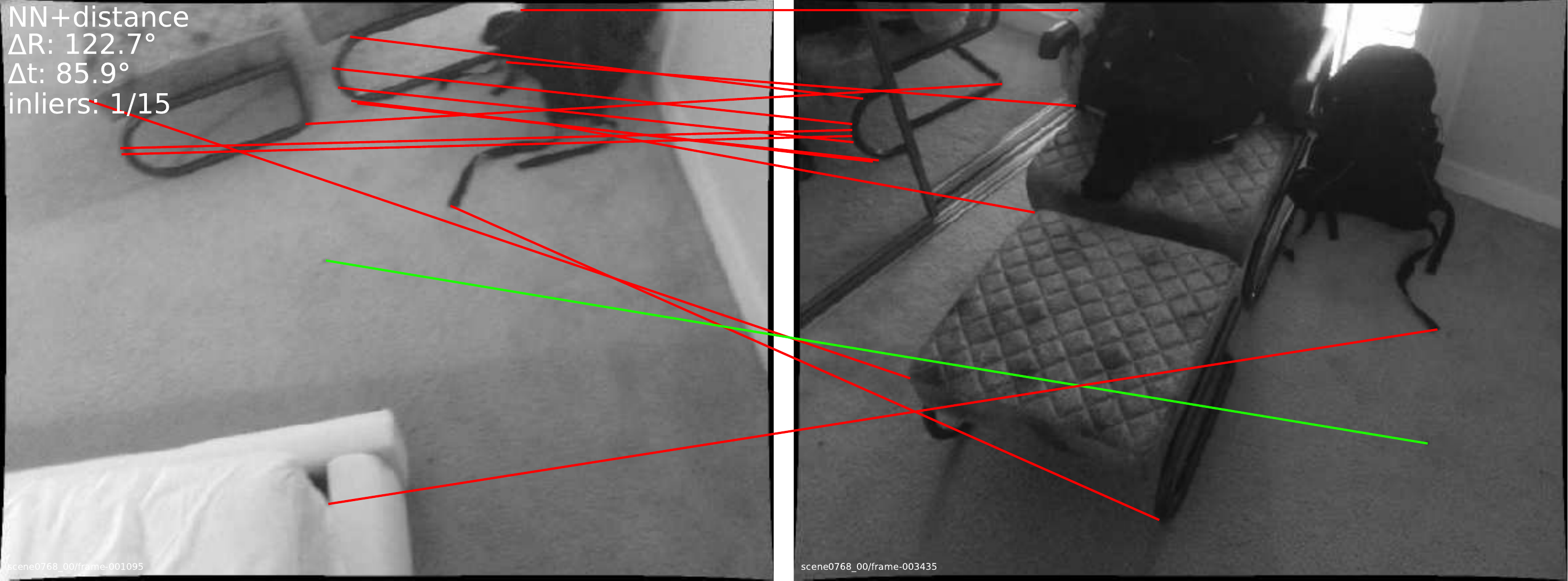}
    
    \vspace{.5mm}
    \includegraphics[width=\linewidth]{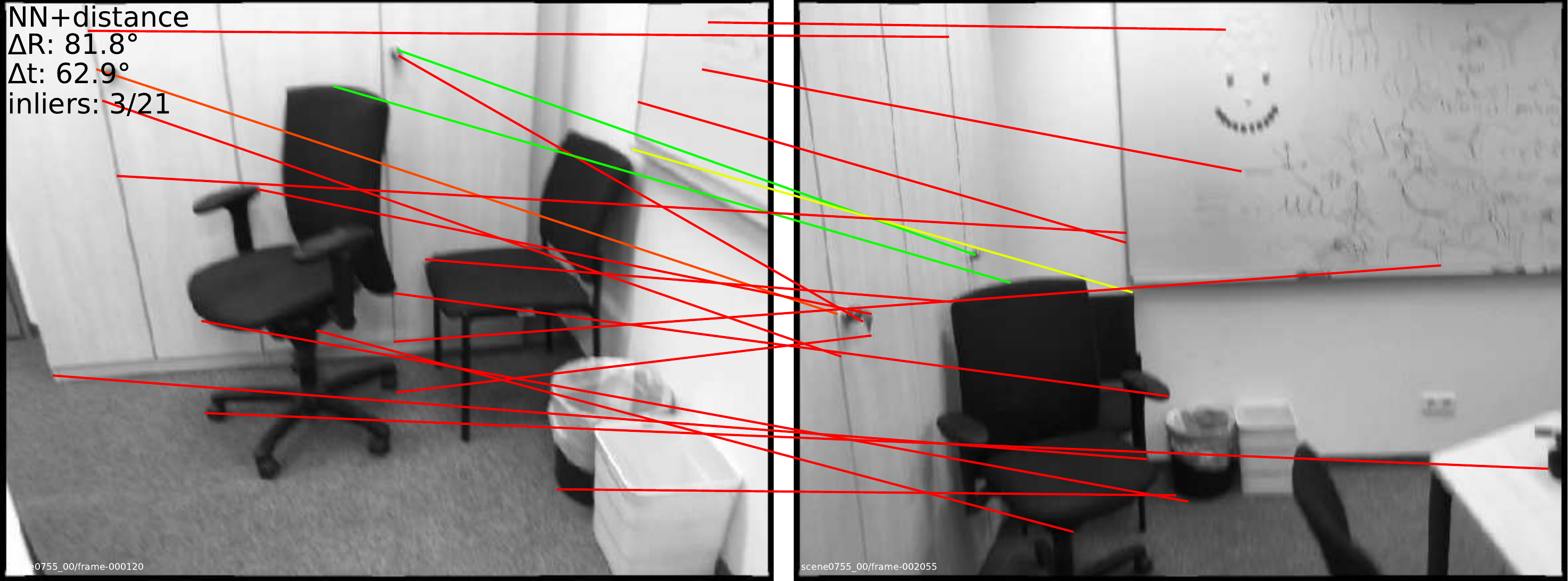}

    \vspace{.5mm}
    \includegraphics[width=\linewidth]{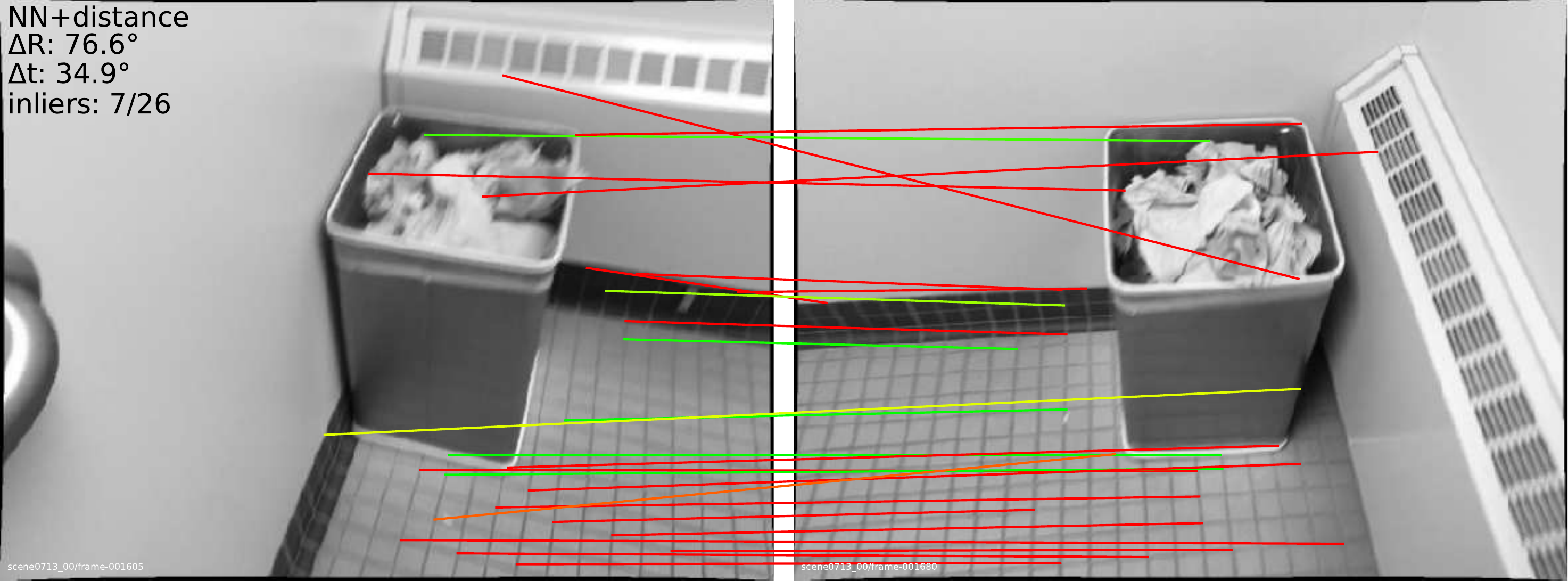}
\end{minipage}%
\hspace{1mm}%
\begin{minipage}{0.32\textwidth}
    \includegraphics[width=\linewidth]{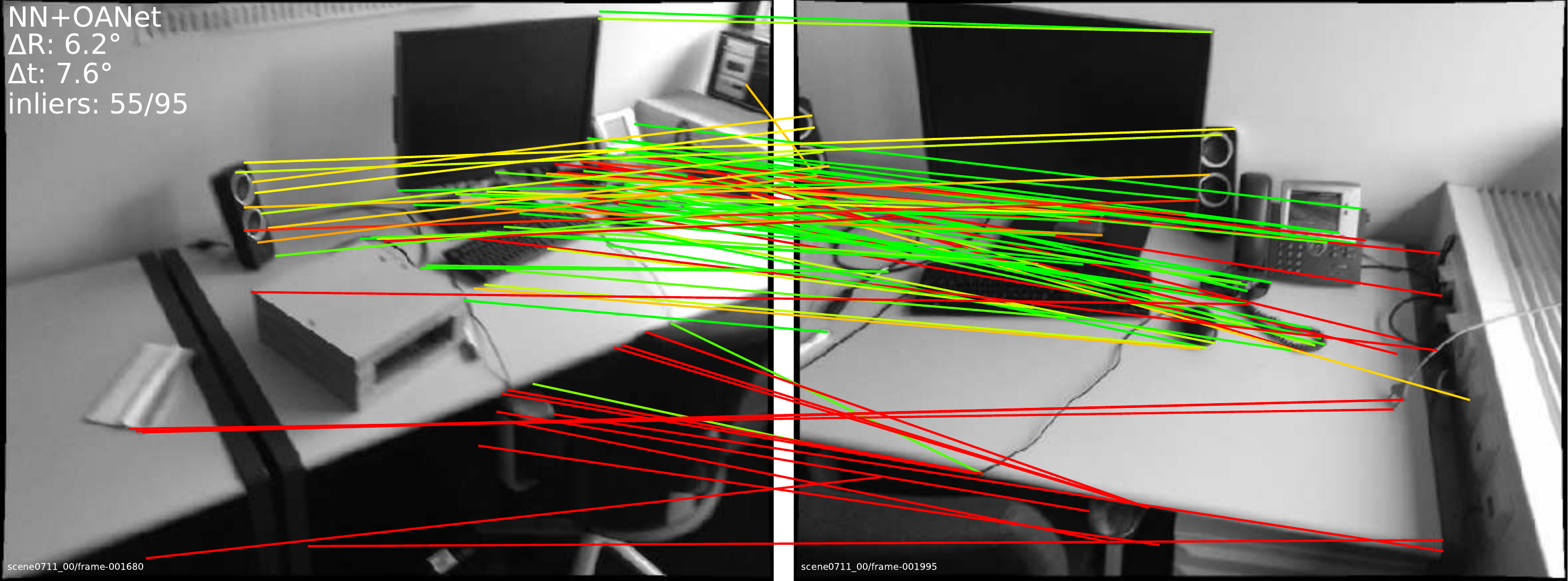}
    
    \vspace{.5mm}
    \includegraphics[width=\linewidth]{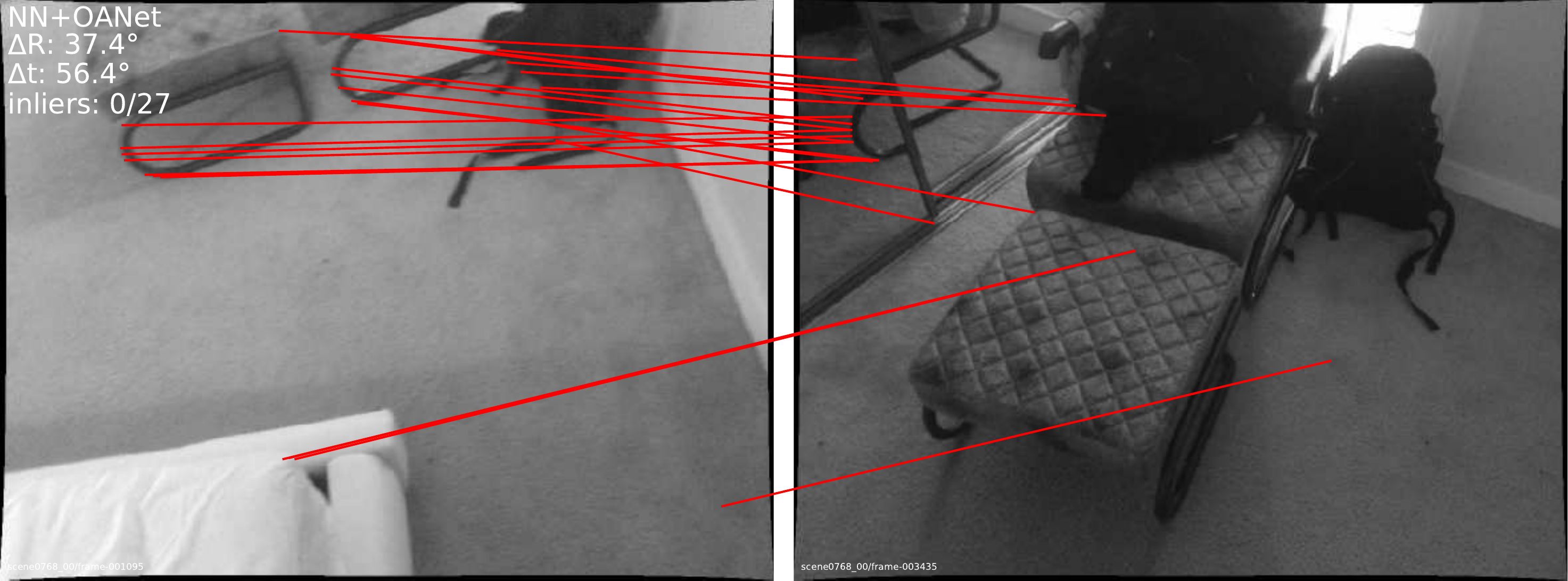}
    
    \vspace{.5mm}
    \includegraphics[width=\linewidth]{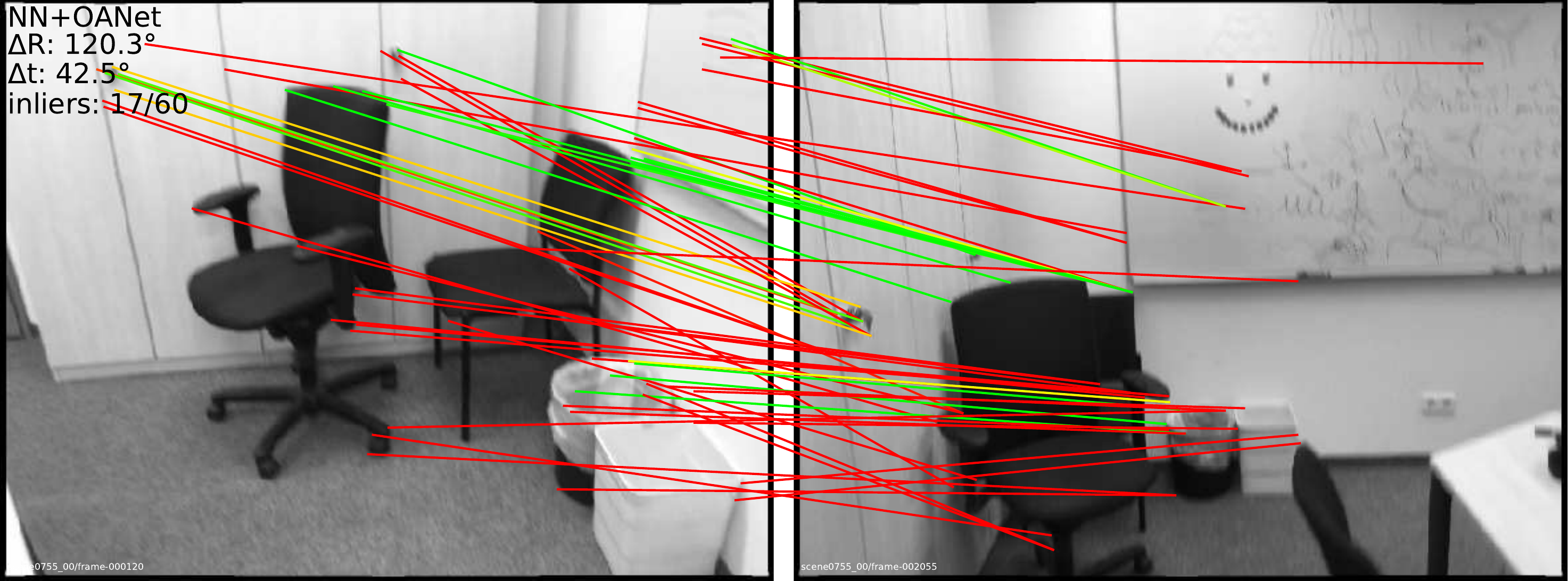}
    
    \vspace{.5mm}
    \includegraphics[width=\linewidth]{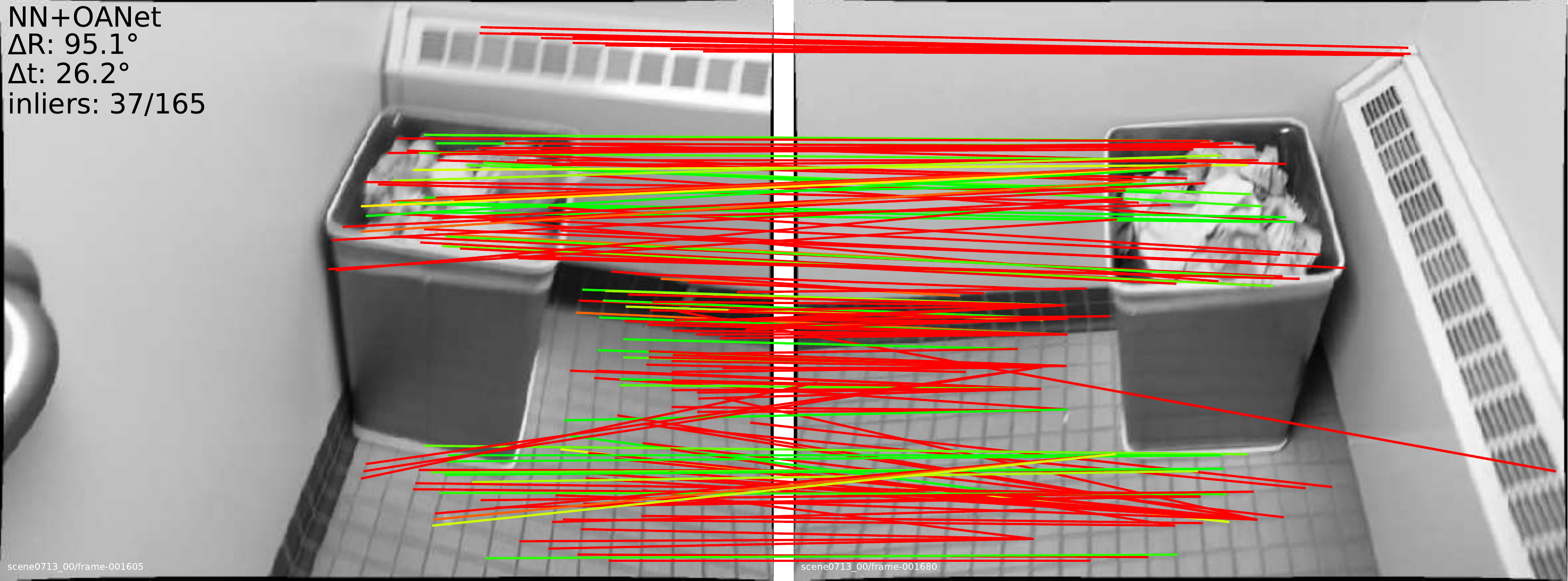}
\end{minipage}%
\hspace{1mm}%
\begin{minipage}{0.32\textwidth}
    \includegraphics[width=\linewidth]{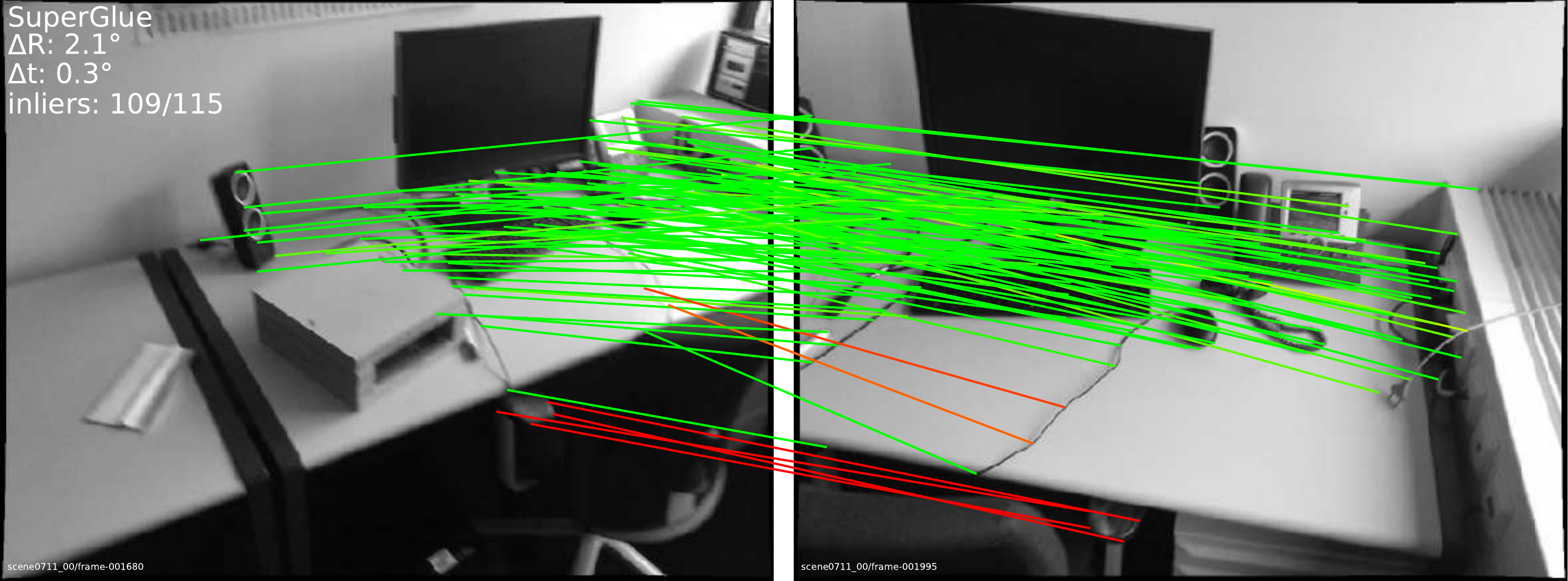}
    
    \vspace{.5mm}
    \includegraphics[width=\linewidth]{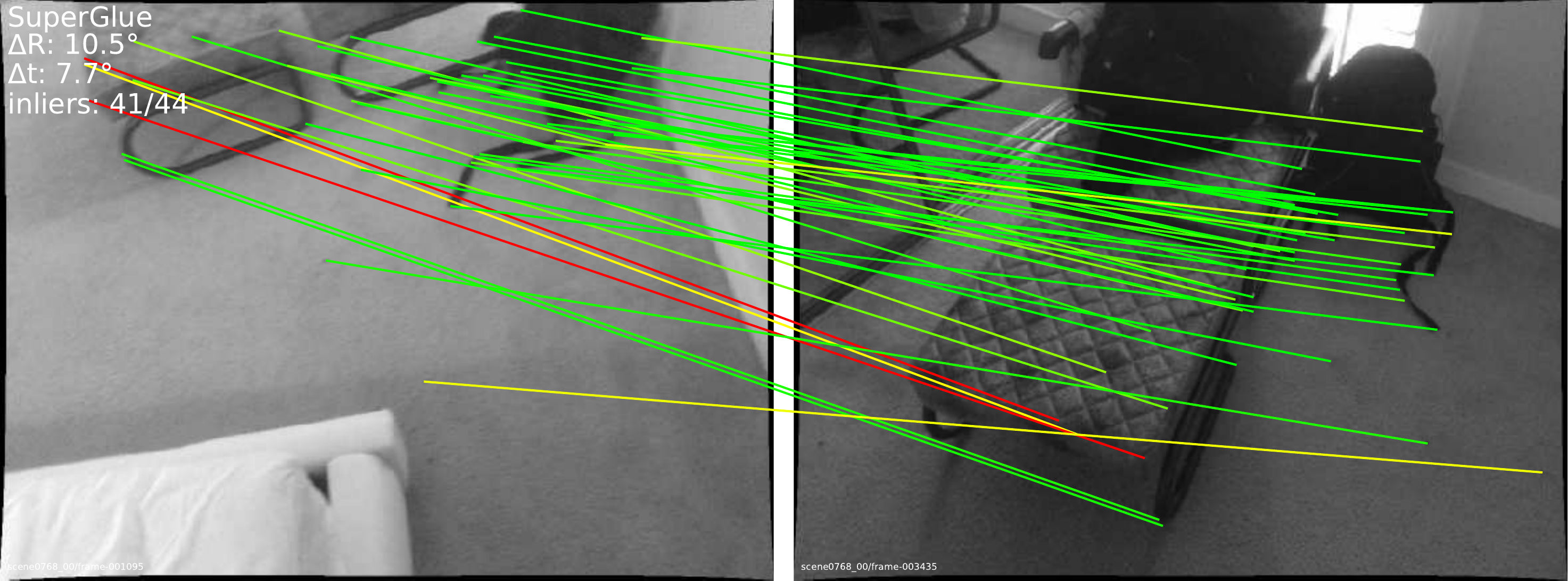}

    \vspace{.5mm}
    \includegraphics[width=\linewidth]{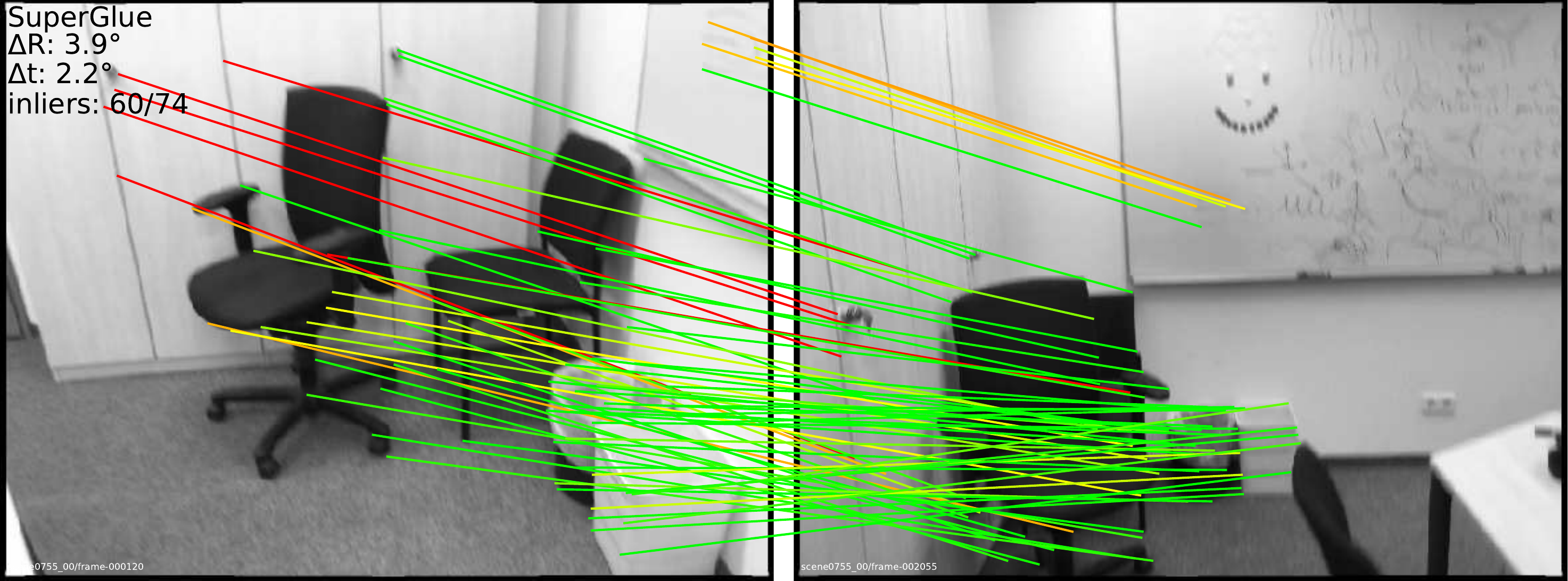}
    
    \vspace{.5mm}
    \includegraphics[width=\linewidth]{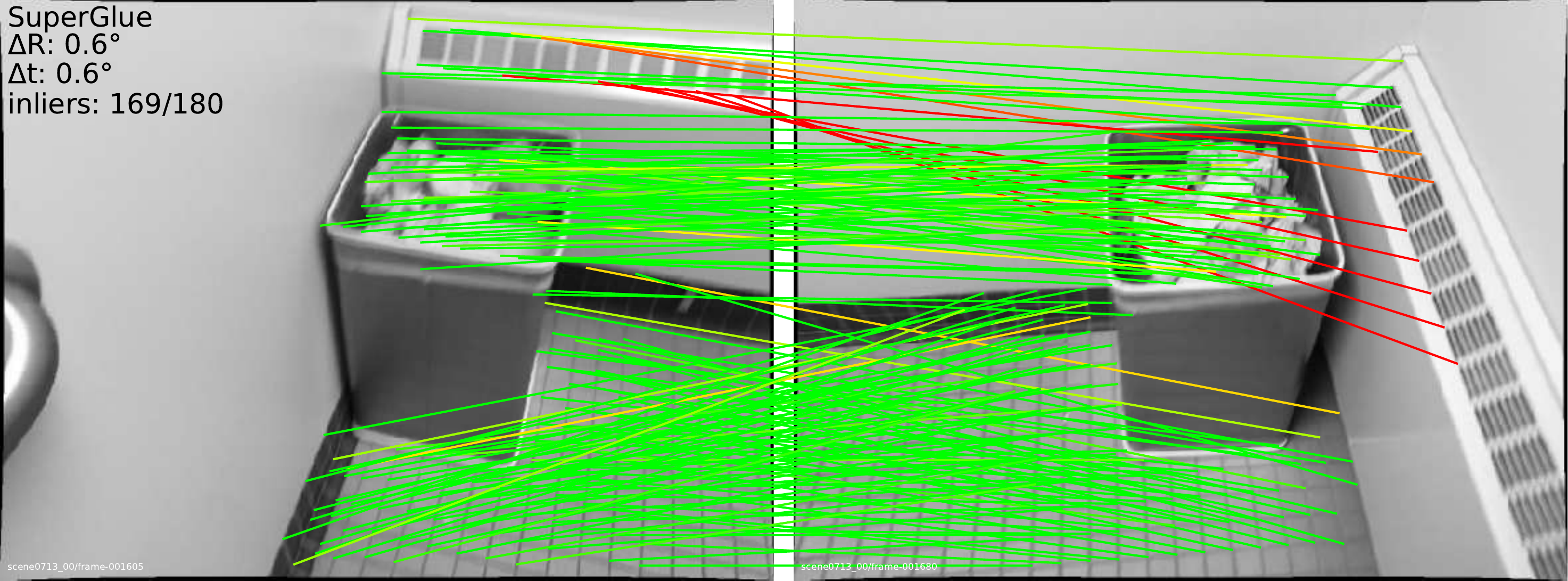}
\end{minipage}

\vspace{1.5mm}
\begin{minipage}{0.02\textwidth}
\rotatebox[origin=c]{90}{Outdoor}
\end{minipage}%
\hfill{\vline width 1pt}\hfill
\hspace{1mm}%
\begin{minipage}{0.32\textwidth}
    \includegraphics[width=\linewidth]{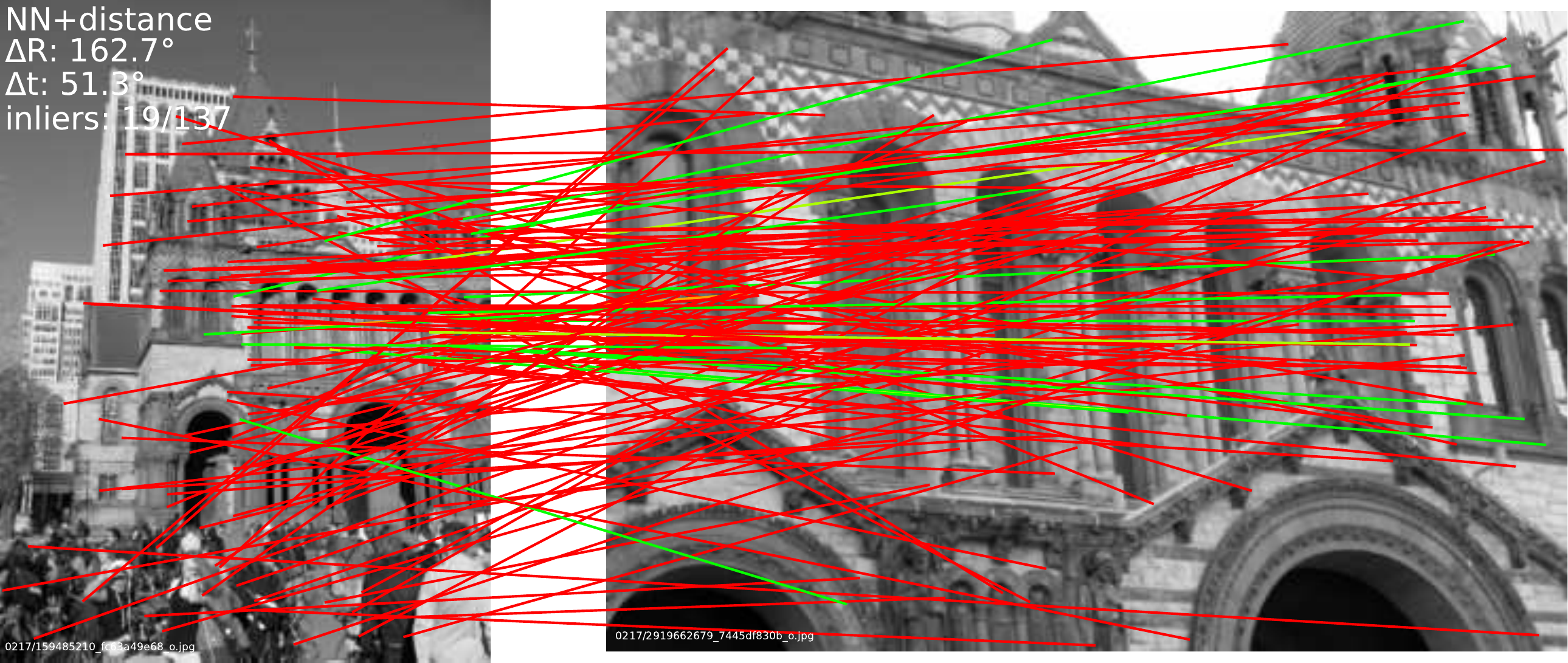}
    
    \vspace{.5mm}
    \includegraphics[width=\linewidth]{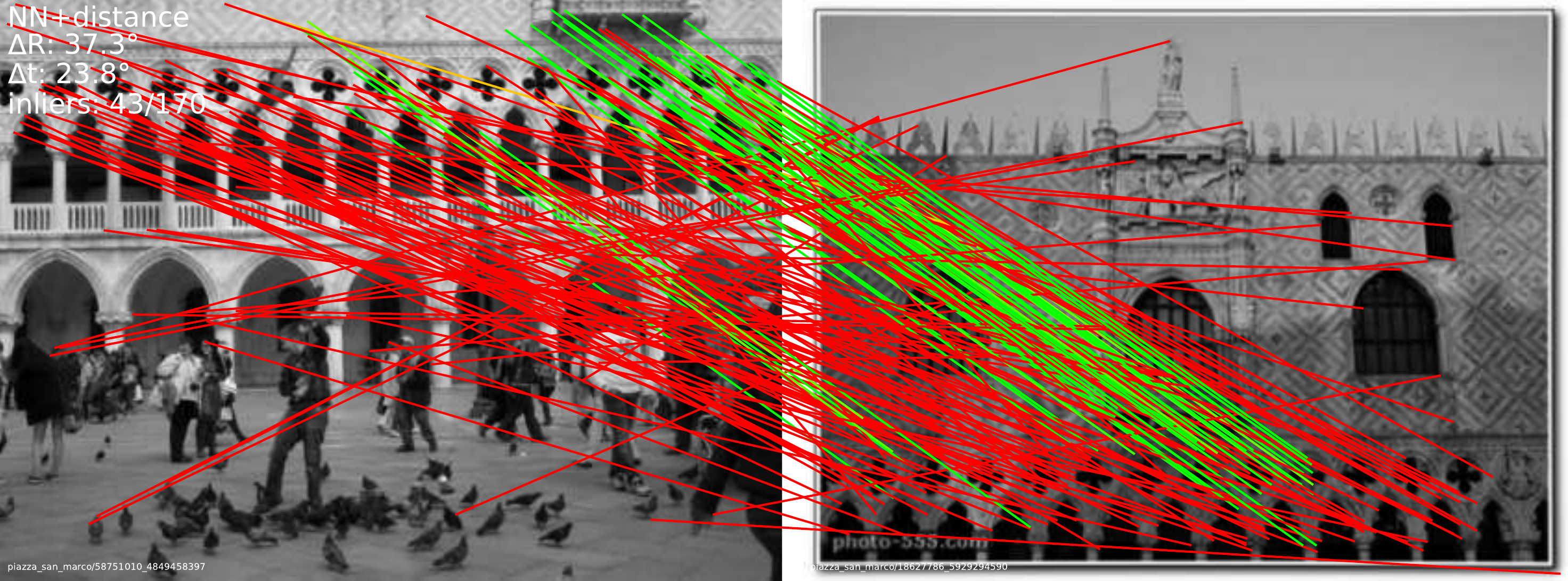}
\end{minipage}%
\hspace{1mm}%
\begin{minipage}{0.32\textwidth}
    \includegraphics[width=\linewidth]{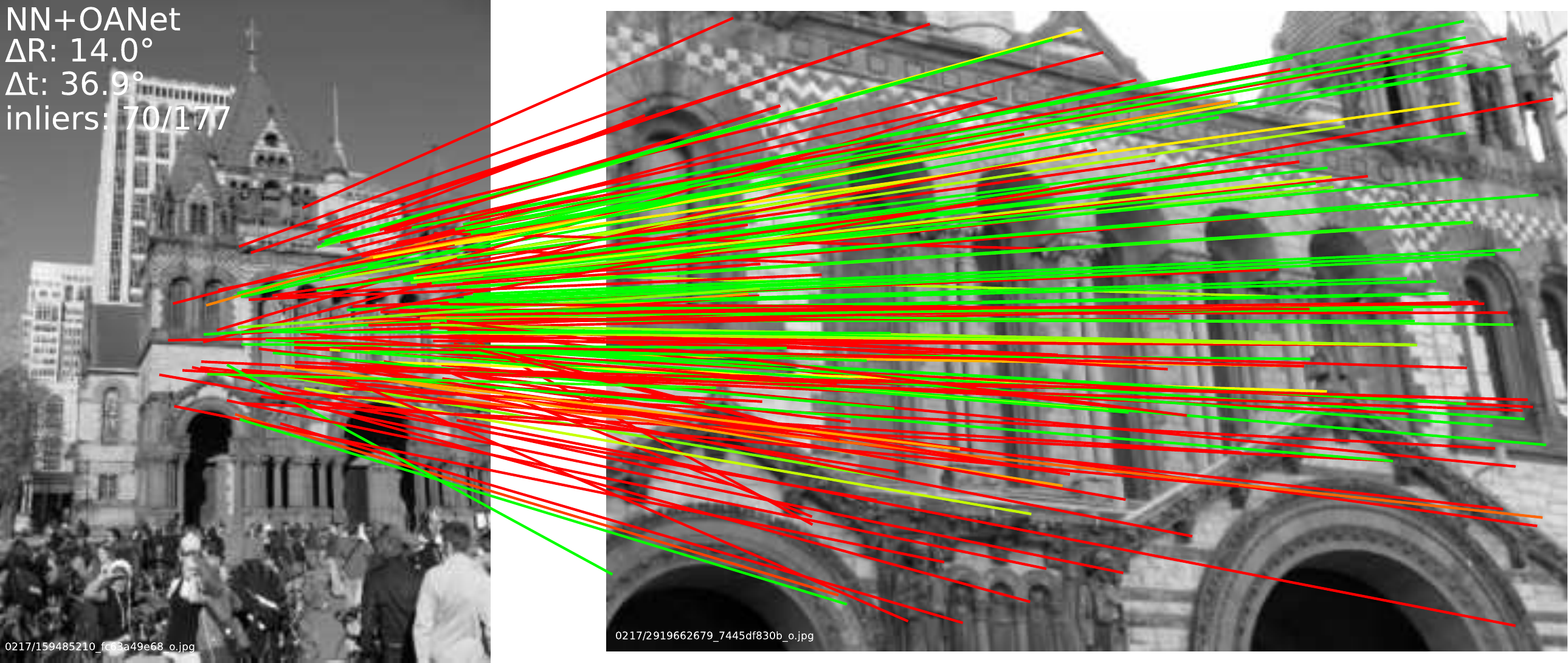}
    
    \vspace{.5mm}
    \includegraphics[width=\linewidth]{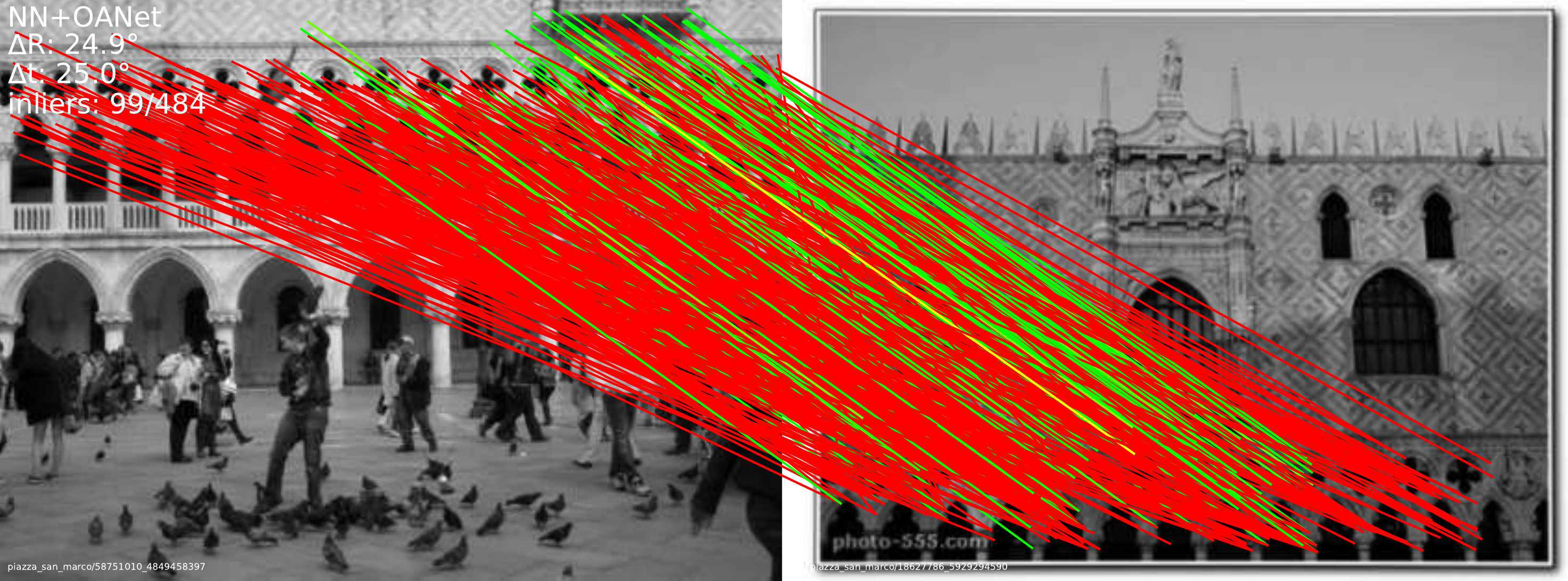}
\end{minipage}%
\hspace{1mm}%
\begin{minipage}{0.32\textwidth}
    \includegraphics[width=\linewidth]{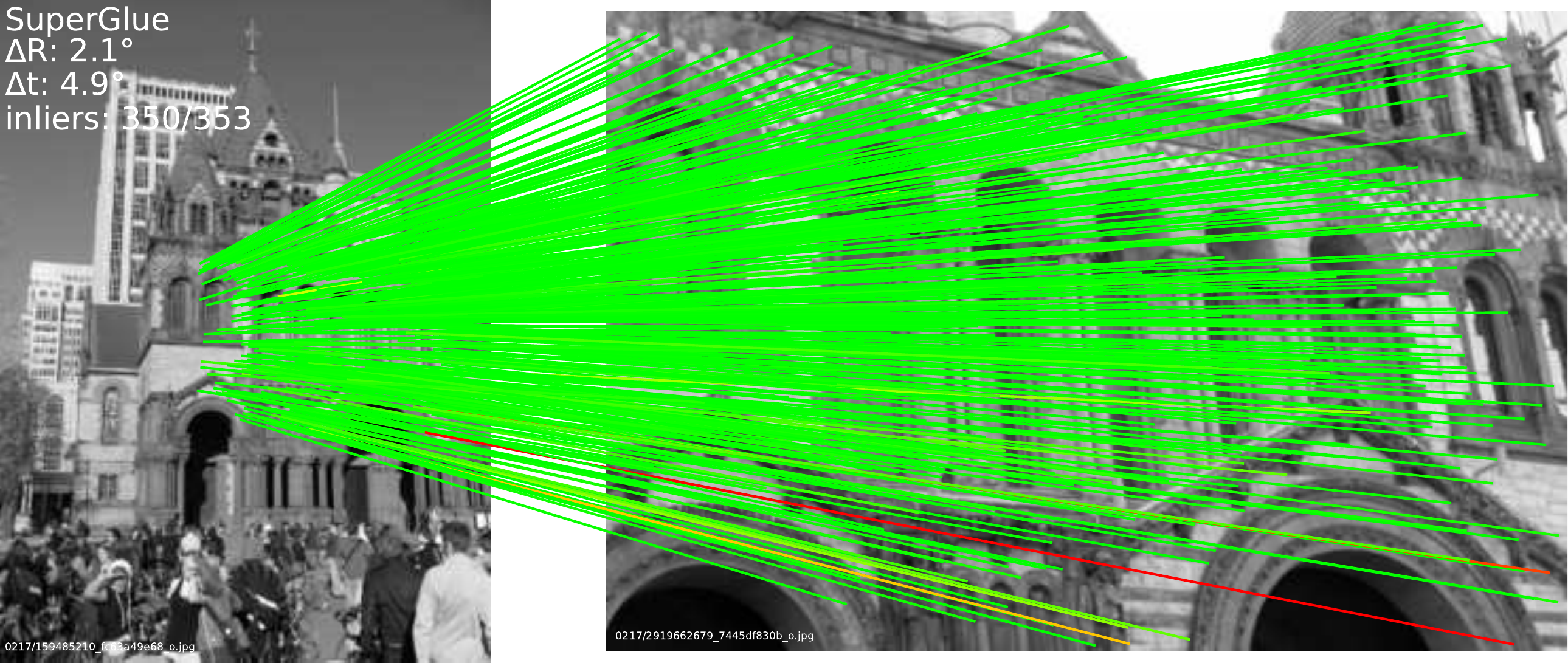}
    
    \vspace{.5mm}
    \includegraphics[width=\linewidth]{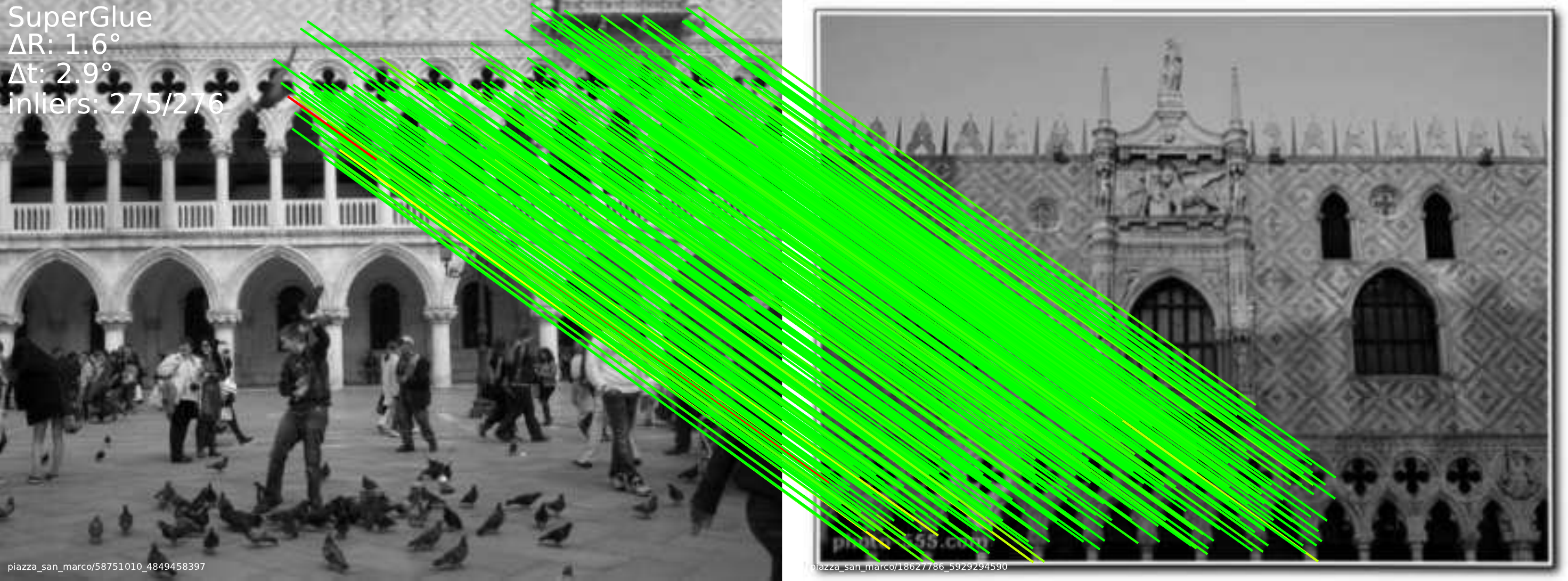}
\end{minipage}

\vspace{1.5mm}
\begin{minipage}{0.02\textwidth}
\rotatebox[origin=c]{90}{Homography}
\end{minipage}%
\hfill{\vline width 1pt}\hfill
\hspace{1mm}%
\begin{minipage}{0.32\textwidth}
    \includegraphics[width=\linewidth]{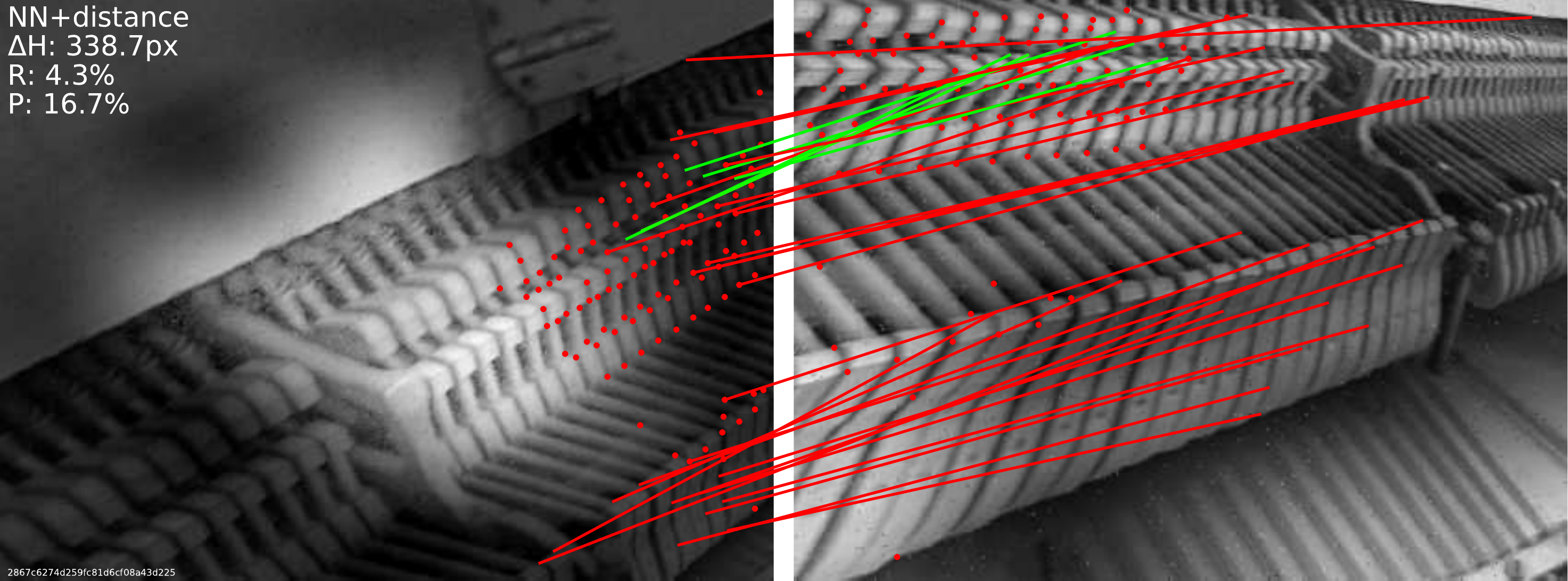}
\end{minipage}%
\hspace{1mm}%
\begin{minipage}{0.32\textwidth}
    \includegraphics[width=\linewidth]{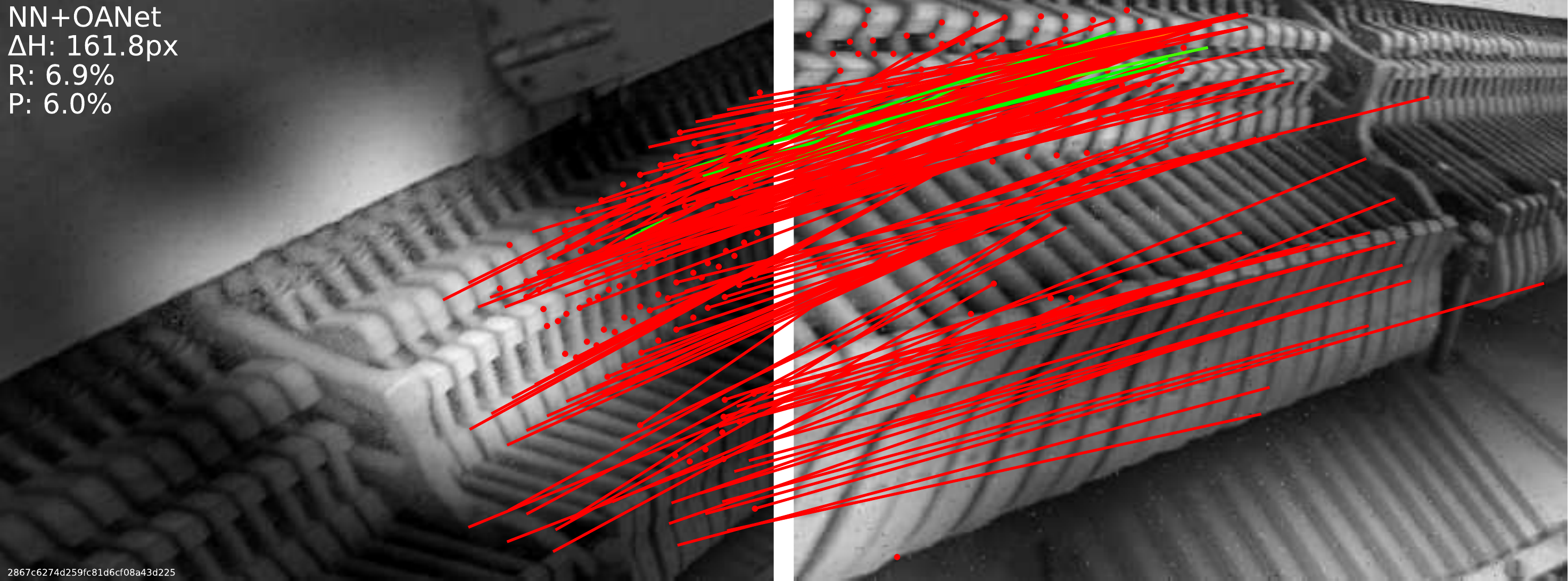}
\end{minipage}%
\hspace{1mm}%
\begin{minipage}{0.32\textwidth}
    \includegraphics[width=\linewidth]{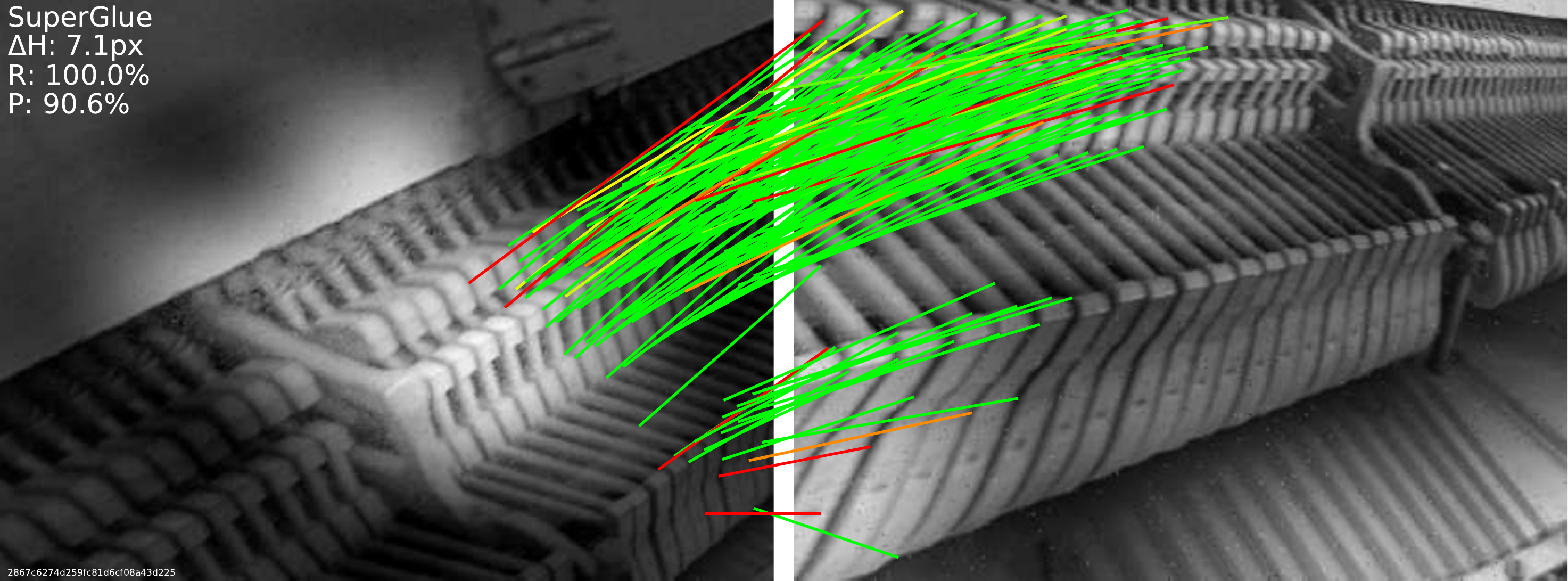}
\end{minipage}
\vspace{-.3cm}
\caption{{\bf Qualitative image matches.} We compare SuperGlue to the Nearest Neighbor (NN) matcher with two outlier rejectors, handcrafted and learned, in three environments. SuperGlue consistently estimates more correct matches ({\color{green}green} lines) and fewer mismatches ({\color{red}red} lines), successfully coping with repeated texture, large viewpoint, and illumination changes.}
\label{fig:qual-all}

\vspace{3mm}
\def\iwidth{0.49}
\begin{minipage}{0.02\textwidth}
\rotatebox[origin=c]{90}{Self}
\end{minipage}%
\hfill{\vline width 1pt}\hfill
\hspace{1mm}%
\begin{minipage}{0.32\textwidth}
    \includegraphics[width=\iwidth\linewidth]{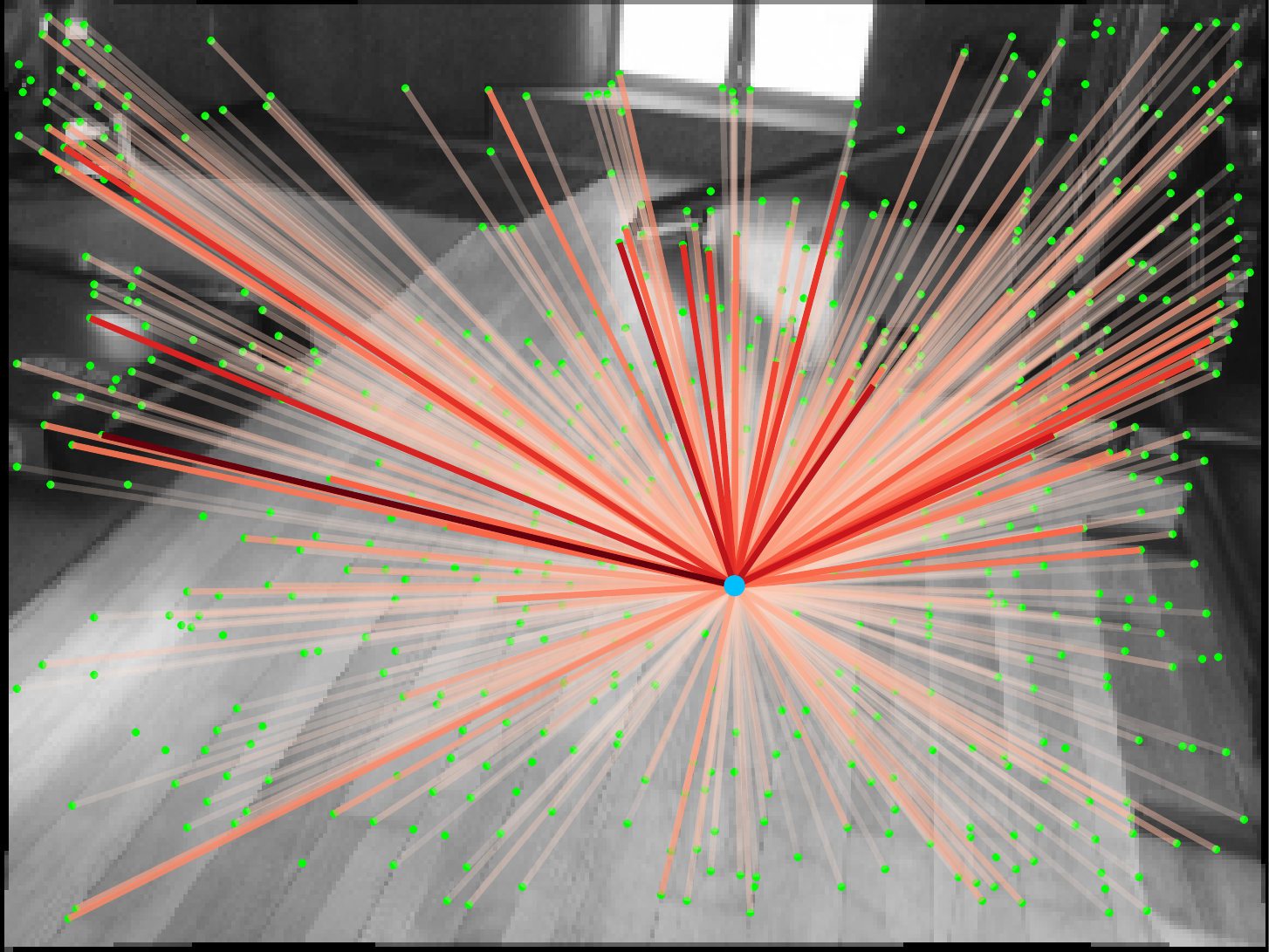}
    \includegraphics[width=\iwidth\linewidth]{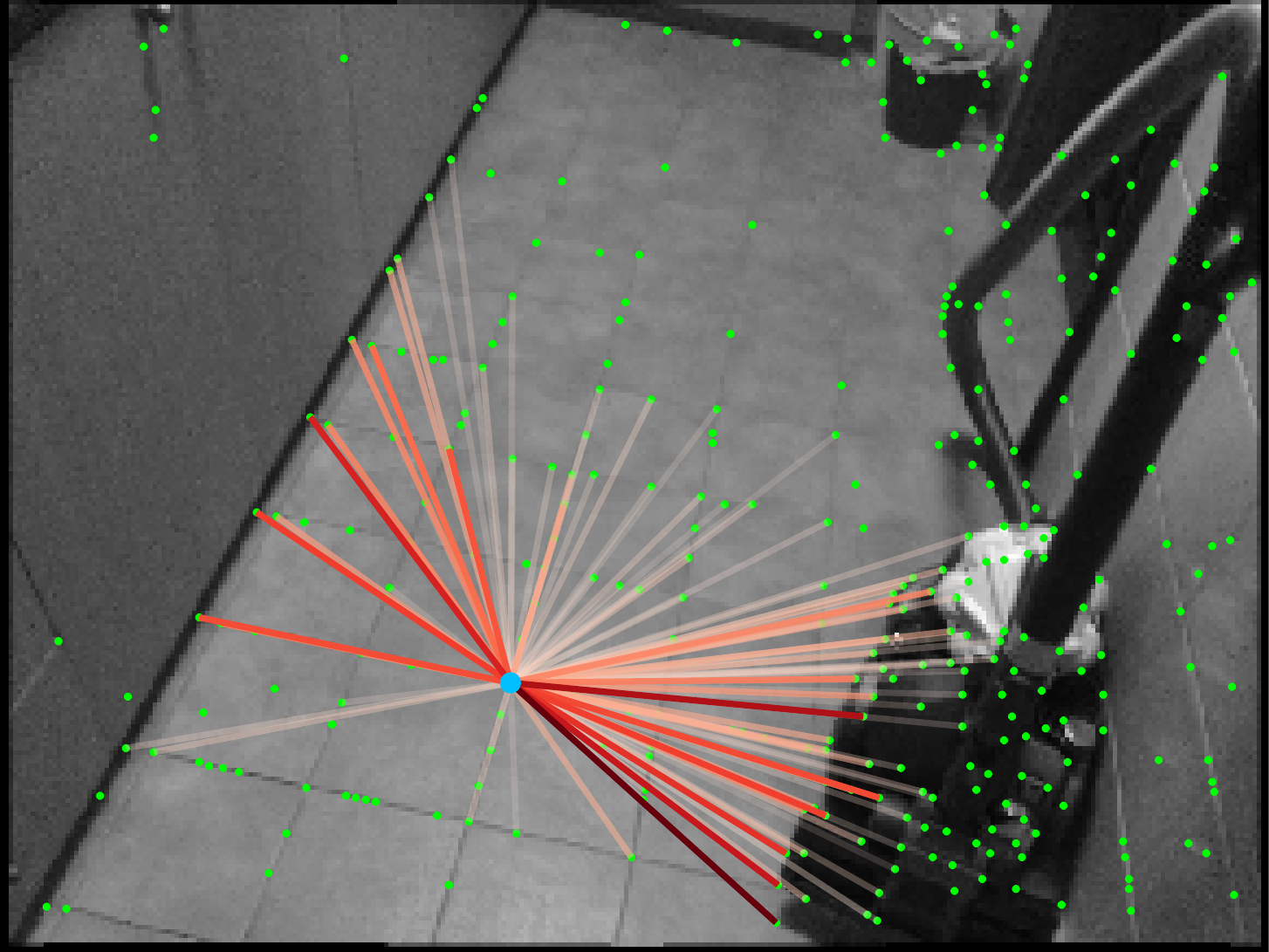}
\end{minipage}%
\hspace{1mm}%
\begin{minipage}{0.32\textwidth}
    \includegraphics[width=\iwidth\linewidth]{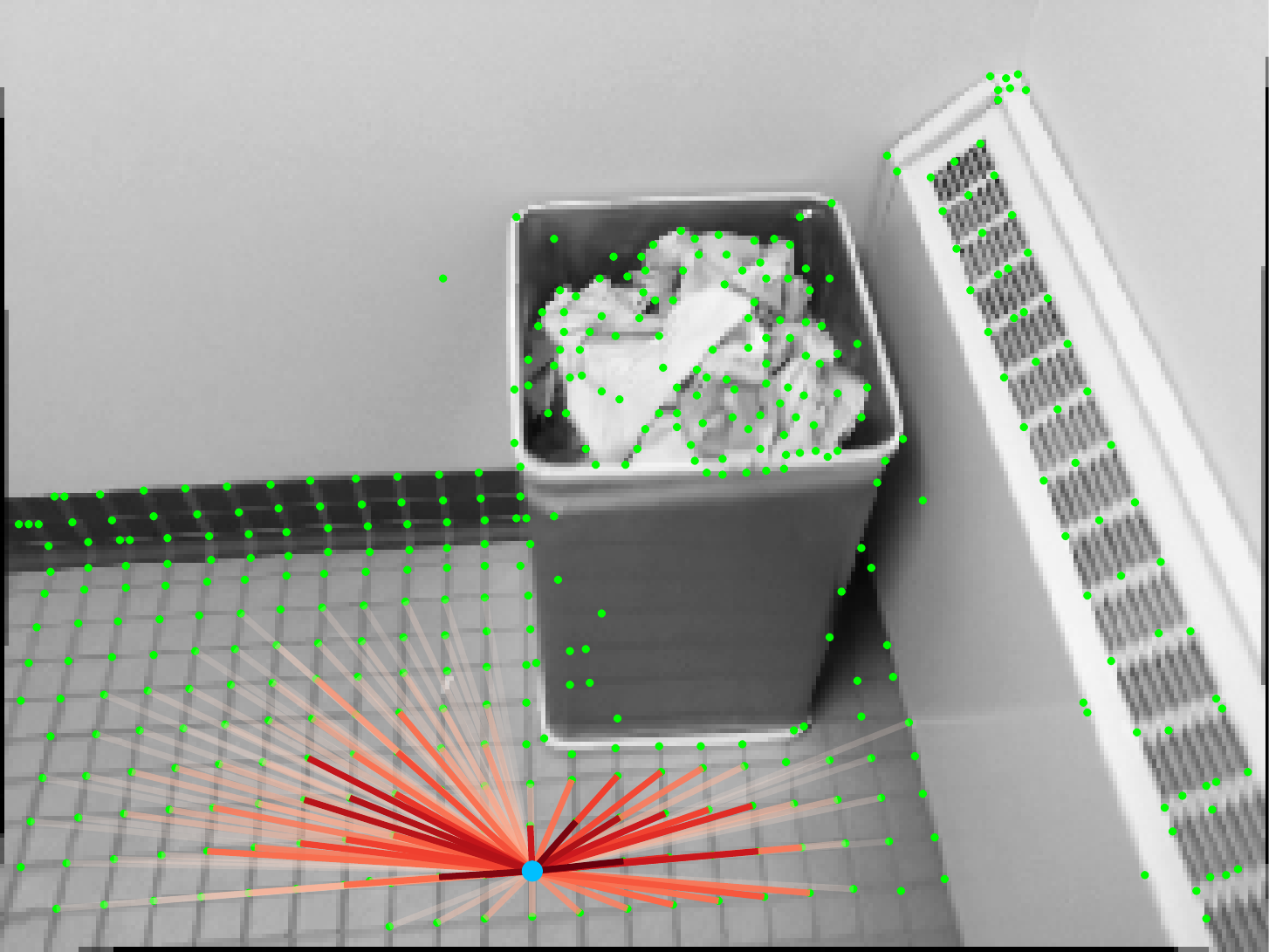}
    \includegraphics[width=\iwidth\linewidth]{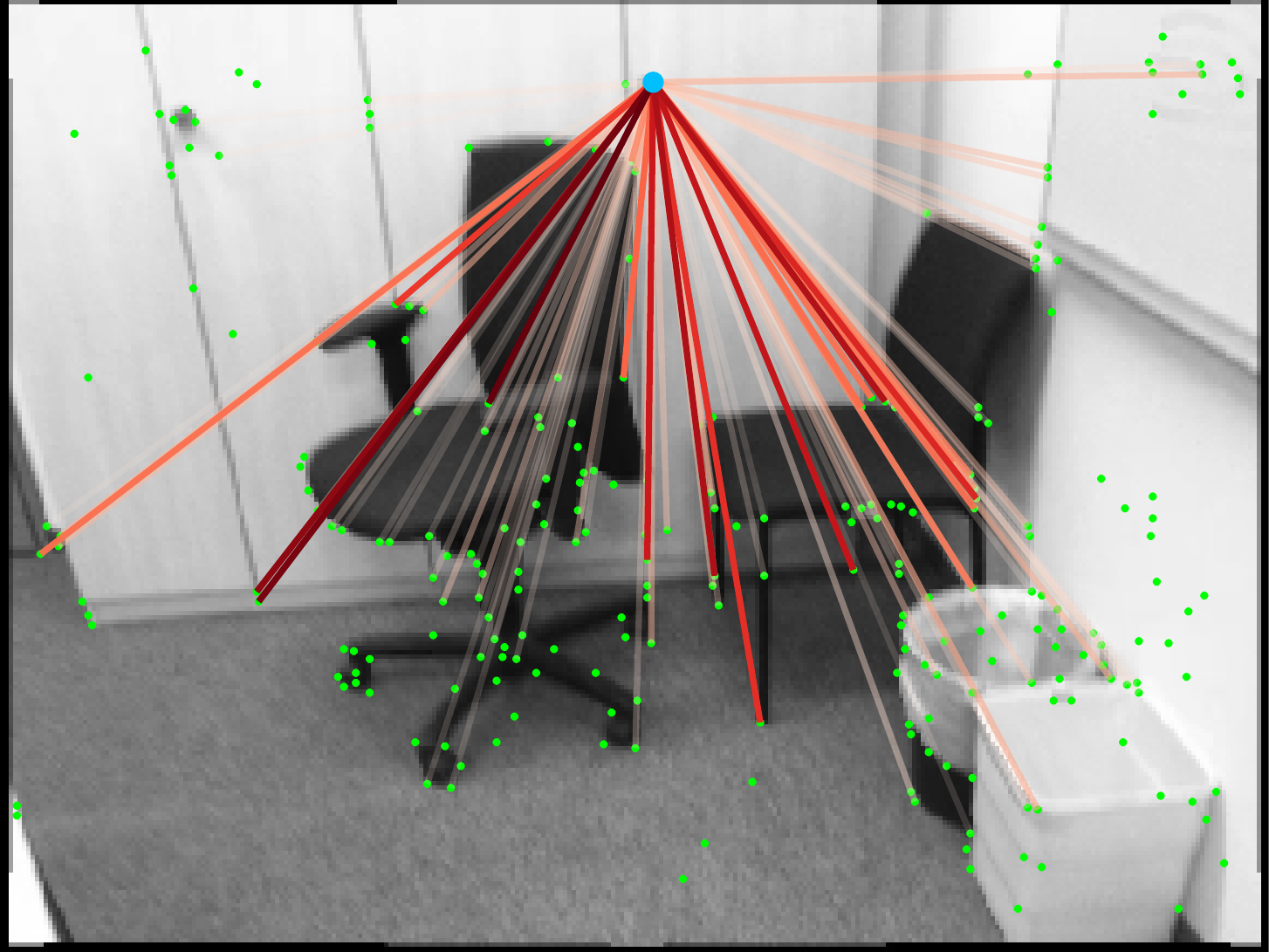}
\end{minipage}%
\hspace{1mm}%
\begin{minipage}[c]{0.32\textwidth}
    \includegraphics[width=\iwidth\linewidth, trim=0 0 1.3cm 0, clip]{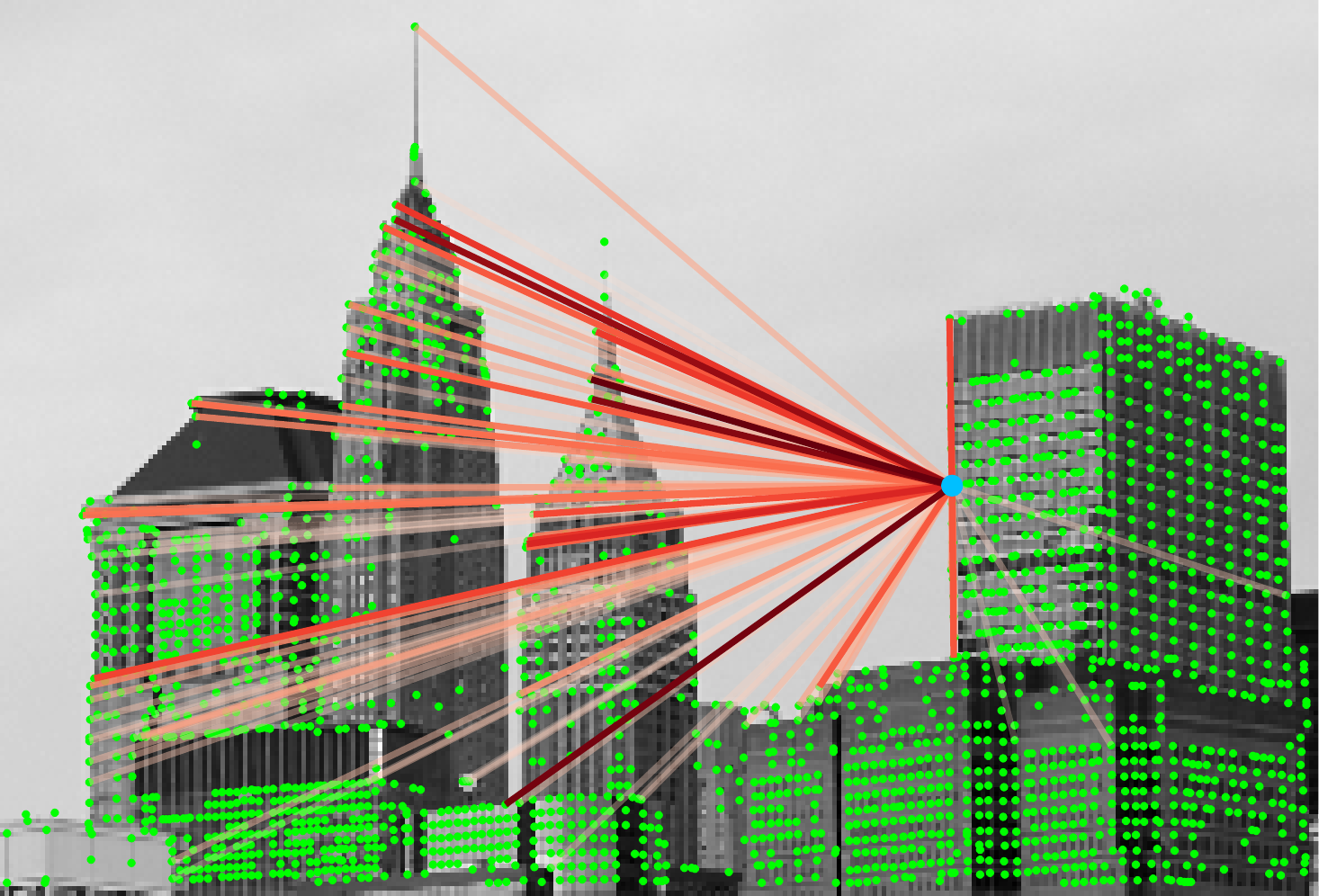}
    \includegraphics[width=\iwidth\linewidth]{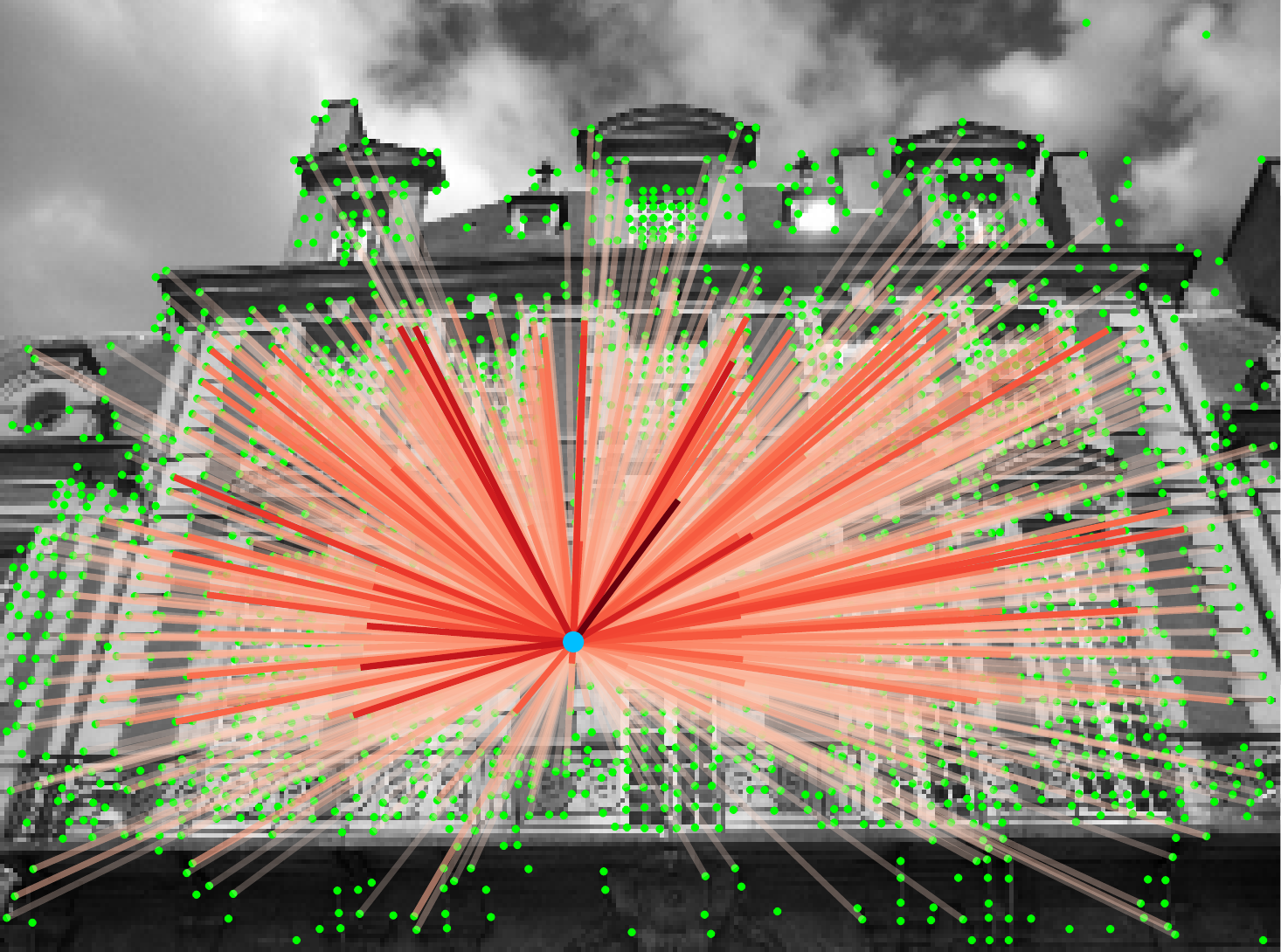}
\end{minipage}

\vspace{.5mm}
\begin{minipage}{0.02\textwidth}
\rotatebox[origin=c]{90}{Cross}
\end{minipage}%
\hfill{\vline width 1pt}\hfill
\hspace{1mm}%
\begin{minipage}{0.32\textwidth}
\includegraphics[width=\linewidth,frame=.4mm -.4mm]{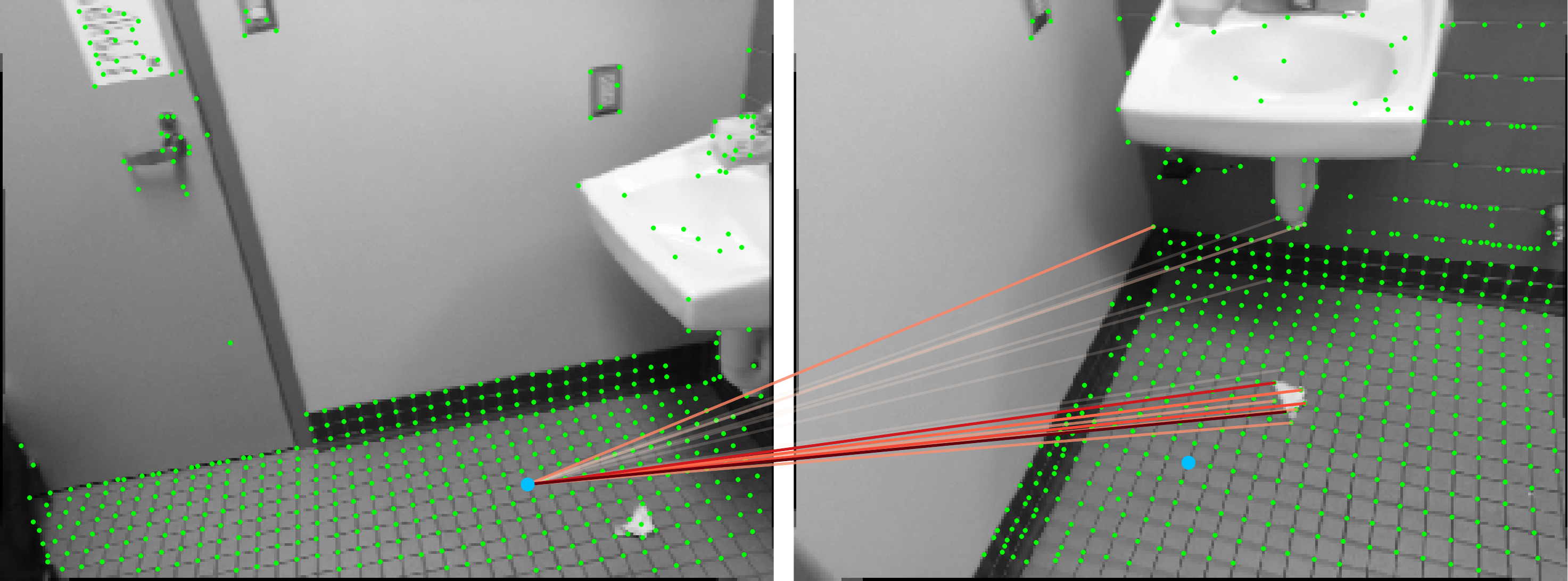}
\end{minipage}%
\hspace{1mm}%
\begin{minipage}{0.32\textwidth}
\includegraphics[width=\linewidth,frame=.4mm -.4mm]{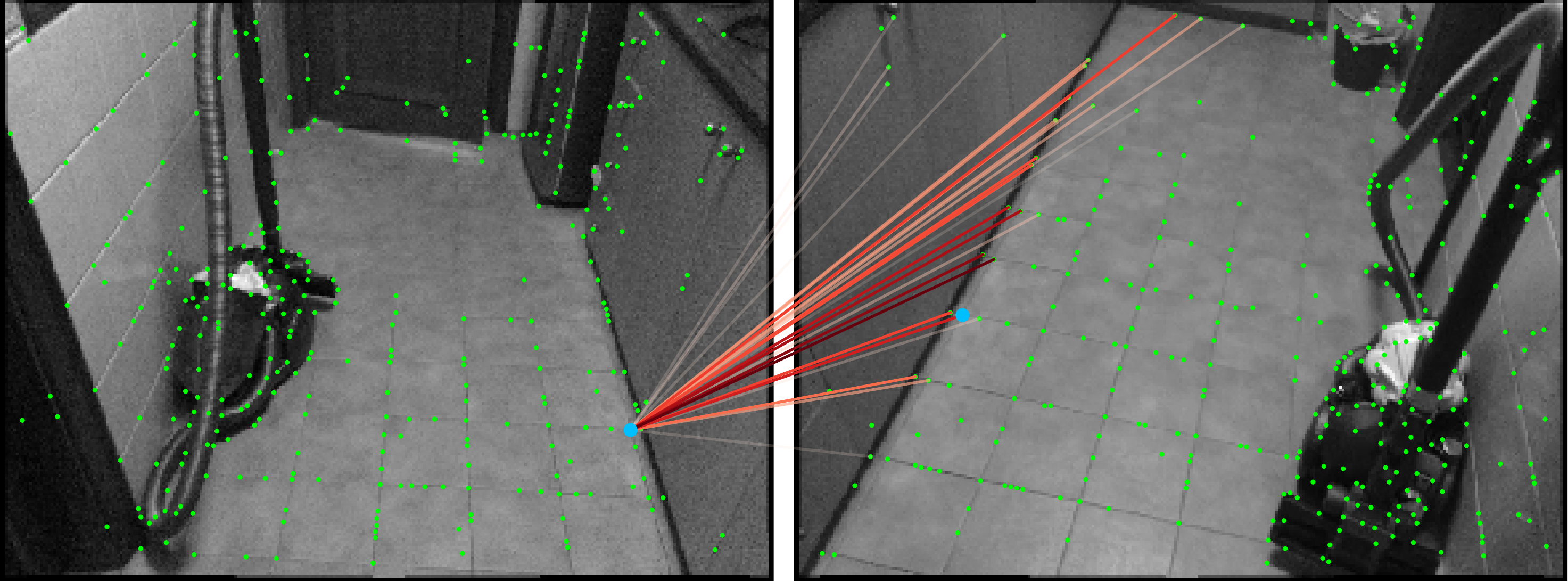}
\end{minipage}%
\hspace{1mm}%
\begin{minipage}{0.32\textwidth}
\includegraphics[width=\linewidth, trim=1.4cm 5mm 1.4cm 5mm, clip,frame=.4mm -.4mm]{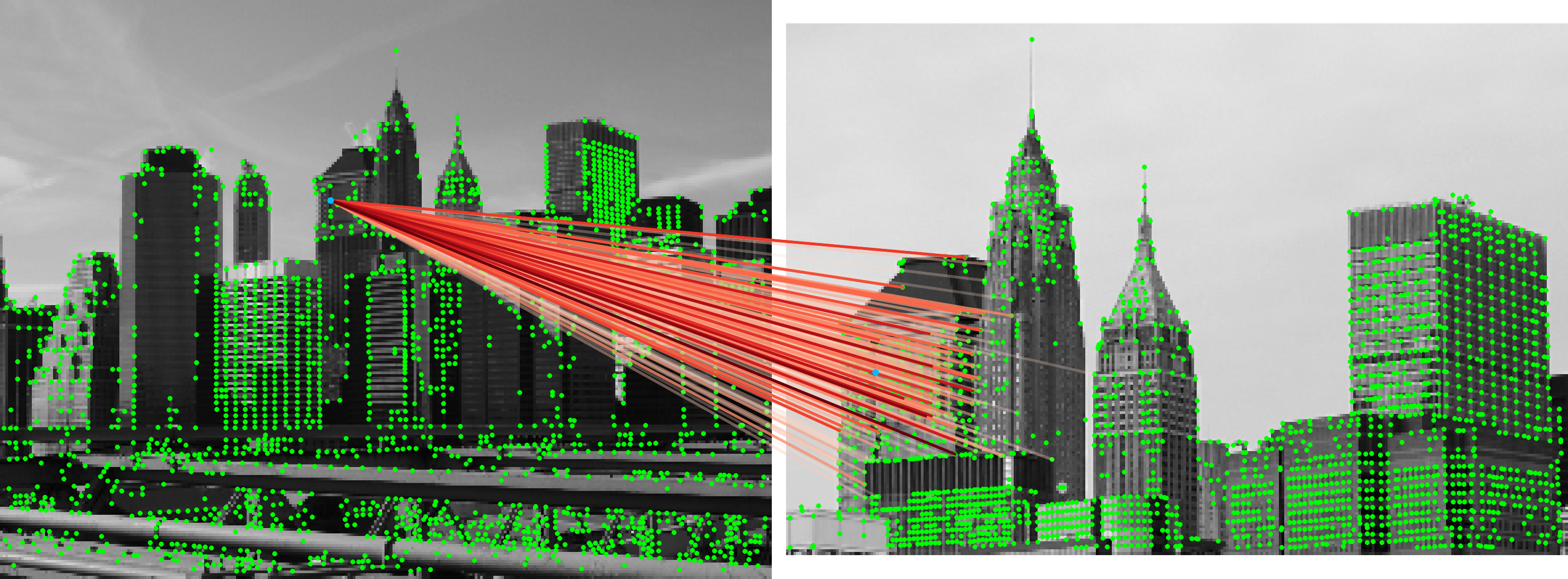}
\end{minipage}
\vspace{-.3cm}
\caption{{\bf Visualizing attention.} We show self- and cross-attention weights $\alpha_{ij}$ at various layers and heads. SuperGlue exhibits a diversity of patterns: it can focus on global or local context, self-similarities, distinctive features, or match candidates.\looseness=-1}
\label{fig:qualitative-attention}
\end{figure*}

\fi 

\ifproceedings
\ifsupponly\appendix

\ifproceedings
\pagestyle{plain}
\begin{strip}
\begin{center}
    \vspace{-1.8cm}
    {\Large \bf SuperGlue: Learning Feature Matching with Graph Neural Networks \par}
    \vspace{.03cm}
    {
      \large
      \lineskip .5em
        Paul-Edouard Sarlin$^{1}$
        \hspace{.05in} Daniel DeTone$^2$ 
        \hspace{.05in} Tomasz Malisiewicz$^2$ 
        \hspace{.05in} Andrew Rabinovich$^2$
    }
    \vspace{-0.3cm}
\end{center}
\end{strip}
\fi
\section*{\supp}
\ifproceedings\else
In the following pages, we present additional experimental details, quantitative results, qualitative examples of SuperGlue in action, detailed timing results, as well as visualizations and analysis of the learned attention patterns.
\fi

\section{Detailed results}
\label{sec:results-supp}

\subsection{Homography estimation}
\label{sec:homography-supp}
\PAR{Qualitative results:} A full page of qualitative results of SuperGlue matching on synthetic and real homographies can be seen in Figure~\ref{fig:supp-homography-qualitative}.

\PAR{Synthetic dataset:} We take a more detailed look at the homography evaluation from Section~\ref{sec:homography}. Figure~\ref{fig:supp-homography-plots} shows the match precision at several correctness pixel thresholds and the cumulative error curve of homography estimation. SuperGlue dominates across all pixel correctness thresholds.\ifproceedings\blfootnote{$^1$ ETH Zurich}\blfootnote{$^2$ Magic Leap, Inc.}\fi%

\begin{figure}[!h]
    \centering
    \includegraphics[width=1\linewidth, trim=0cm 0cm 0cm 0, clip]{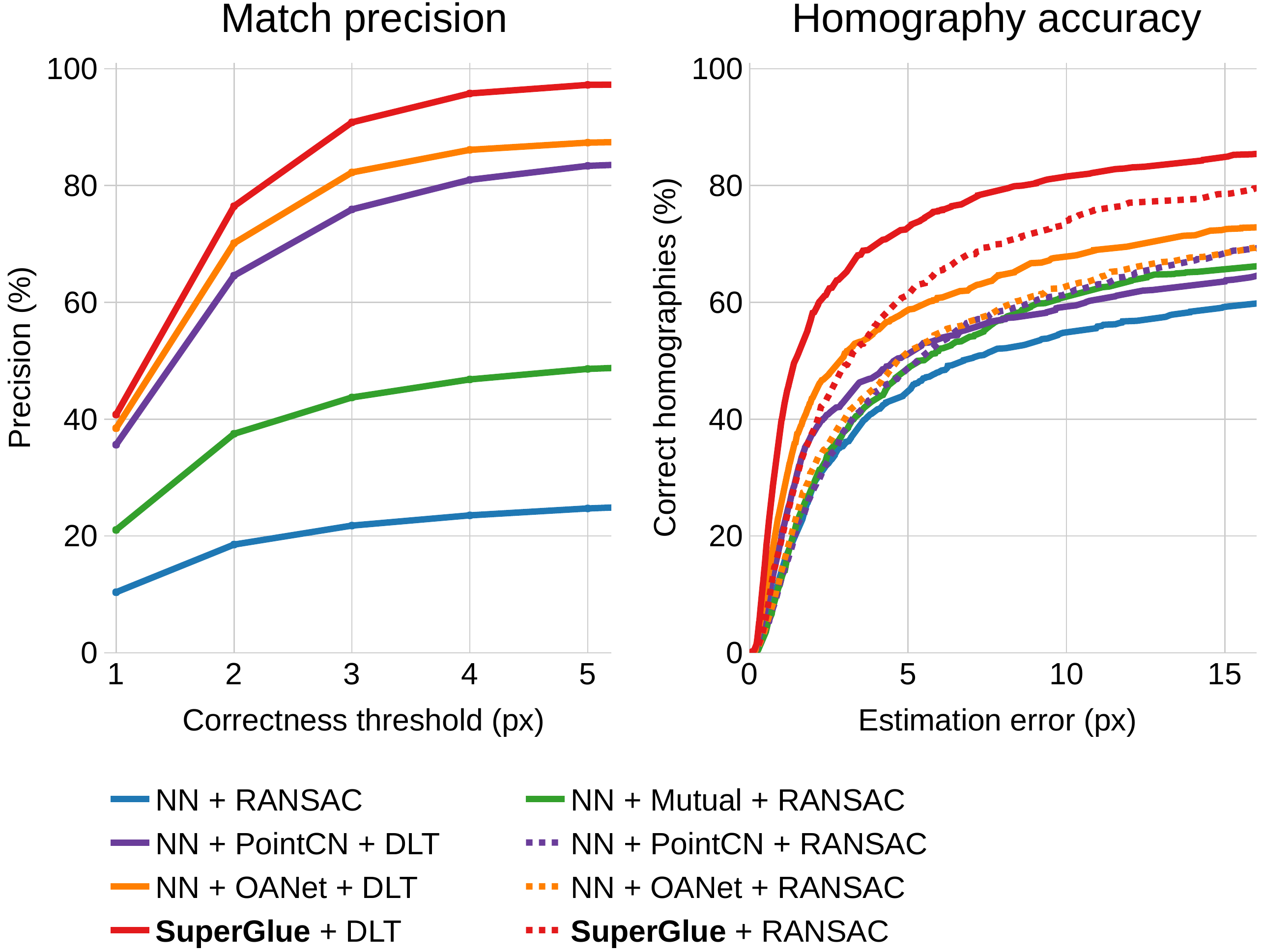}%
    \vspace{-.1cm}%
    \caption{\textbf{Details of the homography evaluation.}
    SuperGlue exhibits higher precision and homography accuracy at all thresholds. High precision results in more accurate estimation with DLT than with RANSAC.
    }
    \label{fig:supp-homography-plots}
\end{figure}

\begin{table}[!h]
\centering
\scriptsize{
\setlength\tabcolsep{3.0pt}
\begin{tabular}{llcccc}
    \toprule
    \multirow{2}{1cm}[-.4em]{Local features} & \multirow{2}{*}[-.4em]{Matcher} &
    \multicolumn{2}{c}{Viewpoint} & \multicolumn{2}{c}{Illumination} \\
    \cmidrule(lr){3-4}\cmidrule(lr){5-6}
    && P & R & P & R\\
    \midrule
    \multirow{5}{*}{SuperPoint}
    & NN & 39.7 & 81.7 & 51.1 & 84.9\\
    & NN + mutual & 65.6 & 77.1 & 74.2 & 80.7\\
    & NN + PointCN & 87.6 & 80.7 & 94.5 & 82.6\\
    & NN + OANet & 90.4 & 81.2 & \b{96.3} & 83.5\\
    & \b{SuperGlue} & \b{91.4} & \b{95.7} & 89.1 & \b{91.7}\\
    \bottomrule
\end{tabular}
}
\vspace{-.05in}
\caption{\textbf{Generalization to real data.} We show the precision (P) and recall (R) of the methods trained on our synthetic homography dataset (see Section~\ref{sec:homography}) on the viewpoint and illumination subsets of the HPatches dataset. While trained on synthetic homographies, SuperGlue generalizes well to real data.}
\label{tab:supp-hpatches}
\end{table}

\PAR{HPatches:} We assess the generalization ability of SuperGlue on real data with the HPatches~\cite{balntas2017hpatches} dataset, as done in previous works~\cite{superpoint, revaud2019r2d2}. This dataset depicts planar scenes with ground truth homographies and contains 295 image pairs with viewpoint changes and 285 pairs with illumination changes. We evaluate the models trained on the synthetic dataset (see Section~\ref{sec:homography}). The HPatches experiment is summarized in Table~\ref{tab:supp-hpatches}. As previously observed in the synthetic homography experiments, SuperGlue has significantly higher recall than all matchers relying on the NN search.
We attribute the remaining gap in recall to several challenging pairs for which SuperPoint does not detect enough repeatable keypoints.
Nevertheless, synthetic-dataset trained SuperGlue generalizes well to real data.

\subsection{Indoor pose estimation}
\label{sec:indoor-supp}
\PAR{Qualitative results:} More visualizations of matches computed by SuperGlue on indoor images are shown in Figure~\ref{fig:supp-scannet-qualitative}, and highlight
the extreme difficulty of the wide-baseline image pairs that constitute our evaluation dataset.

\begin{figure}[b]
\vspace{-4mm}
    \centering
    \includegraphics[width=1.0\linewidth, trim=0cm 0cm 0cm 0, clip]{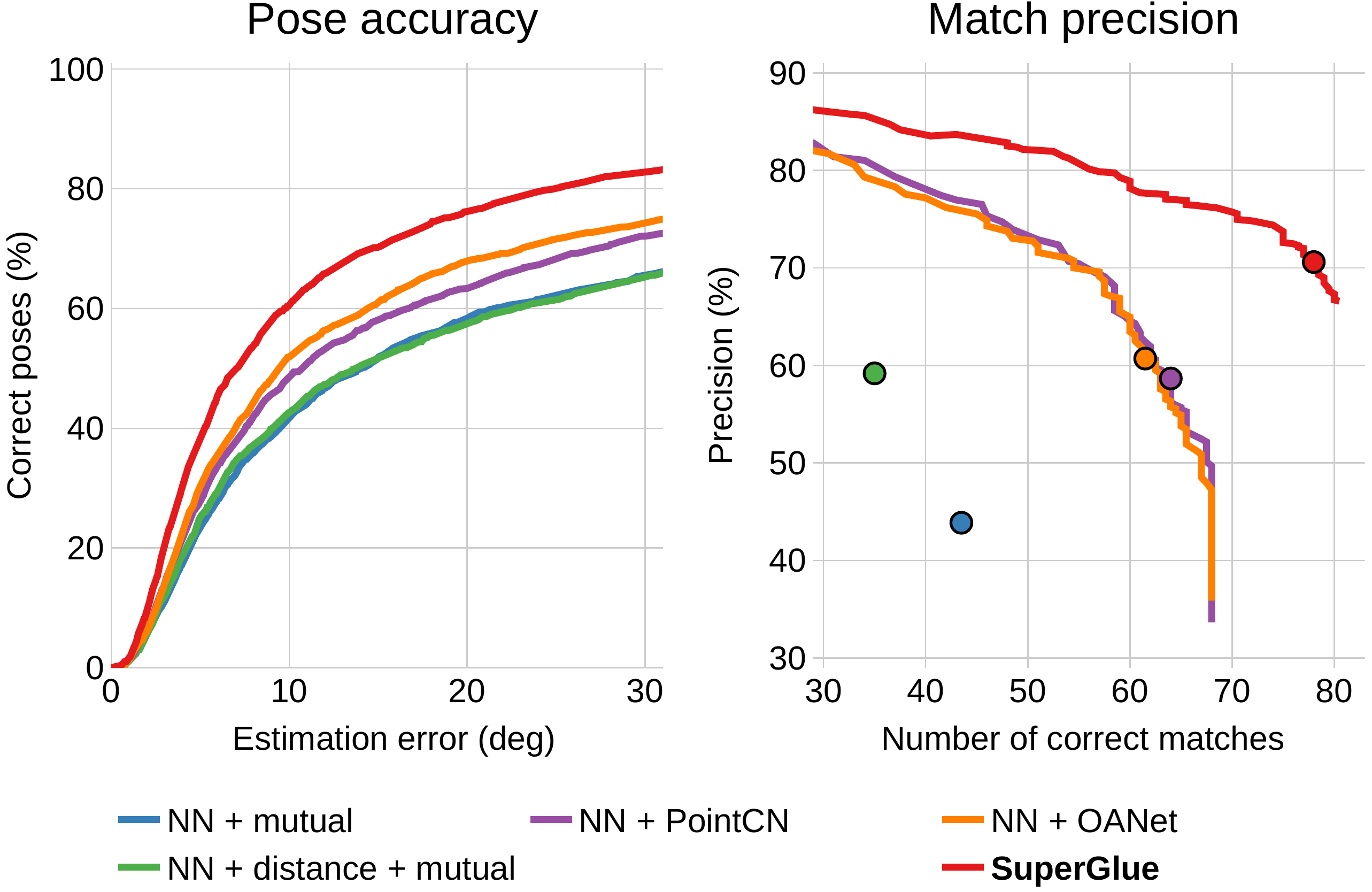}%
    \vspace{-.1cm}%
    \caption{\textbf{Details of the ScanNet evaluation.}
    Poses estimated with SuperGlue are more accurate at all error thresholds.
    SuperGlue offers the best trade-off between precision and number of correct matches, which are both critical for accurate and robust pose estimation.
    }
    \label{fig:supp-scannet-plots}
\end{figure}

\PAR{ScanNet:} We present more details regarding the results on ScanNet (Section~\ref{sec:indoor}), only analyzing the methods which use SuperPoint local features.  Figure~\ref{fig:supp-scannet-plots} plots the cumulative pose estimation error curve and the trade-off between precision and number of correct matches. We compute the correctness from the reprojection error (using the ground truth depth and a threshold of 10 pixels), and, for keypoints with invalid depth, from the symmetric epipolar error. We obtain curves by varying the confidence thresholds of PointCN, OANet, and SuperGlue. At evaluation, we use the original value 0.5 for the former two, and 0.2 for SuperGlue.

\begin{figure*}[b]
\vspace{-.1cm}
\centering
\def\iwidth{2.65cm}
\begin{minipage}{\linewidth}
\centering
\includegraphics[height=\iwidth]{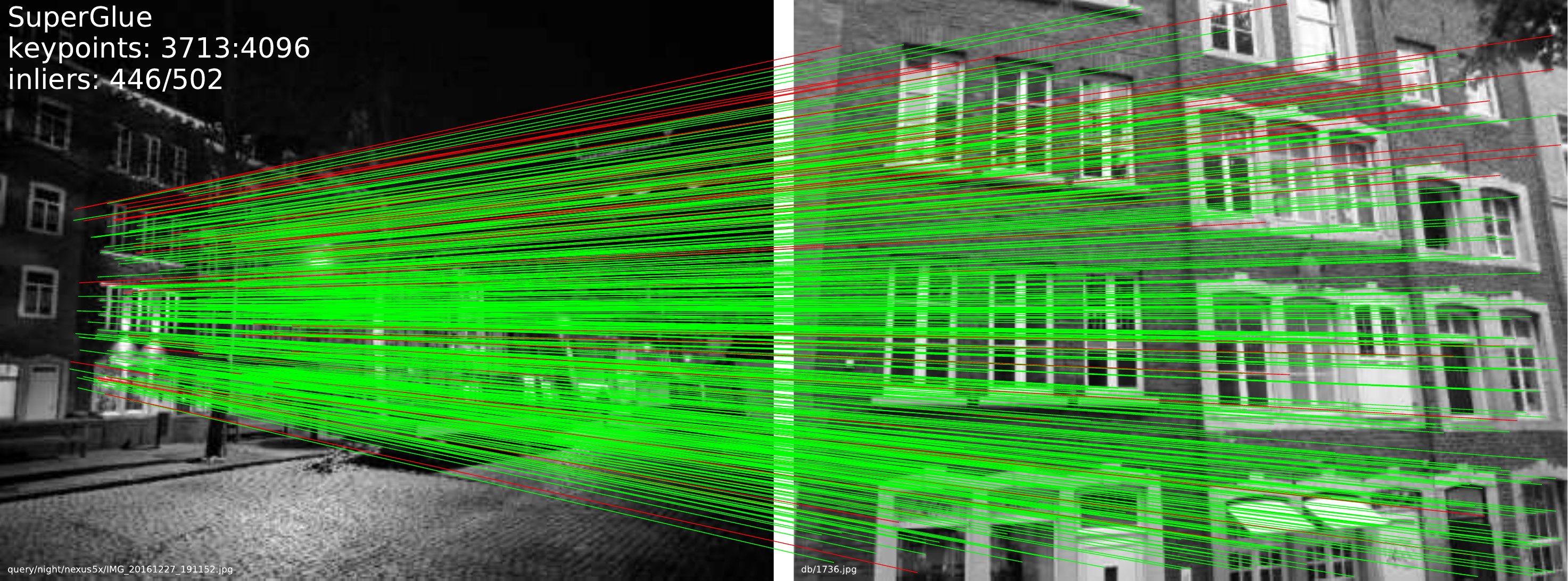}%
\hspace{2mm}%
\includegraphics[height=\iwidth]{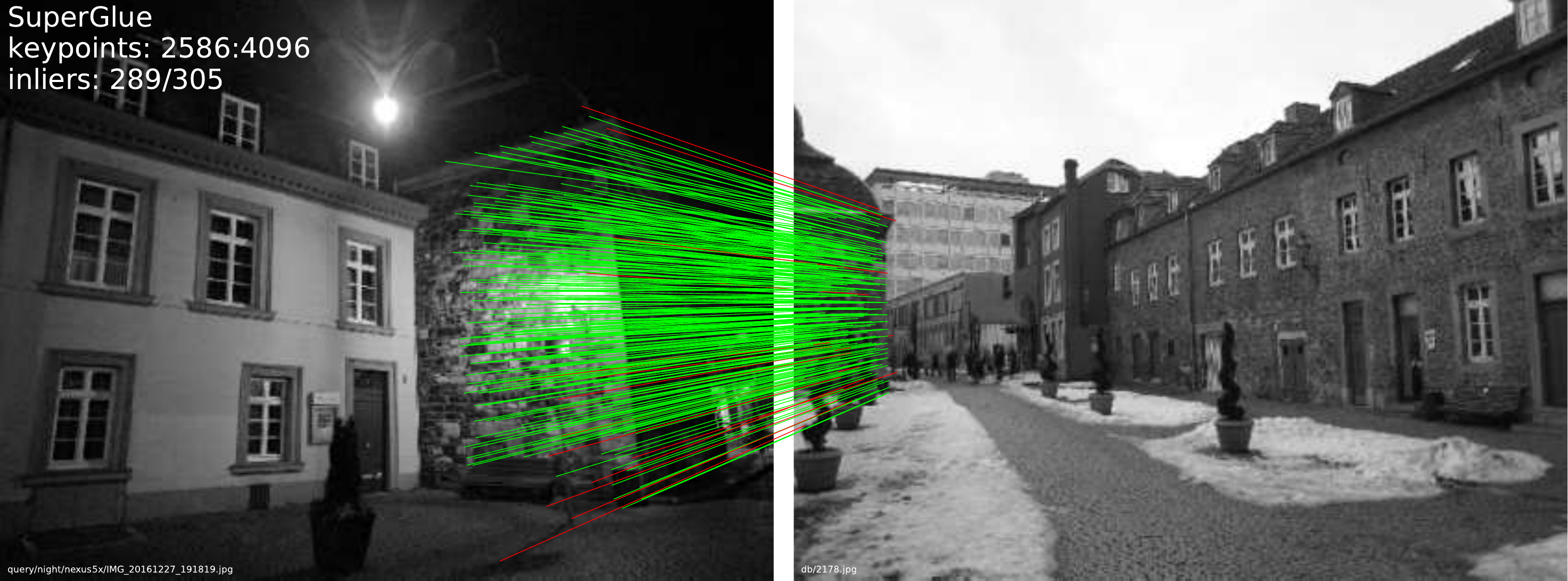}%
\end{minipage}

\vspace{1mm}
\begin{minipage}{\linewidth}
\centering
\includegraphics[height=\iwidth]{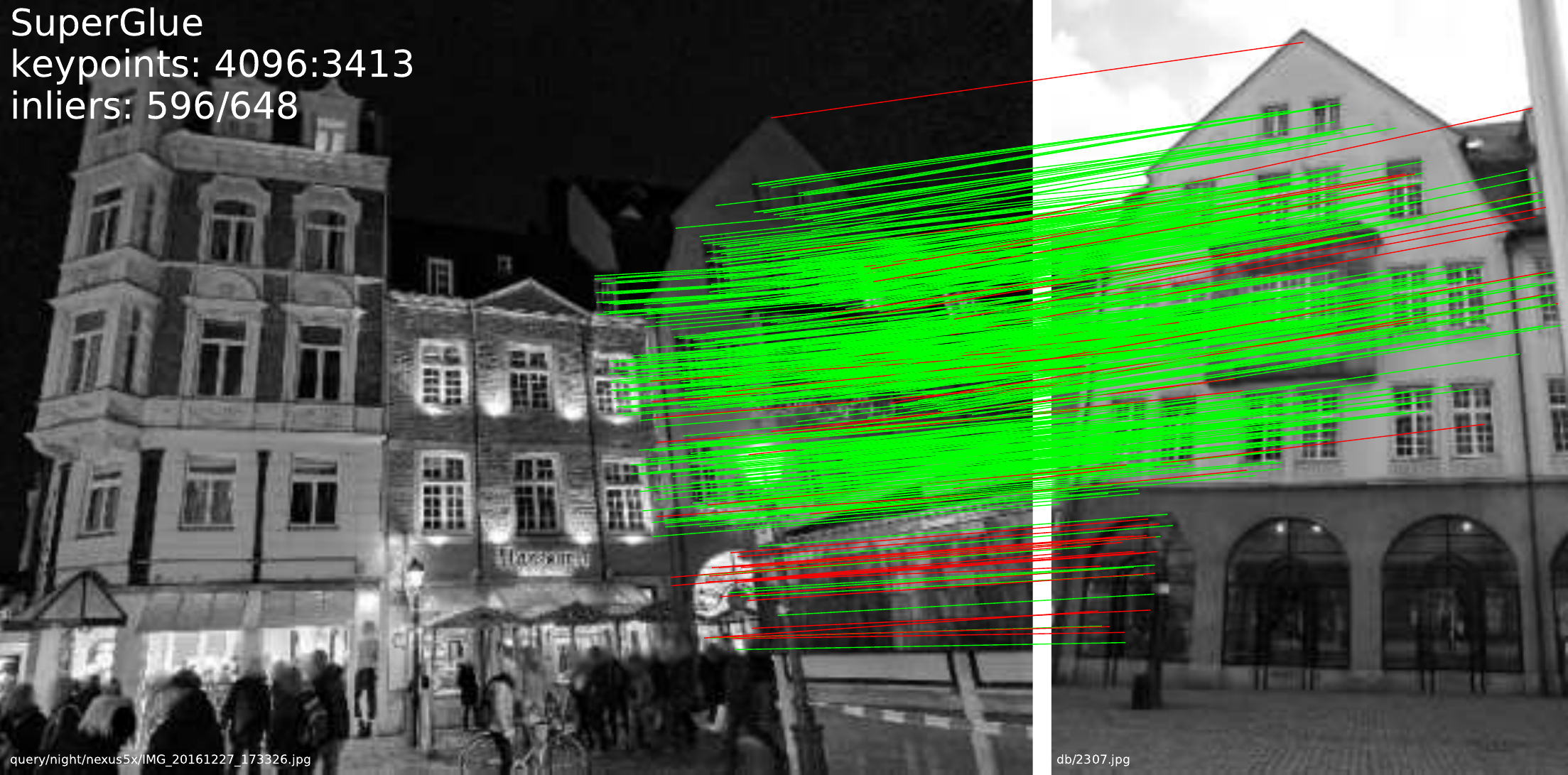}%
\hspace{2mm}%
\includegraphics[height=\iwidth]{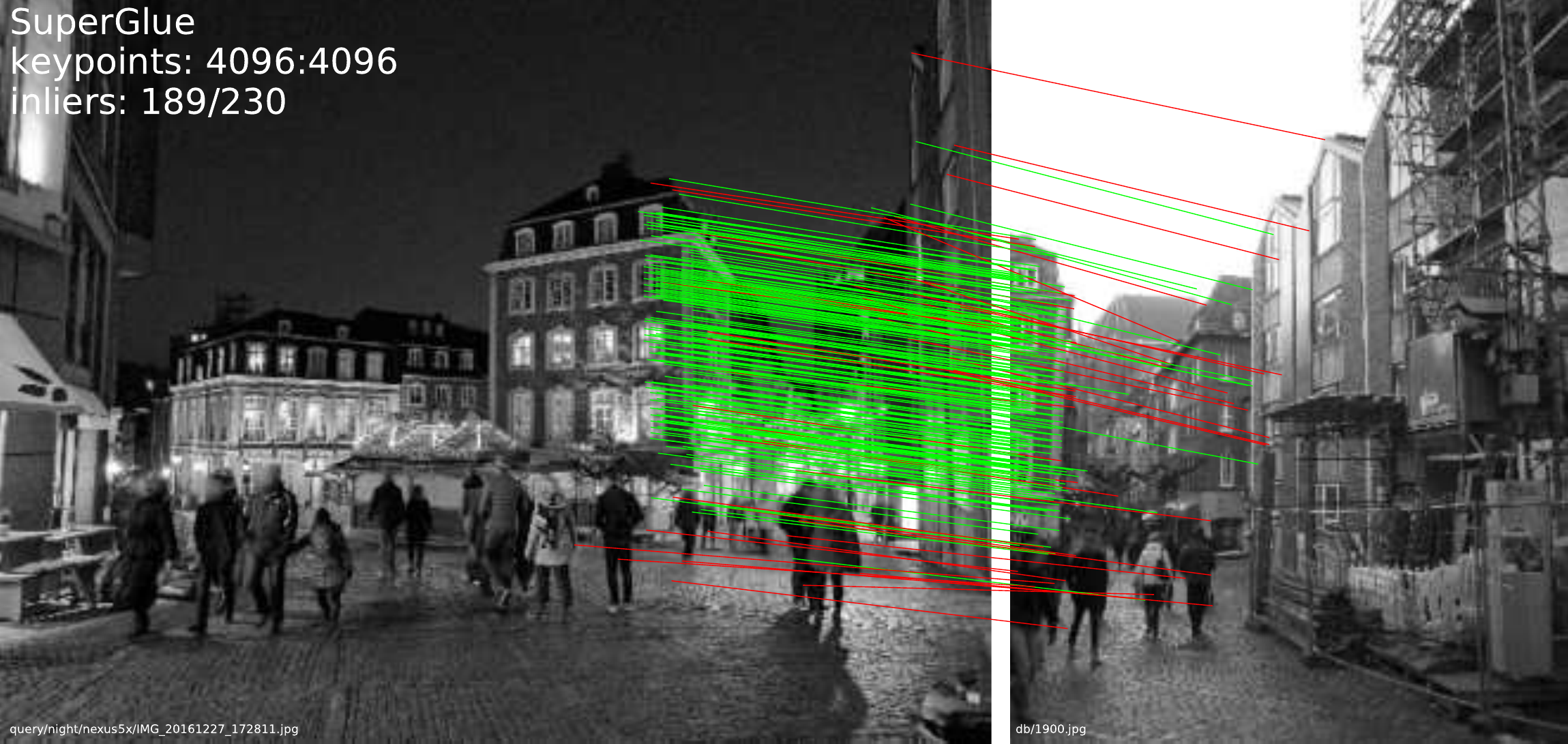}%
\hspace{2mm}%
\includegraphics[height=\iwidth]{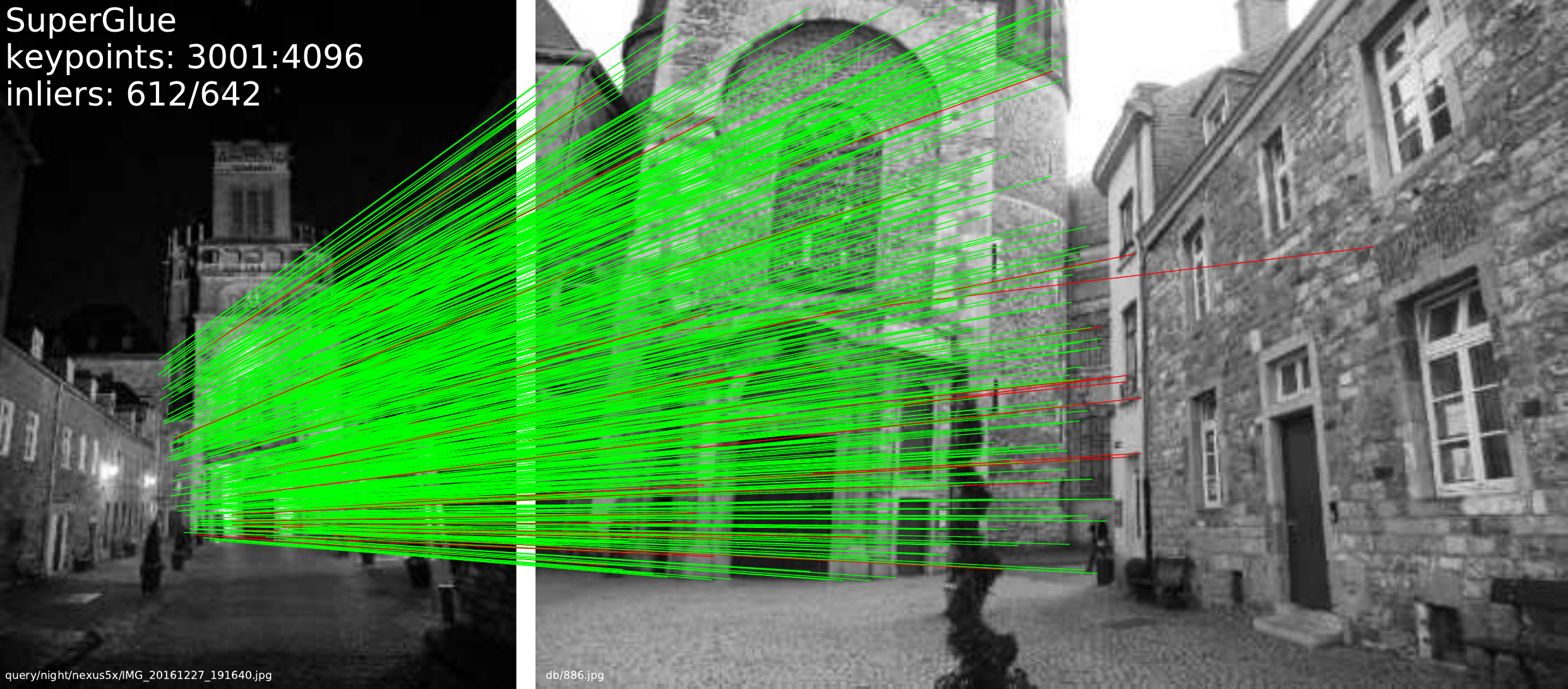}%
\end{minipage}
\vspace{-.1in}
\caption{{\bf Matching challenging day-night pairs with SuperGlue.} We show predicted correspondences between night-time queries and day-time databases images of the Aachen Day-Night dataset. The correspondences are colored as RANSAC inliers in {\color{green}green} or outliers in {\color{red}red}. Although the outdoor training set has few night images, SuperGlue generalizes well to such extreme illumination changes. Moreover, it can accurately match building facades with repeated patterns like windows.
}%
\label{fig:aachen}%
\vspace{-.2cm}
\end{figure*}

\subsection{Outdoor pose estimation}
\vspace{-.15cm}
\label{sec:outdoor-supp}

\PAR{Qualitative results:} Figure~\ref{fig:supp-outdoor-qualitative} shows additional results on the Phototourism test set and the MegaDepth validation set.

\begin{table}[tb]
\vspace{-.3cm}
\centering
\scriptsize{
\setlength\tabcolsep{4.0pt}
\begin{tabular}{llcccccc}
    \toprule
    \multirow{2}{1cm}[-.4em]{Local features} & \multirow{2}{*}[-.4em]{Matcher}
    & \multicolumn{3}{c}{Exact AUC} & \multicolumn{3}{c}{Approx.\ AUC~\cite{zhang2019learning}} \\
    \cmidrule(lr){3-5}
    \cmidrule(lr){6-8}
    && 5\degree & 10\degree & 20\degree & 5\degree & 10\degree & 20\degree\\
    \midrule
    \multirow{1}{*}{ContextDesc}
    & NN + ratio test & 26.09 & 45.52 & 63.07 & 53.00 & 63.13 & 73.00\\
    \midrule
    \multirow{4}{*}{SIFT}
    & NN + ratio test & 24.09 & 40.71 & 58.14 & 45.12 & 55.81 & 67.20\\
    & NN + OANet* & 28.76 & 48.42 & 66.18 & 55.50 & 65.94 & 76.17\\
    & NN + OANet & 29.15 & 48.12 & 65.08 & 55.06 & 64.97 & 74.83\\
    & \b{SuperGlue} & \b{30.49} & \b{51.29} & \b{69.72} & \b{59.25} & \b{70.38} & \b{80.44}\\
    \midrule
    \multirow{3}{*}{SuperPoint}
    & NN + mutual & 16.94 & 30.39 & 45.72 & 35.00 & 43.12 & 54.05\\
    & NN + OANet & 26.82 & 45.04 & 62.17 & 50.94 & 61.41 & 71.77\\
    & \b{SuperGlue} & \b{38.72} & \b{59.13} & \b{75.81} & \b{67.75} & \b{77.41} & \b{85.70}\\
    \bottomrule
\end{tabular}
}
\vspace{-.05in}
\caption{\textbf{Outdoor pose estimation on YFCC100M pairs.} The evaluation is performed on the same image pairs as in OANet~\cite{zhang2019learning} using both their approximate and our exact AUC. SuperGlue consistently improves over the baselines when using either SIFT and SuperPoint.}
\label{tab:supp-yfcc}
\vspace{-.35cm}
\end{table}

\PAR{YFCC100M:} While the PhotoTourism~\cite{imwchallenge2019} and Zhang \etal's~\cite{zhang2019learning} test sets are both based on YFCC100M~\cite{thomee2016yfcc100m}, they use different scenes and pairs. For the sake of comparability, we also evaluate SuperGlue on the same evaluation pairs as in OANet~\cite{zhang2019learning}, using their evaluation metrics.  We include an OANet model (*) retrained on their training set (instead of MegaDepth) using root-normalized SIFT. The results are shown in Table~\ref{tab:supp-yfcc}.

As observed in Section~\ref{sec:outdoor} when evaluating on the PhotoTourism dataset, SuperGlue consistently improves over all baselines for both SIFT and SuperPoint. For SIFT, the improvement over OANet is decreased, which we attribute to the significantly higher overlap and lower difficulty of the pairs used by~\cite{zhang2019learning}. While the approximate AUC tends to overestimate the accuracy, it results in an identical ranking of the methods. The numbers for OANet with SIFT and SuperPoint are consistent with the ones reported in their paper.\looseness=-1

\begin{table}[tb]
\vspace{-.3cm}
\centering
\scriptsize{
\setlength\tabcolsep{3.0pt}
\begin{tabular}{ccccc}
    \toprule
    \multirow{2}{*}[-.4em]{Method} &
    \multicolumn{3}{c}{Correctly localized queries (\%)} &
    \multirow{2}{*}[-.4em]{\# features} \\
    \cmidrule(lr){2-4}
    & .5m/2\degree & 1m/5\degree & 5m/10\degree \\
    \midrule
    R2D2~\cite{revaud2019r2d2} & \b{46.9} & 66.3 & \b{88.8} & 20k \\
    D2-Net~\cite{dusmanu2019d2} & 45.9 & 68.4 & \b{88.8} & 15k \\
    UR2KID~\cite{yang2020ur2kid} & \b{46.9} & 67.3 & \b{88.8} & 15k \\
    SuperPoint+NN+mutual & 43.9 & 59.2 & 76.5 & 4k \\
    \b{SuperPoint+SuperGlue} & 45.9 & \b{70.4} & \b{88.8} & 4k \\
    \bottomrule
\end{tabular}
}
\vspace{-.05in}
\caption{\textbf{Visual localization on Aachen Day-Night.} 
SuperGlue significantly improves the performance of SuperPoint for localization, reaching new state-of-the-art results with comparably fewer keypoints.
}
\label{tab:aachen}
\vspace{-.3cm}
\end{table}

\section{SuperGlue for visual localization}
\PAR{Visual localization:} While two-view relative pose estimation is an important fundamental problem, advances in image matching can directly benefit practical tasks like visual localization~\cite{sattler2018benchmarking, sarlin2019coarse}, which aims at estimating the absolute pose of a query image with respect to a 3D model. Moreover, real-world localization scenarios exhibit significantly higher scene diversity and more challenging conditions, such as larger viewpoint and illumination changes, than phototourism datasets of popular landmarks.

\PAR{Evaluation:} The Aachen Day-Night benchmark~\cite{sattler2012image, sattler2018benchmarking} evaluates local feature matching for day-night localization. We extract up to 4096 keypoints per images with SuperPoint, match them with SuperGlue, triangulate an SfM model from posed day-time database images, and register night-time query images with the 2D-2D matches and COLMAP~\cite{schoenberger2016sfm}. The evaluation server\footnote{\url{https://www.visuallocalization.net/}} computes the percentage of queries localized within several distance and orientation thresholds. As reported in Table~\ref{tab:aachen}, SuperPoint+SuperGlue performs similarly or better than all existing approaches despite using significantly fewer keypoints. Figure~\ref{fig:aachen} shows challenging day-night image pairs.

\section{Timing and model parameters}
\label{sec:supp-timings}

\PAR{Timing:} We measure the run-time of SuperGlue and its two major blocks, the Graph Neural Network and the Optimal Matching Layer, for different numbers of keypoints per image. The measurements are performed on an NVIDIA GeForce GTX 1080 GPU across 500 runs. See Figure~\ref{fig:timings}.

\PAR{Model Parameters:}  The Keypoint Encoder MLP has 5 layers, mapping positions to dimensions of size $(32,64,128,256,D)$, yielding 100k parameters. Each layer has the three projection matrices, and an extra ${\bf W}^O$ to deal with the multi-head output. The message update MLP has 2 layers and maps to dimensions $(2D,D)$. Both MLPs use BatchNorm and ReLUs. Each layer has 0.66M parameters. SuperGlue has 18 layers, with a total of 12M parameters.

\begin{figure}[t]
\centering
{\footnotesize\fontfamily{phv}\selectfont SuperGlue inference time}
\includegraphics[width=1\linewidth, trim=0cm 0cm 0cm 0cm, clip]{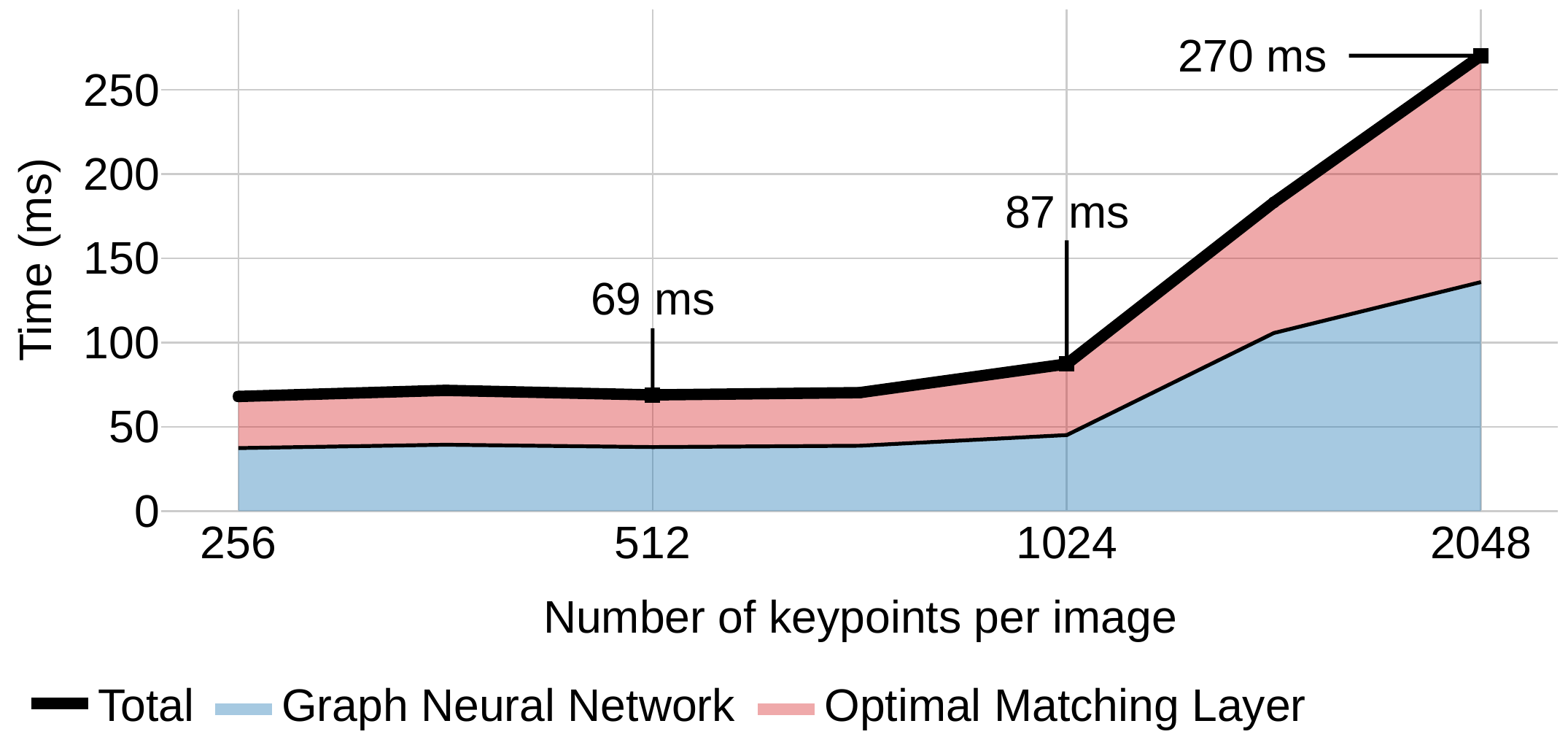}%
\vspace{-1mm}
\caption{\textbf{SuperGlue detailed inference time.} SuperGlue's two main blocks, the Graph Neural Network and the Optimal Matching Layer, have similar computational costs. For 512 and 1024 keypoints per image, SuperGlue runs at 14.5 and 11.5 FPS, respectively.}%
\label{fig:timings}%
\vspace{-3mm}
\end{figure}

\section{Analyzing attention}
\label{sec:sup:attention}

\PAR{Quantitative analysis:} We compute the spatial extent of the attention weights -- the \emph{attention span} -- for all layers and all keypoints. The self-attention span corresponds to the distance in pixel space between one keypoint $i$ and all the others $j$, weighted by the attention weight $\alpha_{ij}$, and averaged for all queries. The cross-attention span corresponds to the average distance between the final predicted match and all the attended keypoints $j$. We average the spans over 100 ScanNet pairs and plot in Figure~\ref{fig:supp-attention-span} the minimum across all heads for each layer, with 95\% confidence intervals.

The spans of both self- and cross-attention tend to decrease throughout the layers, by more than a factor of 10 between the first and the last layer. SuperGlue initially attends to keypoints covering a large area of the image, and later focuses on specific locations -- the self-attention attends to a small neighborhood around the keypoint, while the cross-attention narrows its search to the vicinity of the true match. Intermediate layers have oscillating spans, hinting at a more complex process.

\PAR{Qualitative example:} We analyze the attention patterns of a specific example in Figure~\ref{fig:supp-attention-qualitative}. Our observations are consistent with the attention span trends reported in Figure~\ref{fig:supp-attention-span}.

\begin{figure}[t]
\centering
\includegraphics[width=1\linewidth, trim=0cm 0cm 0cm 0, clip]{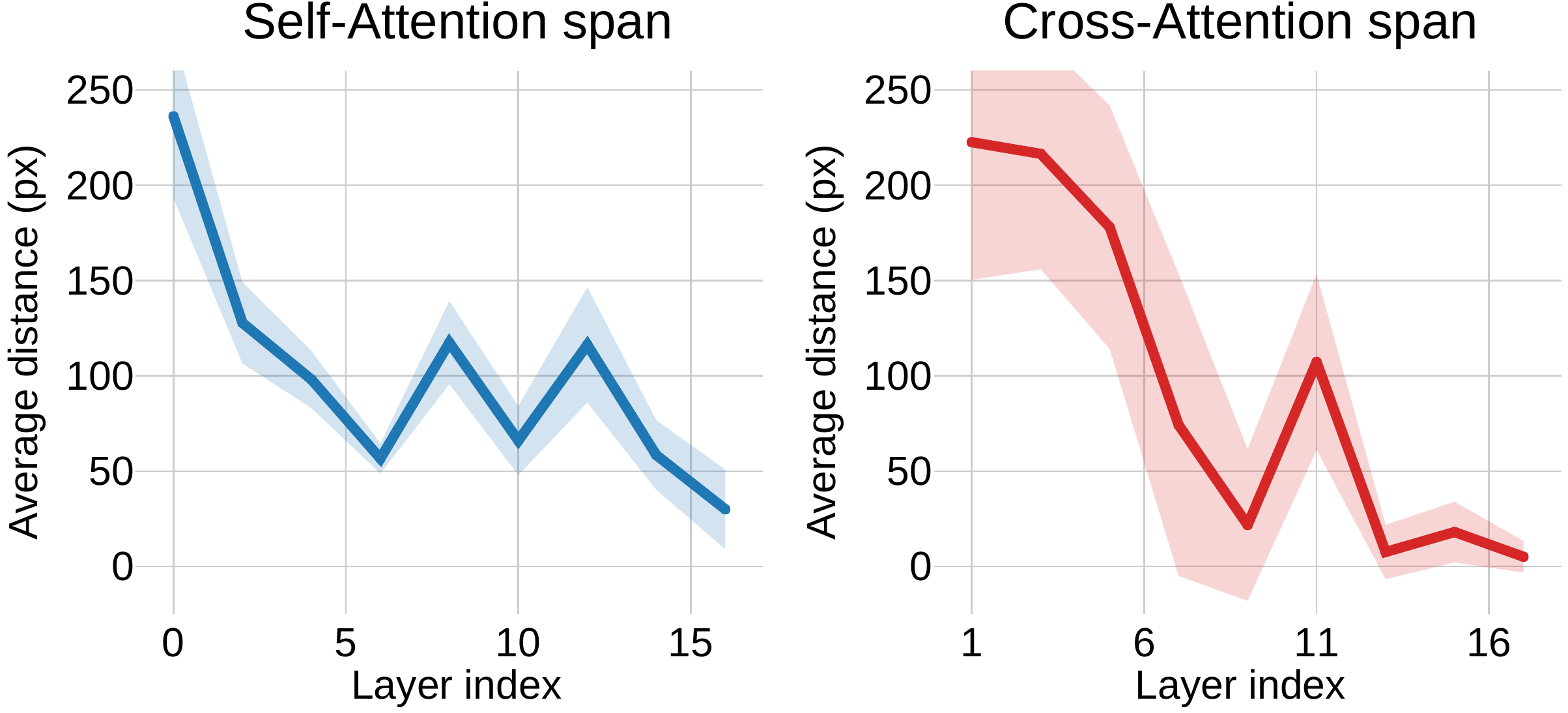}%
\caption{\textbf{Attention spans throughout SuperGlue.}
We plot the attention span, a measure of the attention's spatial dispersion, vs. layer index. For both types of attention, the span tends to decrease deeper in the network as SuperGlue focuses on specific locations. See an example in Figure~\ref{fig:supp-attention-qualitative}.
}%
\label{fig:supp-attention-span}%
\end{figure}

\section{Experimental details}
\label{sec:details-supp}
In this section, we provide details on the training and evaluation of SuperGlue. The trained models and the evaluation code and image pairs are publicly available at {\footnotesize \href{https://github.com/magicleap/SuperGluePretrainedNetwork}{\texttt{github.com/magicleap/SuperGluePretrainedNetwork}}}.

\PAR{Choice of indoor dataset:} Previous works on inlier classification~\cite{moo2018learning, zhang2019learning, brachmann2019neural} evaluate indoor pose estimation on the SUN3D dataset~\cite{xiao2013sun3d}. Camera poses in SUN3D are estimated from SIFT-based sparse SfM, while ScanNet leverages RGB-D fusion and optimization~\cite{dai2017bundlefusion}, resulting in significantly more accurate poses. This makes ScanNet more suitable for generating accurate correspondence labels and evaluating pose estimation. We additionally noticed that the SUN3D image pairs used by Zhang~\etal~\cite{zhang2019learning} have generally small baseline and rotation angle. This makes the essential matrix estimation degenerate~\cite{hartley2003multiple} and the angular translation error ill-defined. In contrast, our ScanNet wide-baseline pairs have significantly more diversity in baselines and rotation, and thus do not suffer from the aforementioned issues.\looseness=-1

\PAR{Homography estimation -- Section~\ref{sec:homography}:} The test set contains 1024 pairs of 640$\times$480 images. Homographies are generated by applying random perspective, scaling, rotation, and translation to the original full-sized images, to avoid bordering artifacts. We evaluate with the 512 top-scoring keypoints detected by SuperPoint with a Non-Maximum Suppression (NMS) radius of 4 pixels. Correspondences are deemed correct if they have a reprojection error lower than 3 pixels. We use the OpenCV function \verb|findHomography| with 3000 iterations and a RANSAC inlier threshold of 3 pixels.

\PAR{Indoor pose estimation -- Section~\ref{sec:indoor}:} The overlap score between two images $A$ and $B$ is the average ratio of pixels in $A$ that are visible in $B$ (and vice versa), after accounting for missing depth values and occlusion (by checking for consistency in the depth). We train and evaluate with pairs that have an overlap score in $[0.4, 0.8]$. For training, we sample at each epoch 200 pairs per scene, similarly as in~\cite{dusmanu2019d2}. The test set is generated by subsampling the sequences by 15 and subsequently randomly sampling 15 pairs for each of the 300 sequences. We resize all ScanNet images and depth maps to 640$\times$480. We detect up to 1024 SuperPoint keypoints (using the publicly available trained model\footnote{\href{https://github.com/magicleap/SuperPointPretrainedNetwork}{\texttt{github.com/magicleap/SuperPointPretrainedNetwork}}} with NMS radius of 4) and 2048 SIFT keypoints (using OpenCV's implementation). Poses are computed by first estimating the essential matrix with OpenCV's \verb|findEssentialMat| and RANSAC with an inlier threshold of 1 pixel divided by the focal length, followed by \verb|recoverPose|. In contrast with previous works~\cite{moo2018learning, zhang2019learning, brachmann2019neural}, we compute a more accurate AUC using explicit integration rather than coarse histograms. The precision (P) is the average ratio of the number of correct matches over the total number of estimated matches. The matching score (MS) is the average ratio of the number of correct matches over the total number of detected keypoints. It does not account for the pair overlap and decreases with the number of covisible keypoints. A match is deemed correct if its epipolar distance is lower than $5\cdot10^{-4}$.

\PAR{Outdoor pose estimation -- Section~\ref{sec:outdoor}:} For training on Megadepth, the overlap score is the ratio of triangulated keypoints that are visible in the two images, as in~\cite{dusmanu2019d2}. We sample pairs with an overlap score in $[0.1, 0.7]$ at each epoch. We evaluate on all 11 scenes of the PhotoTourism dataset and reuse the overlap score based on bounding boxes computed by Ono \etal~\cite{ono2018lf}, with a selection range of $[0.1, 0.4]$. Images are resized so that their longest dimension is equal to 1600 pixels and rotated upright using their EXIF data. We detect 2048 keypoints for both SIFT and SuperPoint (with an NMS radius of 3). The epipolar correctness threshold is here $10^{-4}$. Other evaluation parameters are identical to the ones used for the indoor evaluation.

\PAR{Training of SuperGlue:} For training on homography/indoor/outdoor data, we use the Adam optimizer~\cite{kingma2014adam} with a constant leaning rate of $10^{-4}$ for the first 200k/100k/50k iterations, followed by an exponential decay of 0.999998/0.999992/0.999992 until iteration 900k. When using SuperPoint features, we employ batches with 32/64/16 image pairs and a fixed number of 512/400/1024 keypoints per image. For SIFT features we use 1024 keypoints and 24 pairs. Due to the limited number of training scenes, the outdoor model weights are initialized with the homography model weights. Before the keypoint encoder, the keypoints are normalized by the largest dimension of the image.

Ground truth correspondences $\mathcal M$ and unmatched sets $\mathcal I$ and $\mathcal J$ are generated by first computing the $M \times N$ reprojection matrix between all detected keypoints using the ground truth homography or pose and depth. Correspondences are entries with a reprojection error that is a minimum along both rows and columns, and that is lower than a given threshold: 3, 5, and 3 pixels for homographies, indoor, and outdoor matching respectively. For homographies, unmatched keypoints are simply the ones that do not appear in $\mathcal M$. For indoor and outdoor matching, because of errors in the pose and depth, unmatched keypoints must additionally have a minimum reprojection error larger than 15 and 5 pixels, respectively. This allows us to ignore labels for keypoints whose correspondences are ambiguous, while still providing some supervision through the normalization induced by the Sinkhorn algorithm.

\PAR{Ablation study -- Section~\ref{sec:understanding}:} The ``No Graph Neural Net'' baseline replaces the Graph Neural Network with a single linear projection, but retains the Keypoint Encoder and the Optimal Matching Layer. The ``No cross-attention'' baseline replace all cross-attention layers by self-attention: it has the same number of parameters as the full model, and acts like a Siamese network. The ``No positional encoding'' baseline simply removes the Keypoint Encoder and only uses the visual descriptors as input.

\PAR{End-to-end training -- Section~\ref{sec:understanding}:} Two copies of SuperPoint, for detection and description, are initialized with the original weights. The detection network is frozen and gradients are propagated through the descriptor network only, flowing from SuperGlue - no additional losses are used.

\begin{figure*}[ht!]
\vspace{-2mm}
\centering
\def\iwidth{0.315}
\begin{minipage}{\iwidth\textwidth}
    \centering
    \small{SuperPoint + NN + distance}
\end{minipage}%
\hspace{1mm}%
\begin{minipage}{\iwidth\textwidth}
    \centering
    \small{SuperPoint + NN + OANet}
\end{minipage}%
\hspace{1mm}%
\begin{minipage}{\iwidth\textwidth}
    \centering
    \small{SuperPoint + \b{SuperGlue}}
\end{minipage}%

\begin{minipage}{0.02\textwidth}
\rotatebox[origin=c]{90}{Synthetic}
\end{minipage}%
\hfill{\vline width 1pt}\hfill
\hspace{1mm}%
\begin{minipage}{\iwidth\textwidth}
    \includegraphics[width=\linewidth]{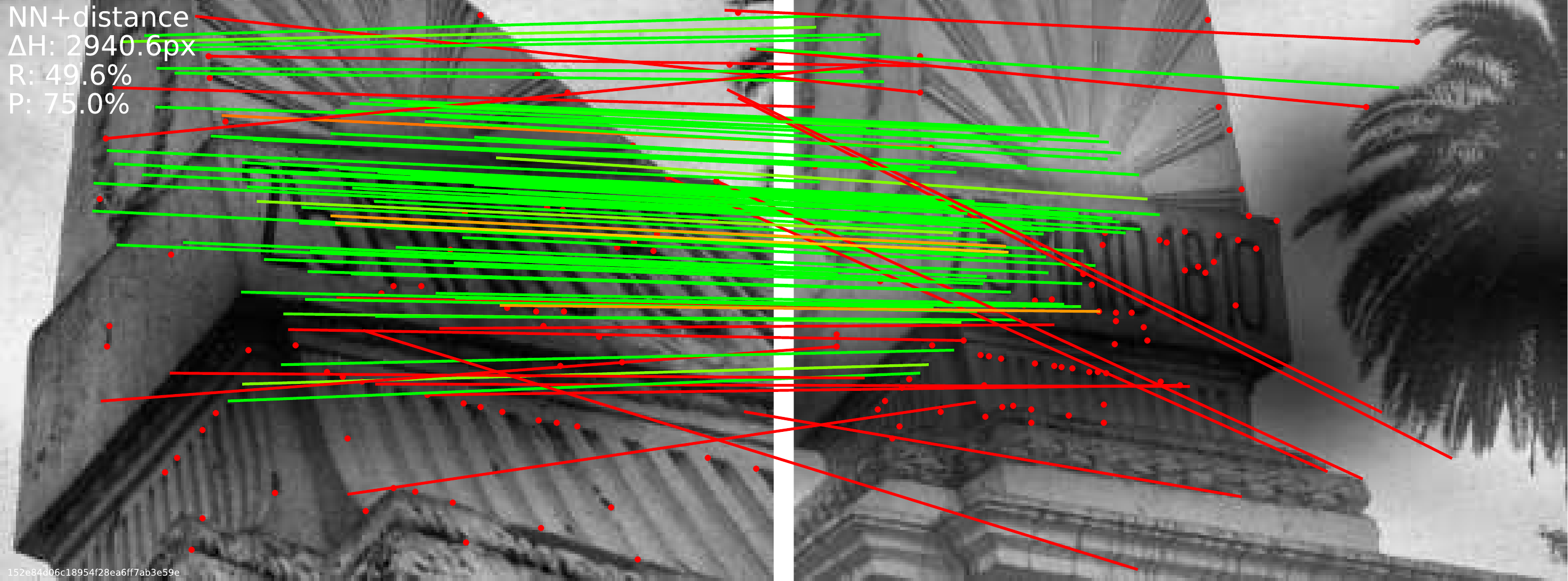}
    
    \vspace{.5mm}
    \includegraphics[width=\linewidth]{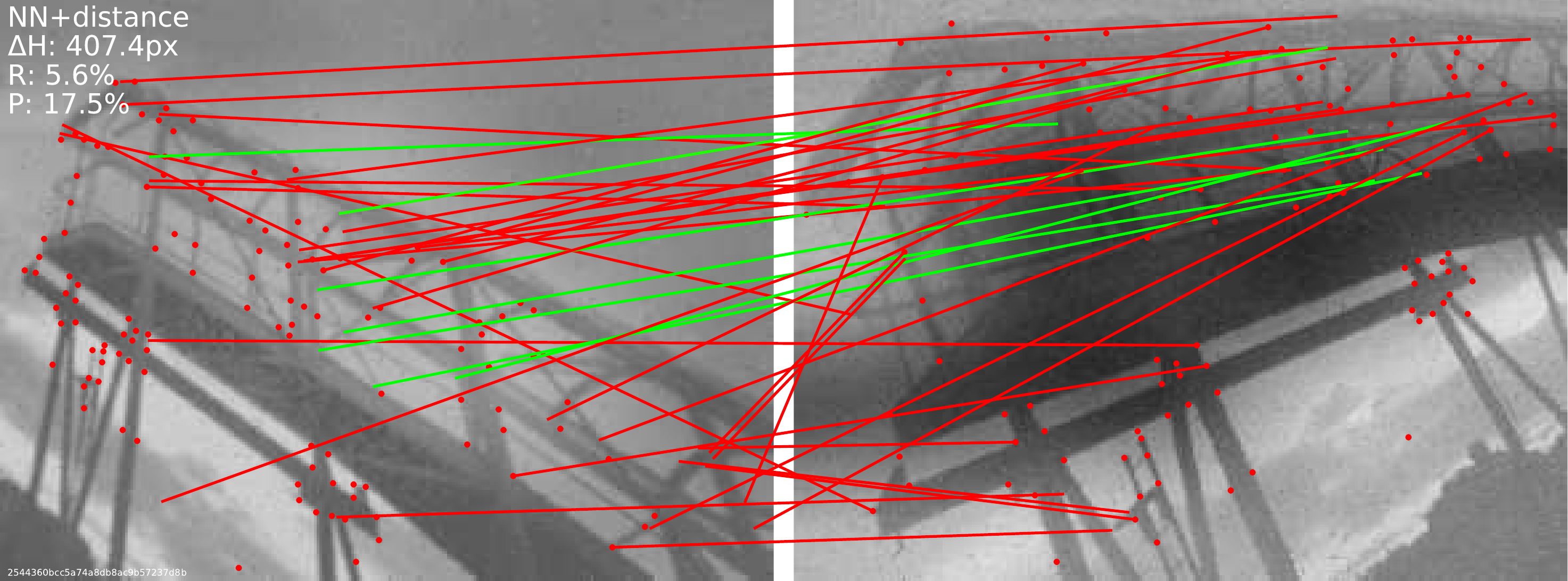}
    
    \vspace{.5mm}
    \includegraphics[width=\linewidth]{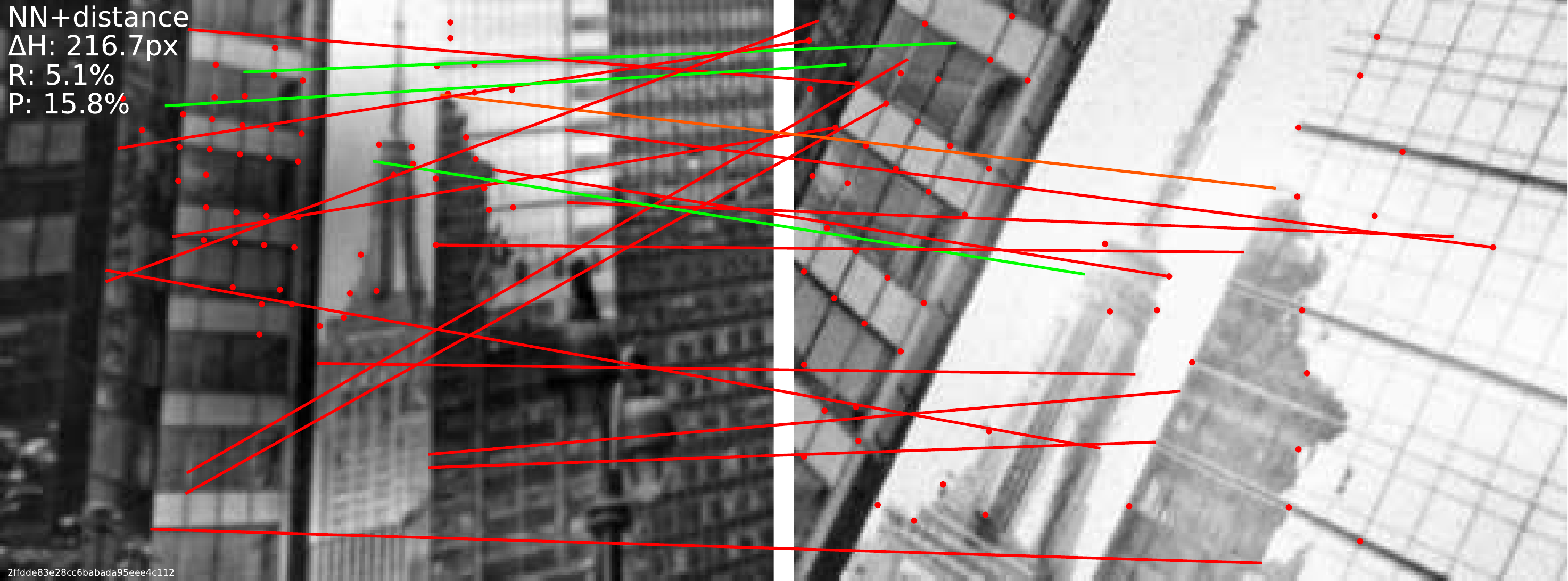}
    
    \vspace{.5mm}
    \includegraphics[width=\linewidth]{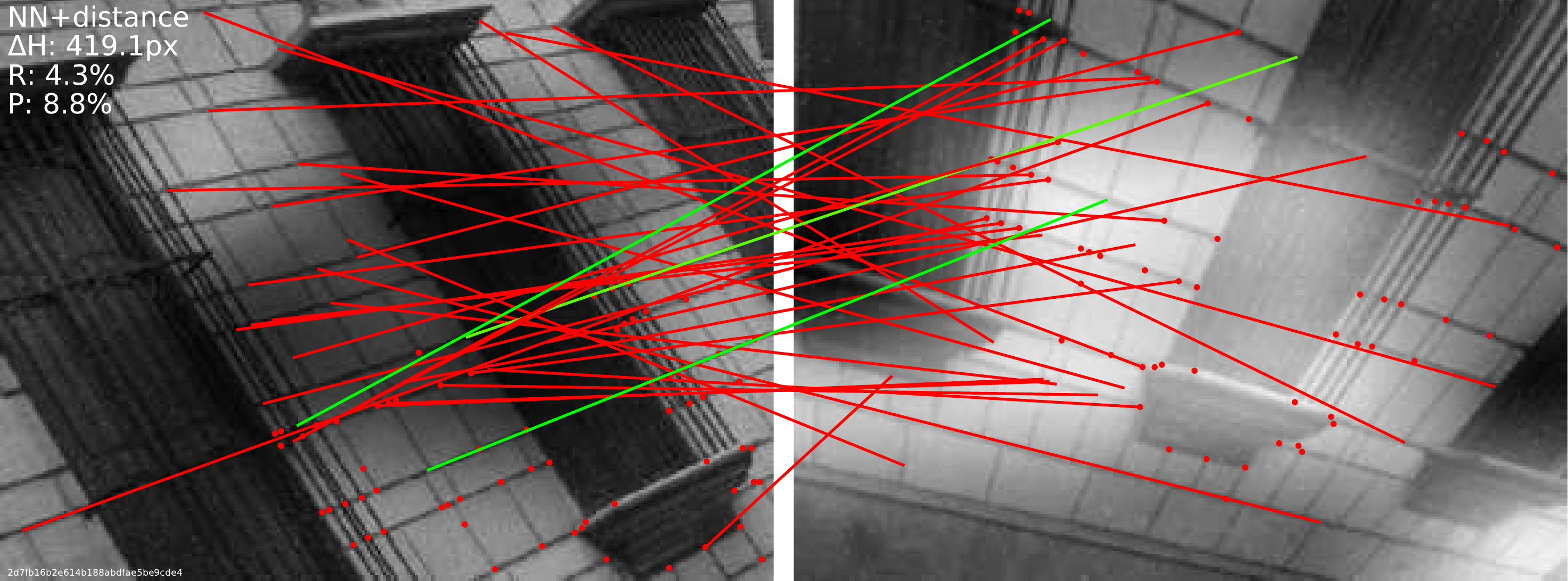}
    
    \vspace{.5mm}
    \includegraphics[width=\linewidth]{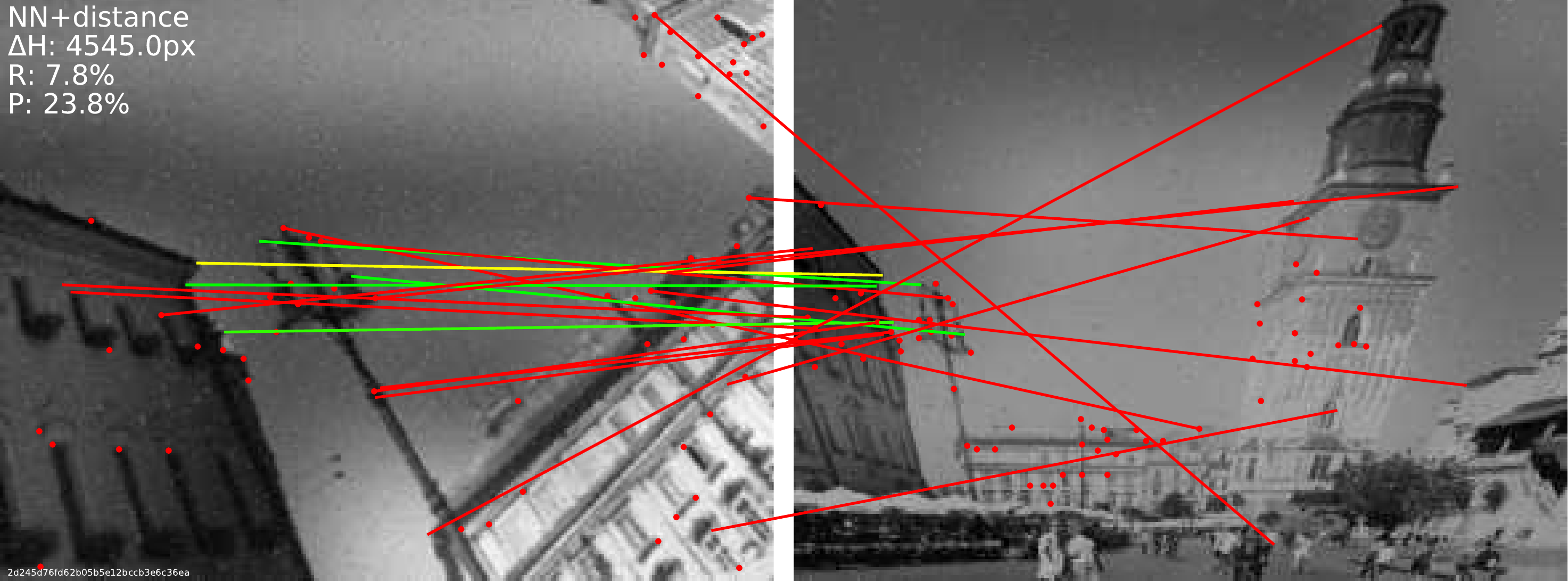}
    
    \vspace{.5mm}
    \includegraphics[width=\linewidth]{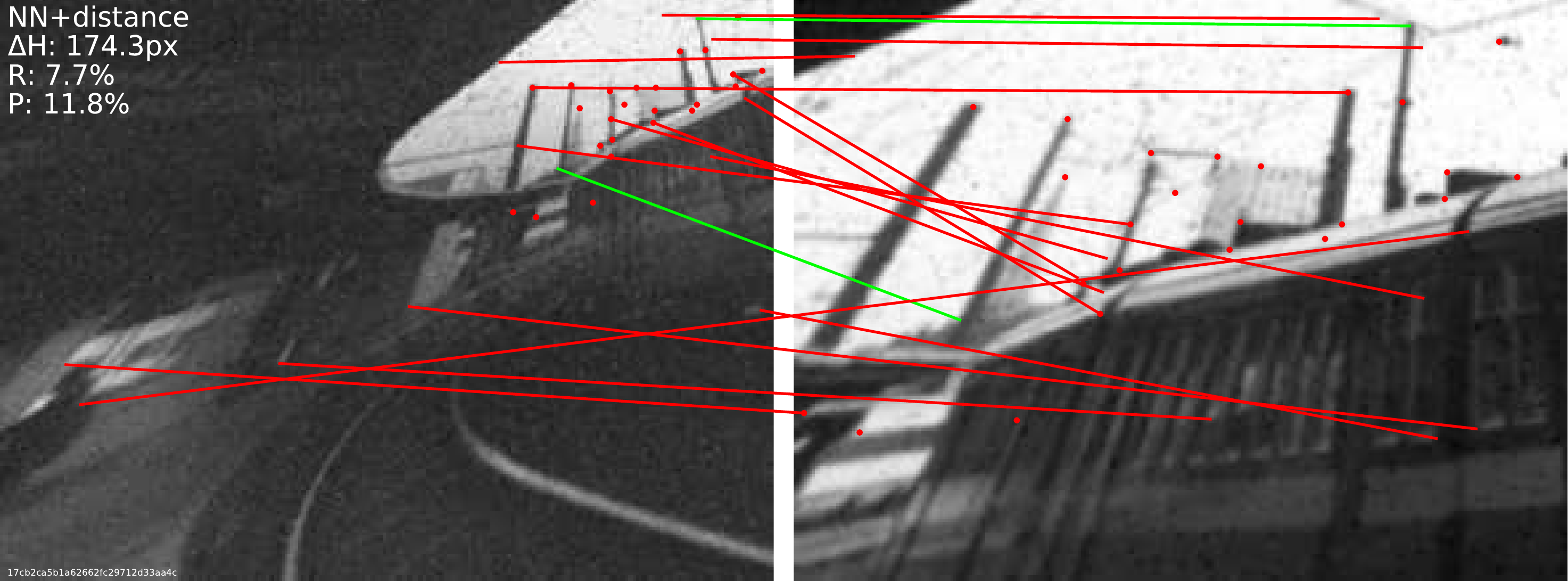}
    
    \vspace{.5mm}
    \includegraphics[width=\linewidth]{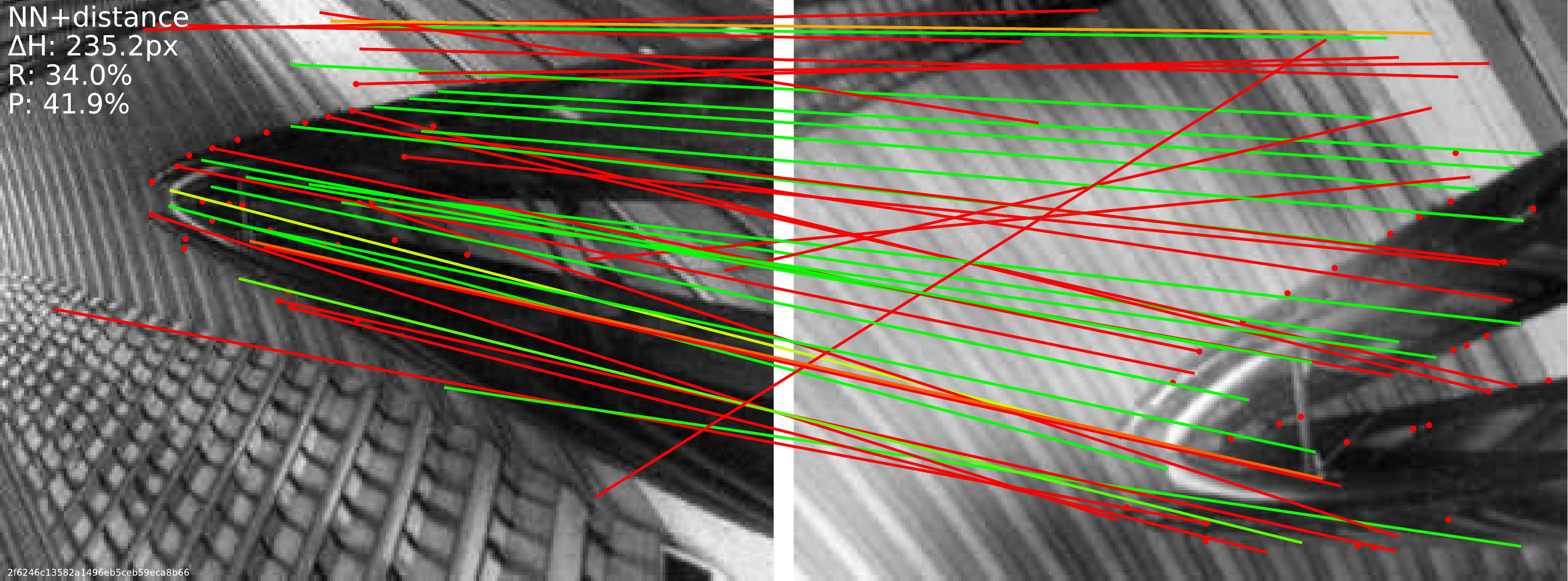}
\end{minipage}%
\hspace{1mm}%
\begin{minipage}{\iwidth\textwidth}
    \includegraphics[width=\linewidth]{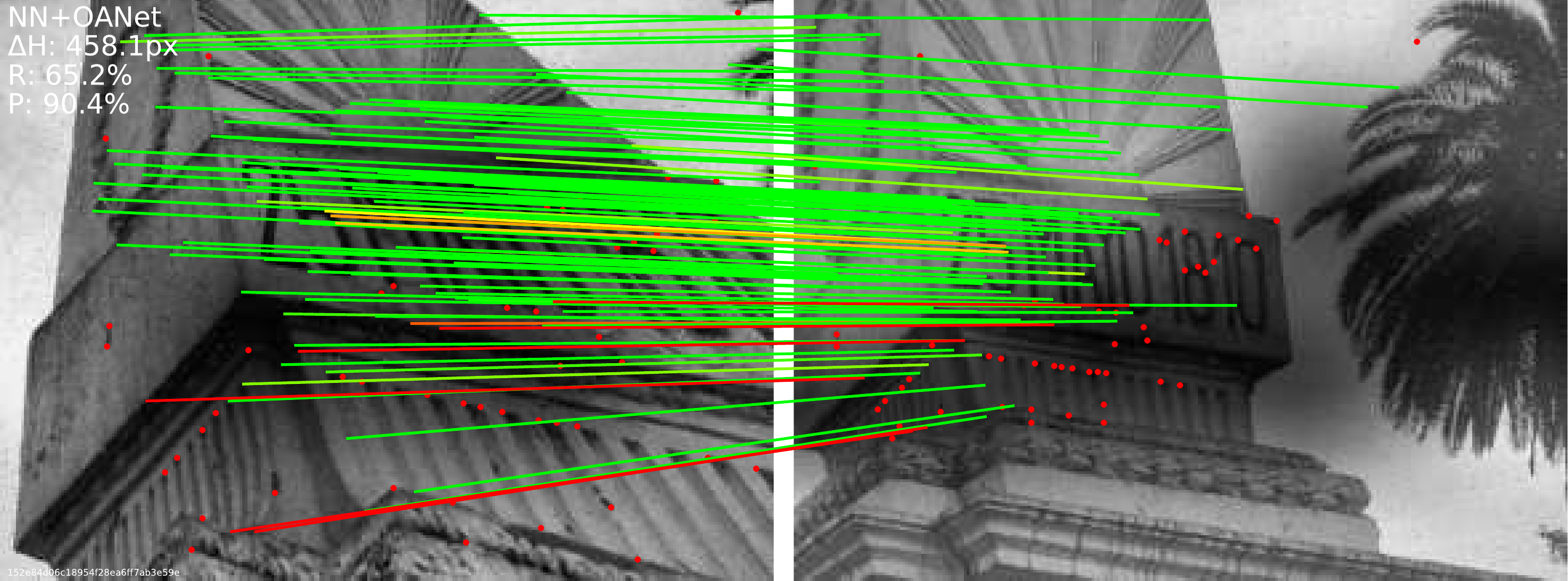}
    
    \vspace{.5mm}
    \includegraphics[width=\linewidth]{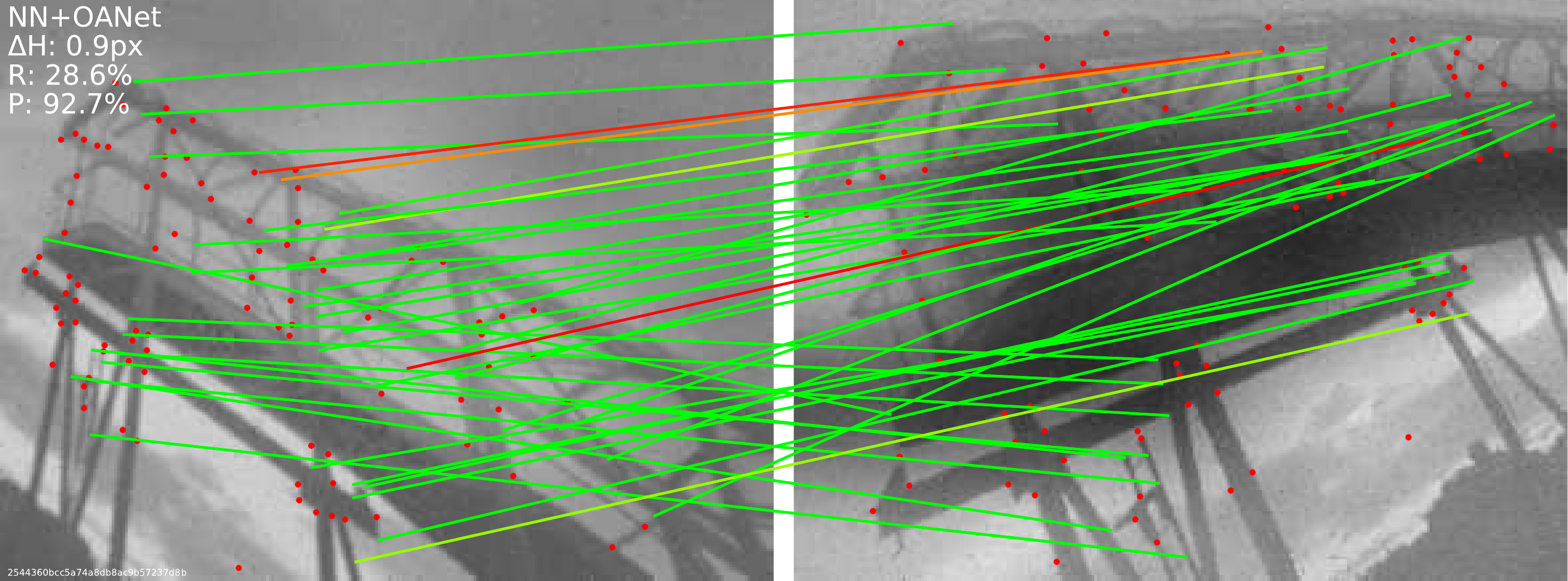}
    
    \vspace{.5mm}
    \includegraphics[width=\linewidth]{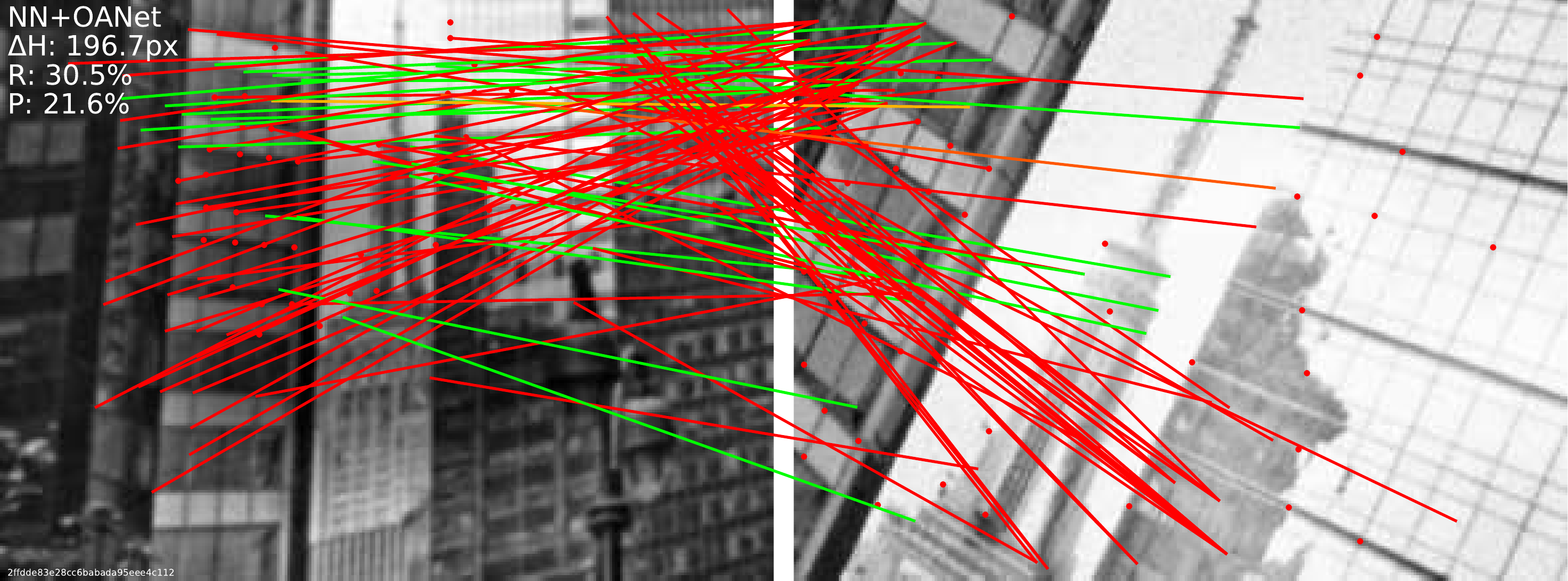}
    
    \vspace{.5mm}
    \includegraphics[width=\linewidth]{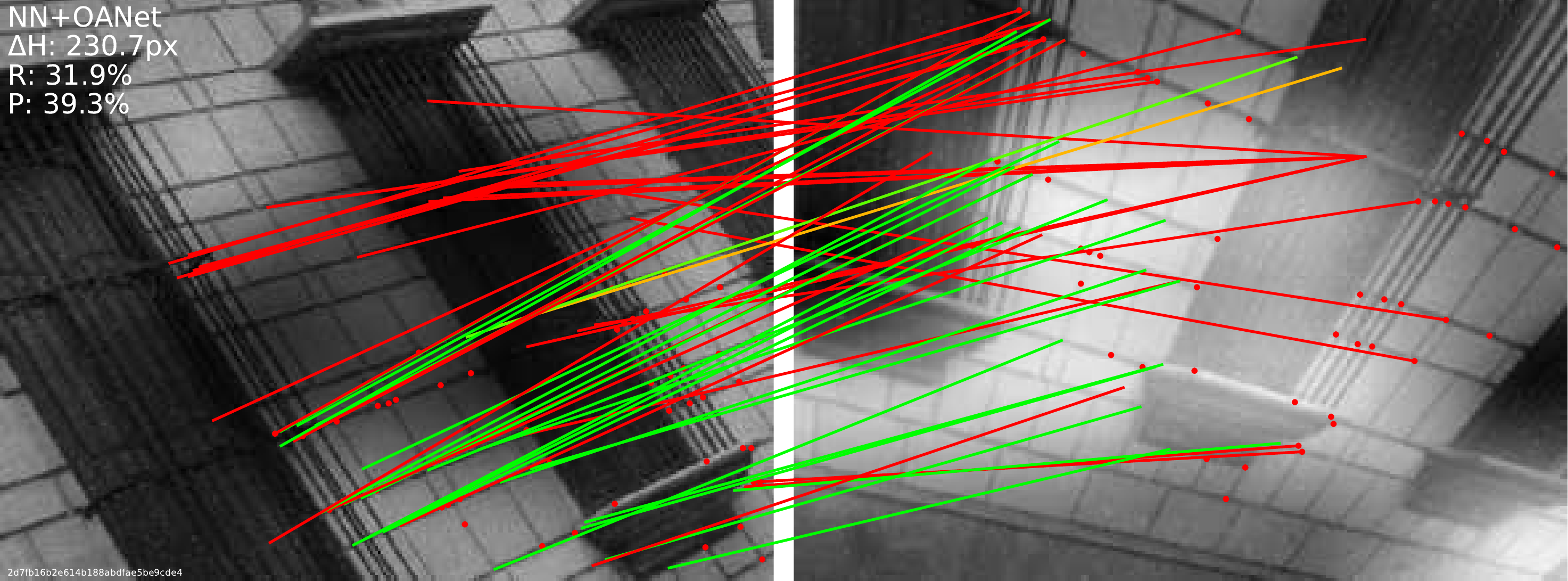}
    
    \vspace{.5mm}
    \includegraphics[width=\linewidth]{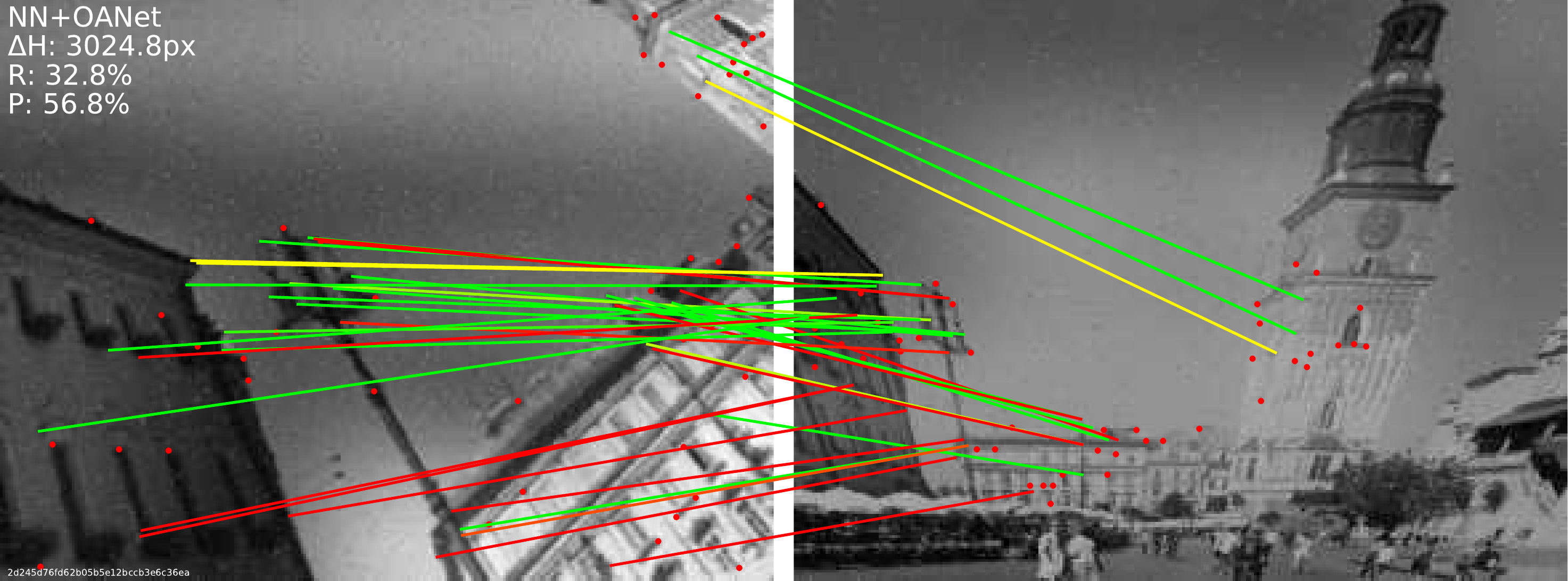}
    
    \vspace{.5mm}
    \includegraphics[width=\linewidth]{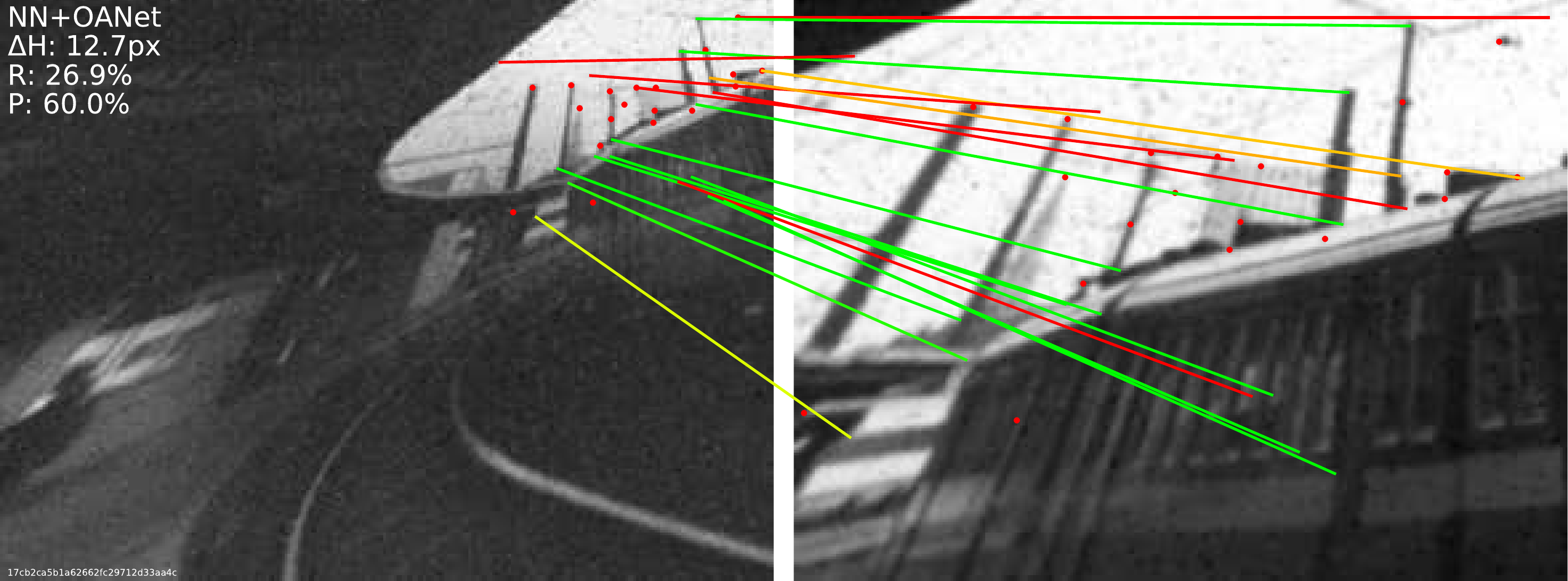}
    
    \vspace{.5mm}
    \includegraphics[width=\linewidth]{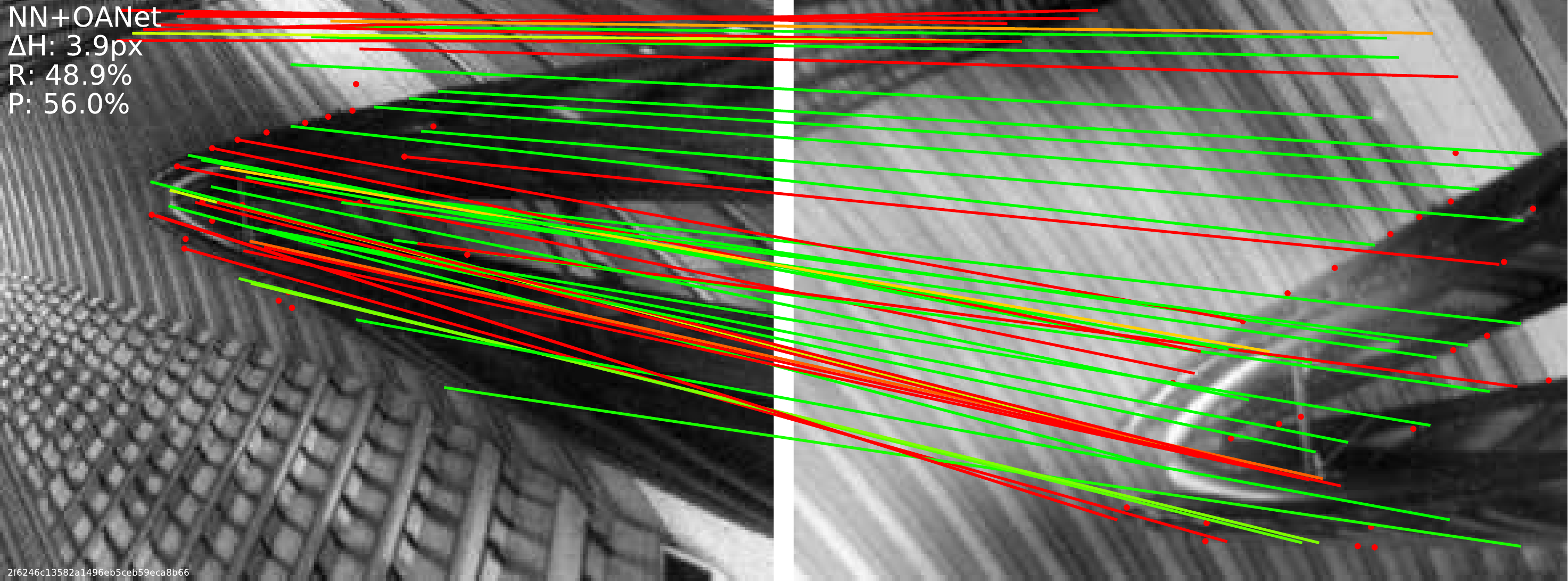}
\end{minipage}%
\hspace{1mm}%
\begin{minipage}{\iwidth\textwidth}
    \includegraphics[width=\linewidth]{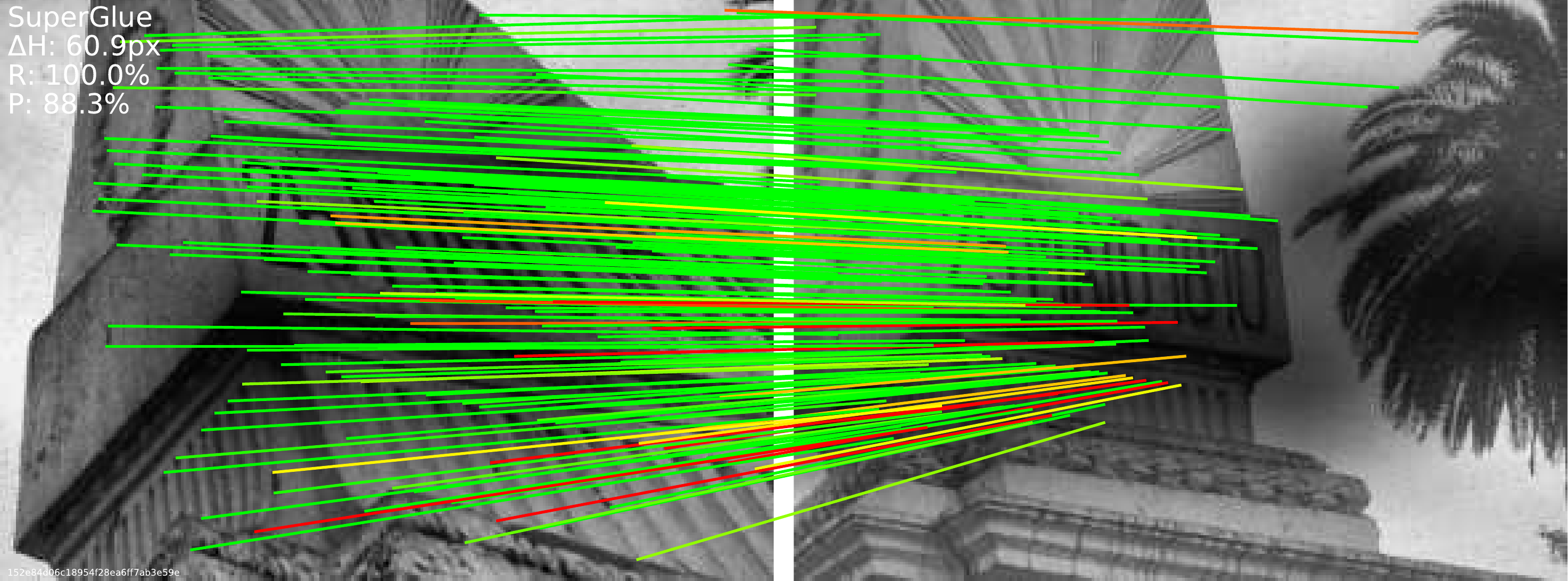}
    
    \vspace{.5mm}
    \includegraphics[width=\linewidth]{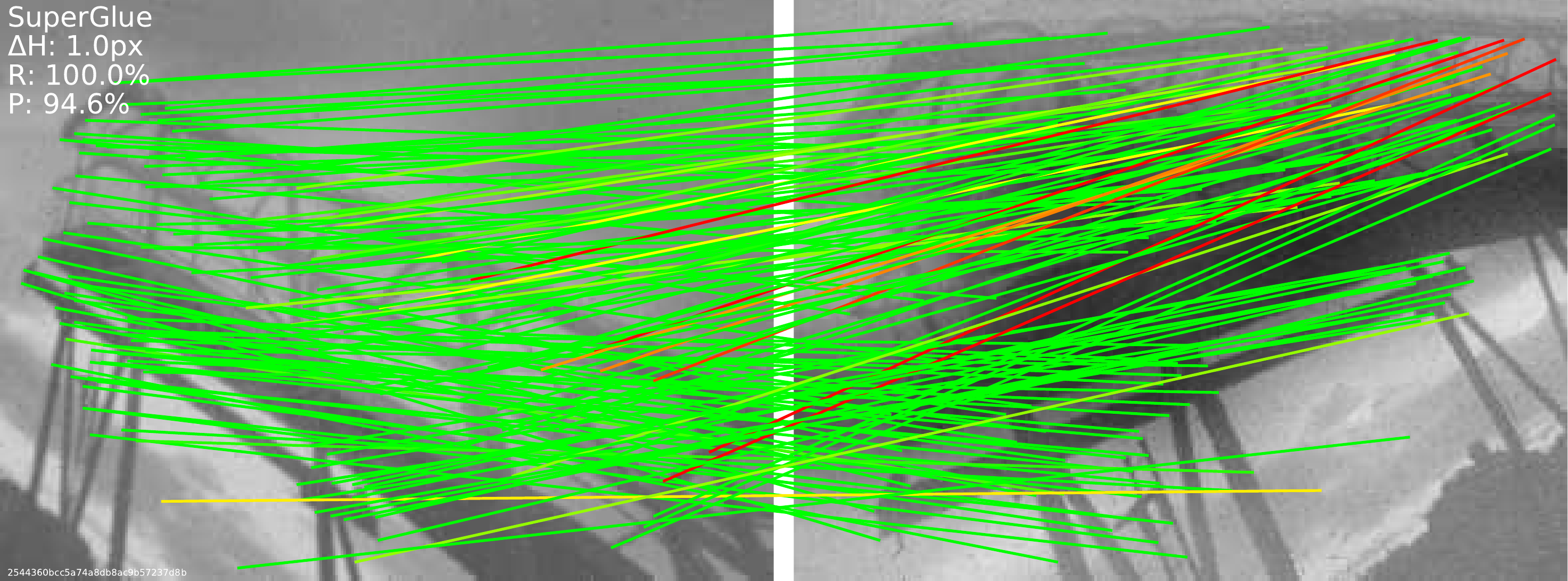}
    
    \vspace{.5mm}
    \includegraphics[width=\linewidth]{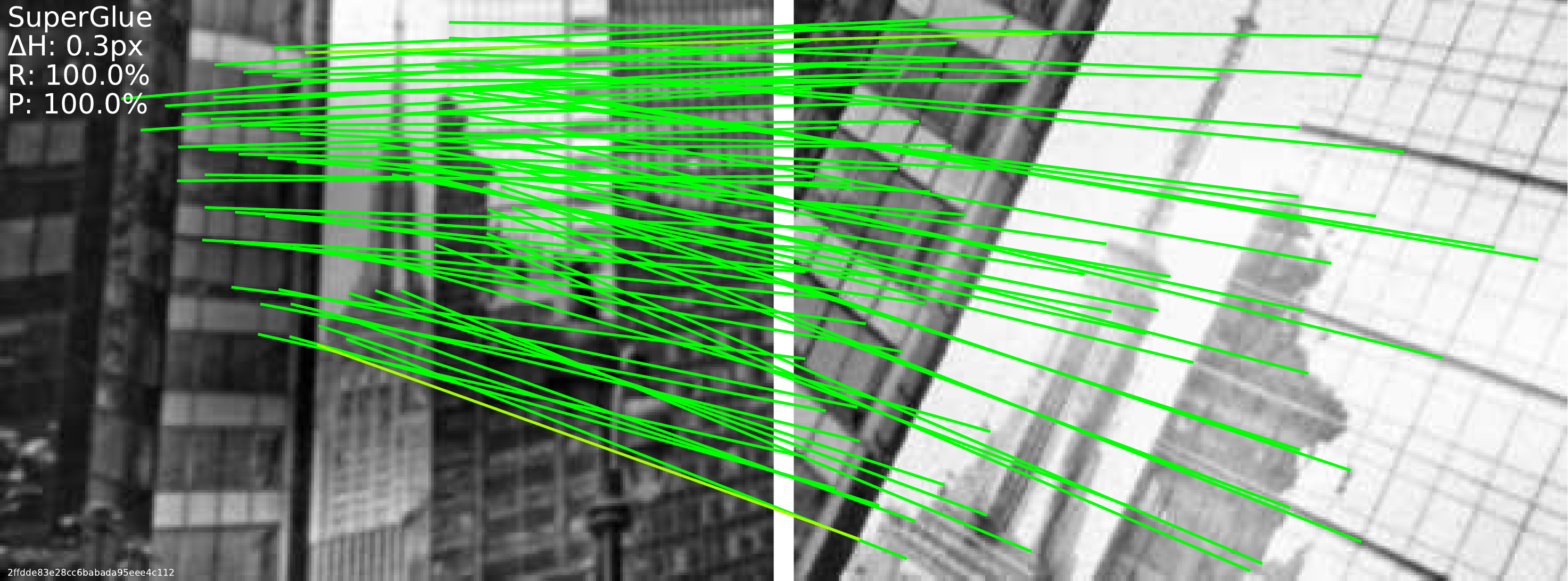}
    
    \vspace{.5mm}
    \includegraphics[width=\linewidth]{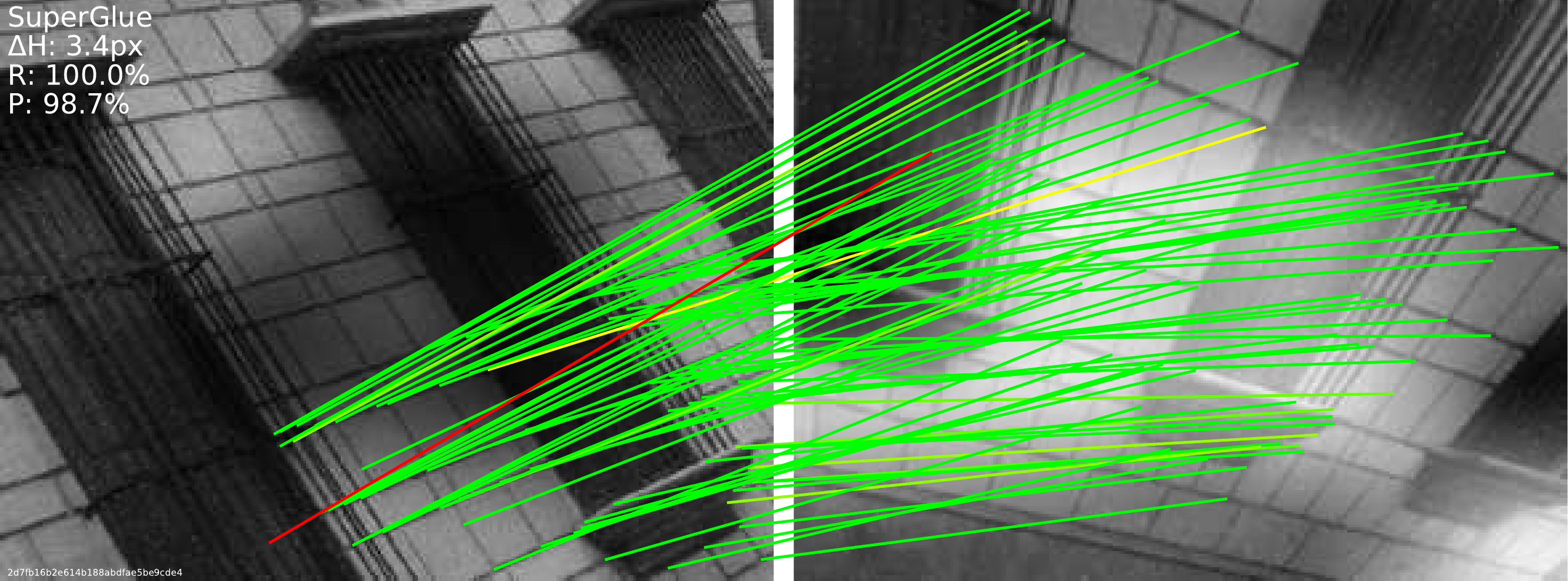}
    
    \vspace{.5mm}
    \includegraphics[width=\linewidth]{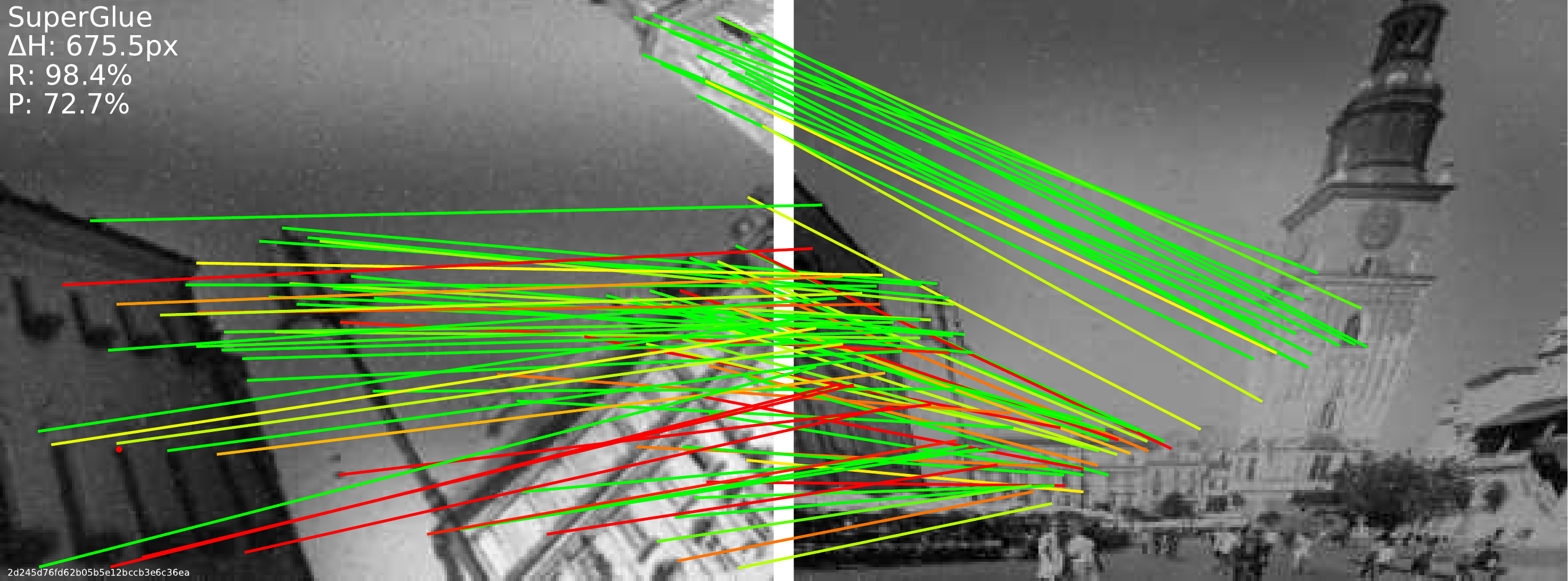}
    
    \vspace{.5mm}
    \includegraphics[width=\linewidth]{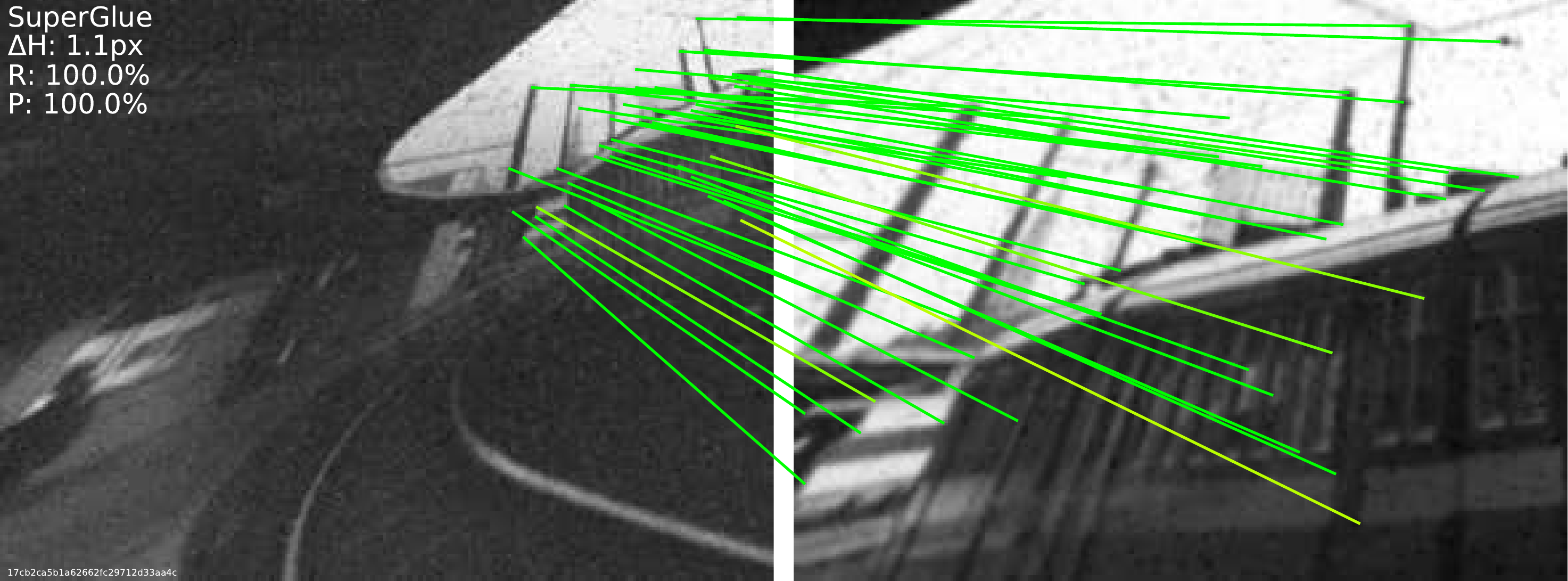}
    
    \vspace{.5mm}
    \includegraphics[width=\linewidth]{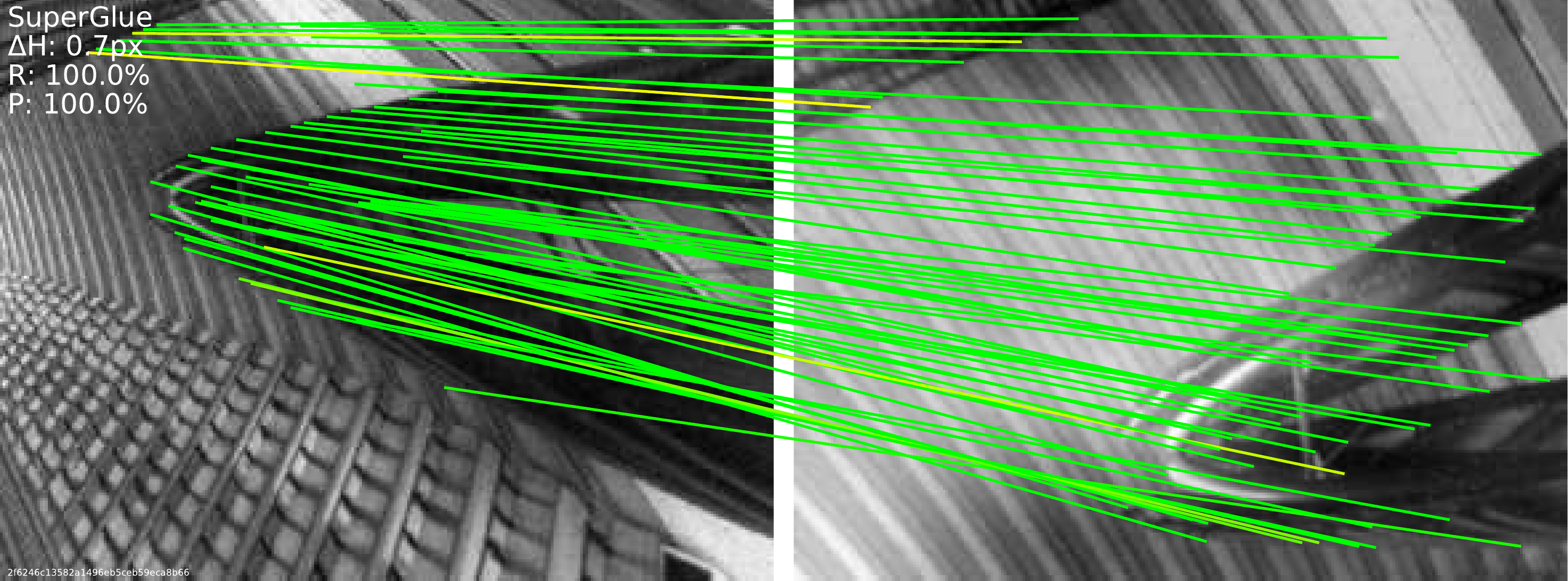}
\end{minipage}

\vspace{1.5mm}
\begin{minipage}{0.02\textwidth}
\rotatebox[origin=c]{90}{HPatches}
\end{minipage}%
\hfill{\vline width 1pt}\hfill
\hspace{1mm}%
\begin{minipage}{\iwidth\textwidth}
    \includegraphics[width=\linewidth]{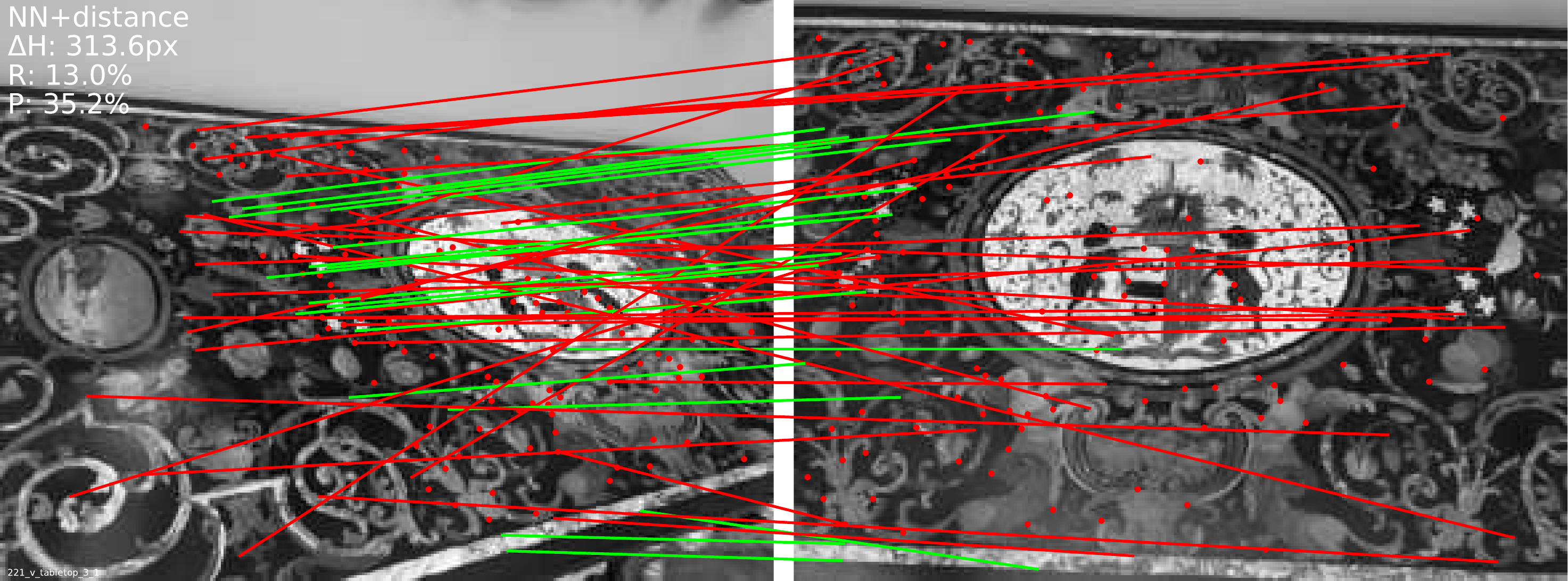}
    
    \vspace{.5mm}
    \includegraphics[width=\linewidth]{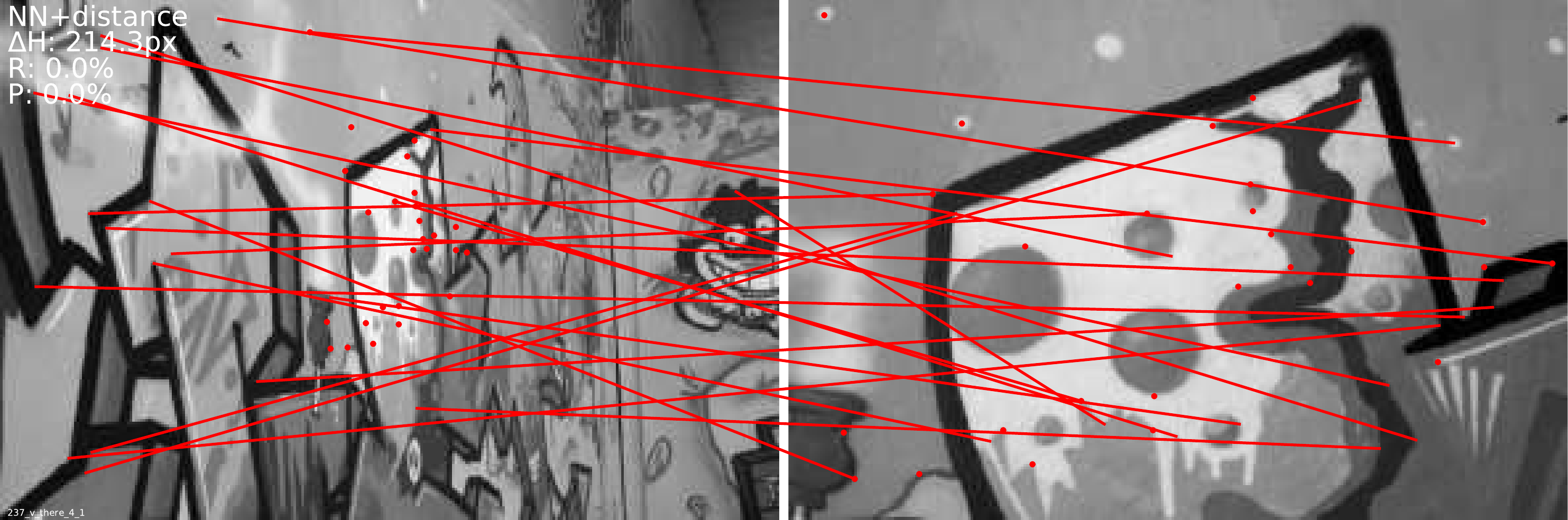}
\end{minipage}%
\hspace{1mm}%
\begin{minipage}{\iwidth\textwidth}
    \includegraphics[width=\linewidth]{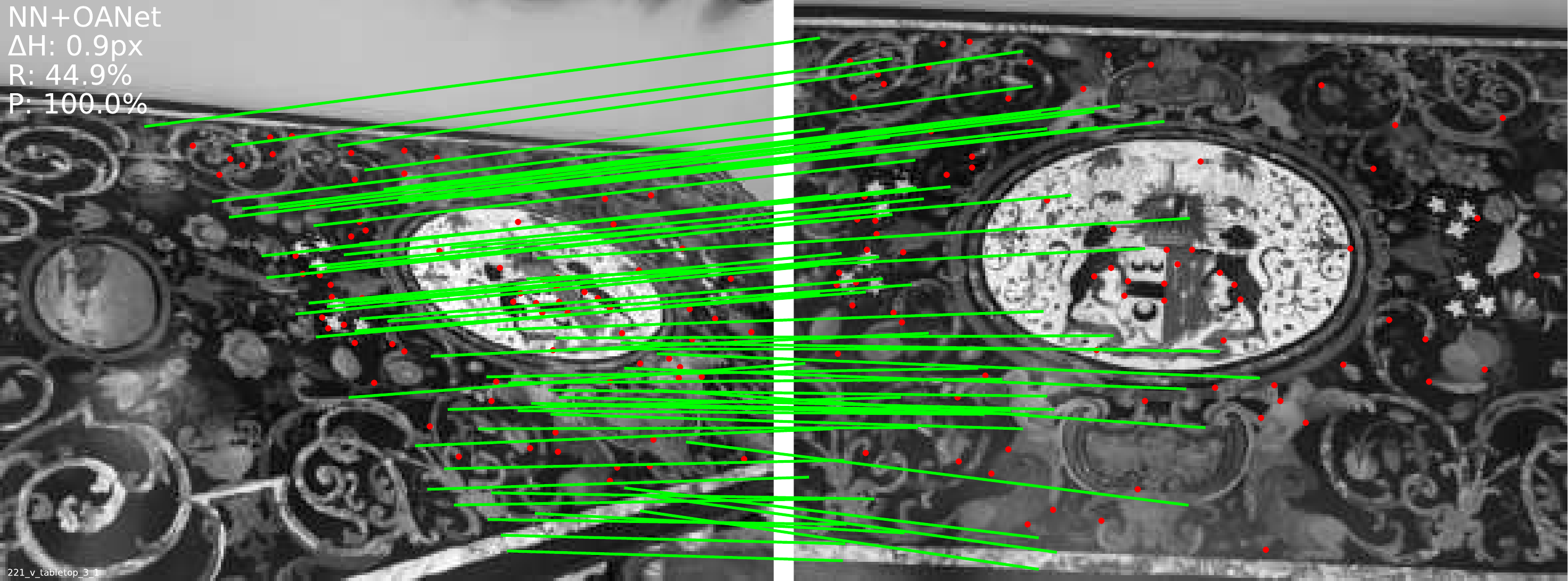}
    
    \vspace{.5mm}
    \includegraphics[width=\linewidth]{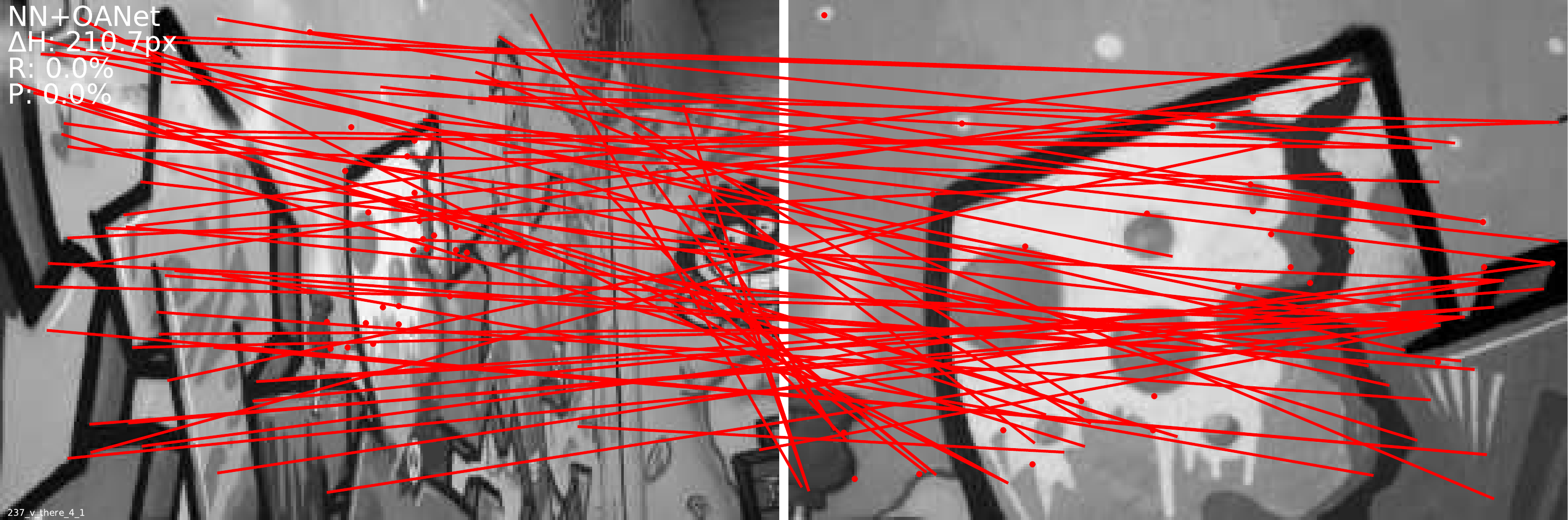}
\end{minipage}%
\hspace{1mm}%
\begin{minipage}{\iwidth\textwidth}
    \includegraphics[width=\linewidth]{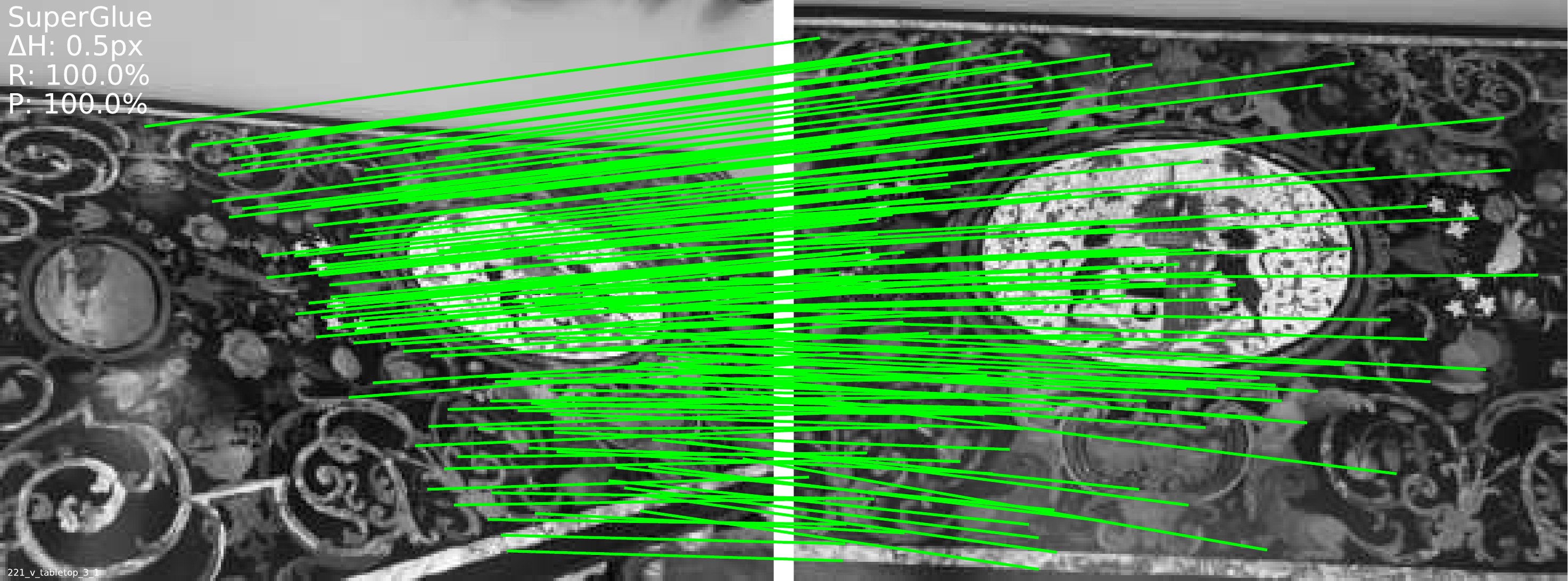}
    
    \vspace{.5mm}
    \includegraphics[width=\linewidth]{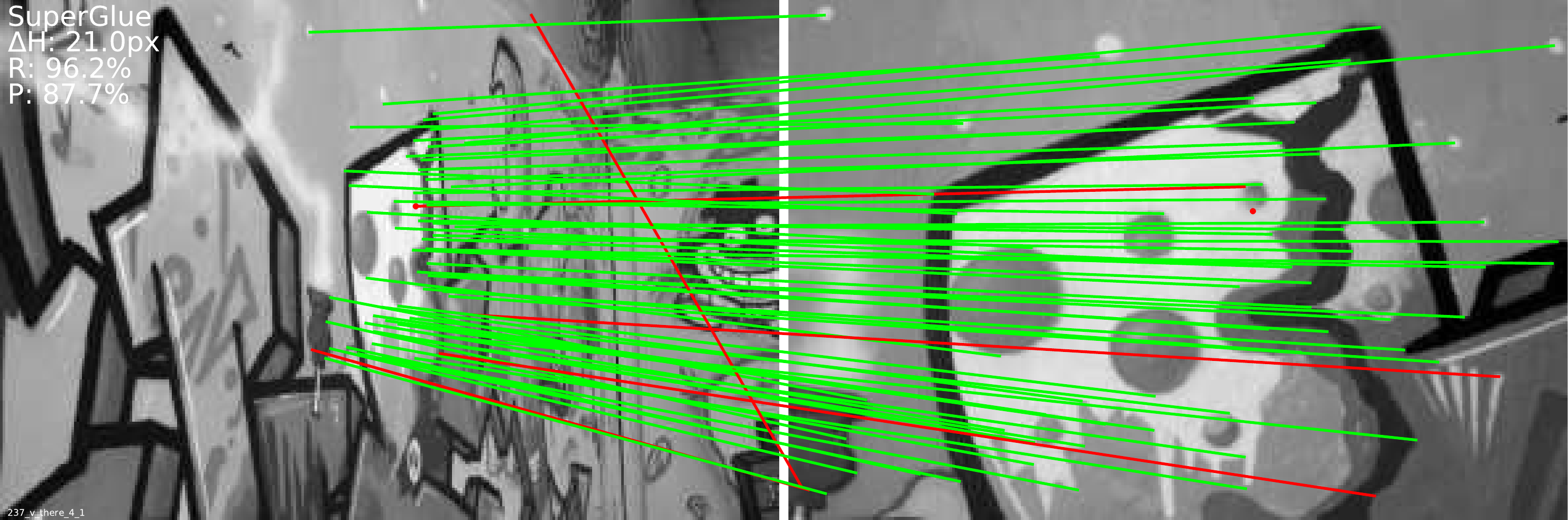}
\end{minipage}

\vspace{1.5mm}
\begin{minipage}{0.02\textwidth}
\rotatebox[origin=c]{90}{Webcam}
\end{minipage}%
\hfill{\vline width 1pt}\hfill
\hspace{1mm}%
\begin{minipage}{\iwidth\textwidth}
    \includegraphics[width=\linewidth]{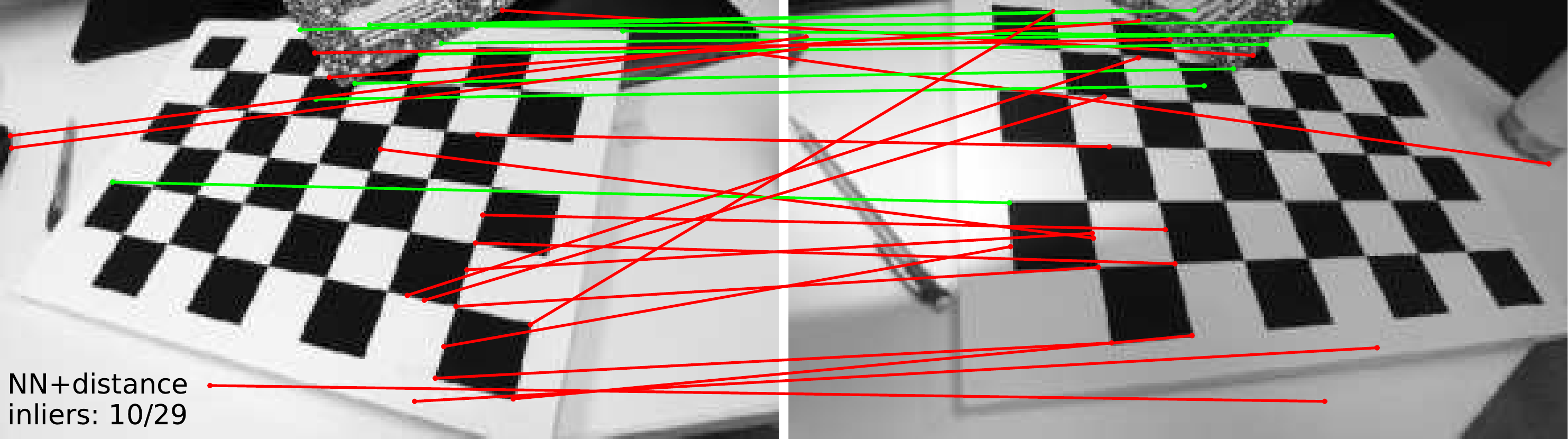}
\end{minipage}%
\hspace{1mm}%
\begin{minipage}{\iwidth\textwidth}
    \includegraphics[width=\linewidth]{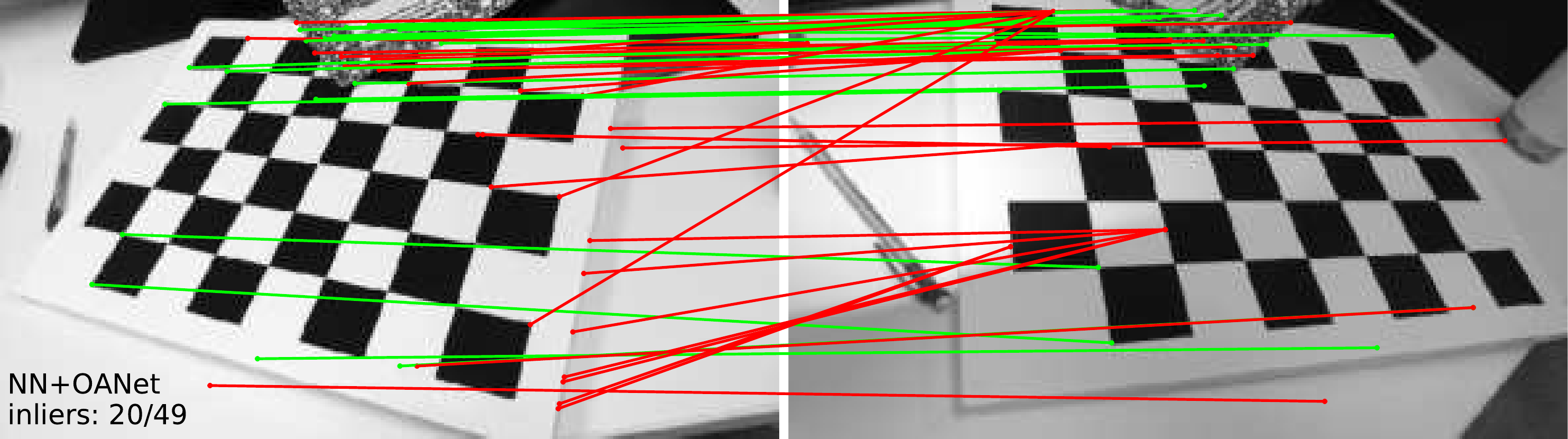}
\end{minipage}%
\hspace{1mm}%
\begin{minipage}{\iwidth\textwidth}
    \includegraphics[width=\linewidth]{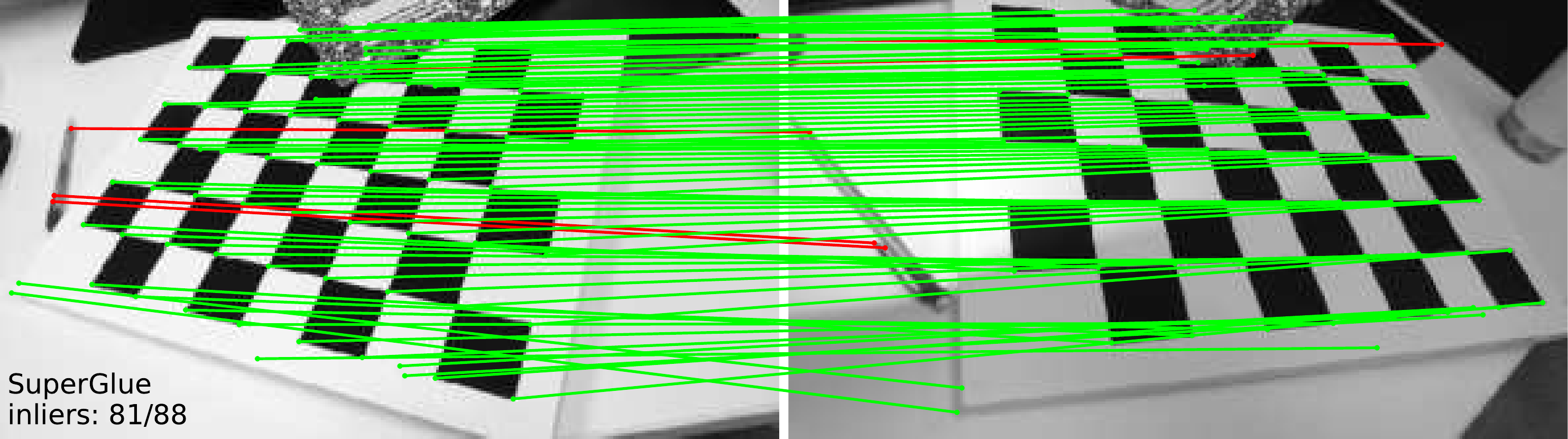}
\end{minipage}

\vspace{-.1cm}
\caption{{\bf More homography examples.} We show point correspondences on our synthetic dataset (see Section~\ref{sec:homography}), on real image pairs from HPatches (see \supp~\ref{sec:homography-supp}), and a checkerboard image captured by a webcam. SuperGlue consistently estimates more correct matches ({\color{green}green} lines) and fewer mismatches ({\color{red}red} lines), successfully coping with repeated texture, large viewpoint, and illumination changes.}
\label{fig:supp-homography-qualitative}
\end{figure*}

\begin{figure*}[ht!]
\vspace{-2mm}
\centering
\def\iwidth{0.315}
\begin{minipage}{\iwidth\textwidth}
    \centering
    \small{SuperPoint + NN + distance}
\end{minipage}%
\hspace{1mm}%
\begin{minipage}{\iwidth\textwidth}
    \centering
    \small{SuperPoint + NN + OANet}
\end{minipage}%
\hspace{1mm}%
\begin{minipage}{\iwidth\textwidth}
    \centering
    \small{SuperPoint + \b{SuperGlue}}
\end{minipage}%

\begin{minipage}{0.02\textwidth}
\rotatebox[origin=c]{90}{Difficult}
\end{minipage}%
\hfill{\vline width 1pt}\hfill
\hspace{1mm}%
\begin{minipage}{\iwidth\textwidth}
    \includegraphics[width=\linewidth]{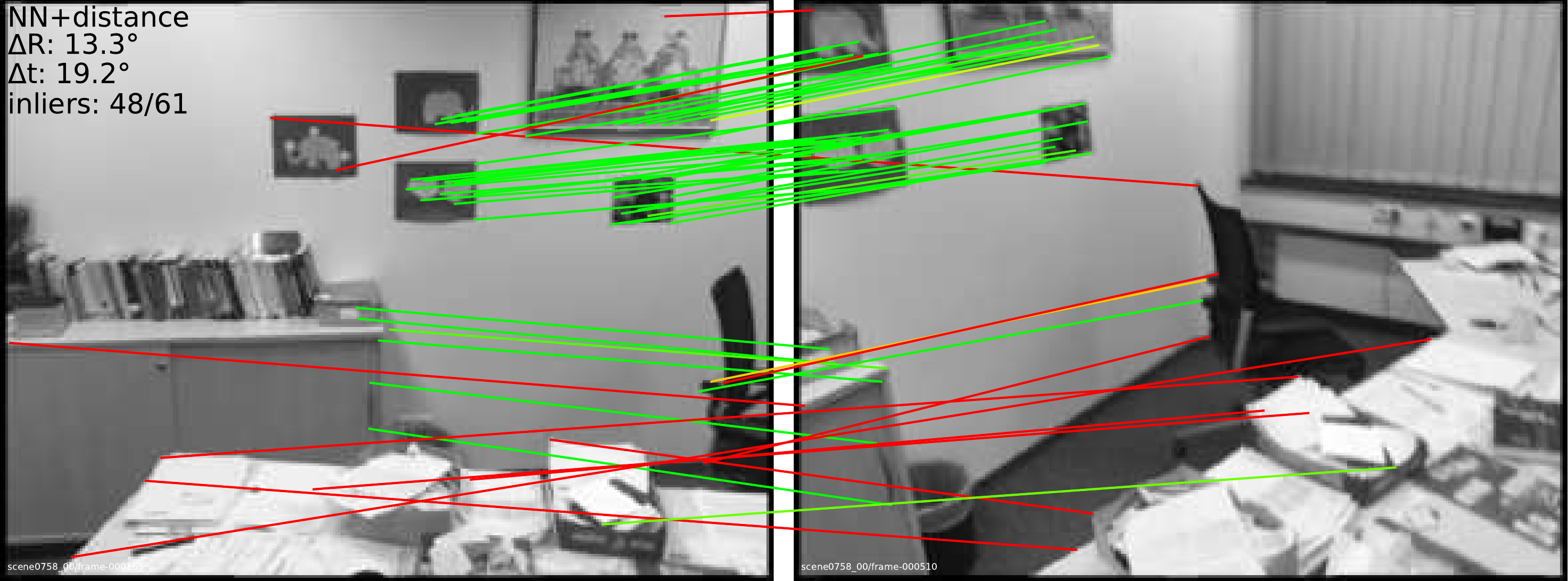}
    
    \vspace{.5mm}
    \includegraphics[width=\linewidth]{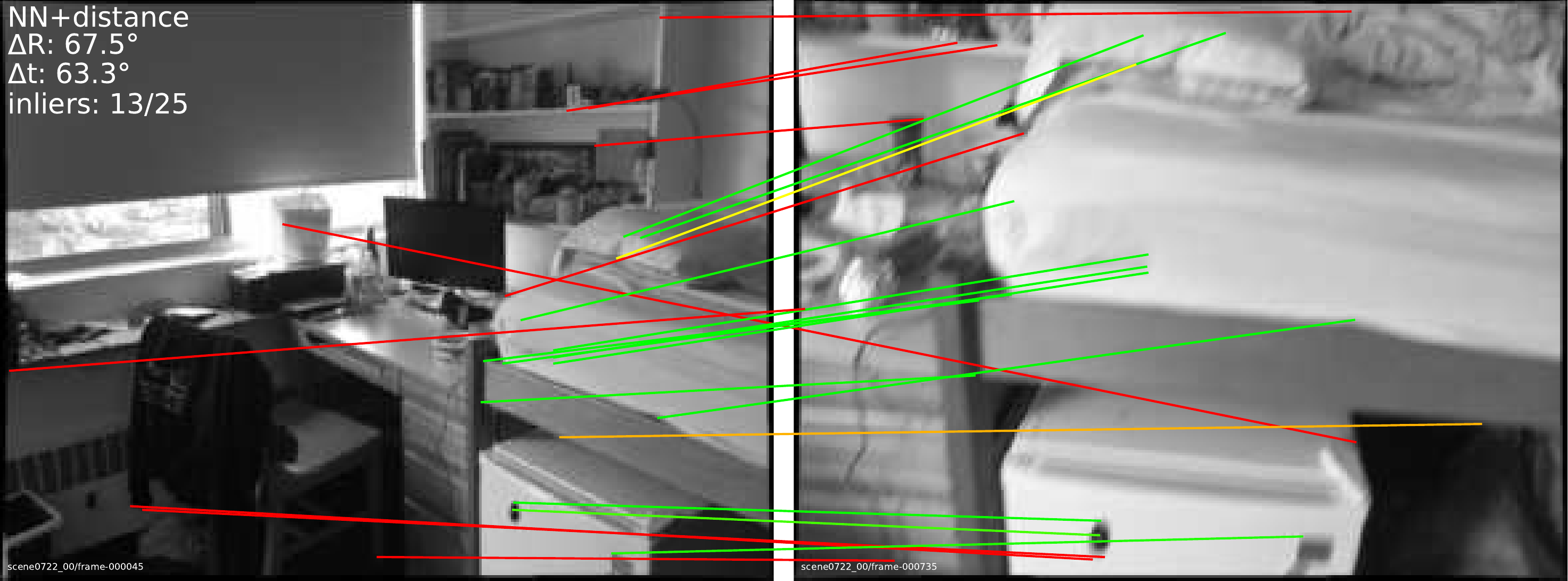}
\end{minipage}%
\hspace{1mm}%
\begin{minipage}{\iwidth\textwidth}
    \includegraphics[width=\linewidth]{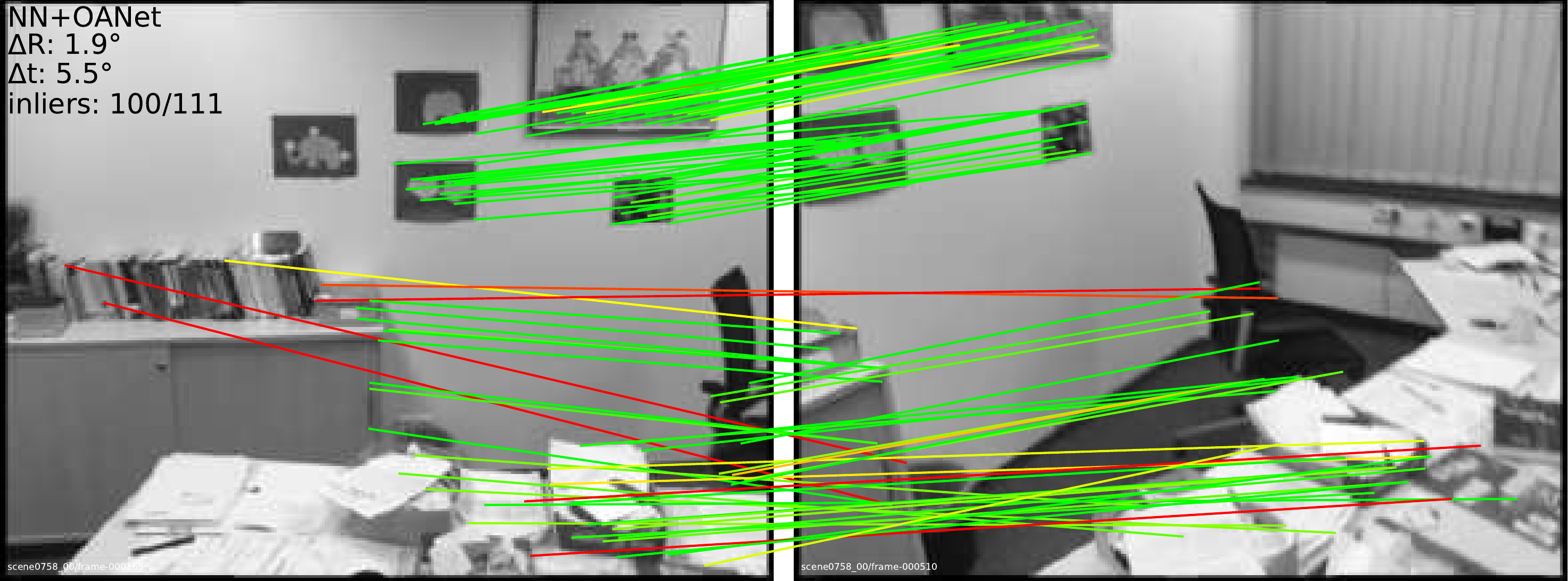}
    
    \vspace{.5mm}
    \includegraphics[width=\linewidth]{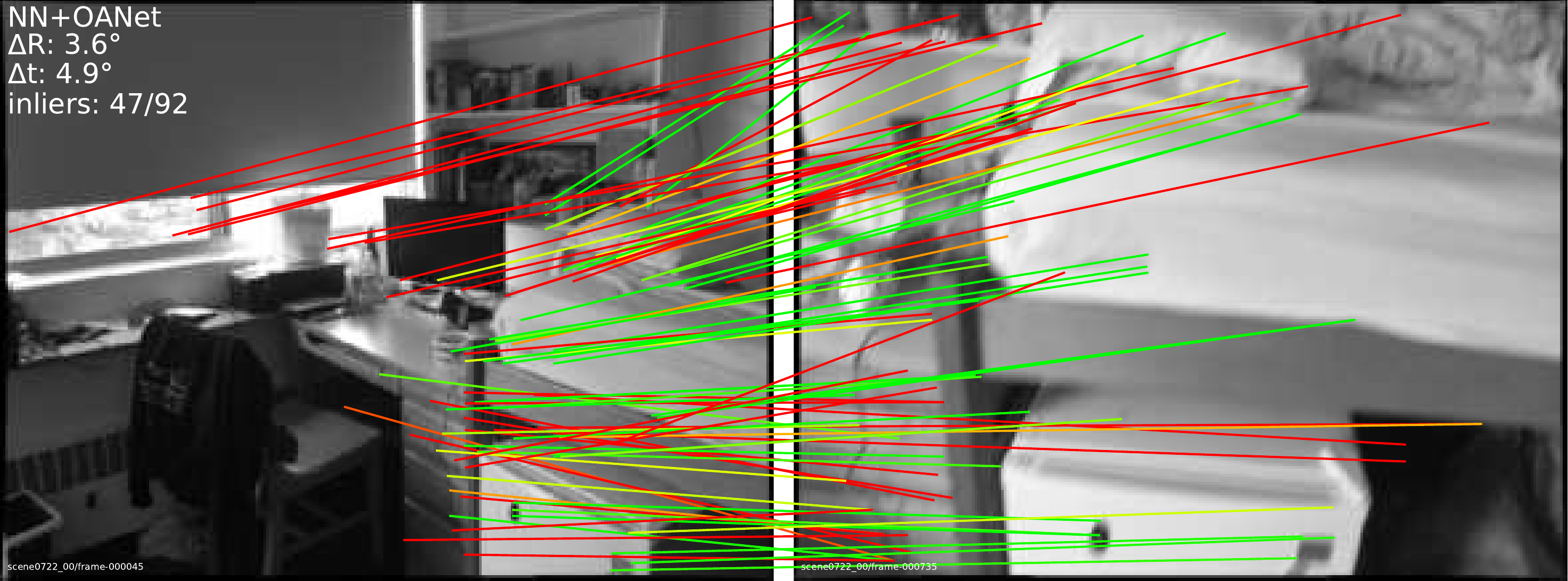}
\end{minipage}%
\hspace{1mm}%
\begin{minipage}{\iwidth\textwidth}
    \includegraphics[width=\linewidth]{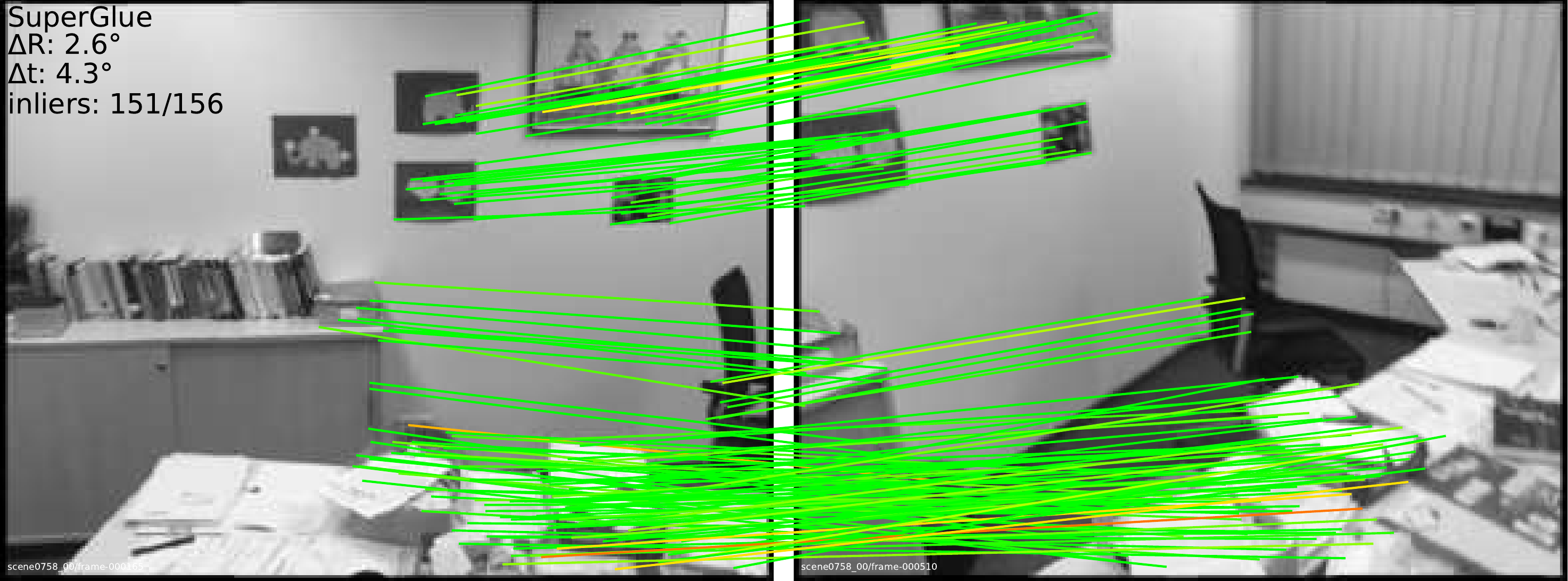}
    
    \vspace{.5mm}
    \includegraphics[width=\linewidth]{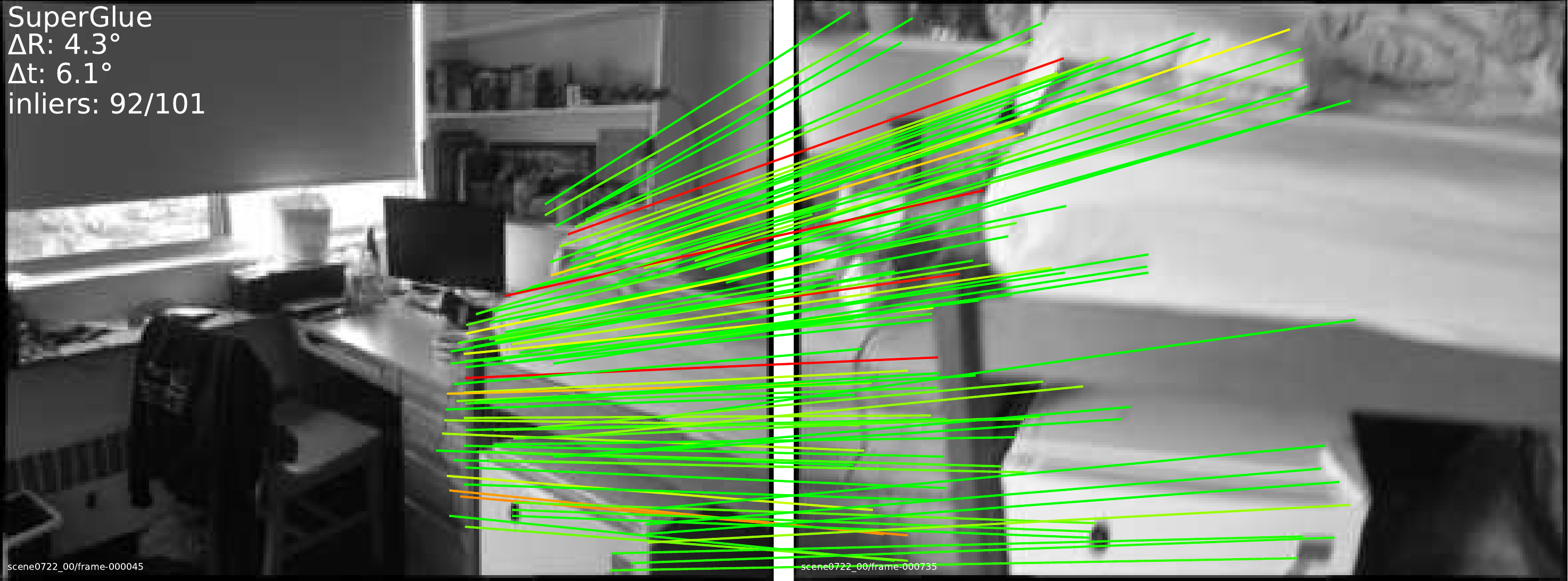}
\end{minipage}

\vspace{1.5mm}
\begin{minipage}{0.02\textwidth}
\rotatebox[origin=c]{90}{Very difficult}
\end{minipage}%
\hfill{\vline width 1pt}\hfill
\hspace{1mm}%
\begin{minipage}{\iwidth\textwidth}
    \includegraphics[width=\linewidth]{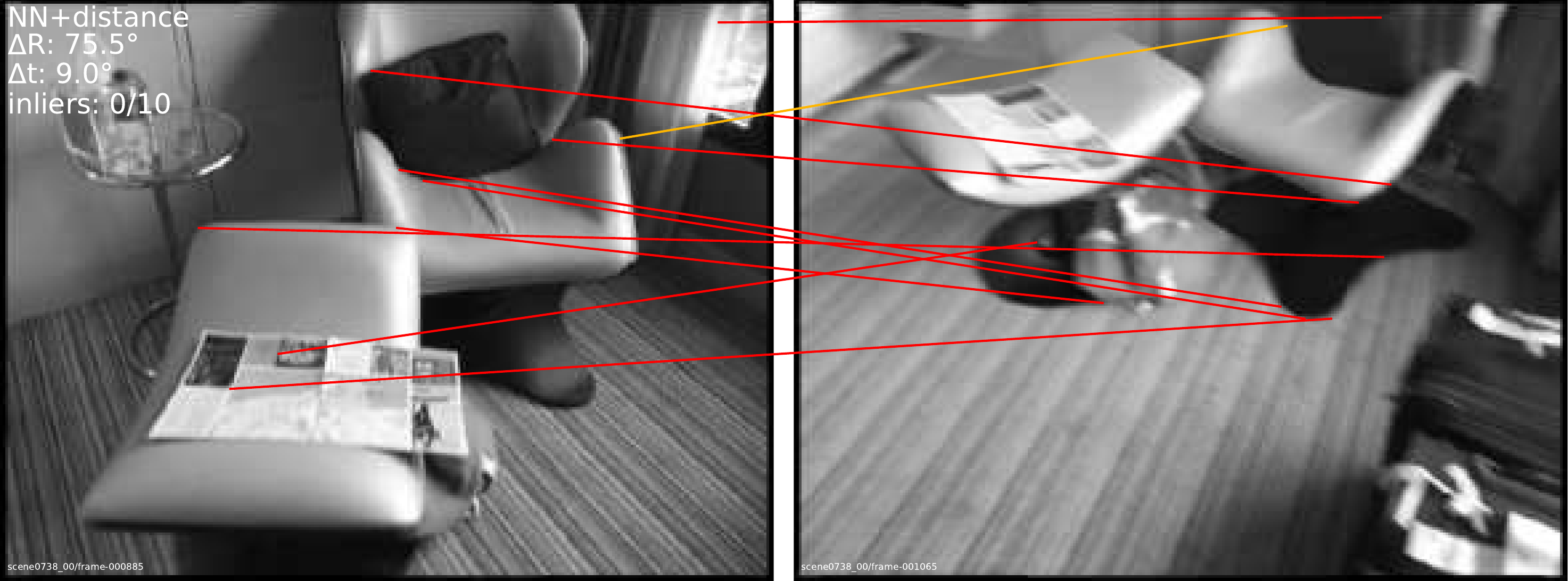}
    
    \vspace{.5mm}
    \includegraphics[width=\linewidth]{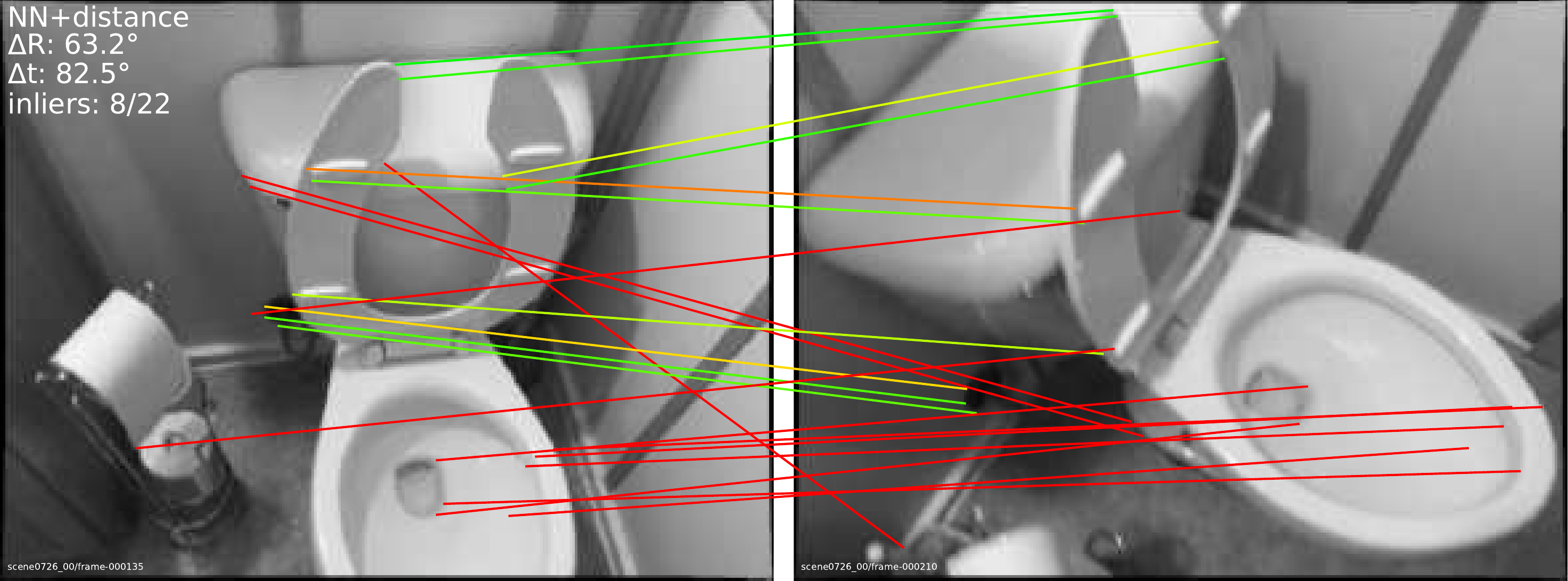}
    
    \vspace{.5mm}
    \includegraphics[width=\linewidth]{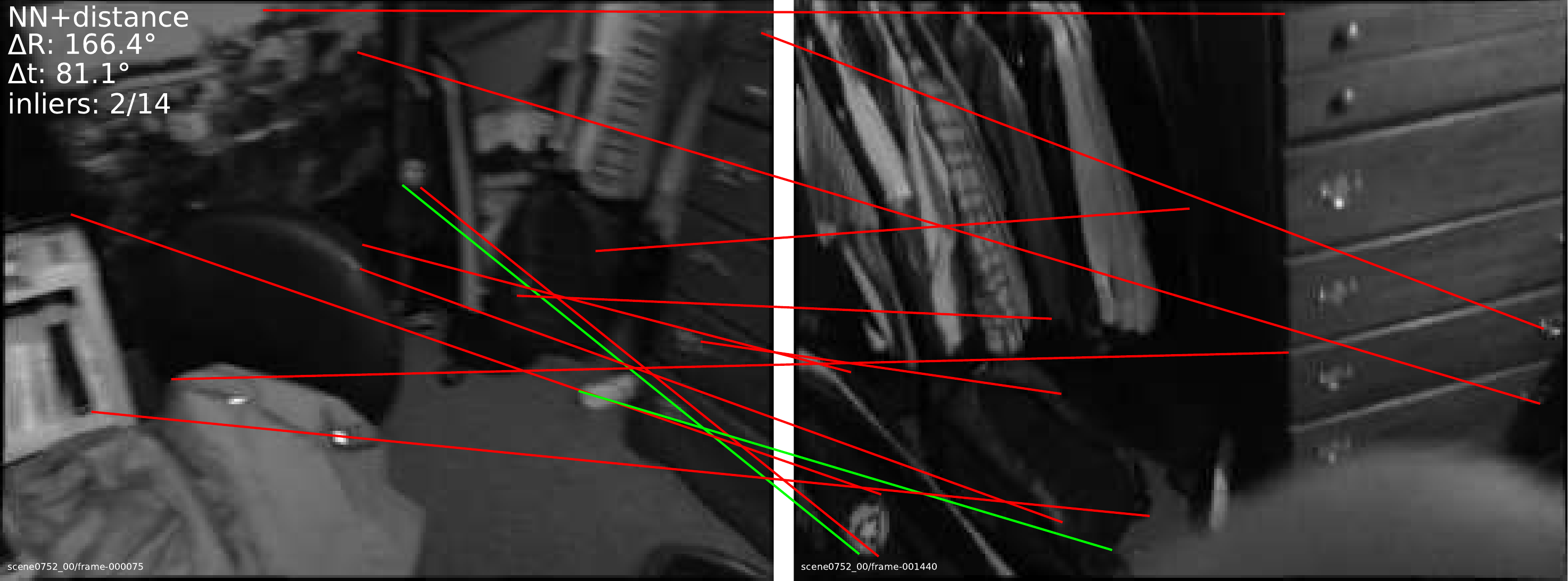}
    
    \vspace{.5mm}
    \includegraphics[width=\linewidth]{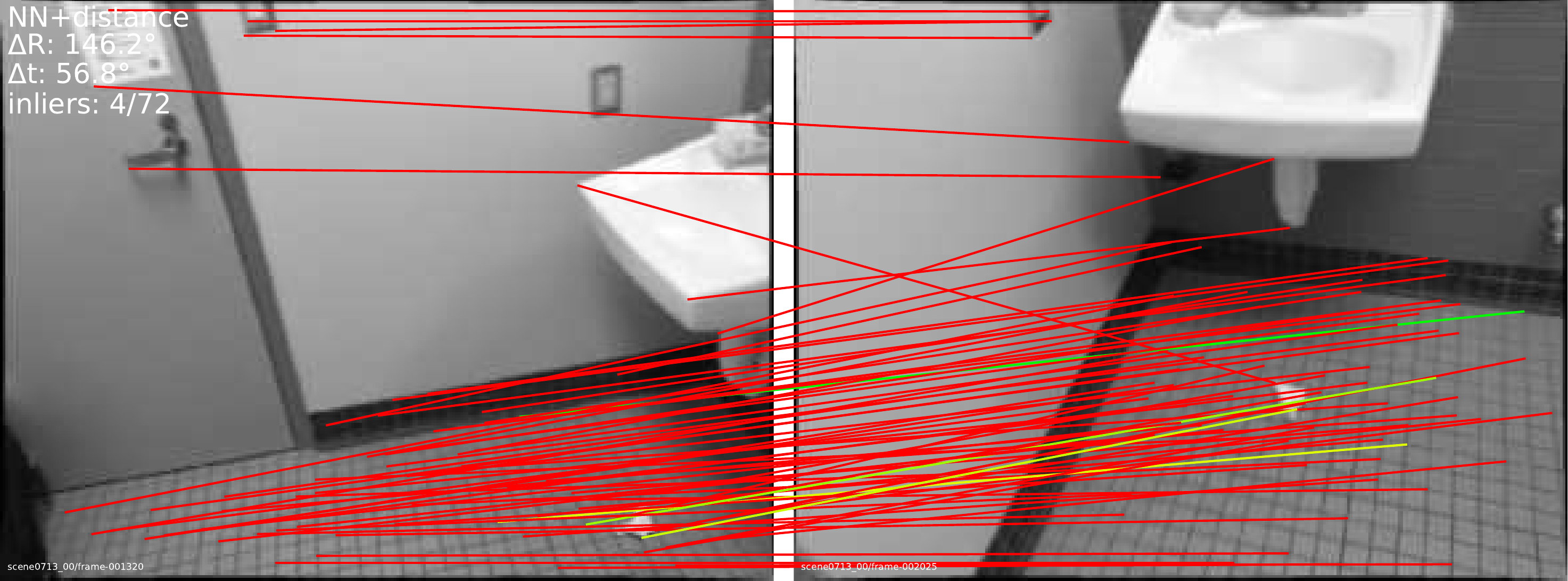}
    
    \vspace{.5mm}
    \includegraphics[width=\linewidth]{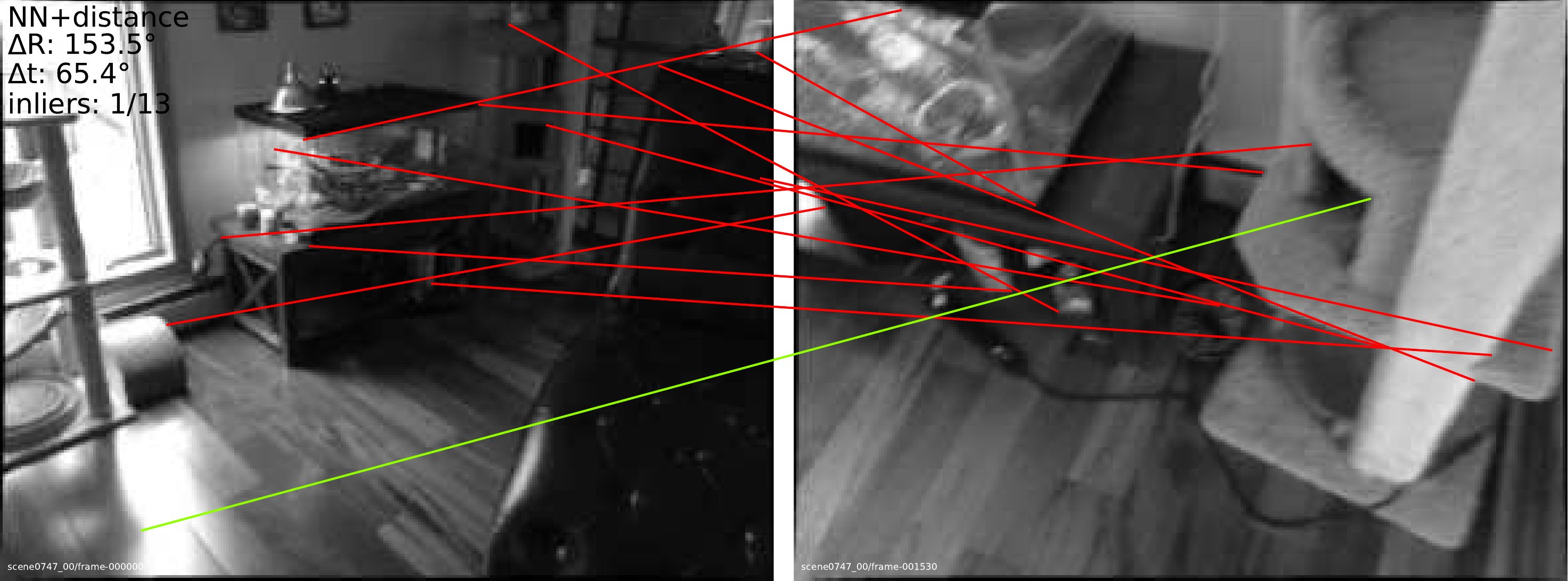}
\end{minipage}%
\hspace{1mm}%
\begin{minipage}{\iwidth\textwidth}
    \includegraphics[width=\linewidth]{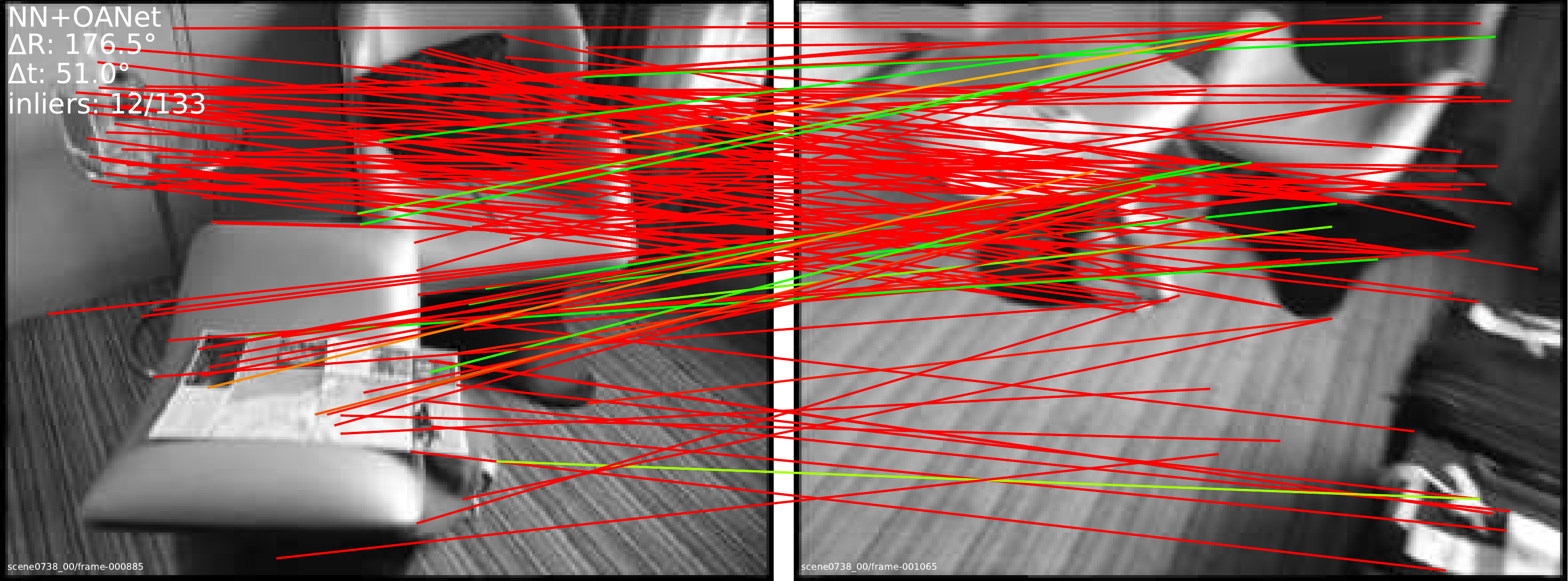}
    
    \vspace{.5mm}
    \includegraphics[width=\linewidth]{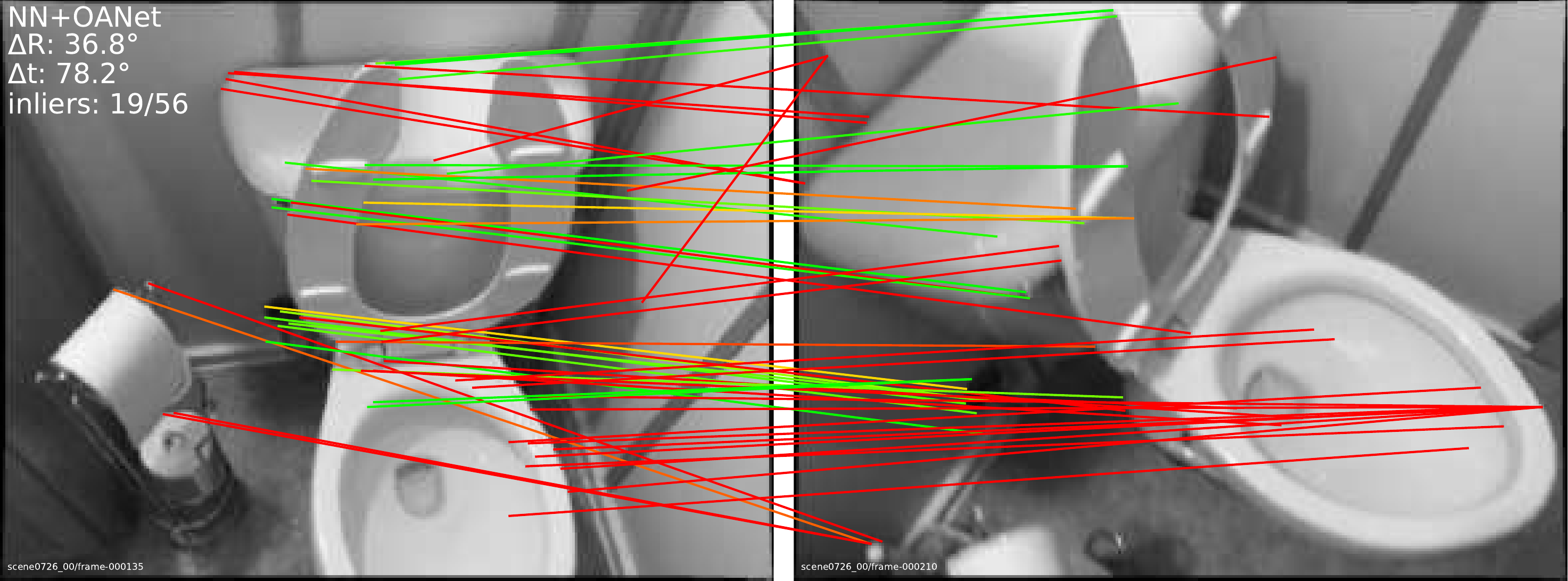}
    
    \vspace{.5mm}
    \includegraphics[width=\linewidth]{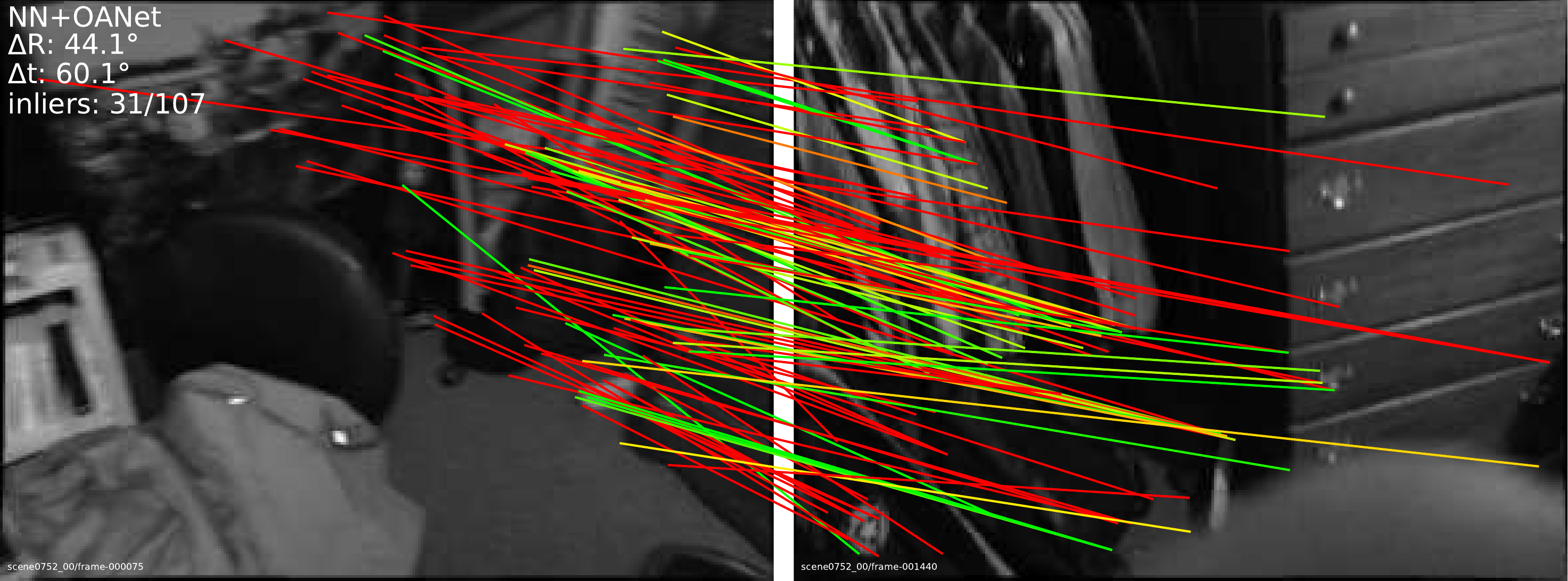}

    \vspace{.5mm}
    \includegraphics[width=\linewidth]{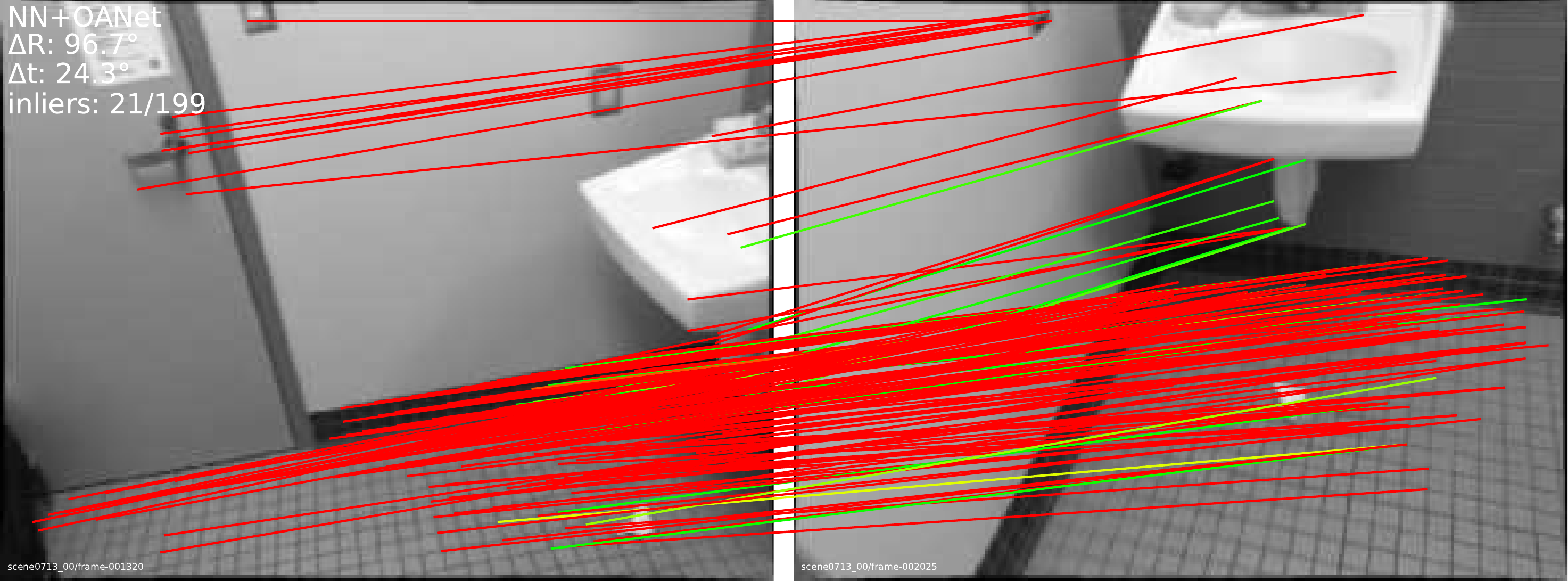}
    
    \vspace{.5mm}
    \includegraphics[width=\linewidth]{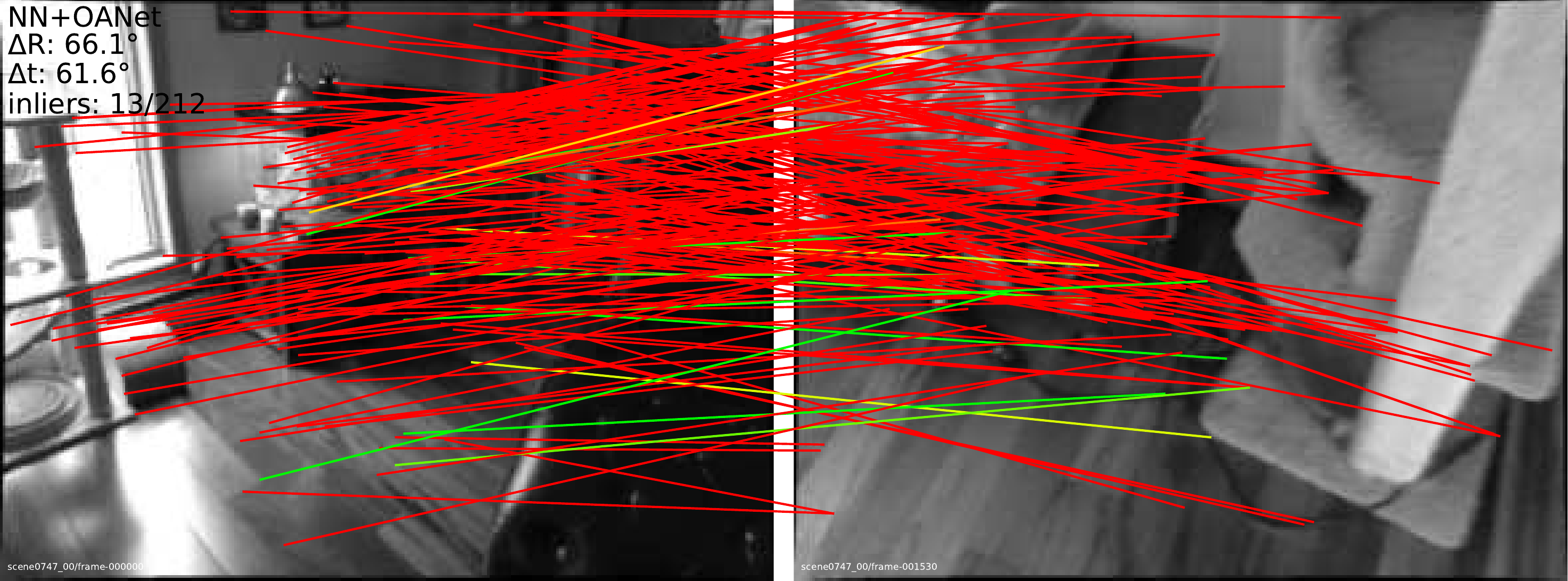}
\end{minipage}%
\hspace{1mm}%
\begin{minipage}{\iwidth\textwidth}
    \includegraphics[width=\linewidth]{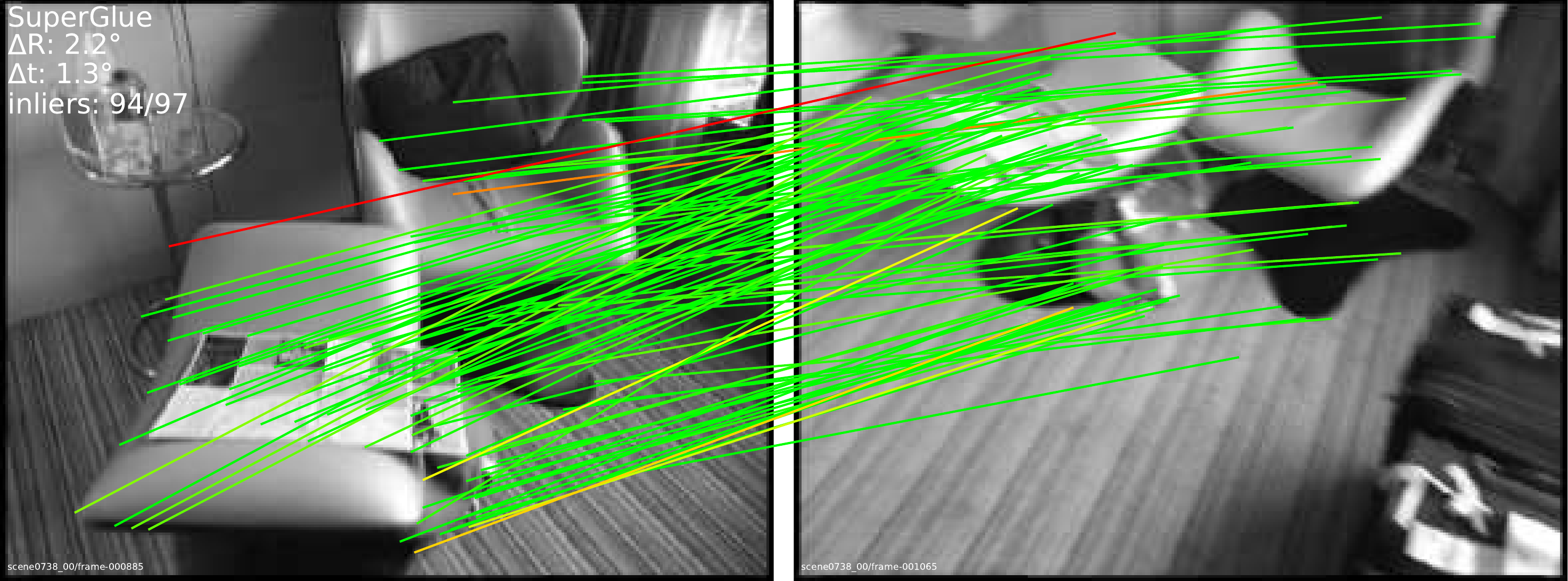}
    
    \vspace{.5mm}
    \includegraphics[width=\linewidth]{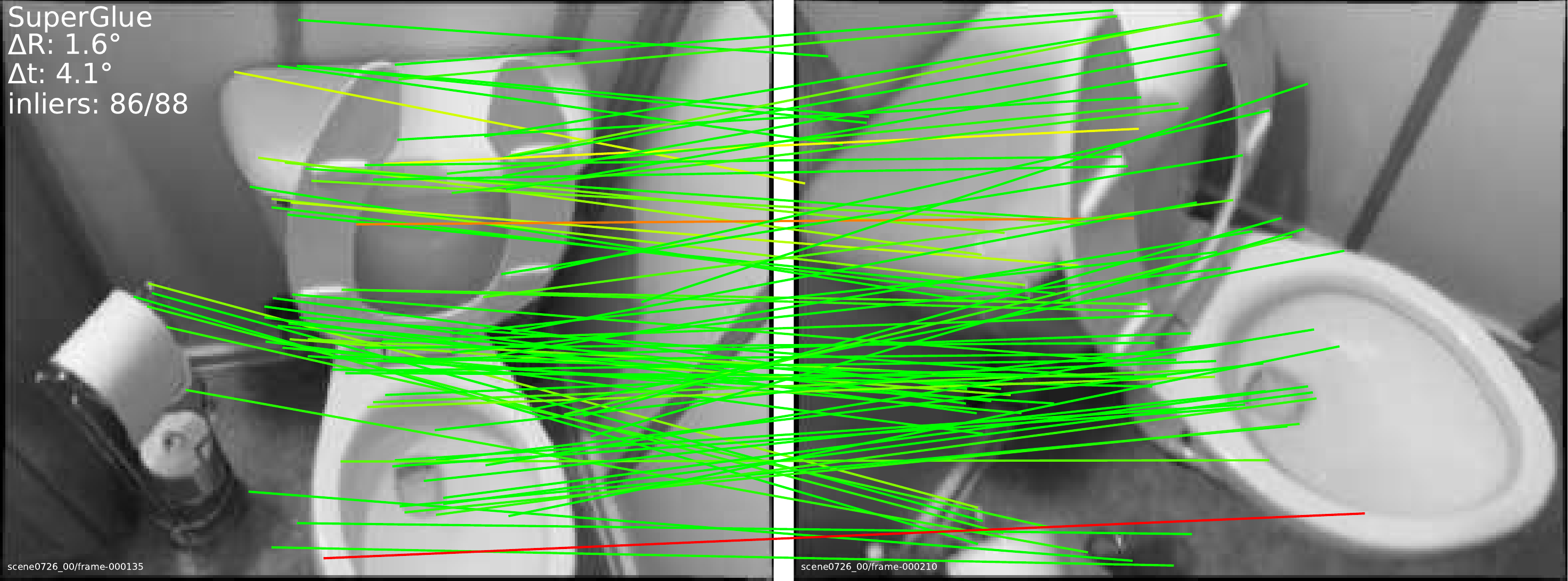}
    
    \vspace{.5mm}
    \includegraphics[width=\linewidth]{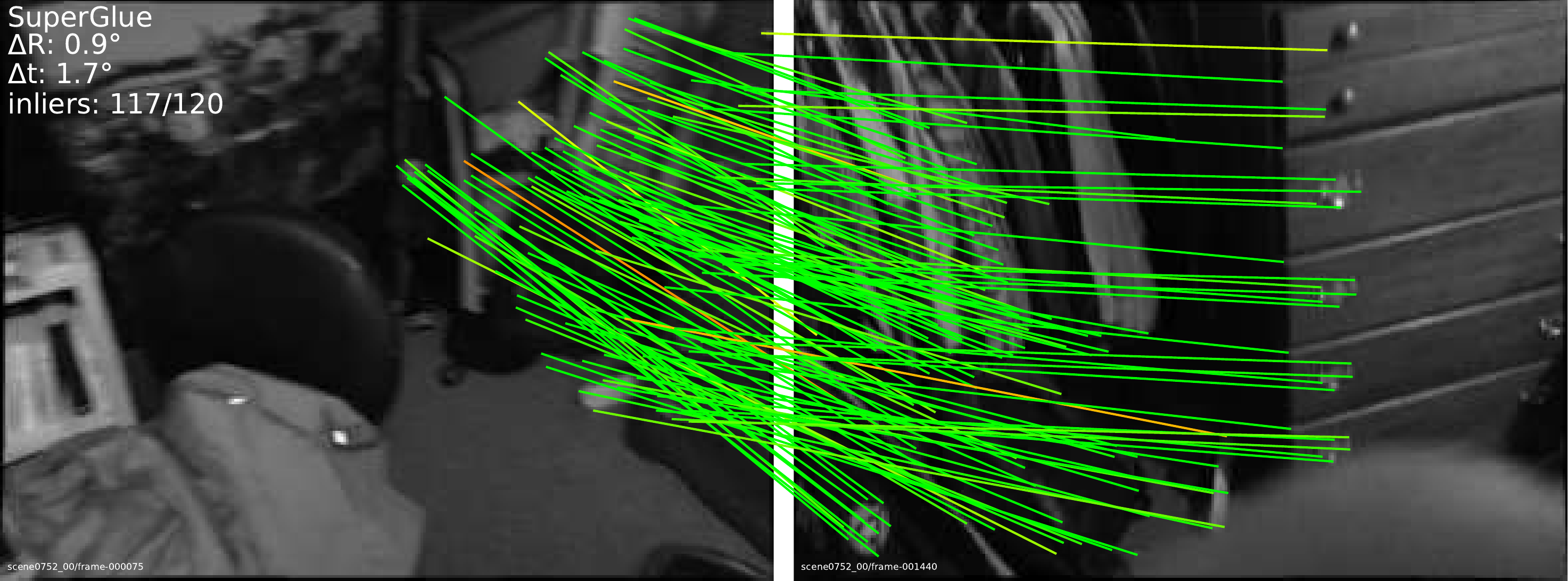}

    \vspace{.5mm}
    \includegraphics[width=\linewidth]{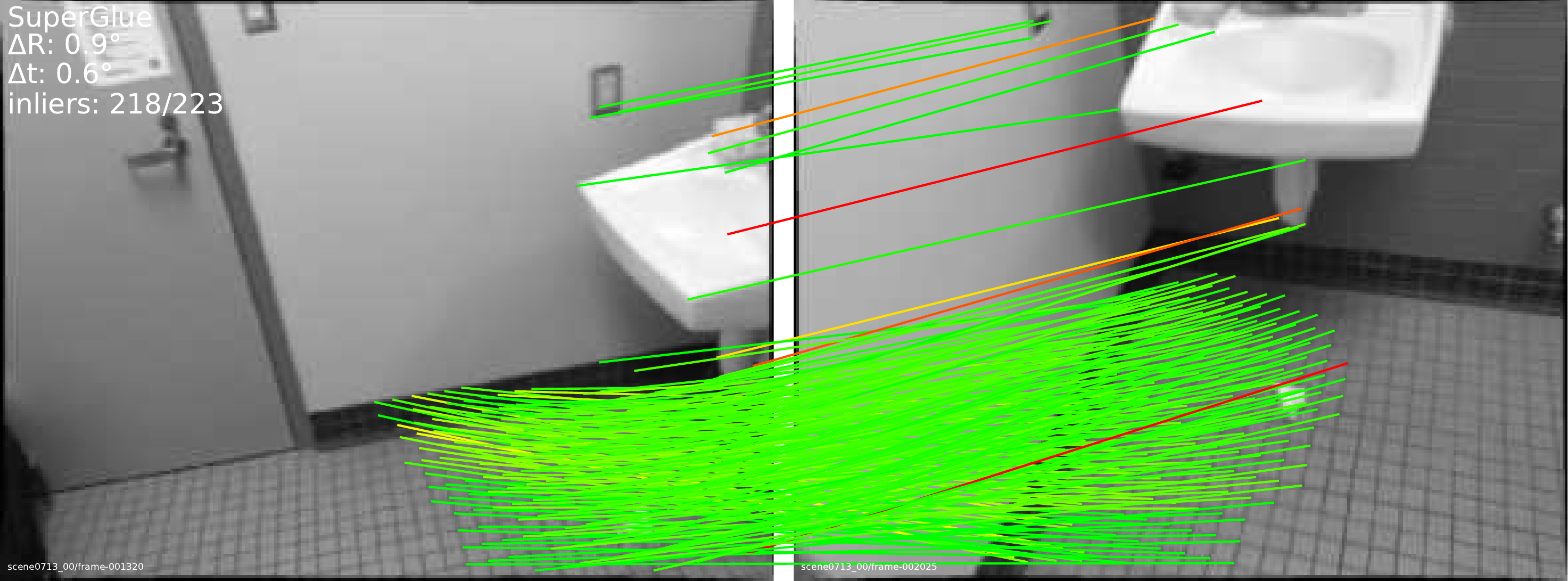}
    
    \vspace{.5mm}
    \includegraphics[width=\linewidth]{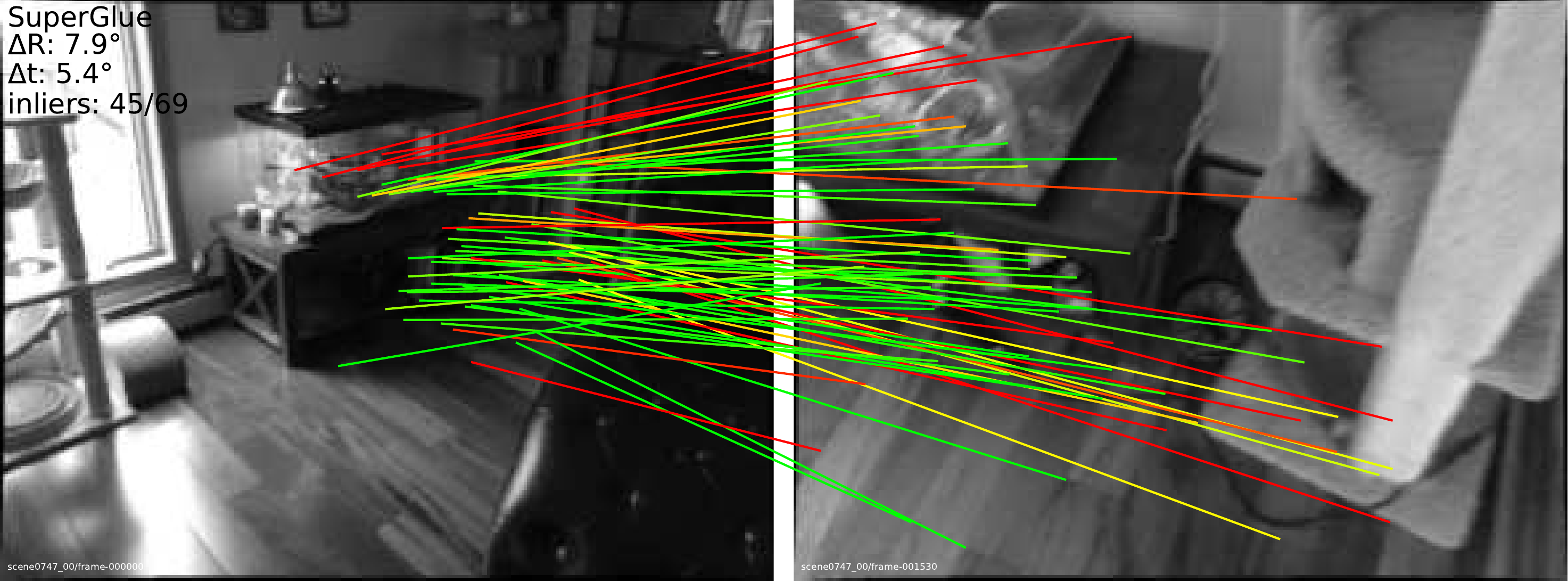}
\end{minipage}

\vspace{1.5mm}
\begin{minipage}{0.02\textwidth}
\rotatebox[origin=c]{90}{Too difficult}
\end{minipage}%
\hfill{\vline width 1pt}\hfill
\hspace{1mm}%
\begin{minipage}{\iwidth\textwidth}
    \includegraphics[width=\linewidth]{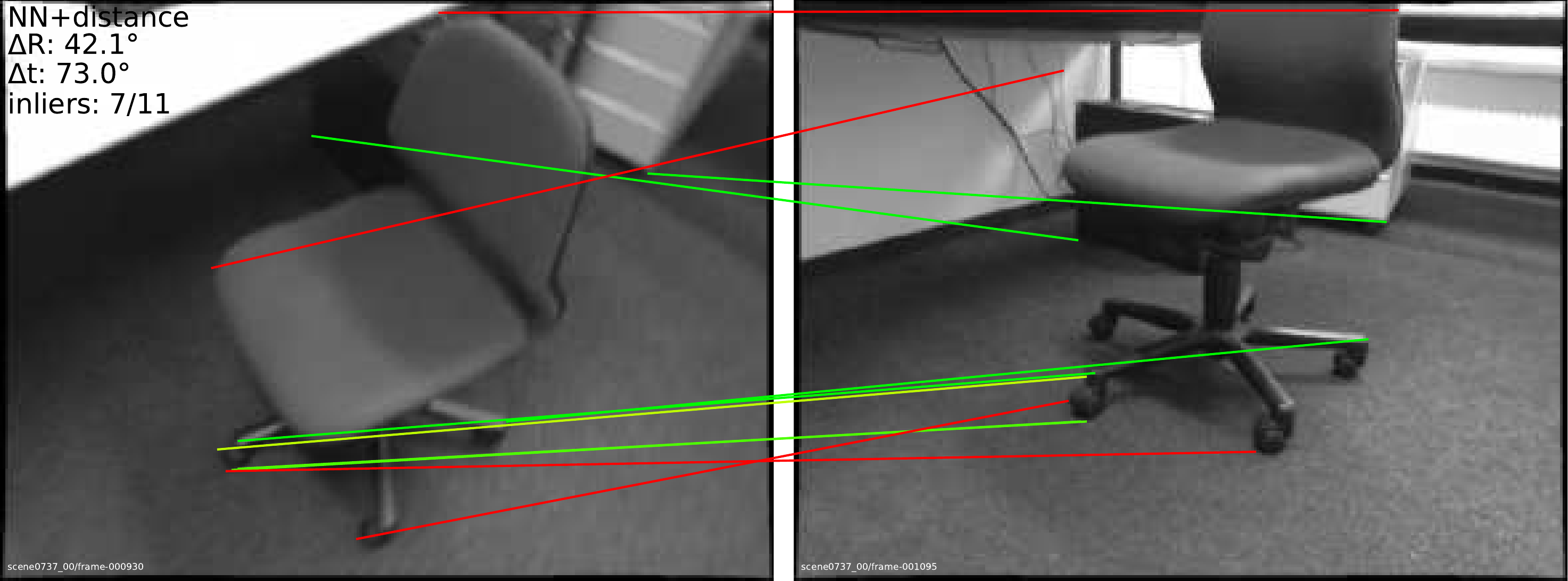}
    
    \vspace{.5mm}
    \includegraphics[width=\linewidth]{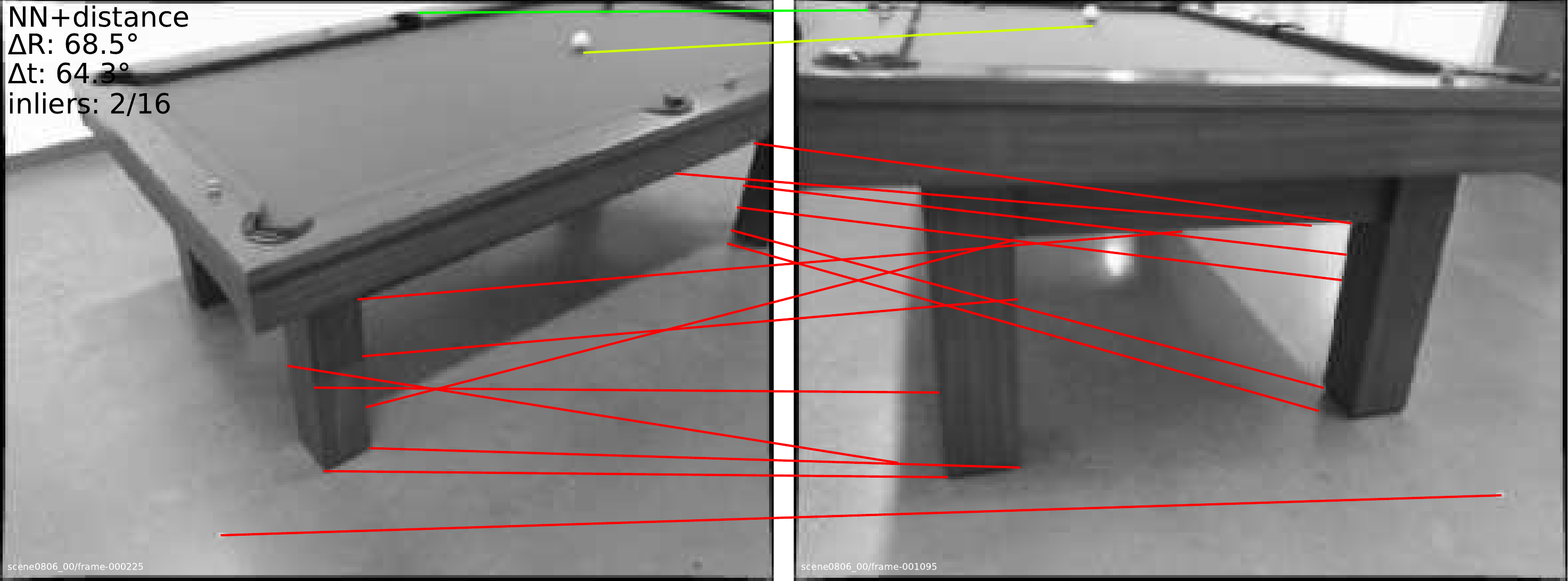}
    
    \vspace{.5mm}
    \includegraphics[width=\linewidth]{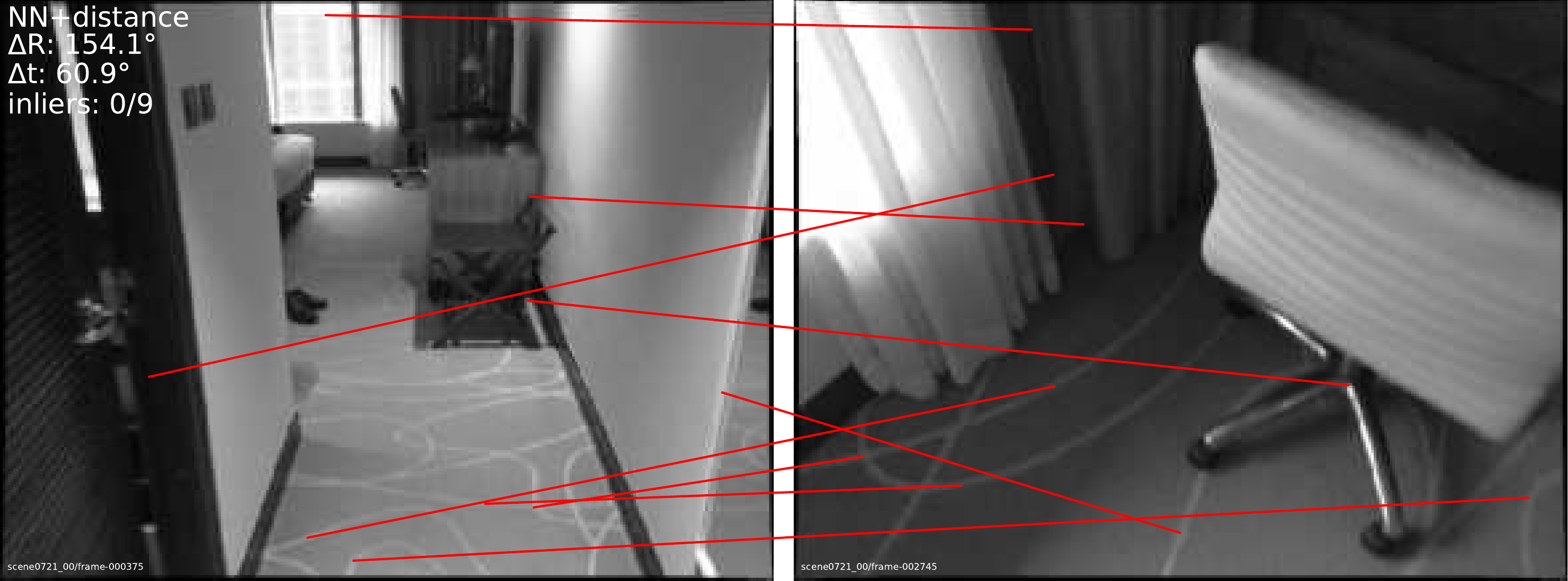}
\end{minipage}%
\hspace{1mm}%
\begin{minipage}{\iwidth\textwidth}
    \includegraphics[width=\linewidth]{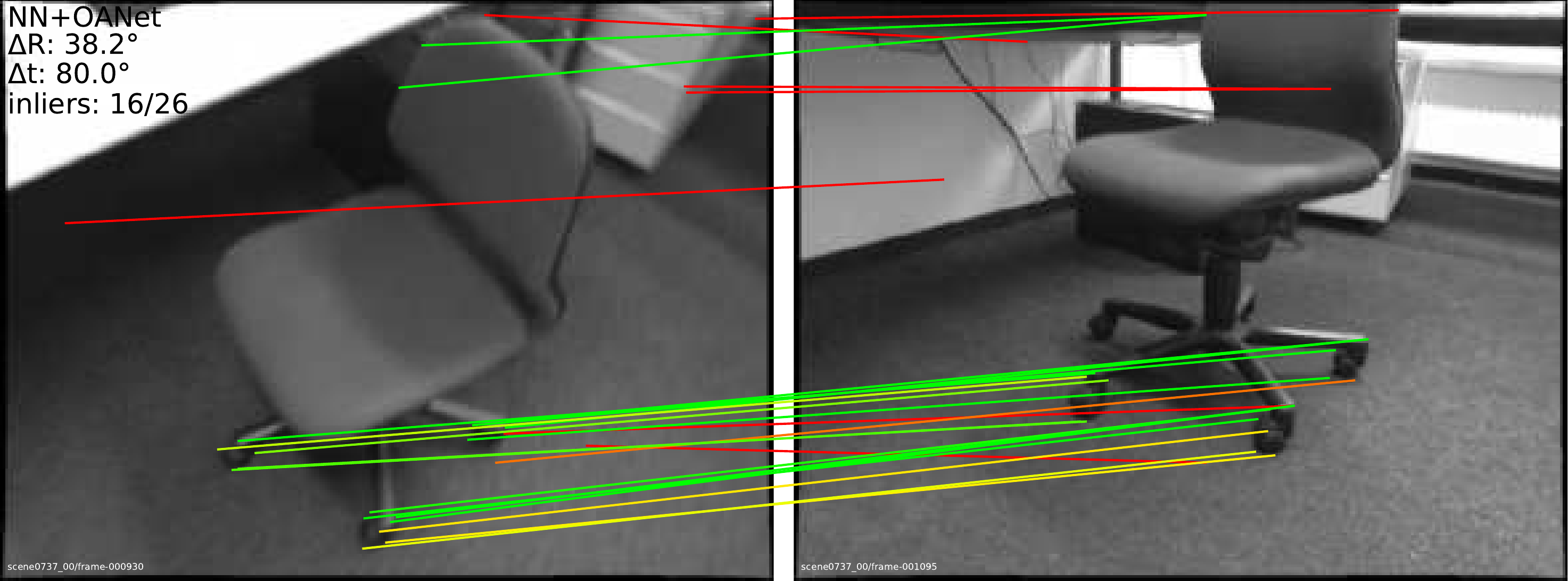}
    
    \vspace{.5mm}
    \includegraphics[width=\linewidth]{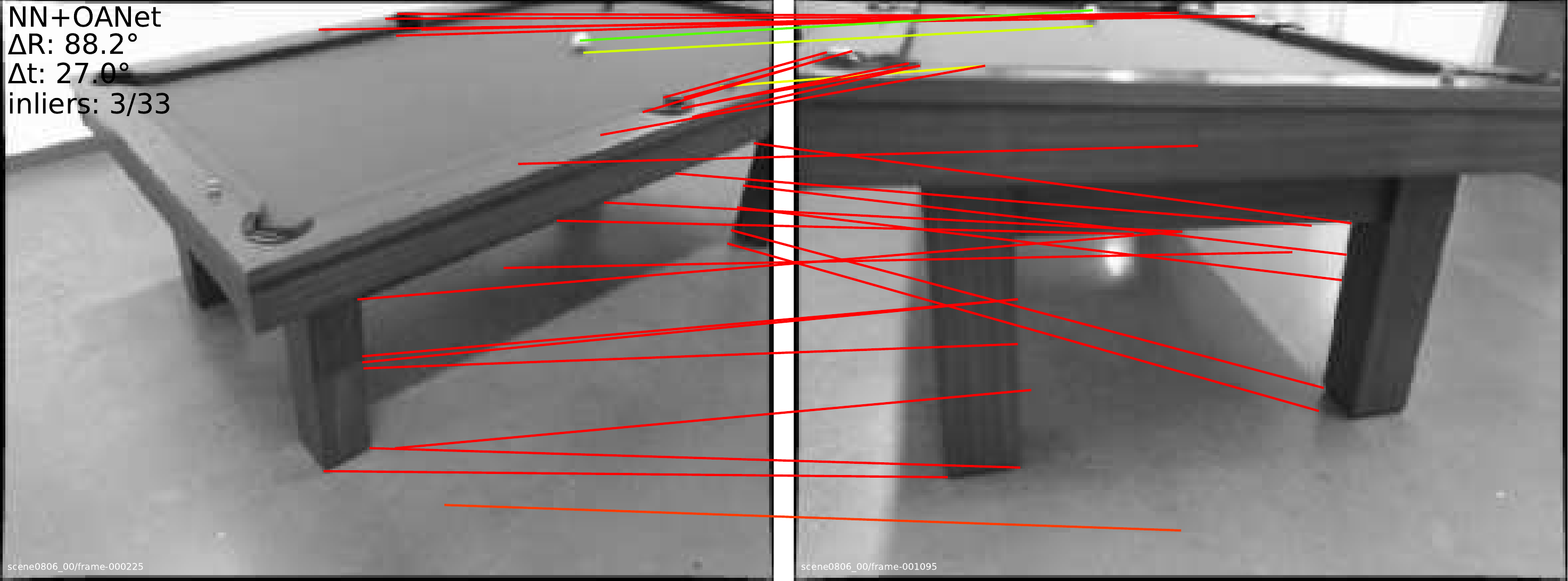}
    
    \vspace{.5mm}
    \includegraphics[width=\linewidth]{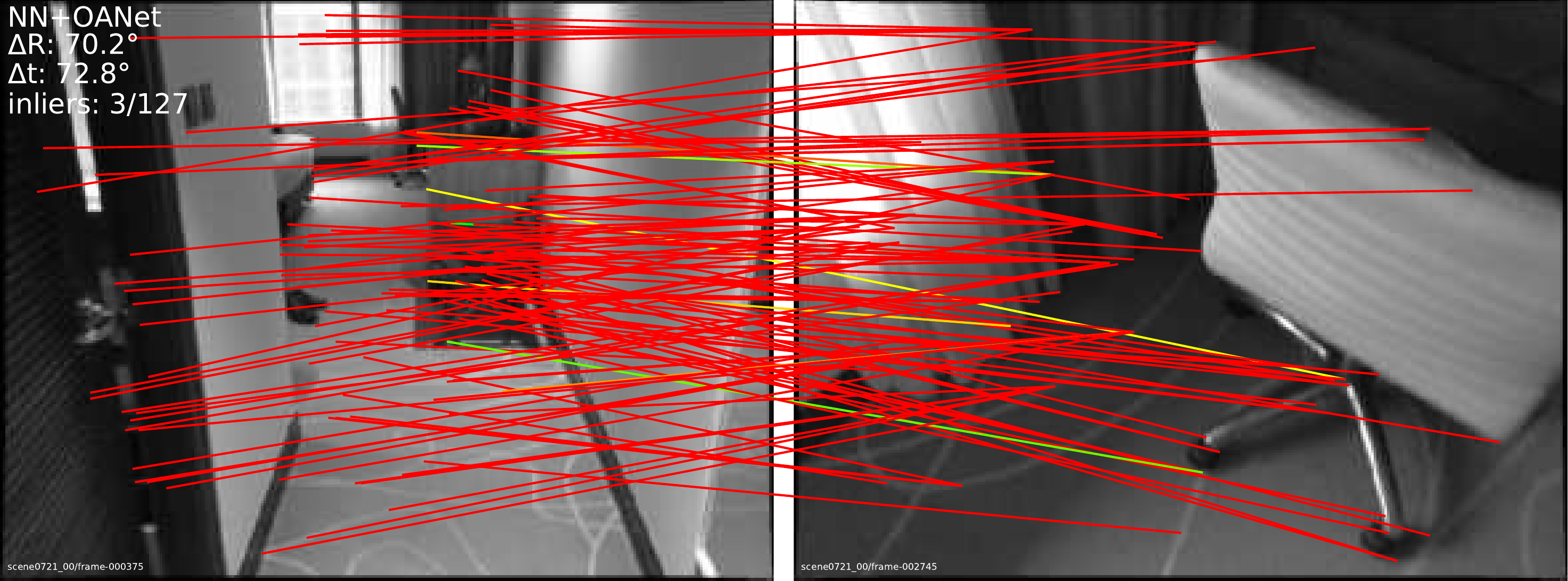}
\end{minipage}%
\hspace{1mm}%
\begin{minipage}{\iwidth\textwidth}
    \includegraphics[width=\linewidth]{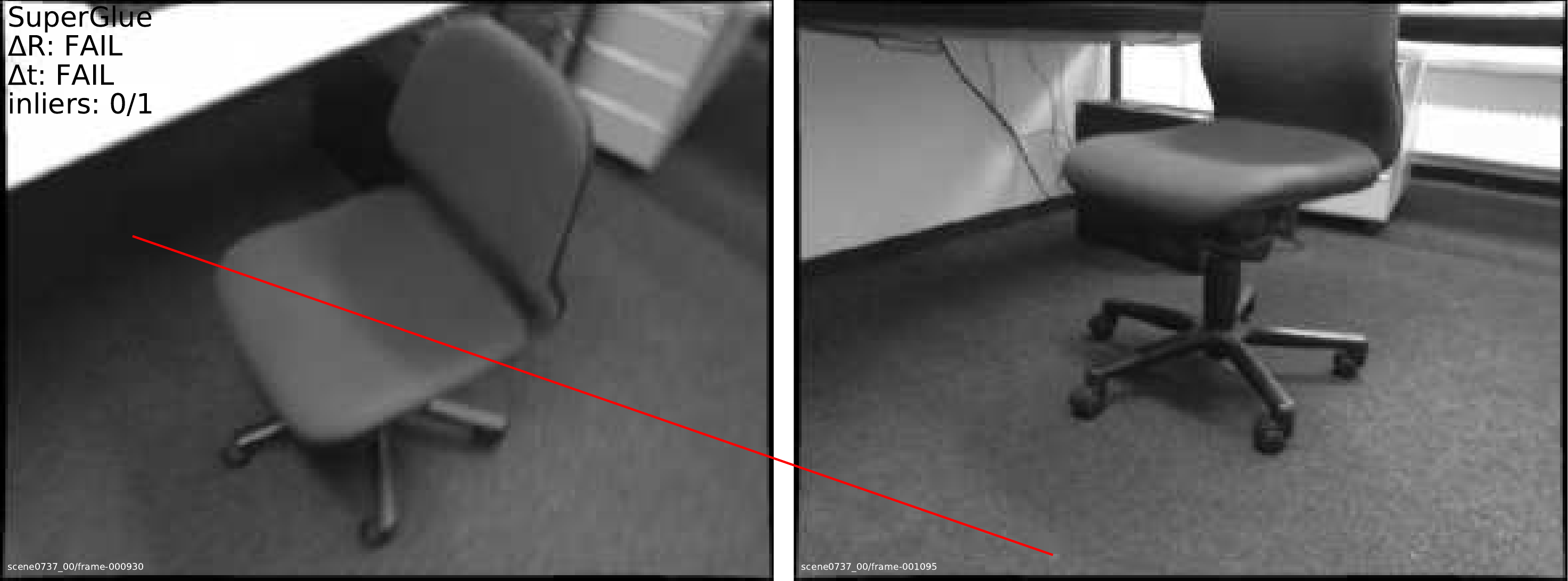}
    
    \vspace{.5mm}
    \includegraphics[width=\linewidth]{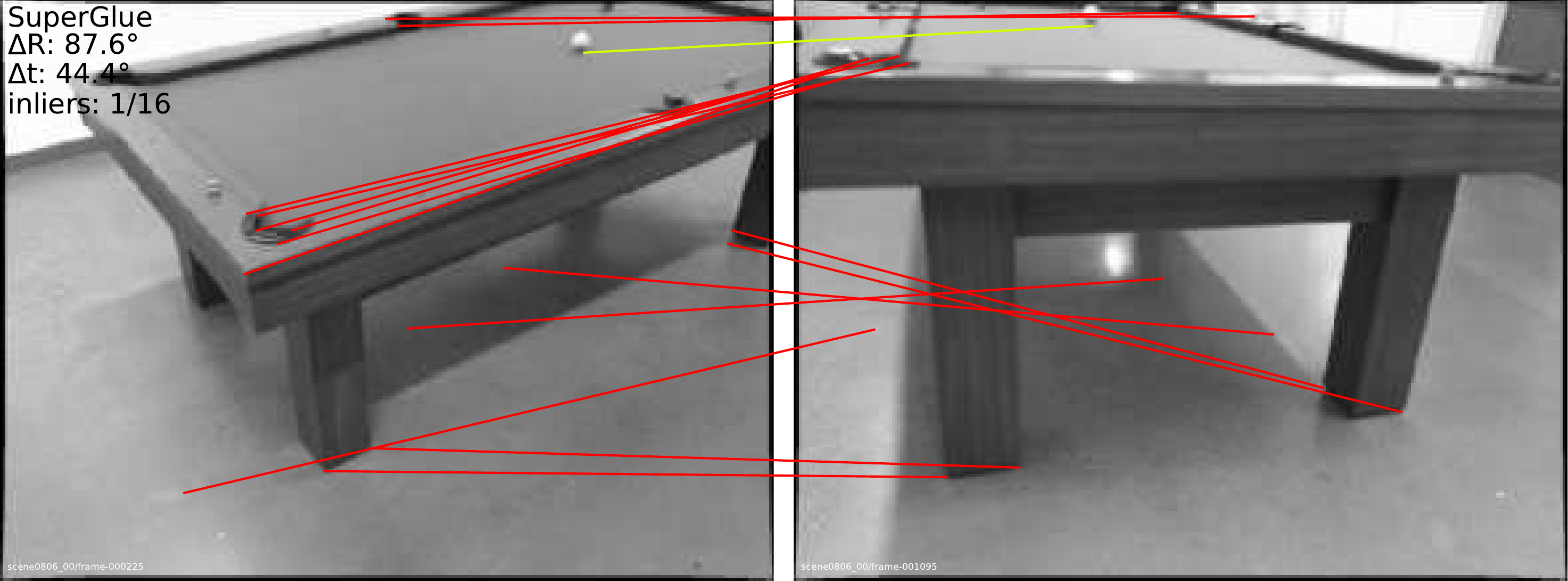}
    
    \vspace{.5mm}
    \includegraphics[width=\linewidth]{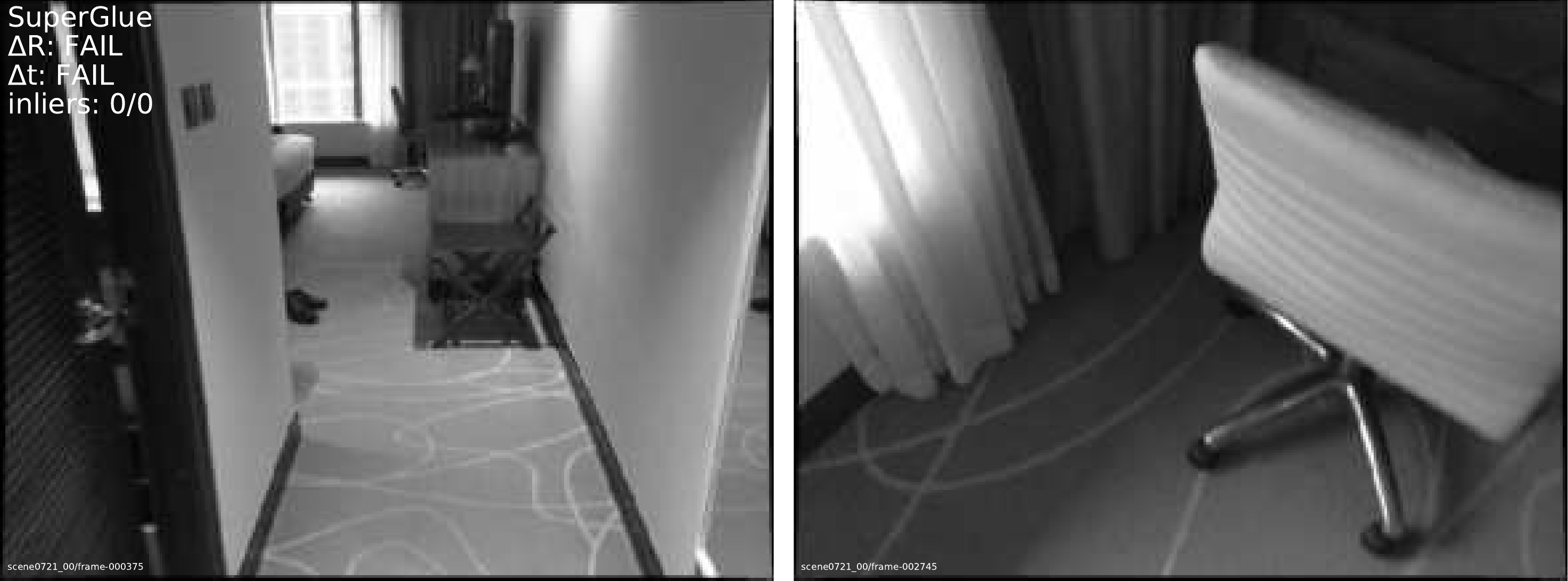}
\end{minipage}

\vspace{-.3cm}
\caption{{\bf More indoor examples.} We show both {\bf Difficult} and {\bf Very Difficult} ScanNet indoor examples for which SuperGlue works well, and three {\bf Too Difficult} examples where it fails, either due to unlikely motion or lack of repeatable keypoints (last two rows). Correct matches are {\color{green}green} lines and mismatches are {\color{red}red} lines. See details in Section~\ref{sec:indoor}. }
\label{fig:supp-scannet-qualitative}
\end{figure*}

\begin{figure*}[ht!]
\centering
\def\iwidth{0.31}
\begin{minipage}{\iwidth\textwidth}
    \centering
    \small{SuperPoint + NN + distance}
\end{minipage}%
\hspace{1mm}%
\begin{minipage}{\iwidth\textwidth}
    \centering
    \small{SuperPoint + NN + OANet}
\end{minipage}%
\hspace{1mm}%
\begin{minipage}{\iwidth\textwidth}
    \centering
    \small{SuperPoint + \b{SuperGlue}}
\end{minipage}%

\begin{minipage}{\iwidth\textwidth}
    \includegraphics[width=\linewidth]{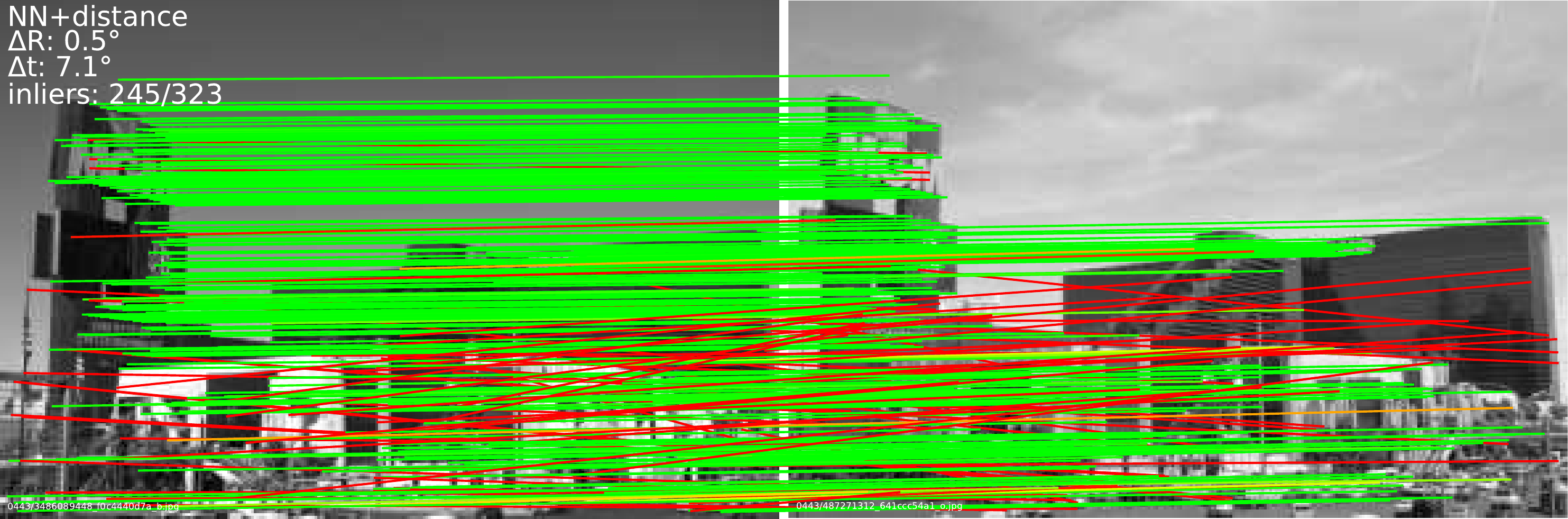}
    
    \vspace{.5mm}
    \includegraphics[width=\linewidth]{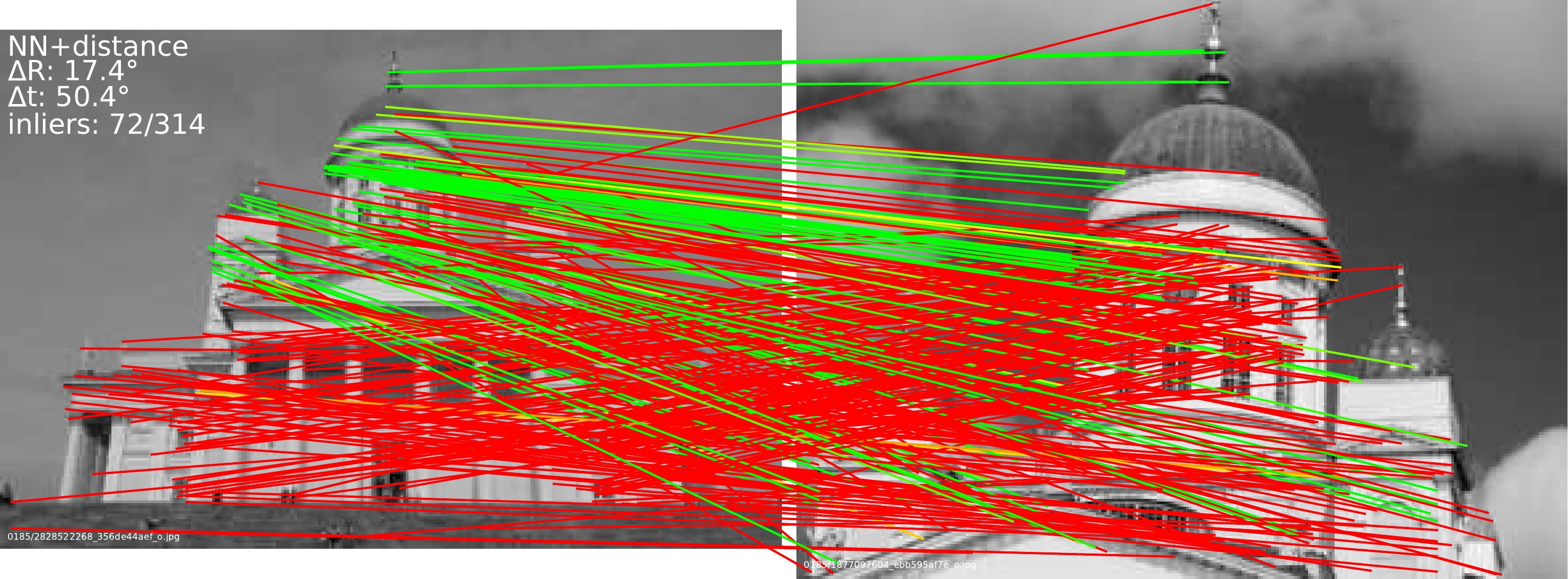}
    
    \vspace{.5mm}
    \includegraphics[width=\linewidth]{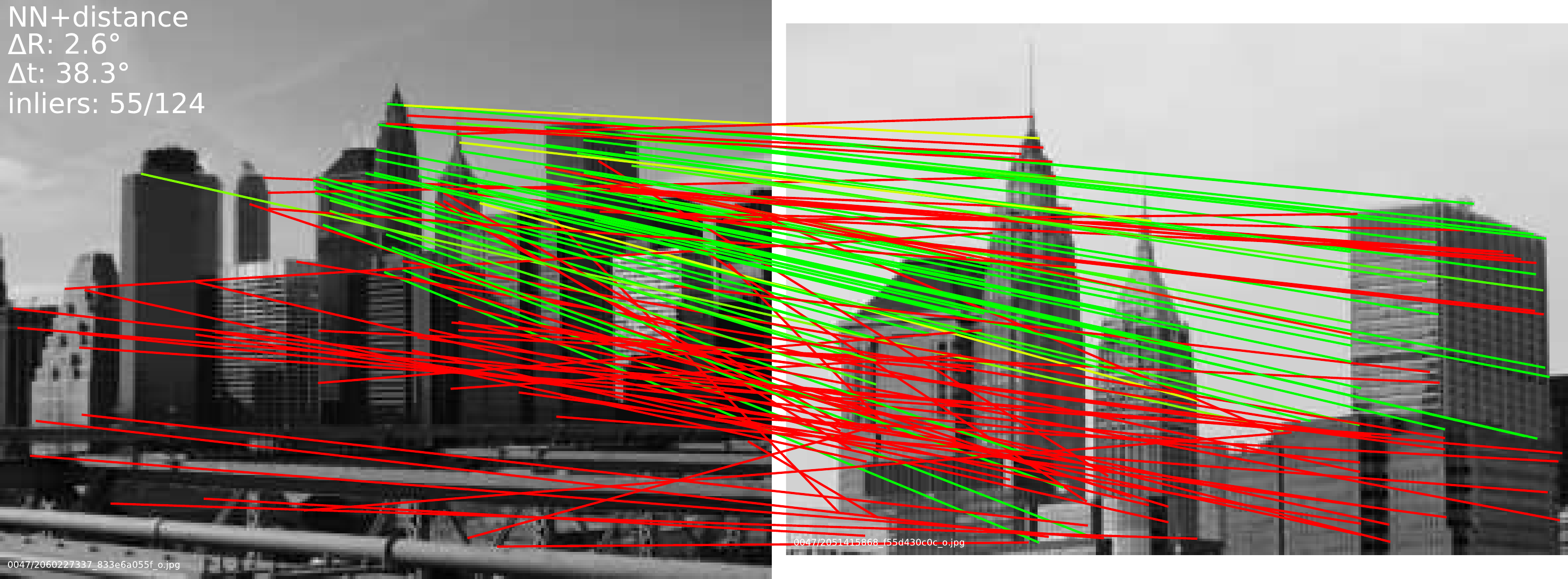}
    
    \vspace{.5mm}
    \includegraphics[width=\linewidth]{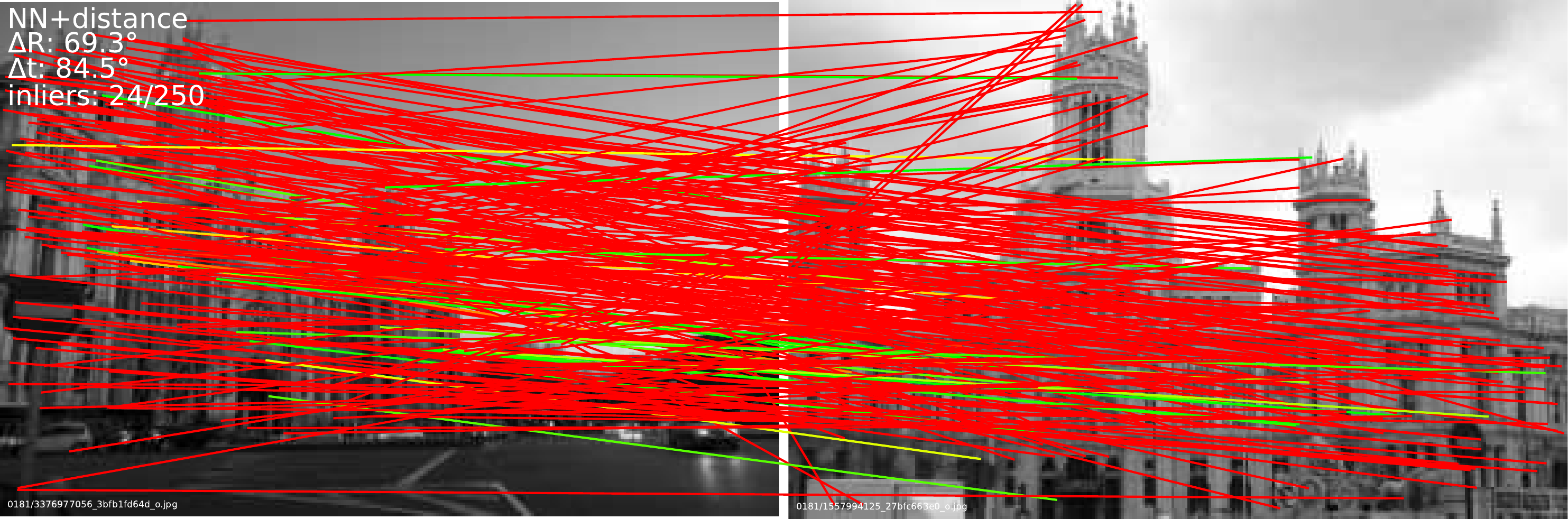}
\end{minipage}%
\hspace{1mm}%
\begin{minipage}{\iwidth\textwidth}
    \includegraphics[width=\linewidth]{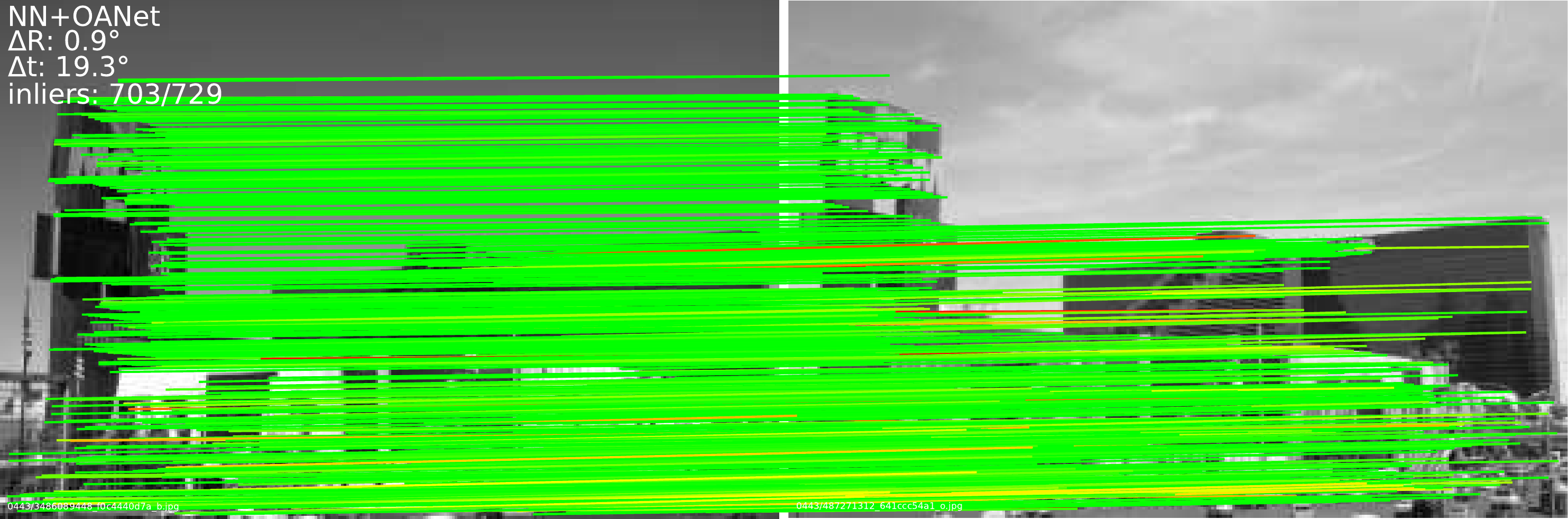}
    
    \vspace{.5mm}
    \includegraphics[width=\linewidth]{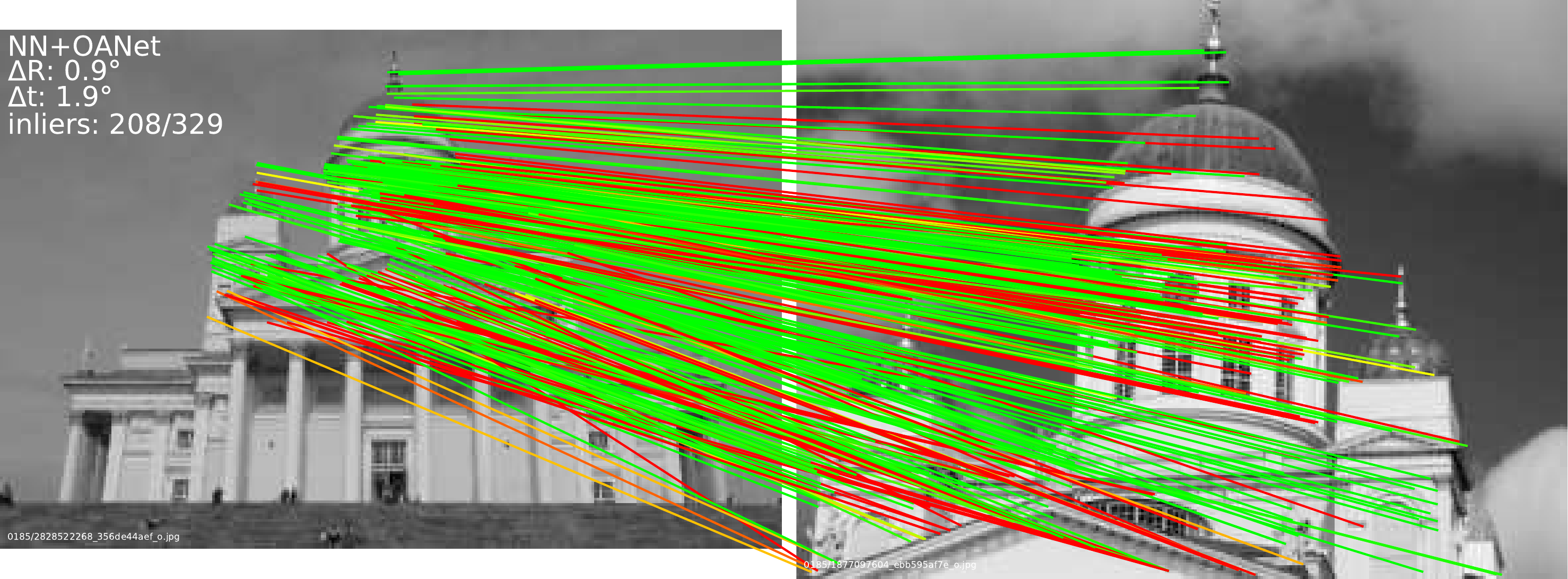}
    
    \vspace{.5mm}
    \includegraphics[width=\linewidth]{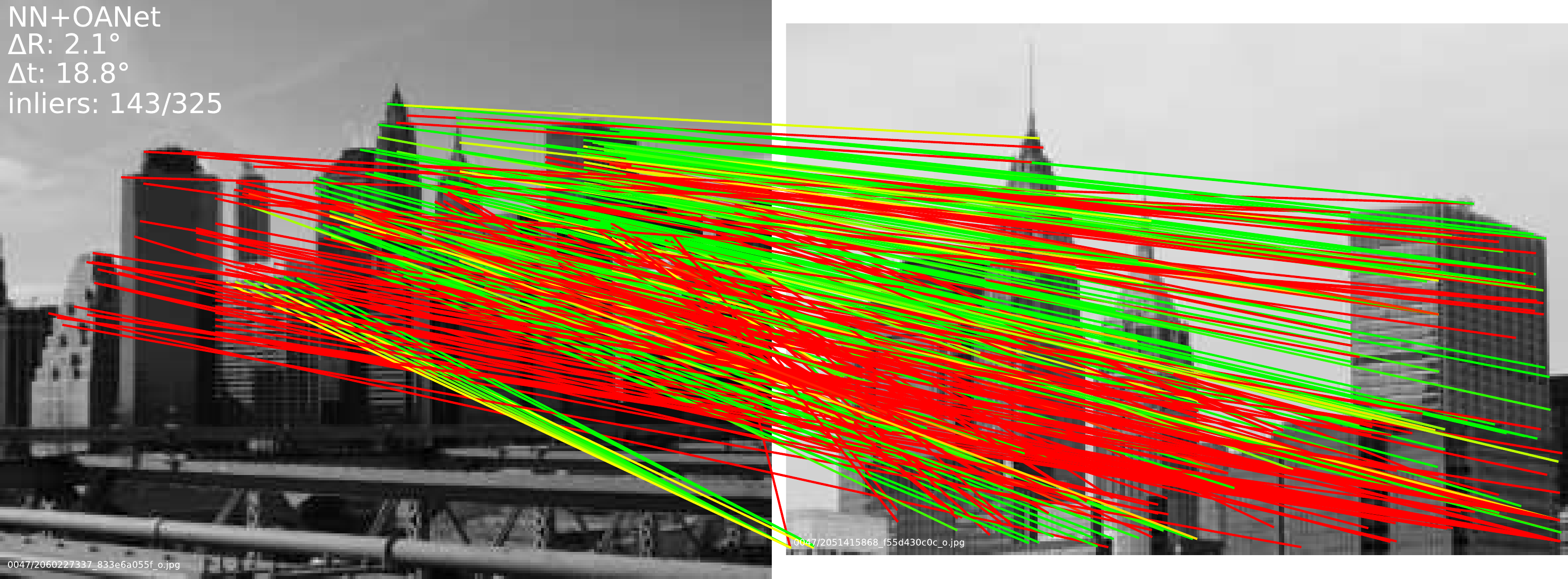}
    
    \vspace{.5mm}
    \includegraphics[width=\linewidth]{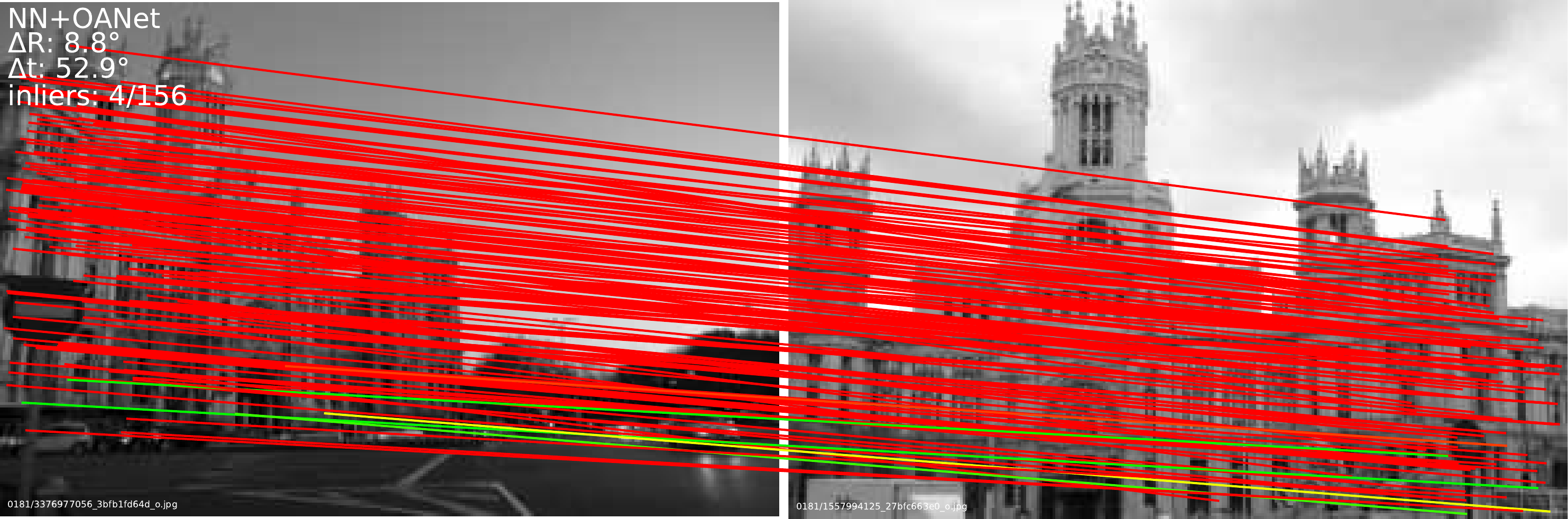}
\end{minipage}%
\hspace{1mm}%
\begin{minipage}{\iwidth\textwidth}
    \includegraphics[width=\linewidth]{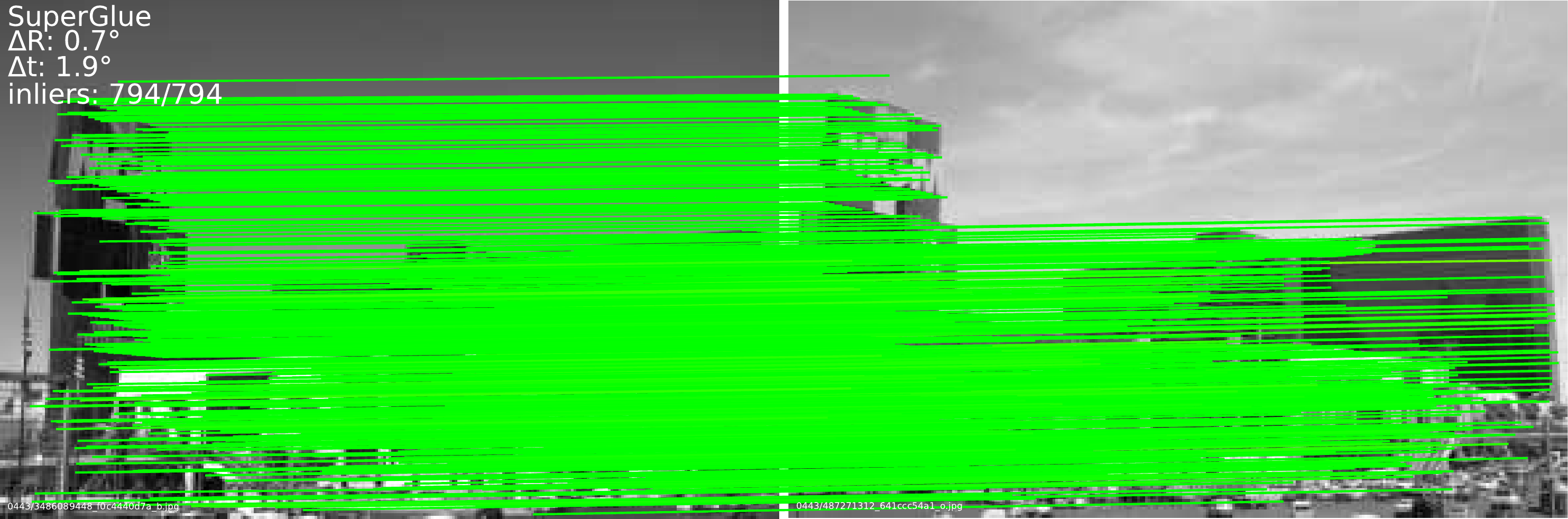}
    
    \vspace{.5mm}
    \includegraphics[width=\linewidth]{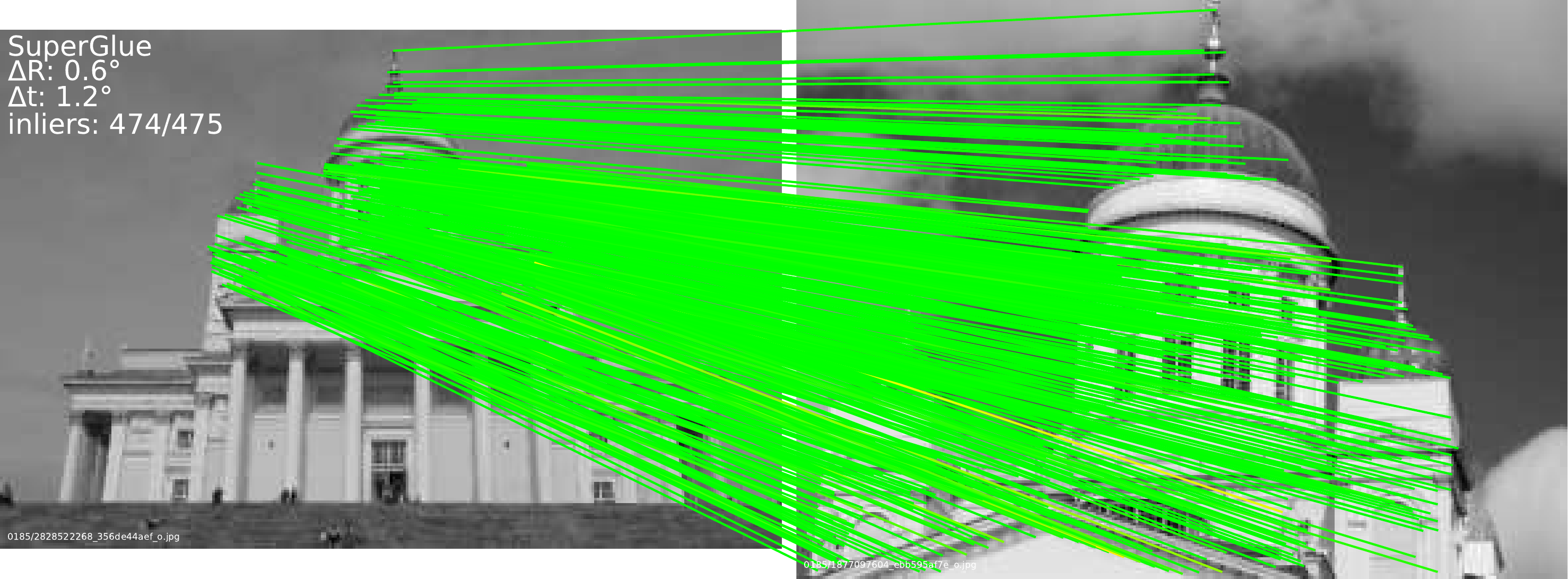}
    
    \vspace{.5mm}
    \includegraphics[width=\linewidth]{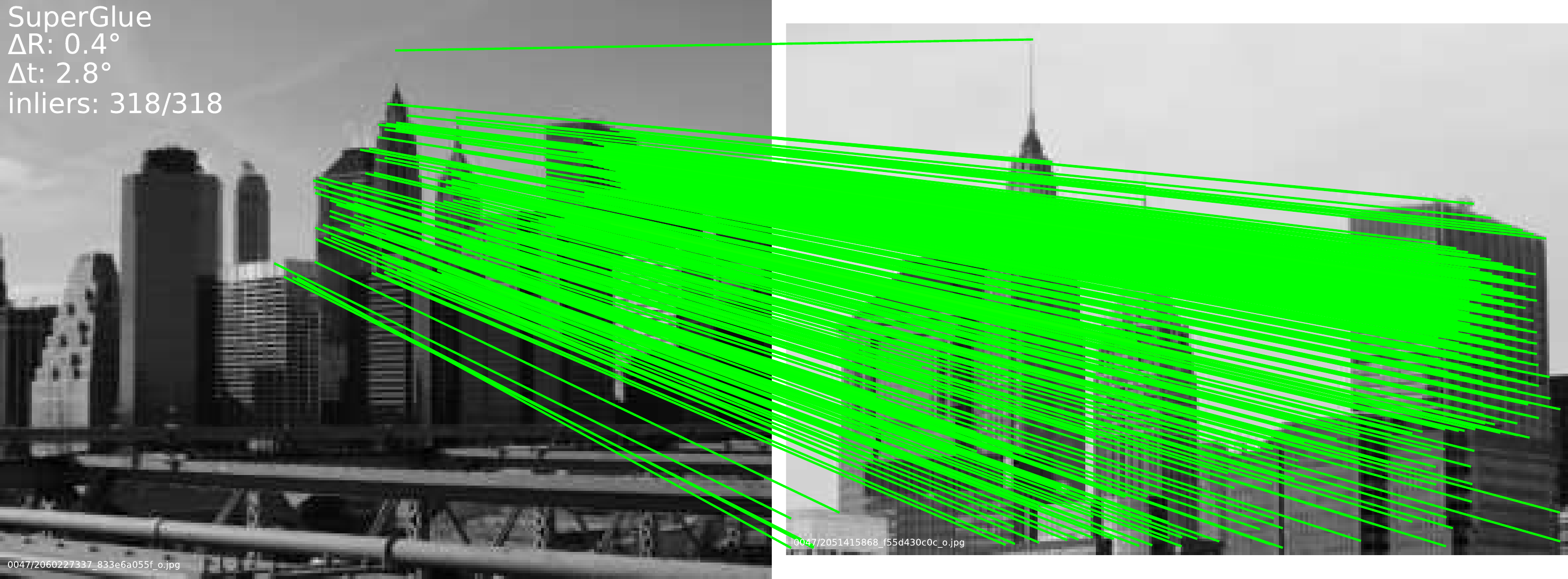}
    
    \vspace{.5mm}
    \includegraphics[width=\linewidth]{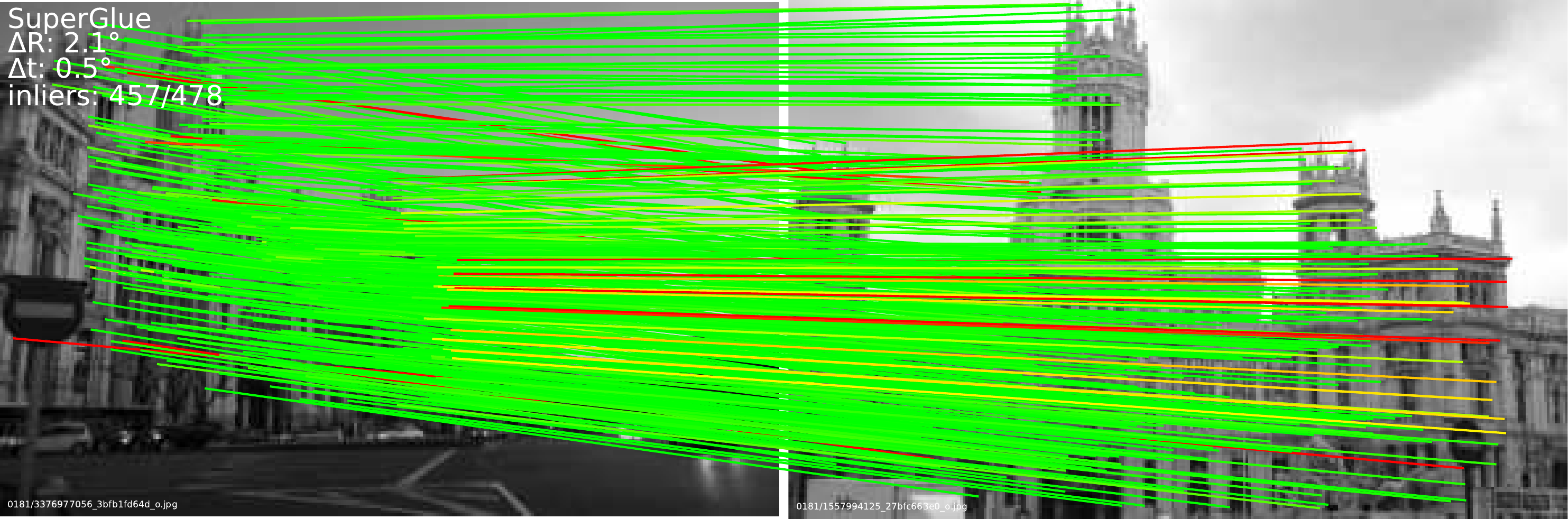}
\end{minipage}

\begin{minipage}{\iwidth\textwidth}
    \includegraphics[width=\linewidth]{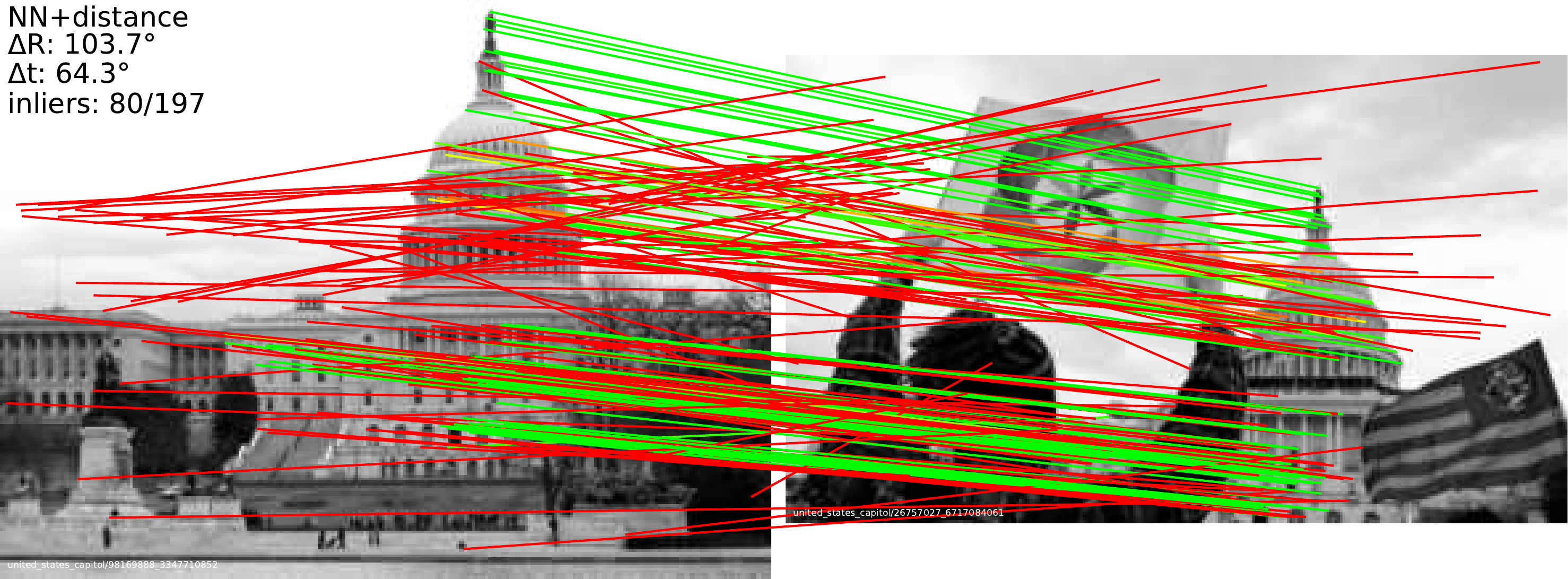}
    
    \vspace{.5mm}
    \includegraphics[width=\linewidth]{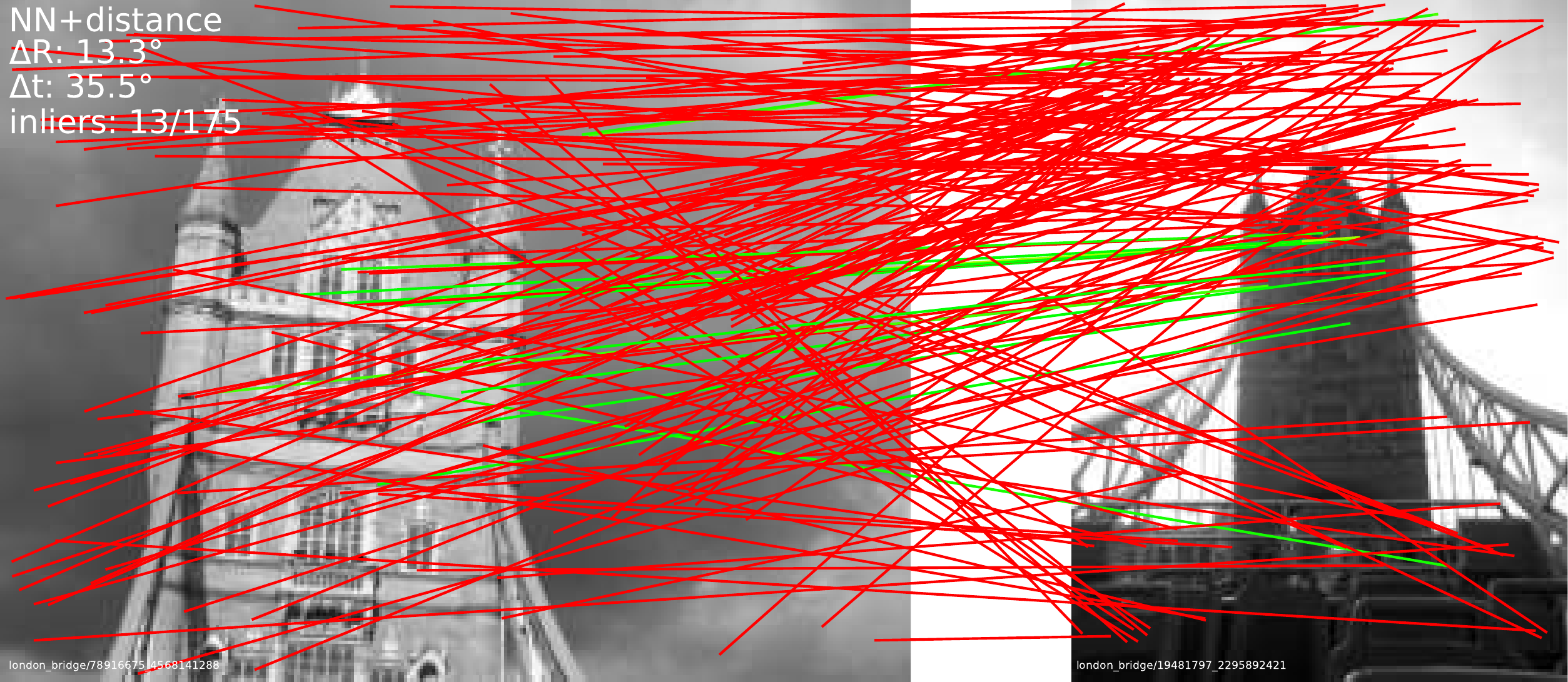}
    
    \vspace{.5mm}
    \includegraphics[width=\linewidth]{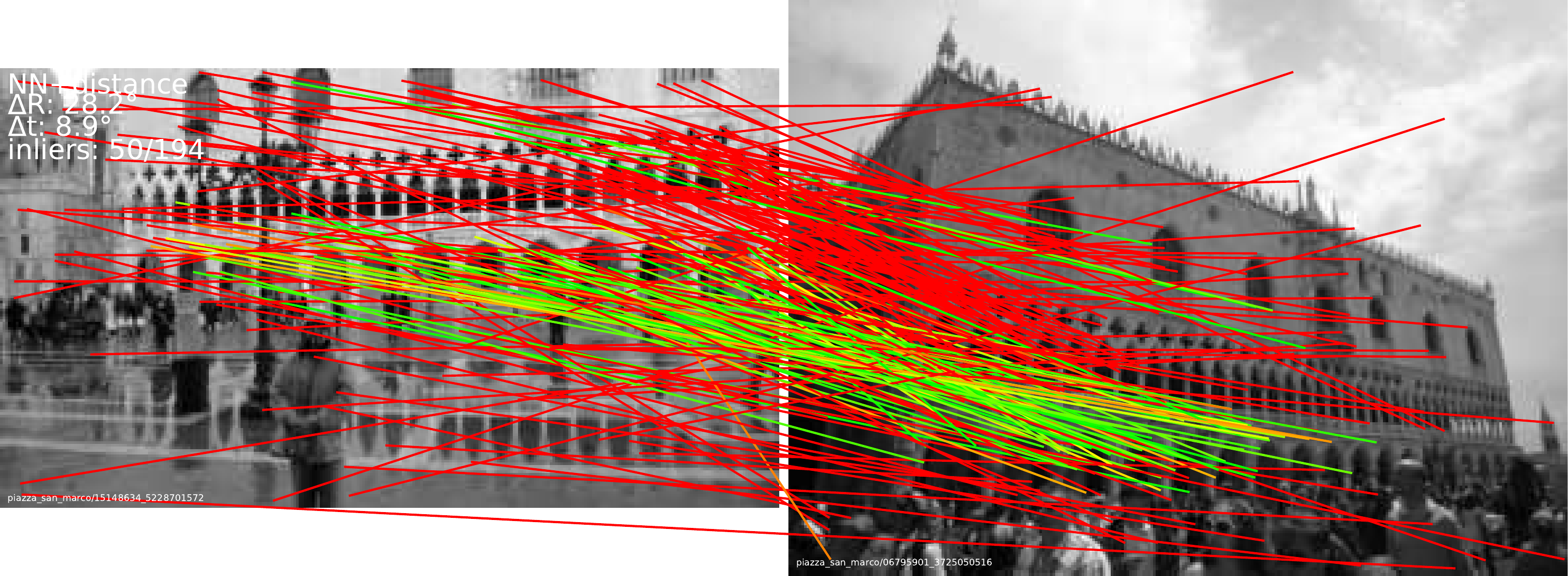}
    
    \vspace{.5mm}
    \includegraphics[width=\linewidth]{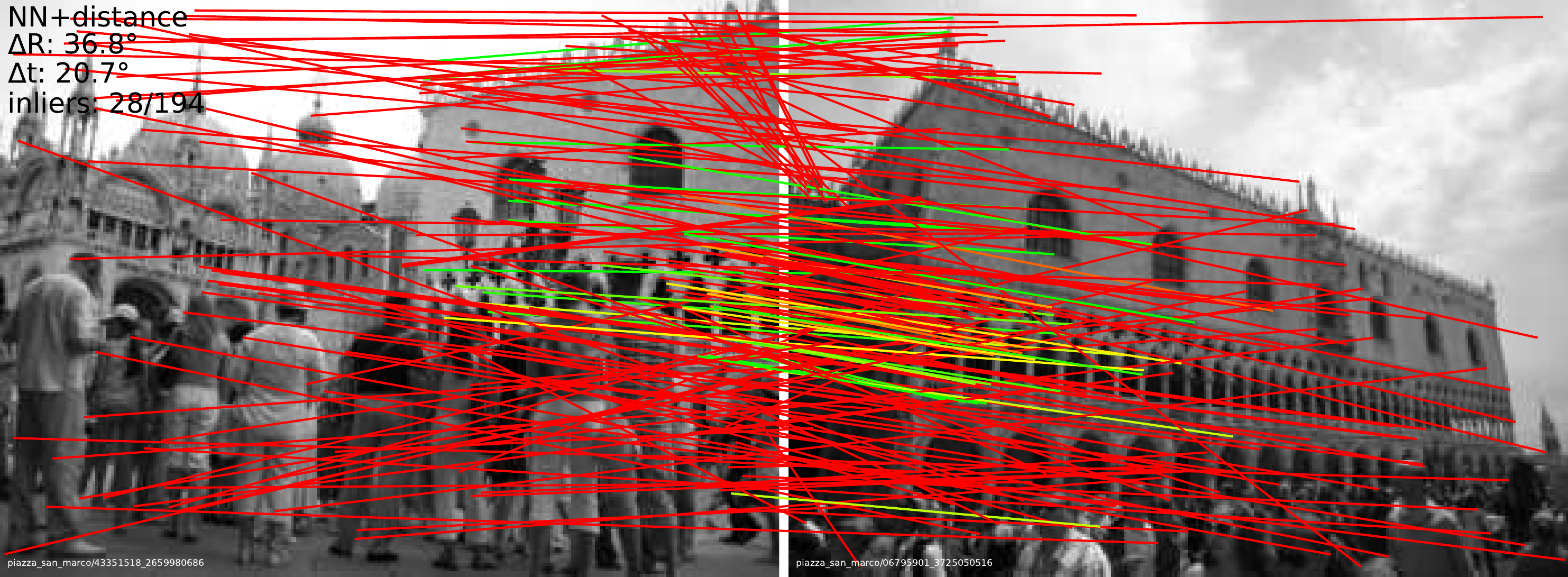}
    
    \vspace{.5mm}
    \includegraphics[width=\linewidth]{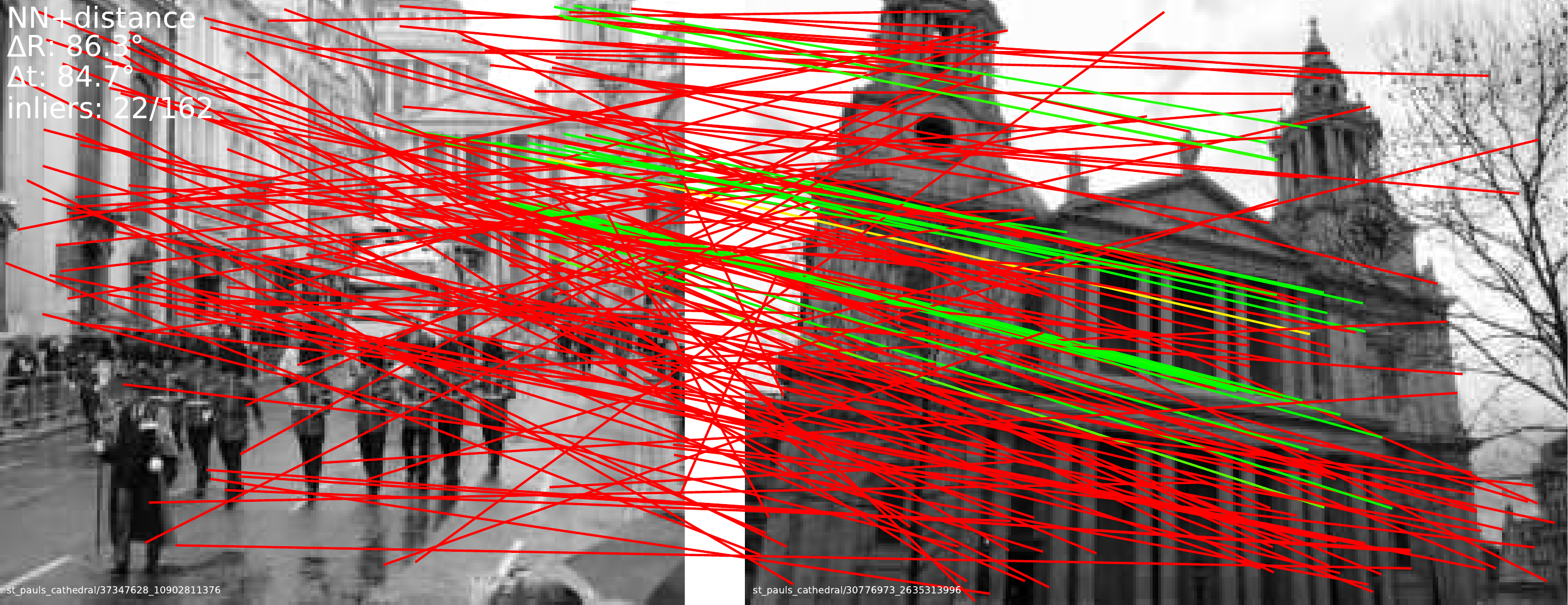}
    
    \vspace{.5mm}
    \includegraphics[width=\linewidth]{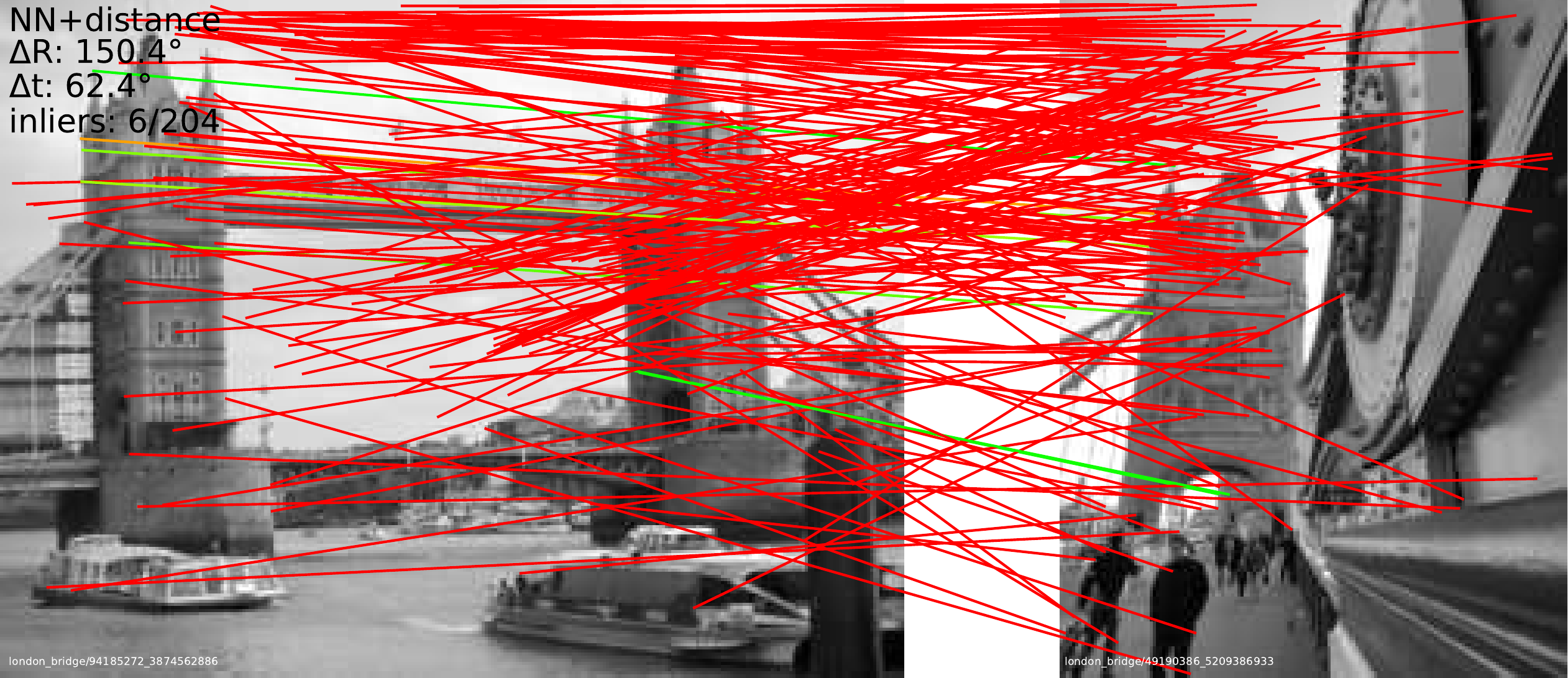}
\end{minipage}%
\hspace{1mm}%
\begin{minipage}{\iwidth\textwidth}
    \includegraphics[width=\linewidth]{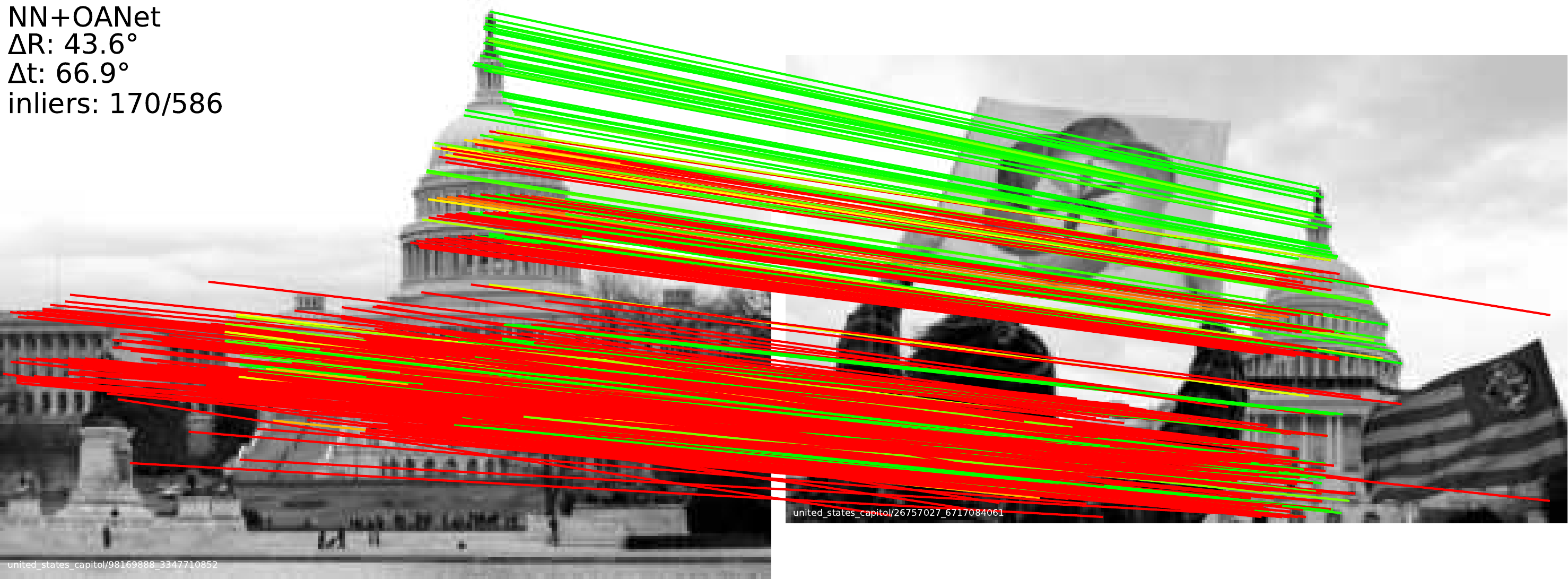}
    
    \vspace{.5mm}
    \includegraphics[width=\linewidth]{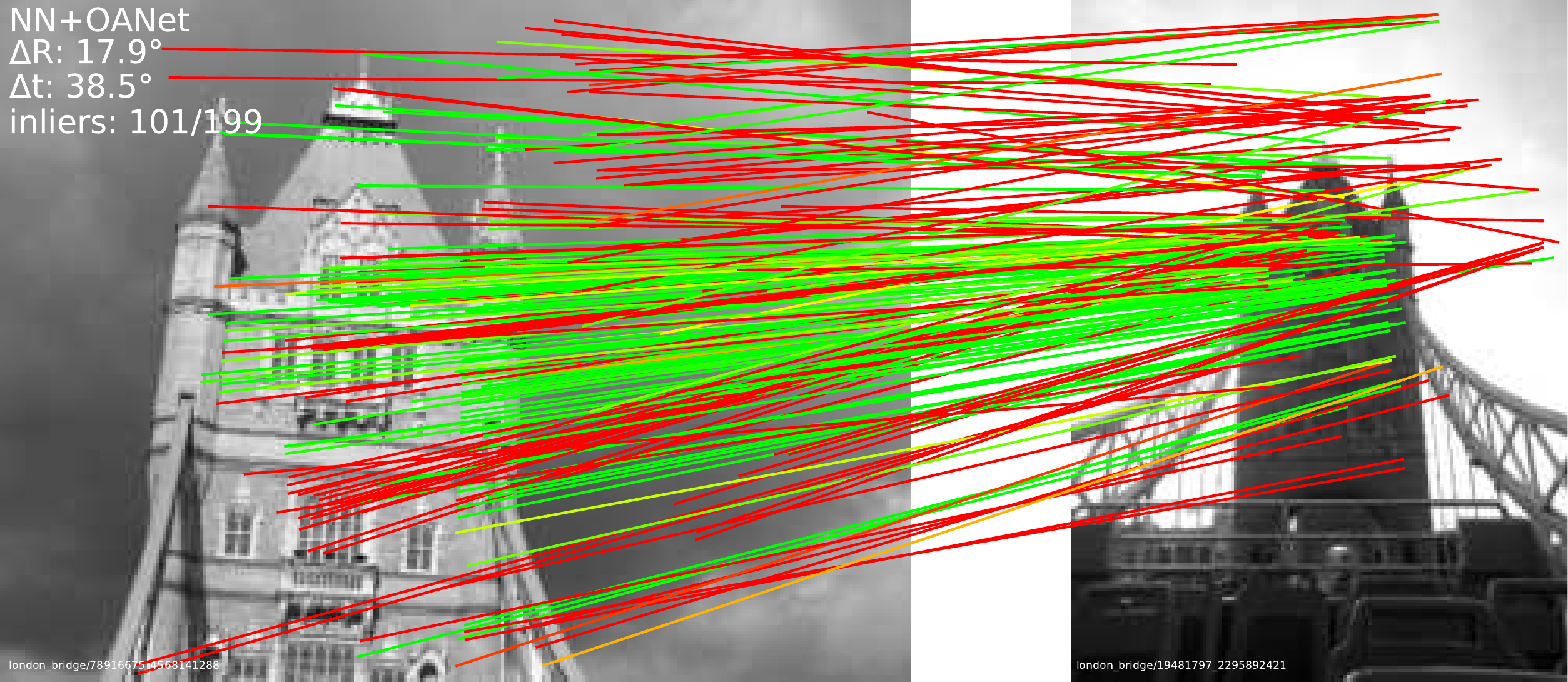}
    
    \vspace{.5mm}
    \includegraphics[width=\linewidth]{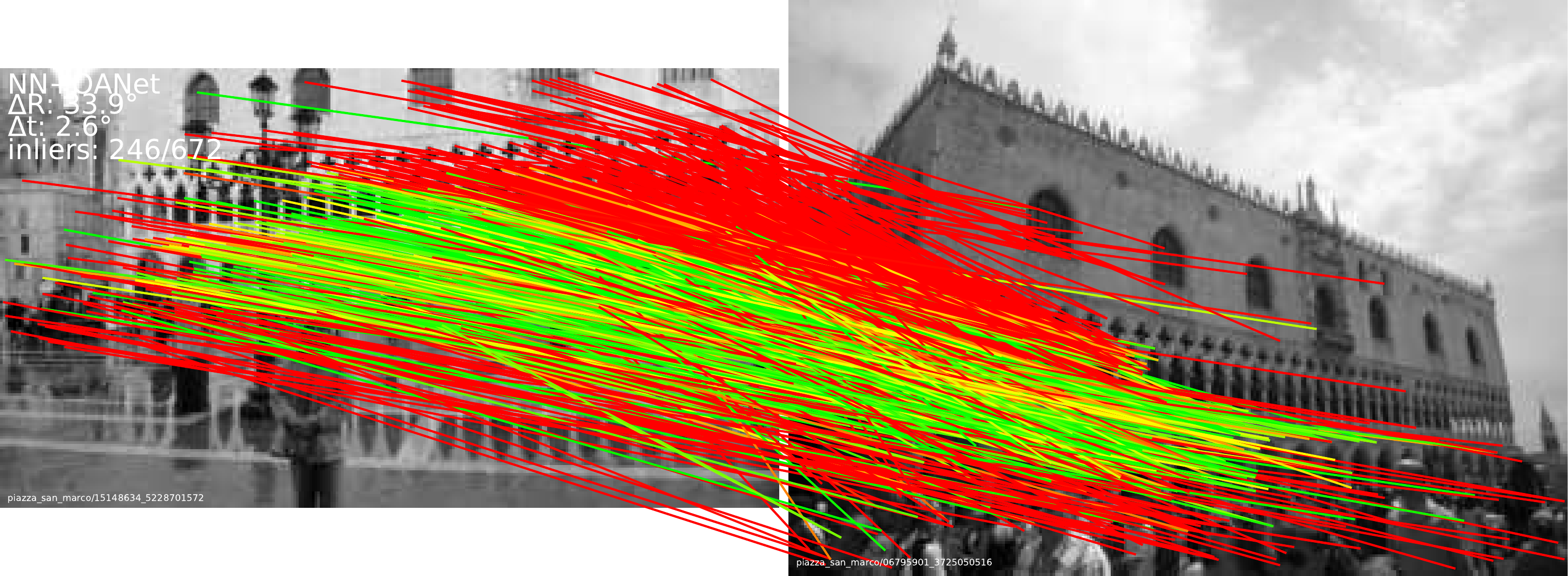}
    
    \vspace{.5mm}
    \includegraphics[width=\linewidth]{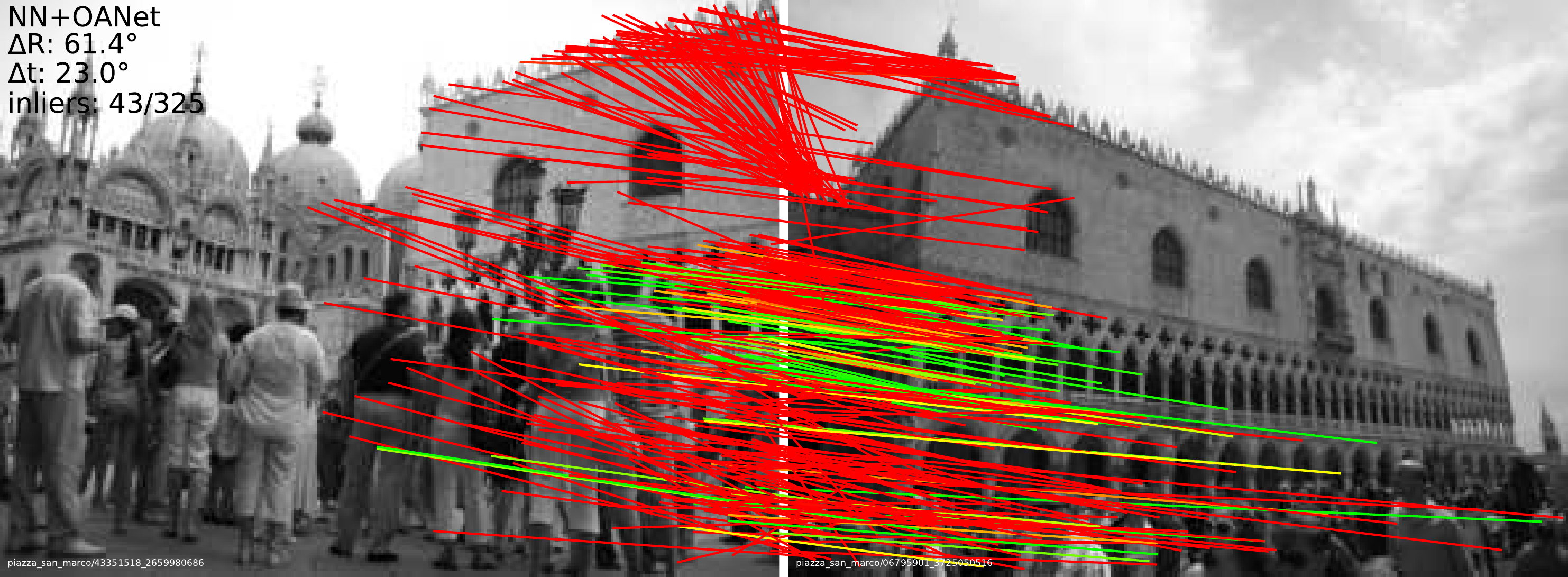}
    
    \vspace{.5mm}
    \includegraphics[width=\linewidth]{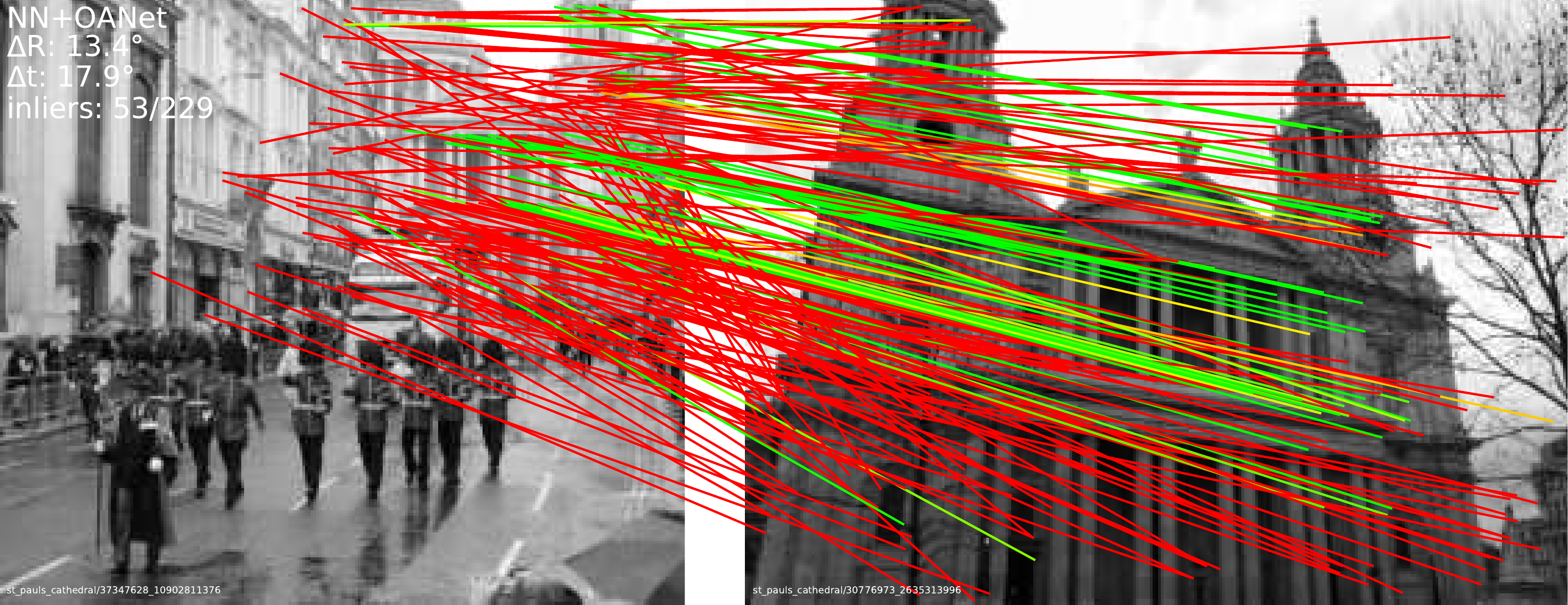}
    
    \vspace{.5mm}
    \includegraphics[width=\linewidth]{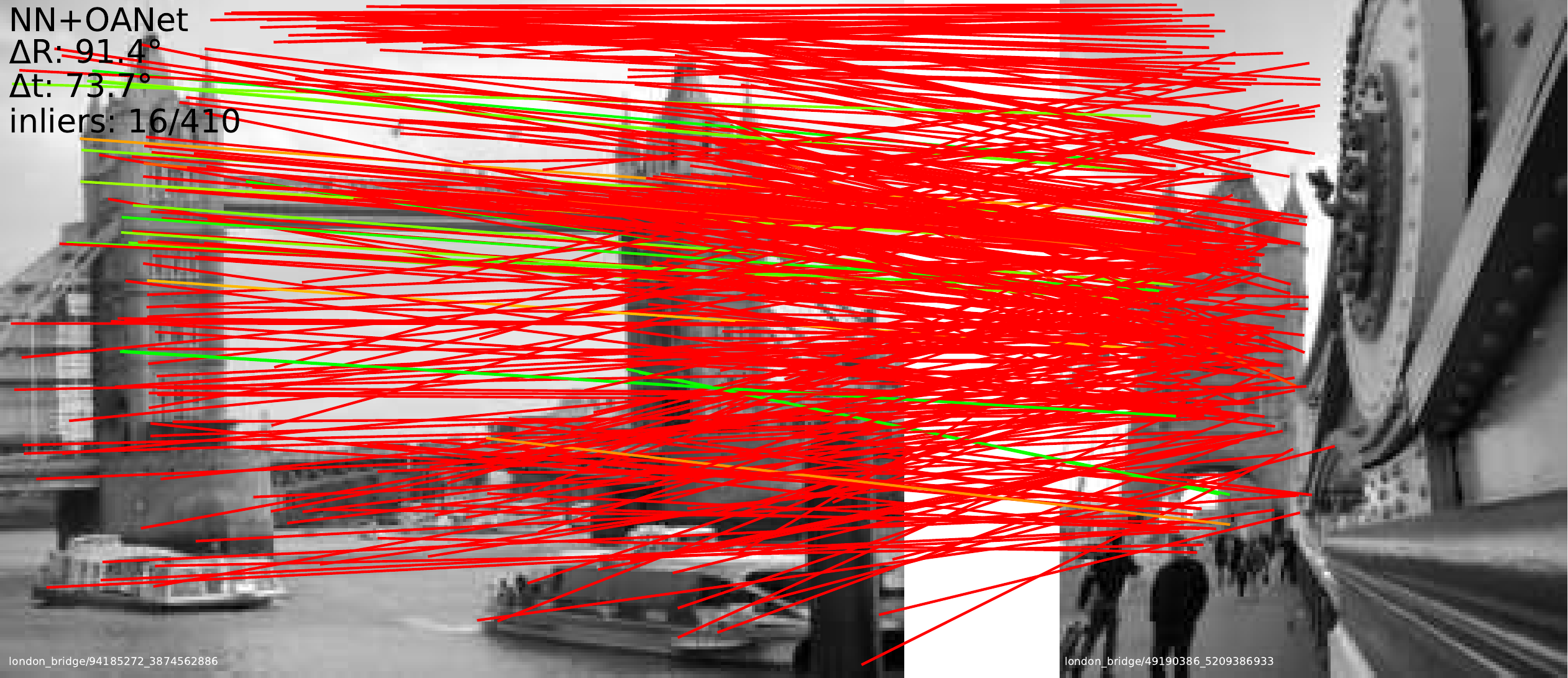}
\end{minipage}%
\hspace{1mm}%
\begin{minipage}{\iwidth\textwidth}
    \includegraphics[width=\linewidth]{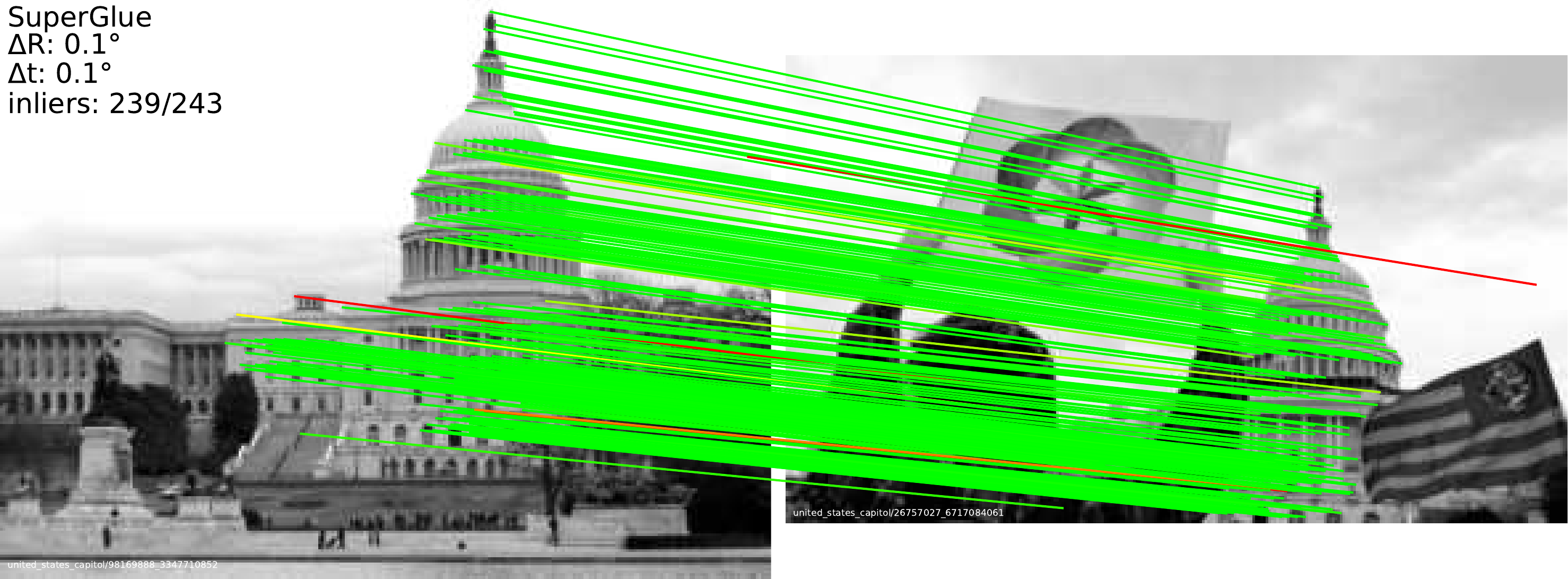}
    
    \vspace{.5mm}
    \includegraphics[width=\linewidth]{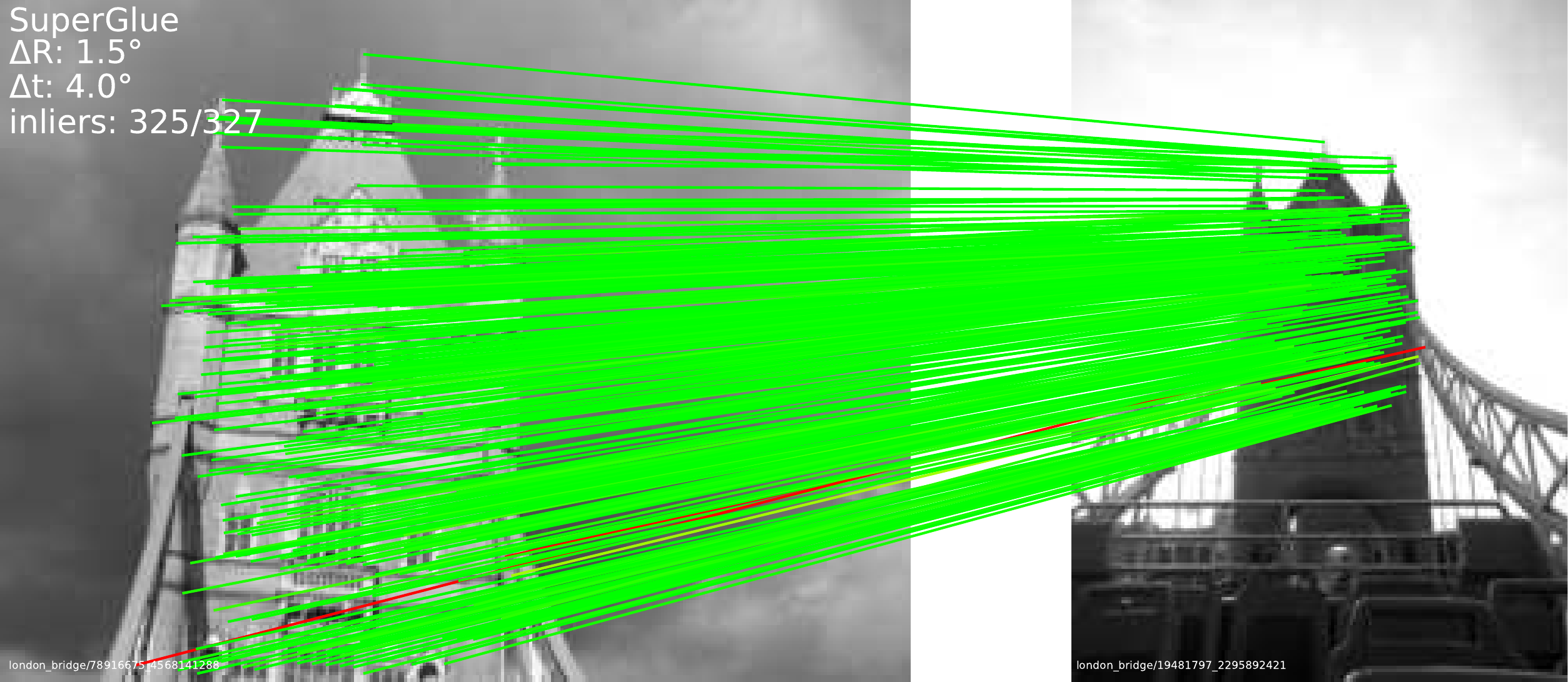}
    
    \vspace{.5mm}
    \includegraphics[width=\linewidth]{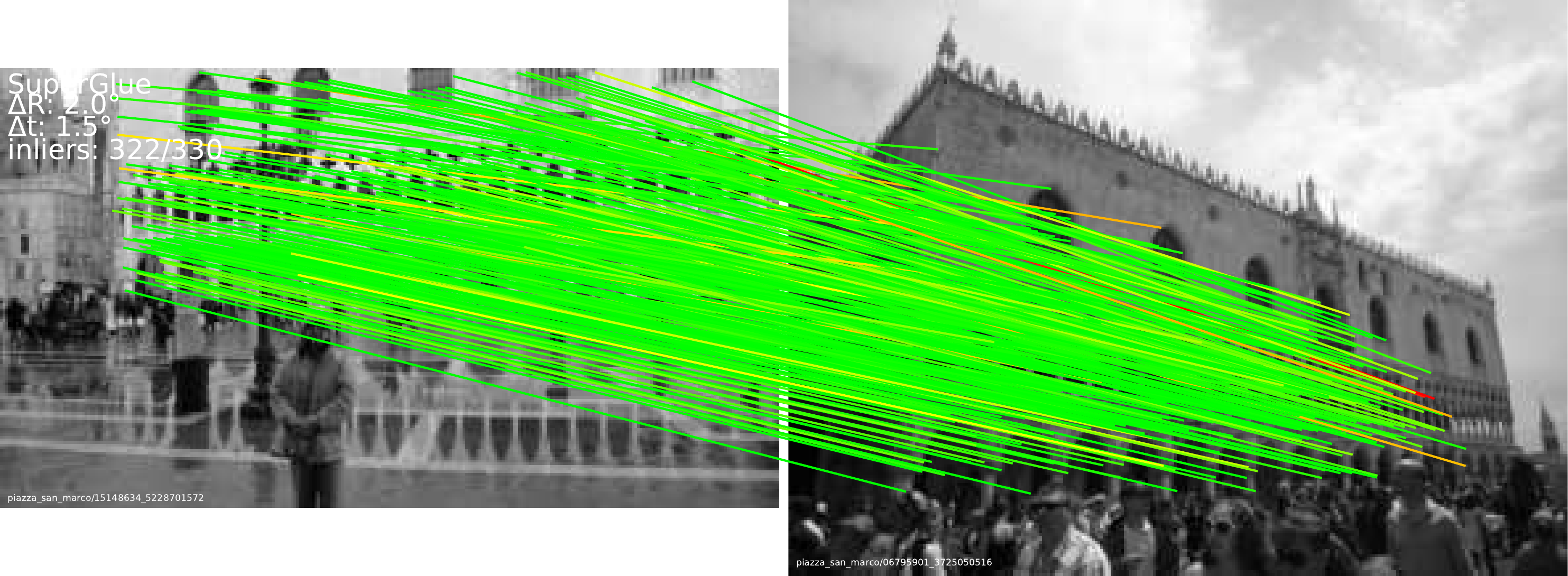}
    
    \vspace{.5mm}
    \includegraphics[width=\linewidth]{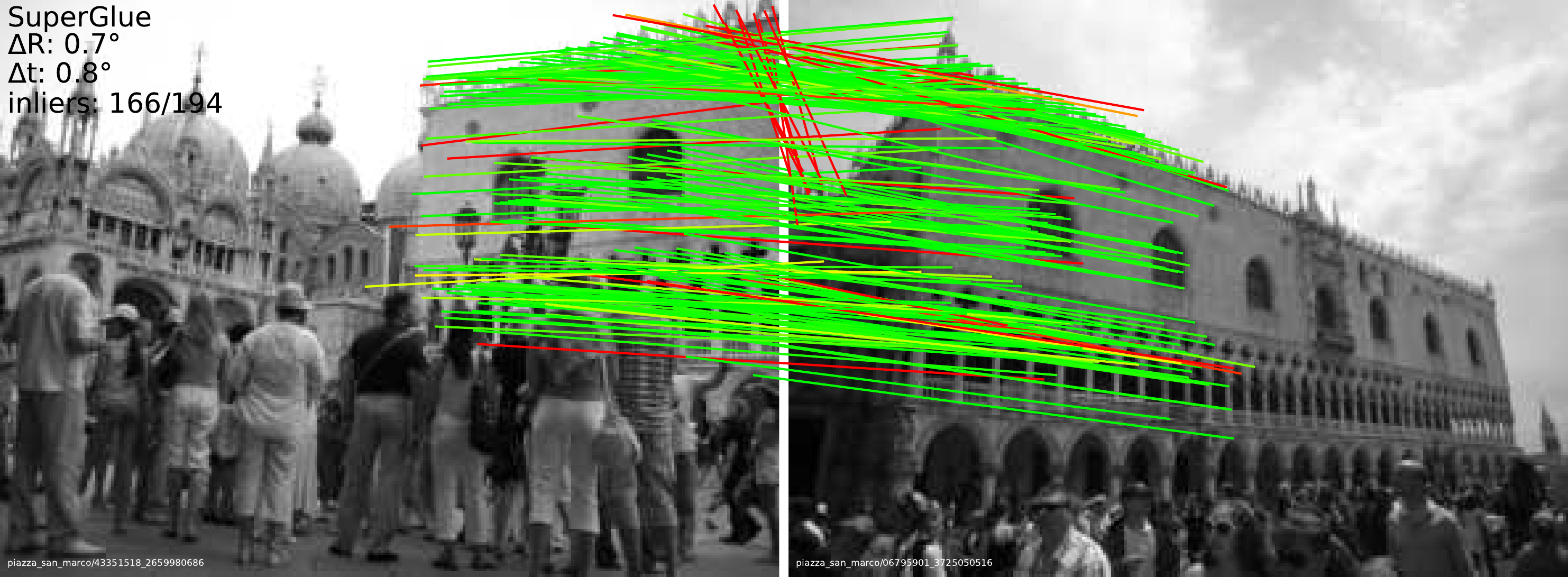}
    
    \vspace{.5mm}
    \includegraphics[width=\linewidth]{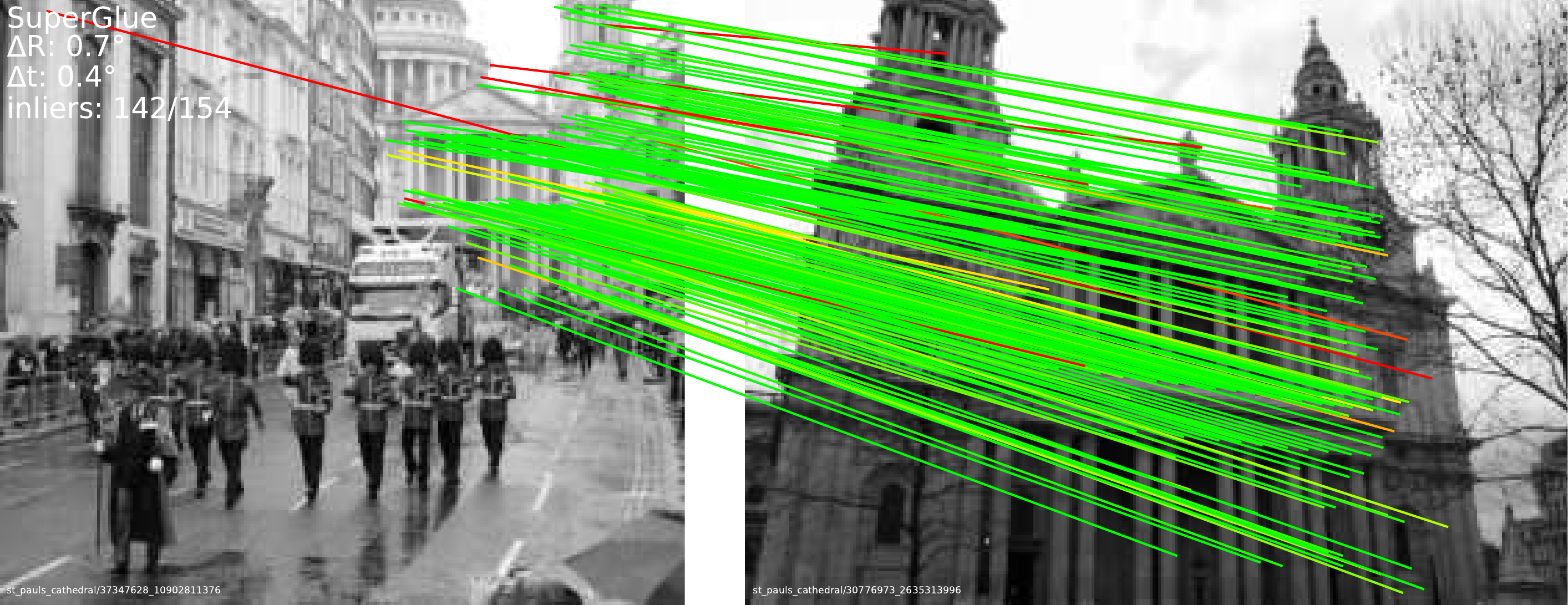}
    
    \vspace{.5mm}
    \includegraphics[width=\linewidth]{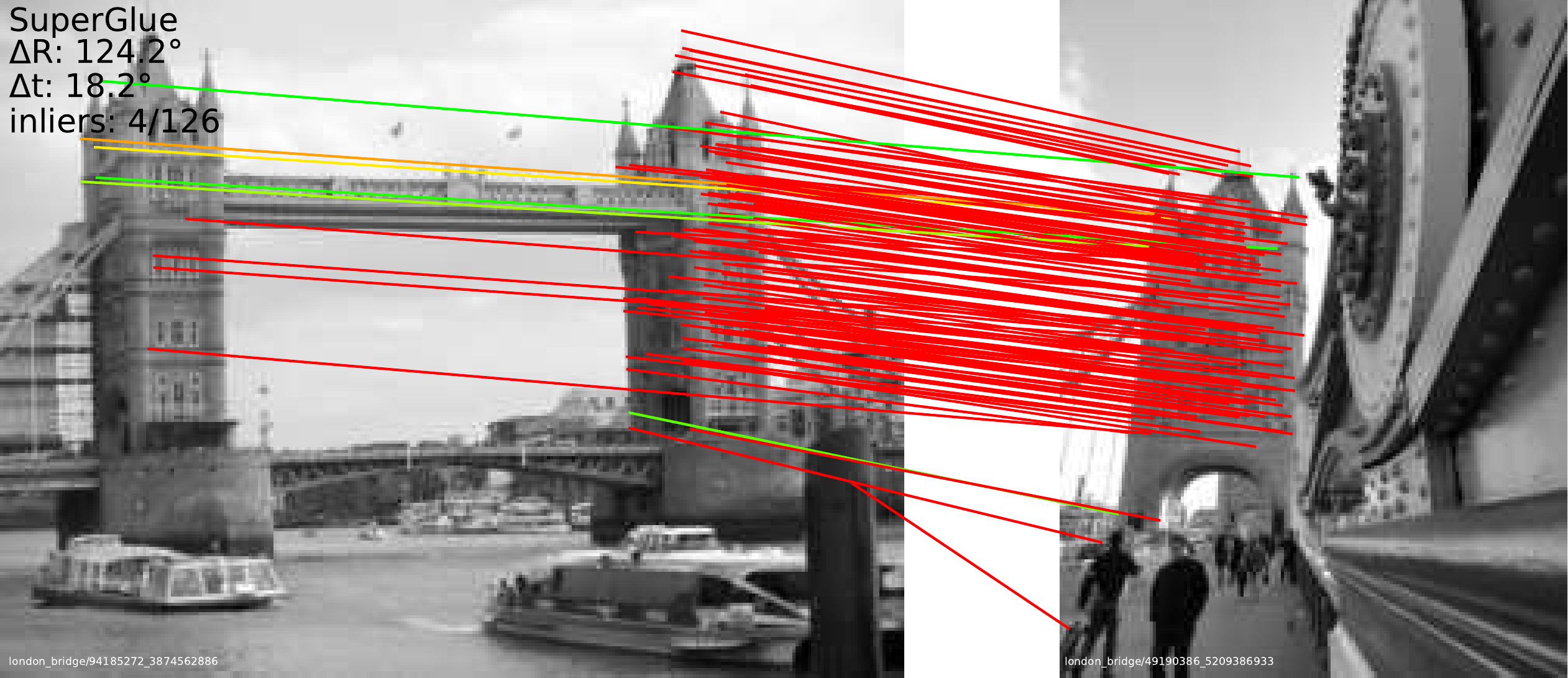}
\end{minipage}

\vspace{-.3cm}
\caption{{\bf More outdoor examples.} We show results on the MegaDepth validation and the PhotoTourism test sets. Correct matches are {\color{green}green} lines and mismatches are {\color{red}red} lines. The last row shows a failure case, where SuperGlue focuses on the incorrect self-similarity. See details in Section~\ref{sec:outdoor}.}
\label{fig:supp-outdoor-qualitative}
\end{figure*}

\begin{figure*}[ht!]
\centering
\def\iwidth{0.445}
\def\circOne{\raisebox{.5pt}{\textcircled{\raisebox{-.9pt} {1}}}}
\def\circTwo{\raisebox{.5pt}{\textcircled{\raisebox{-.9pt} {2}}}}
\def\circThree{\raisebox{.5pt}{\textcircled{\raisebox{-.9pt} {3}}}}
\begin{minipage}{\iwidth\textwidth}
    \centering
    \small{Estimated correspondences}
\end{minipage}%
\hspace{5mm}%
\begin{minipage}{\iwidth\textwidth}
    \centering
    \small{Keypoints}
\end{minipage}

\begin{minipage}{\iwidth\textwidth}
    \includegraphics[width=\linewidth]{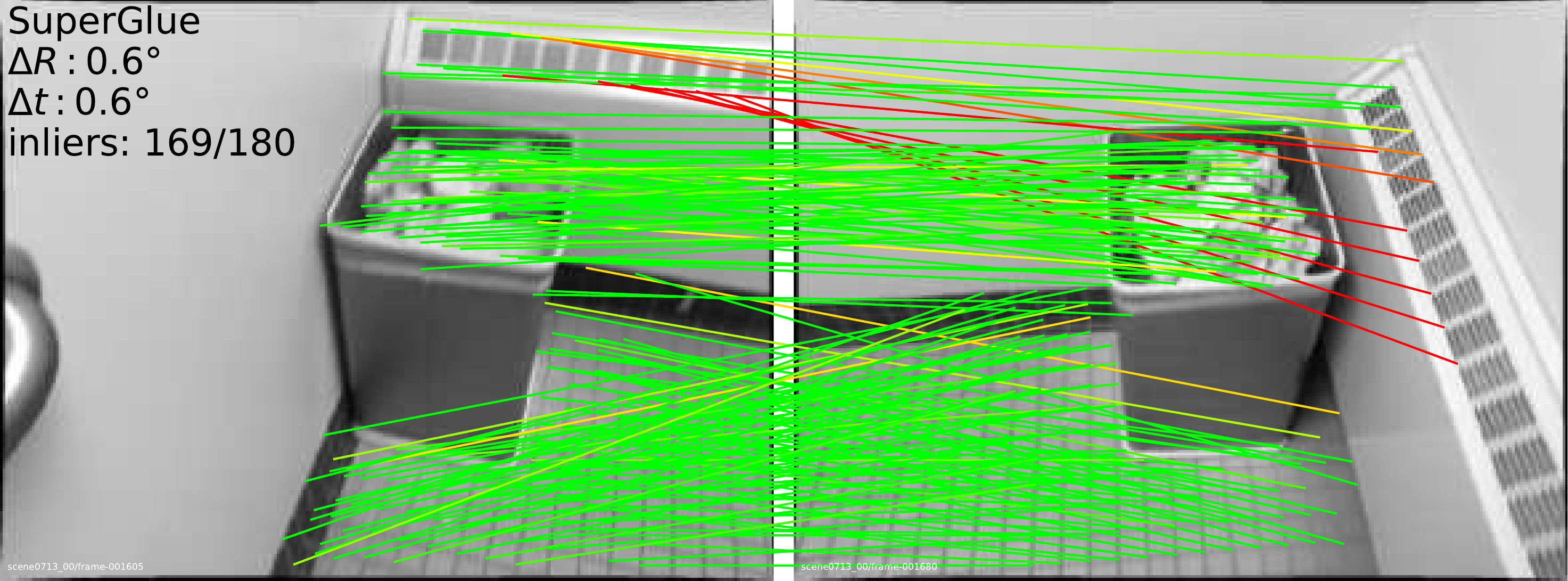}
\end{minipage}%
\hspace{5mm}%
\begin{minipage}{\iwidth\textwidth}
    \includegraphics[width=\linewidth]{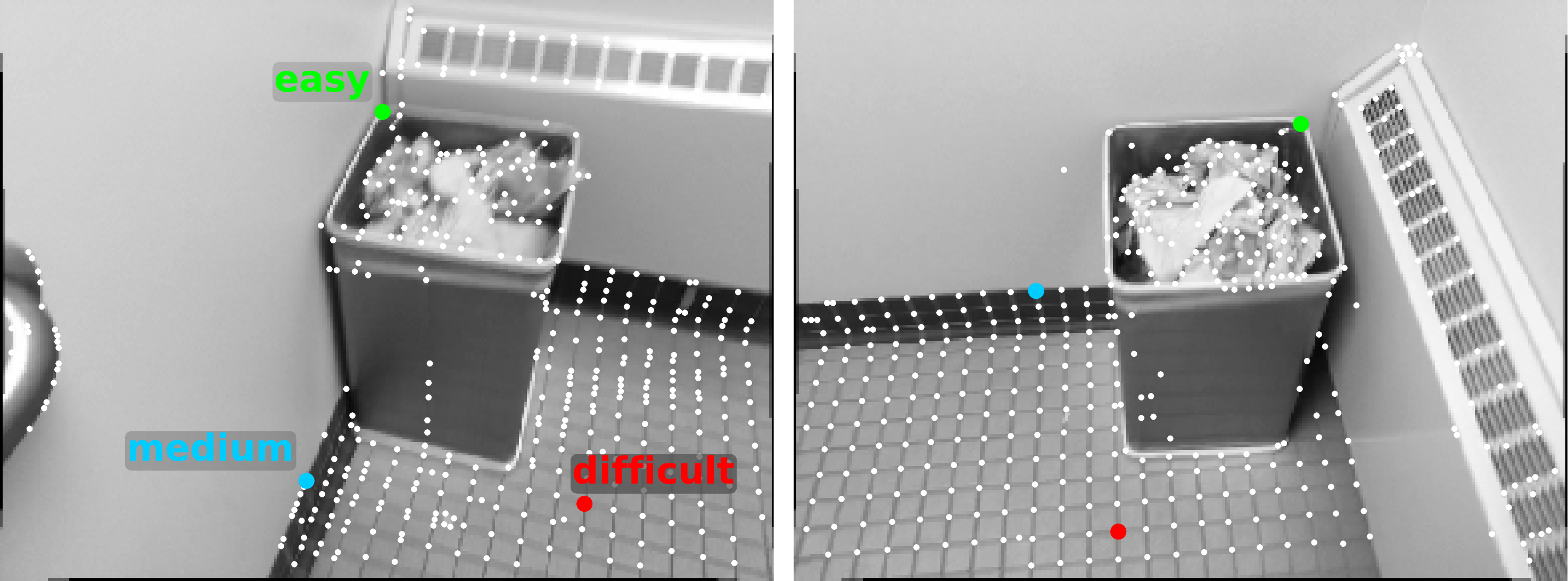}
\end{minipage}
\vspace{3mm}

\begin{minipage}{\iwidth\textwidth}
    \centering
    \small{Self-Attention}
\end{minipage}%
\hspace{5mm}%
\begin{minipage}{\iwidth\textwidth}
    \centering
    \small{Cross-Attention}
\end{minipage}
\vspace{.6mm}

\begin{minipage}{\iwidth\textwidth}
    \includegraphics[width=\linewidth]{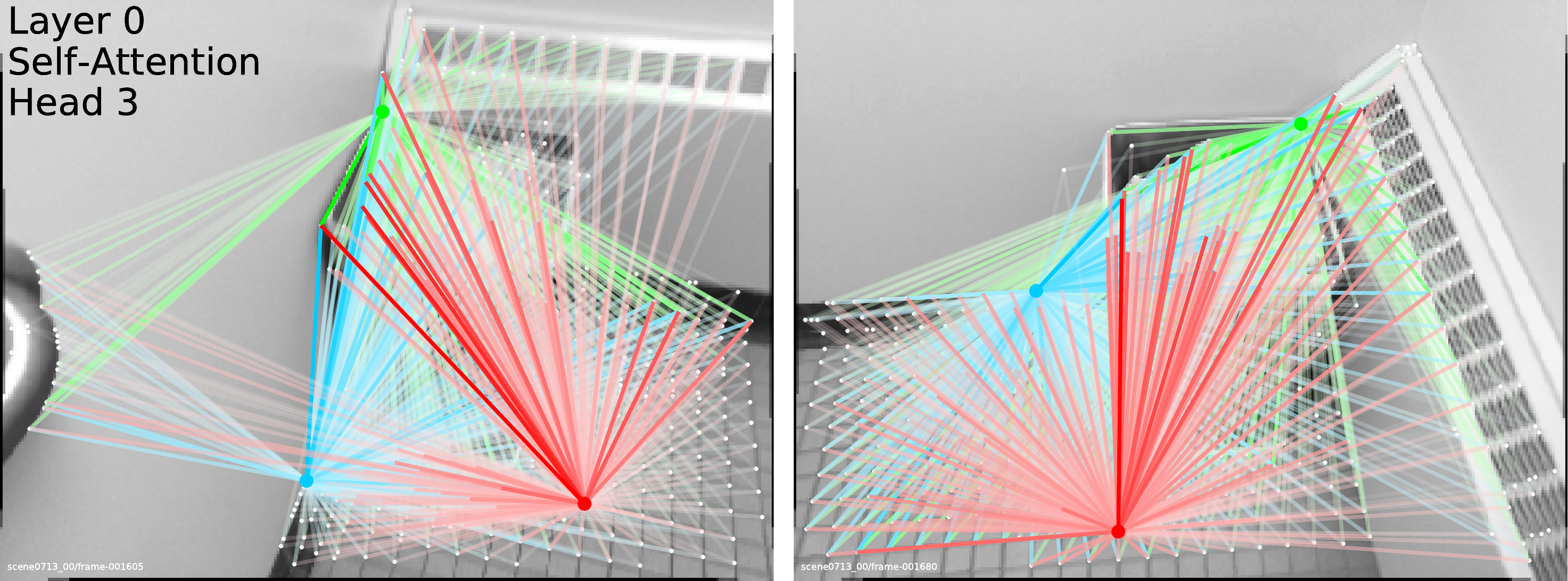}
    
    \vspace{.5mm}
    \includegraphics[width=\linewidth]{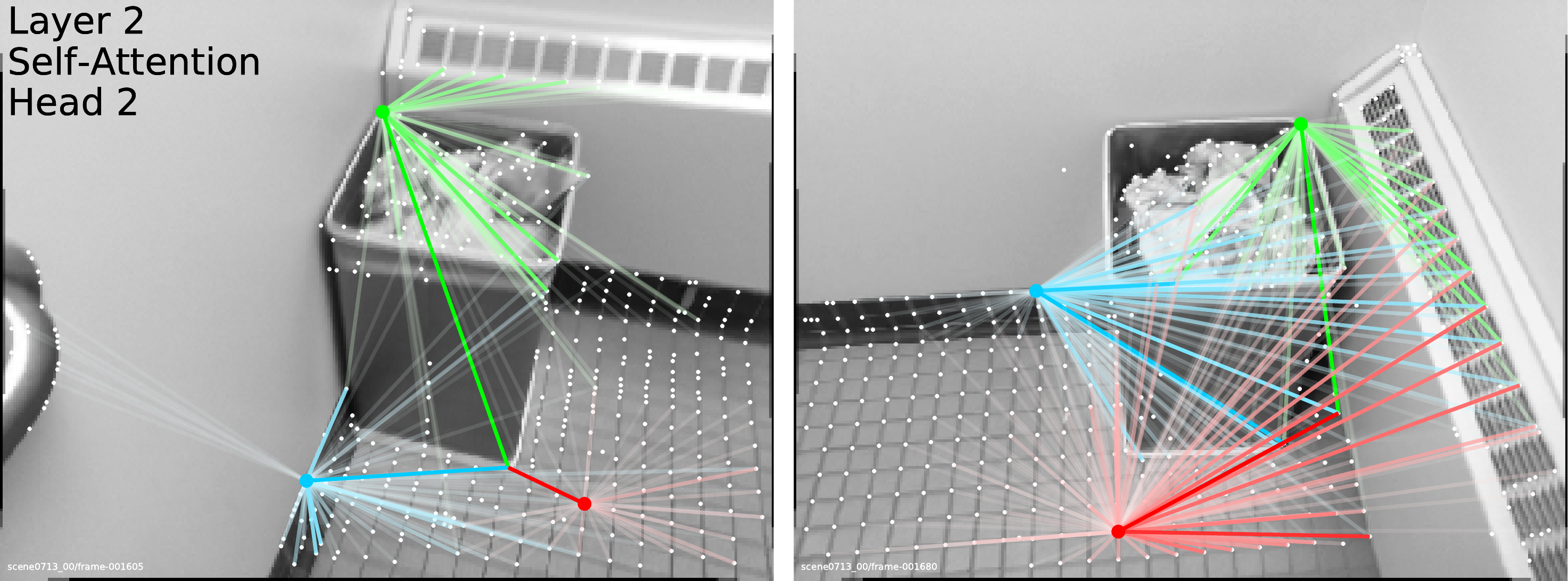}
    
    \vspace{.5mm}
    \includegraphics[width=\linewidth]{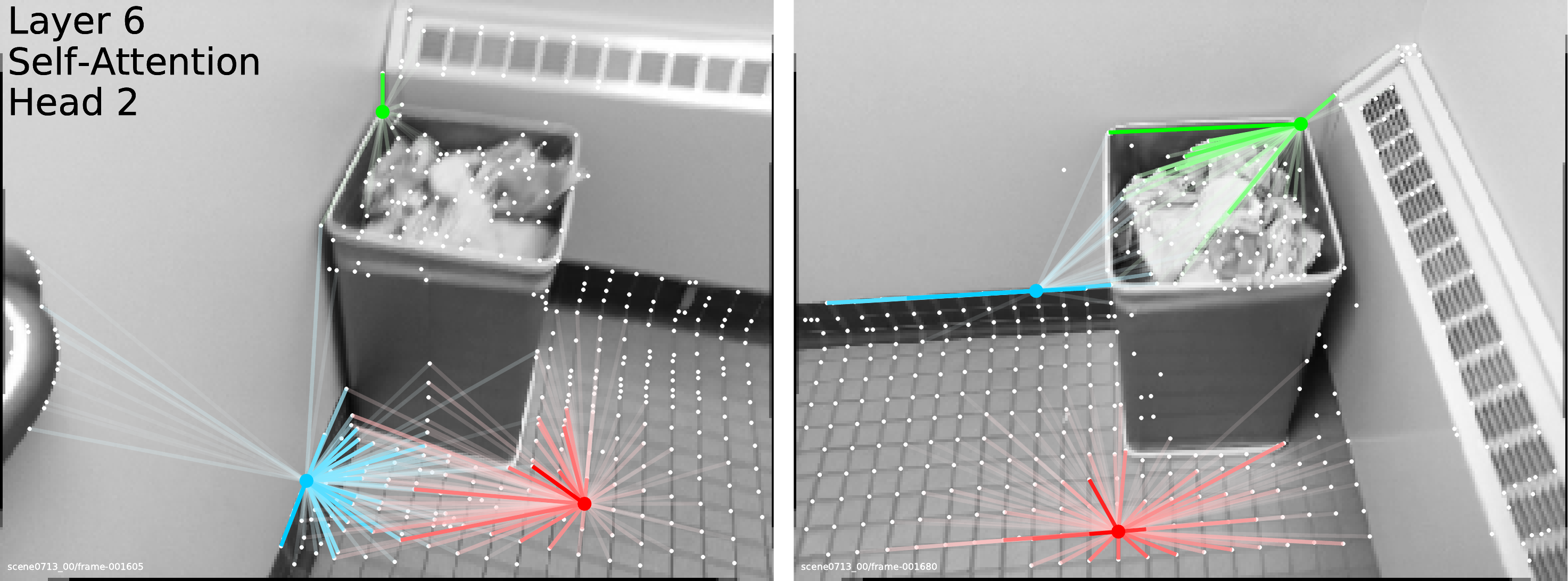}
    
    \vspace{.5mm}
    \includegraphics[width=\linewidth]{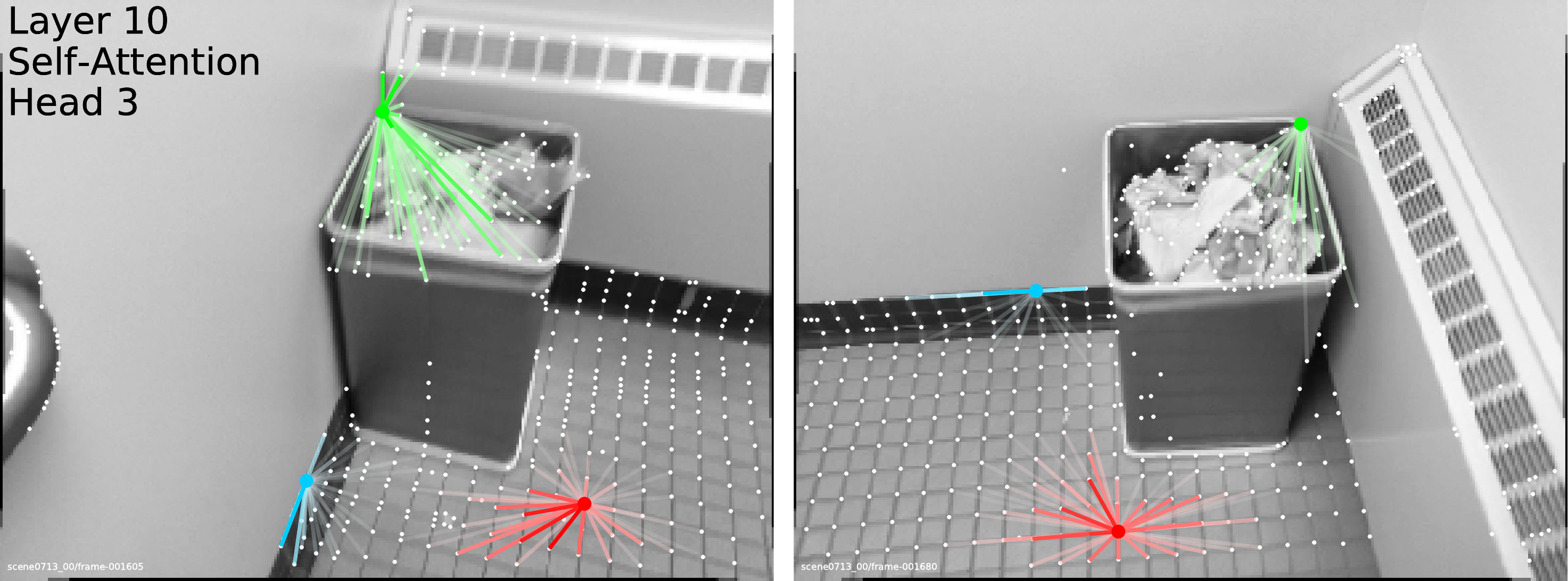}

    \vspace{.5mm}
    \includegraphics[width=\linewidth]{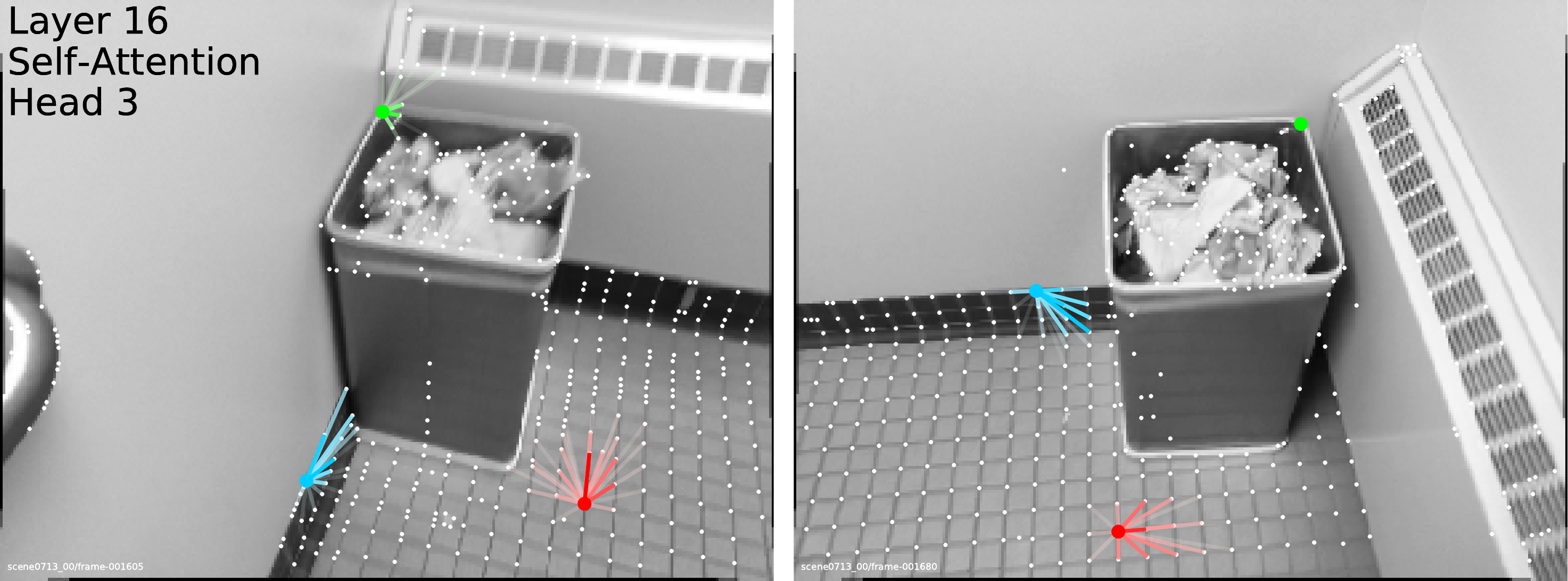}
\end{minipage}%
\hspace{5mm}%
\begin{minipage}{\iwidth\textwidth}
    \includegraphics[width=\linewidth]{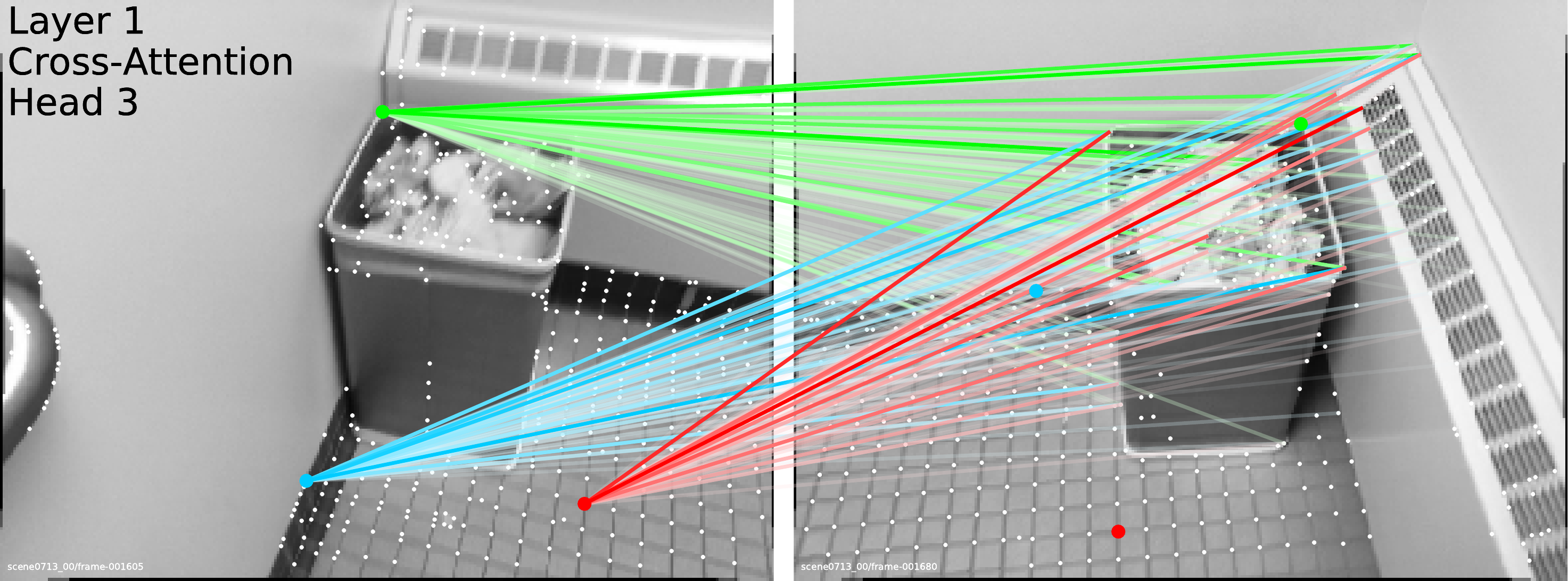}
    
    \vspace{.5mm}
    \includegraphics[width=\linewidth]{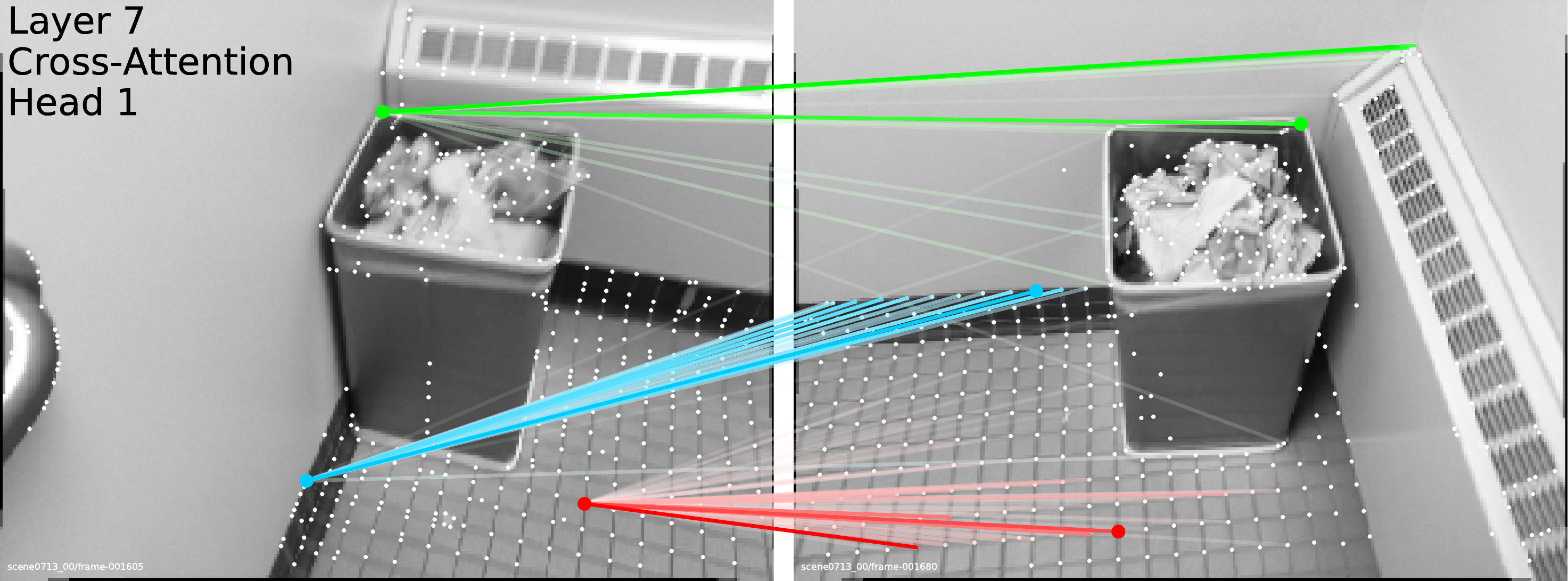}
    
    \vspace{.5mm}
    \includegraphics[width=\linewidth]{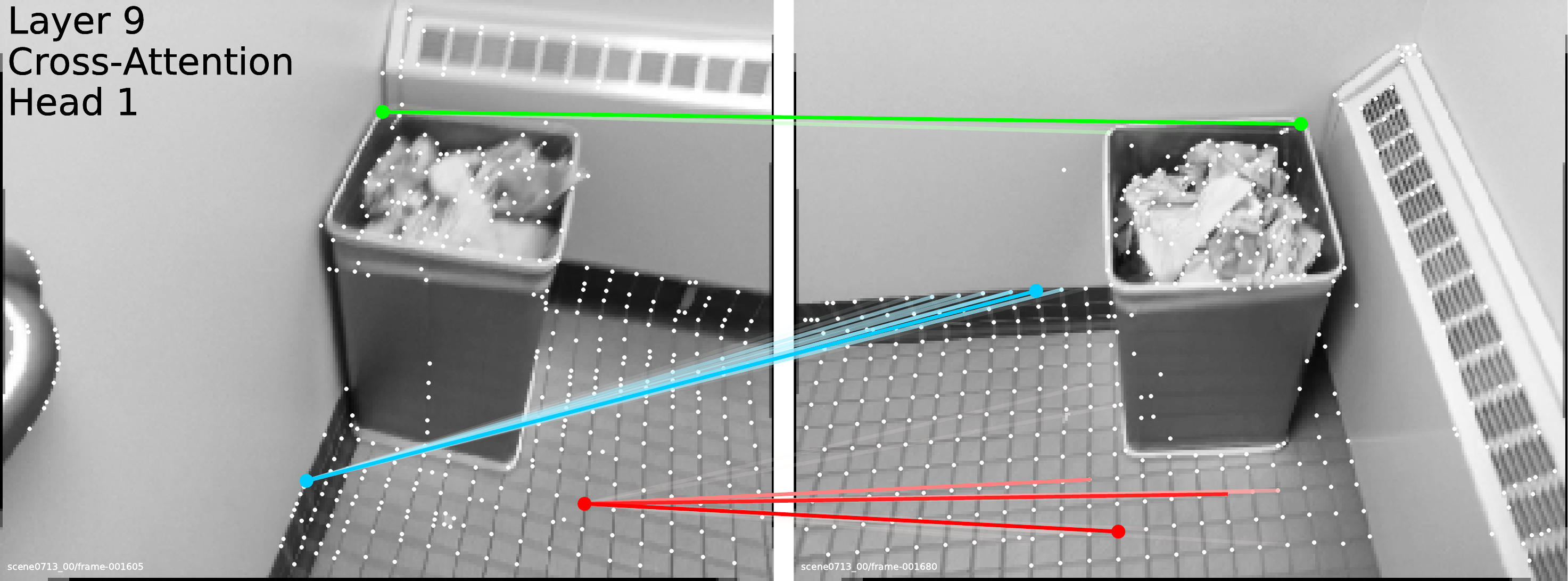}
    
    \vspace{.5mm}
    \includegraphics[width=\linewidth]{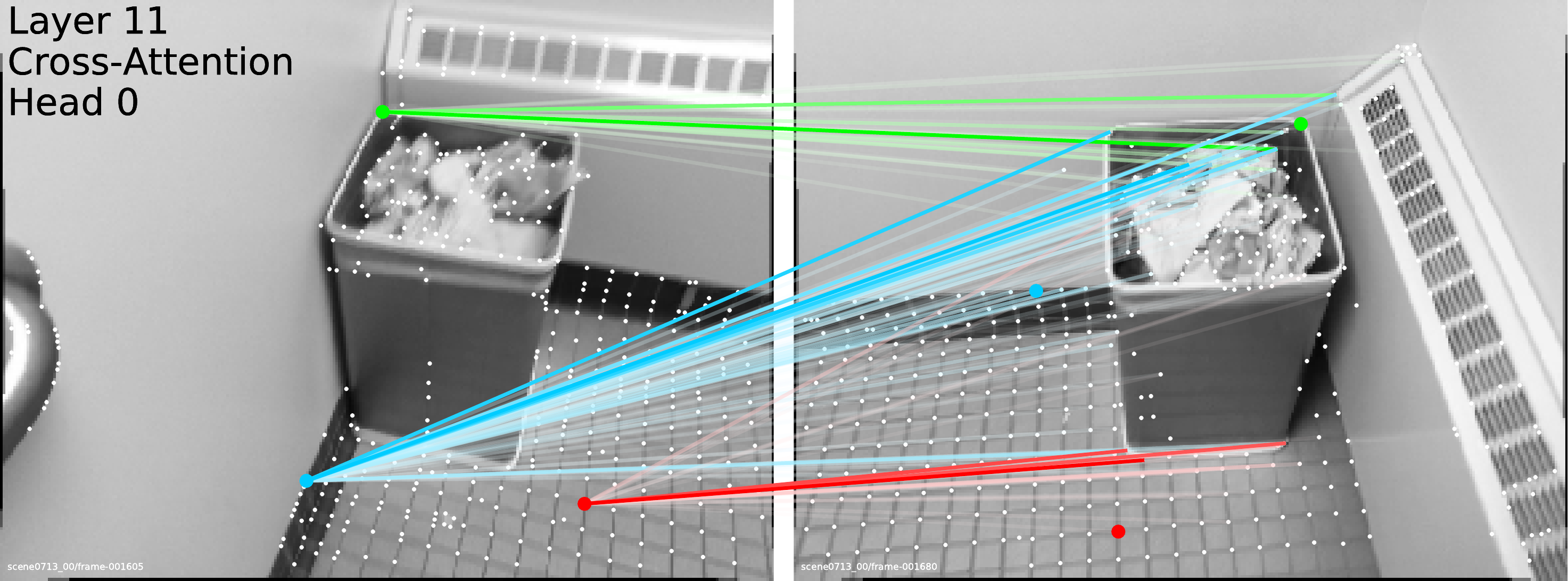}
    
    \vspace{.5mm}
    \includegraphics[width=\linewidth]{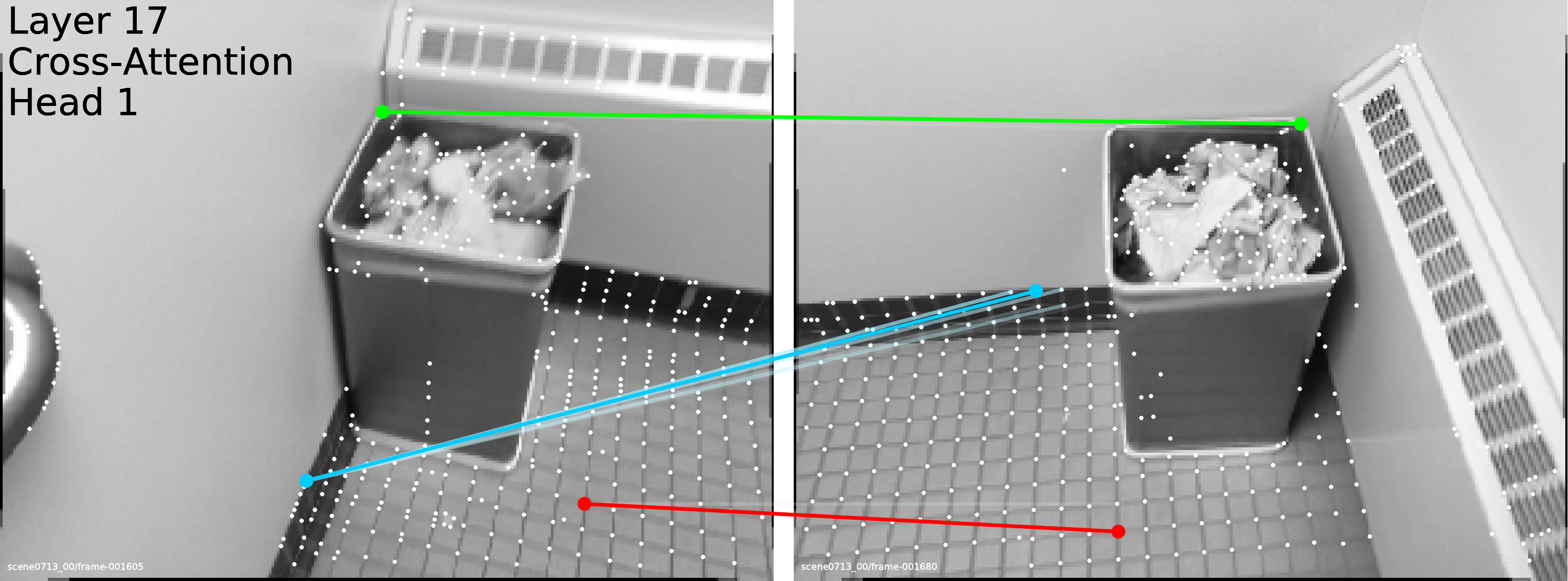}
\end{minipage}

\vspace{-.1cm}
\caption{{\bf Attention patterns across layers.} For this image pair (correctly matched by SuperGlue), we look at three specific keypoints that can be matched with different levels of difficulty: the {\color{green}easy keypoint}, the {\color[rgb]{0,.8,1}medium keypoint}, and the {\color{red}difficult keypoint}.
We visualize self- and cross-attention weights (within images $A$ and $B$, and from $A$ to $B$, respectively) of selected layers and heads, varying the edge opacity with $\alpha_{ij}$.
The self-attention initially attends all over the image (row 1), and gradually focuses on a small neighborhood around each keypoint (last row). Similarly, some cross-attention heads focus on candidate matches, and successively reduce the set that is inspected. The {\color{green}easy keypoint} is matched as early as layer 9, while more difficult ones are only matched at the last layer.
Similarly as in Figure~\ref{fig:supp-attention-span}, the self- and cross-attention spans generally shrink throughout the layers. They however increase in layer 11, which attends to other locations --~seemingly distinctive ones --~that are further away. We hypothesize that SuperGlue attempts to disambiguate challenging matches using additional context.\looseness=-1
}
\label{fig:supp-attention-qualitative}
\end{figure*}\fi
\else
\clearpage

\fi

\clearpage
{\small
\bibliographystyle{ieee_fullname}
\bibliography{mybib}
}

\end{document}